\pdfoutput=1
\documentclass{article}
\usepackage[most]{tcolorbox}
\usepackage{enumitem}
\usepackage{amsmath}
\usepackage{multirow}
\usepackage[table]{xcolor}
\usepackage{graphicx}
\usepackage{tikz}
\usetikzlibrary{arrows.meta,calc}
\usepackage{pgfplots}
\usepgfplotslibrary{groupplots}
\pgfplotsset{compat=1.18}
\usepackage{adjustbox}
\usepackage{makecell}
\usepackage{placeins}
\usepackage{booktabs,titletoc}
\usepackage{listings}
\usepackage{flafter}
\usepackage{algorithm}
\usepackage{algpseudocode}
\tcbuselibrary{listings, skins}

\newtcolorbox{promptbox}[1]{
    colback=gray!8,
    colframe=gray!8,
    coltitle=white,
    colbacktitle=gray!30!black,
    title={\textbf{\small #1}},
    fonttitle=\bfseries\small,
    boxrule=0.4pt,
    arc=2pt,
    left=8pt, right=8pt,
    top=4pt, bottom=6pt,
    titlerule=0pt,
    toptitle=4pt,
    bottomtitle=4pt,
    width=\textwidth,
    enhanced
}

\definecolor{promptInk}{HTML}{252629}
\definecolor{promptSky}{HTML}{E3EDF7}
\definecolor{promptSage}{HTML}{EDF3E7}
\definecolor{promptPeach}{HTML}{F5E3D5}
\definecolor{promptSkyLine}{HTML}{A8BBCD}
\definecolor{promptSageLine}{HTML}{B3BFA8}
\definecolor{promptPeachLine}{HTML}{CBB3A1}

\newtcolorbox{promptblue}[1]{
    colback=promptSky!35!white,
    colframe=promptSkyLine,
    coltitle=promptInk,
    colbacktitle=promptSky,
    title={\large \texttt{#1}},
    fonttitle=\normalfont\large,
    boxrule=0.8pt,
    arc=6pt,
    left=10pt, right=10pt,
    top=6pt, bottom=8pt,
    titlerule=0pt,
    toptitle=6pt,
    bottomtitle=6pt,
    width=\textwidth,
    enhanced,
    breakable
}

\newtcolorbox{promptteal}[1]{
    colback=promptSage!35!white,
    colframe=promptSageLine,
    coltitle=promptInk,
    colbacktitle=promptSage,
    title={\large \texttt{#1}},
    fonttitle=\normalfont\large,
    boxrule=0.8pt,
    arc=6pt,
    left=10pt, right=10pt,
    top=6pt, bottom=8pt,
    titlerule=0pt,
    toptitle=6pt,
    bottomtitle=6pt,
    width=\textwidth,
    enhanced,
    breakable
}

\newtcolorbox{promptorange}[1]{
    colback=promptPeach!35!white,
    colframe=promptPeachLine,
    coltitle=promptInk,
    colbacktitle=promptPeach,
    title={\large \texttt{#1}},
    fonttitle=\normalfont\large,
    boxrule=0.8pt,
    arc=6pt,
    left=10pt, right=10pt,
    top=6pt, bottom=8pt,
    titlerule=0pt,
    toptitle=6pt,
    bottomtitle=6pt,
    width=\textwidth,
    enhanced
}

\newcommand{\method}{\textsc{MedChain-R1}}
\newcommand{\medsg}{MedSG}
\newcommand{\dup}[1]{{\tiny\textcolor{green!55!black}{(+#1)}}}
\newcommand{\ddn}[1]{{\tiny\textcolor{red!75!black}{($-$#1)}}}
\newcommand{\mc}[2]{\makecell{#1\\\textcolor{gray}{#2}}}

\usepackage{iclr2027_conference,times}
\iclrfinalcopy
\usepackage[utf8]{inputenc}
\usepackage[T1]{fontenc}
\usepackage{hyperref}
\usepackage{url}
\usepackage{amsfonts}
\usepackage{nicefrac}
\usepackage{microtype}

\title{Think Before You Locate: Clinical Chain-of-Thought
Supervision for Sequential Medical Image Grounding}

\author{%
\textbf{Adinath Madhavrao Dukre$^{1,2,*}$, Govinda Kolli$^{1,2,*}$, Yifan Lu$^{1}$, Ziyun Zou$^{1}$,} \\
\textbf{Azmine Toushik Wasi$^{3}$, Sara Atito$^{3}$, Behzad Bozorgtabar$^{4}$, Dwarikanath Mahapatra$^{5}$,} \\ \textbf{Muhammad Awais$^{3}$, Imran Razzak$^{1,2}$} \\[4pt]
$^{1}$Mohamed bin Zayed University of Artificial Intelligence, Abu Dhabi, UAE \\
$^{2}$Vitalverse, Abu Dhabi, UAE \\
$^{3}$CVSSP,University of Surrey, UK  \\
$^{5}$Aarhus University, Denmark \\
$^{6}$Khalifa University of Science, Technology and Research, Abu Dhabi, UAE \\
$^{*}$Equal contribution \\[2pt]
\texttt{\{adinath.dukre, imran.razzak\}@mbzuai.ac.ae}
}

\begin{document}

\maketitle
 \lhead{}
\renewcommand{\headrulewidth}{0pt}

\begin{abstract}
Clinicians rarely localize a finding from one image alone. They compare current and prior scans, follow structures across slices, and match regions across views or modalities. Yet vision-language models for sequential medical grounding usually learn only from bounding boxes. Boxes show where the target is, but they do not explain why that region in that image answers the question. We study whether explicit, image-grounded reasoning can improve localization, and which tasks benefit from it. We introduce \method, whose core is MedChain, a corpus of 182,573 filtered rationales spanning eight sequential grounding tasks and ten imaging modalities. A medical VLM (FlemingVL-38B) first drafts a caption-like rationale, and a multimodal refiner (Qwen3.5-397B-A17B) then re-examines the images, question and annotated box to write a box-conditioned explanation with explicit cross-image reasoning. We fine-tune Qwen3.5 on these rationales with deliberative chain-of-thought supervision (dCoT-SFT). Across 4B, 9B, and 27B models, dCoT-SFT sets a new state of the art on MedSG-Bench. The 27B model reaches 77.24 mIoU and 85.20 Acc@0.5, improving over the strongest prior model by 4.69 and 5.49 points. In a paired comparison using fixed 4B checkpoints, dCoT-SFT improves over answer-only SFT by 3.56 mIoU (95\% CI 3.13--4.00). The gains mainly come from change detection and object tracking, while performance drops on three appearance-matching tasks. Replacing a model's own rationale with a rationale from another task reduces mIoU by 12.2 points, showing that localization depends strongly on the reasoning context. Finally, adding GRPO with an IoU reward after dCoT-SFT gives only small and task-dependent gains, which we analyze through controlled multi-seed development experiments.
\end{abstract}

\begin{figure*}[t]
\centering
\begin{minipage}{0.98\linewidth}
\begingroup
\definecolor{ovInk}{HTML}{252629}
\definecolor{ovMuted}{HTML}{60636A}
\definecolor{ovBlue}{HTML}{3B73A7}
\definecolor{ovTeal}{HTML}{476A86}
\definecolor{ovGreen}{HTML}{287548}
\definecolor{ovLine}{HTML}{B9BEC5}
\definecolor{ovSky}{HTML}{E3EDF7}
\definecolor{ovCream}{HTML}{FBF6E2}
\definecolor{ovPeach}{HTML}{F5E3D5}
\definecolor{ovSage}{HTML}{EDF3E7}
\definecolor{ovRose}{HTML}{D988AB}
\definecolor{ovCyan}{HTML}{91C9E1}
\definecolor{ovBarOne}{HTML}{CDDCEB}
\definecolor{ovBarTwo}{HTML}{9FBFDA}
\definecolor{ovBarThree}{HTML}{6E9EC4}
\definecolor{taskCard1}{HTML}{EEF2F5}
\definecolor{taskCard2}{HTML}{EEF2F5}
\definecolor{taskCard3}{HTML}{EEF2F5}
\definecolor{taskCard4}{HTML}{EEF2F5}
\definecolor{taskCard5}{HTML}{EEF2F5}
\definecolor{taskCard6}{HTML}{EEF2F5}
\definecolor{taskCard7}{HTML}{EEF2F5}
\definecolor{taskCard8}{HTML}{EEF2F5}
\resizebox{\linewidth}{!}{%
\begin{tikzpicture}[x=1mm,y=-1mm,
 every node/.style={font=\sffamily\fontsize{7.5}{9.2}\selectfont,text=ovInk,inner sep=0pt,outer sep=0pt},
 body/.style={anchor=north west,align=left},
 tiny/.style={font=\sffamily\fontsize{6.6}{7.7}\selectfont},
 small/.style={font=\sffamily\fontsize{7}{8.3}\selectfont},
 heading/.style={anchor=west,font=\sffamily\bfseries\fontsize{9}{10}\selectfont}]
\path[use as bounding box] (0,0) rectangle (170,169);
\fill[taskCard1,rounded corners=2.5mm] (-1,-1) rectangle (171,98);
\filldraw[fill=ovInk,draw=ovCream,line width=1.6pt] (53,0) rectangle (117,8);
\node[font=\sffamily\bfseries\fontsize{16}{17}\selectfont] at (85,4){\textcolor{ovRose}{Med}\textcolor{ovCyan}{Chain}\textcolor{ovCream}{-R1}};
\node[anchor=west,font=\sffamily\bfseries\fontsize{7.5}{9}\selectfont] at (2,4){(a) Sequential grounding};
\begin{scope}[shift={(0,9)}]
\filldraw[fill=taskCard1,draw=taskCard1!65!ovMuted,line width=.45pt,rounded corners=1.5mm] (0,0) rectangle (41,42);
\node[anchor=west,font=\sffamily\bfseries\fontsize{7.2}{8}\selectfont] at (2,3){1\enspace Object tracking};
\node[anchor=east,font=\sffamily\fontsize{6.5}{7.5}\selectfont,text=ovMuted] at (39,3){MRI};
\node[anchor=north] at (20,10){\adjincludegraphics[trim={0.011\width} {0.881\height} {0.8665\width} {0.017\height},clip,width=26mm,height=10.5mm,keepaspectratio]{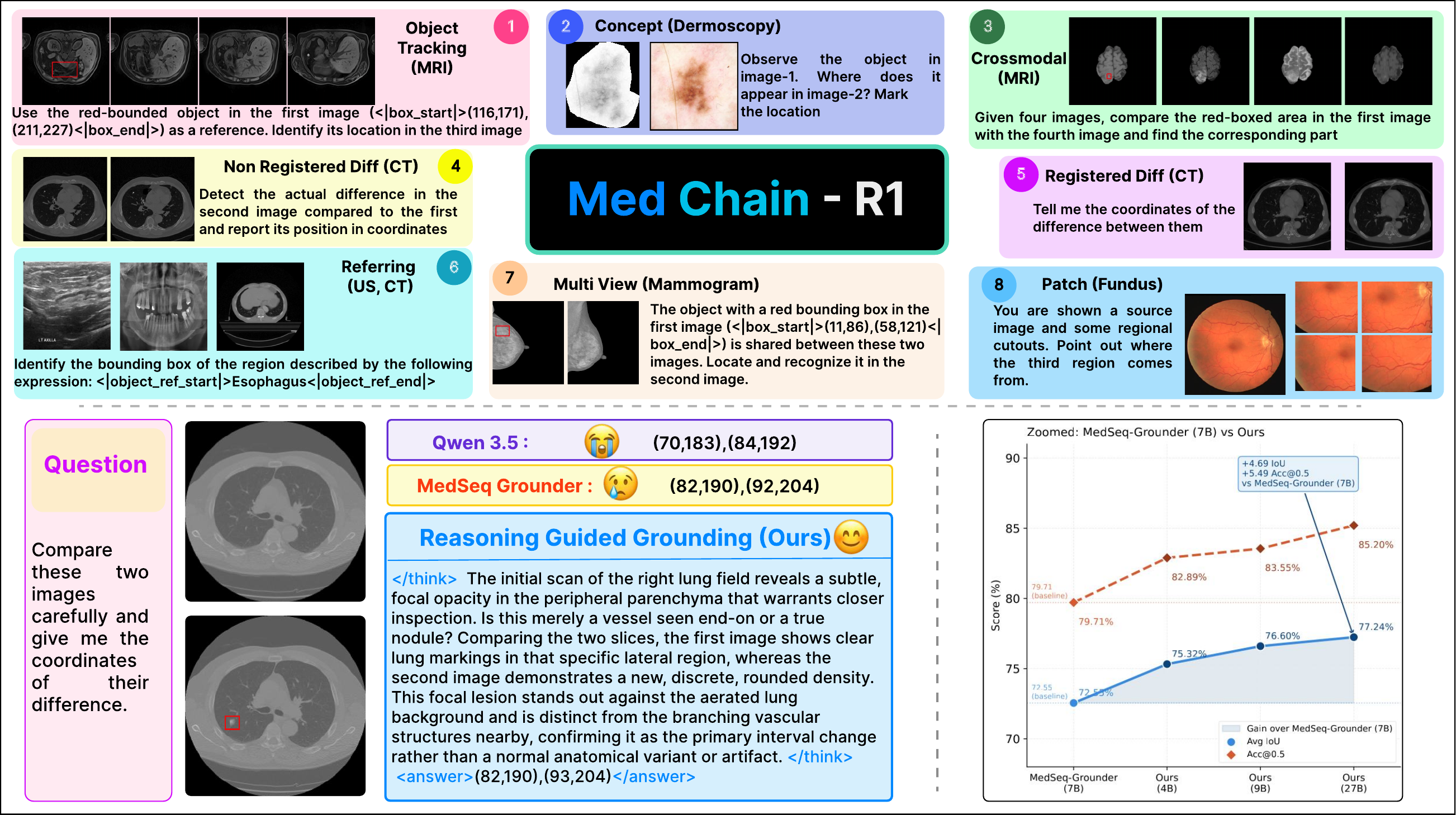}};
\node[anchor=north] at (20,21.5){\adjincludegraphics[trim={0.1335\width} {0.881\height} {0.744\width} {0.017\height},clip,width=26mm,height=10.5mm,keepaspectratio]{fig2.pdf}};
\node[body,text width=38mm,font=\sffamily\fontsize{6.6}{7.7}\selectfont] at (1,34){Use the red-bounded object in the first image ... as a reference. Identify its location in the third image.};
\end{scope}
\begin{scope}[shift={(43,9)}]
\filldraw[fill=taskCard2,draw=taskCard2!65!ovMuted,line width=.45pt,rounded corners=1.5mm] (0,0) rectangle (41,42);
\node[anchor=west,font=\sffamily\bfseries\fontsize{7.2}{8}\selectfont] at (2,3){2\enspace Concept};
\node[anchor=east,font=\sffamily\fontsize{6.5}{7.5}\selectfont,text=ovMuted] at (39,3){Dermoscopy};
\node[anchor=north] at (20,10){\adjincludegraphics[trim={0.385\width} {0.844\height} {0.493\width} {0.05\height},clip,width=40mm,height=22mm,keepaspectratio]{fig2.pdf}};
\node[body,text width=38mm,font=\sffamily\fontsize{6.6}{7.7}\selectfont] at (1,31){Observe the object in image-1. Where does it appear in image-2? Mark the location.};
\end{scope}
\begin{scope}[shift={(86,9)}]
\filldraw[fill=taskCard3,draw=taskCard3!65!ovMuted,line width=.45pt,rounded corners=1.5mm] (0,0) rectangle (41,42);
\node[anchor=west,font=\sffamily\bfseries\fontsize{7.2}{8}\selectfont] at (2,3){3\enspace Cross-modal};
\node[anchor=east,font=\sffamily\fontsize{6.5}{7.5}\selectfont,text=ovMuted] at (39,3){MRI};
\node[anchor=north] at (20,7){\adjincludegraphics[trim={0.732\width} {0.881\height} {0.1415\width} {0.02\height},clip,width=26mm,height=10.5mm,keepaspectratio]{fig2.pdf}};
\node[anchor=north] at (20,18.5){\adjincludegraphics[trim={0.8585\width} {0.881\height} {0.015\width} {0.02\height},clip,width=26mm,height=10.5mm,keepaspectratio]{fig2.pdf}};
\node[body,text width=38mm,font=\sffamily\fontsize{6.6}{7.7}\selectfont] at (1.5,30.5){Given four images, compare the red-boxed area in the first image with the fourth image and find the corresponding part.};
\end{scope}
\begin{scope}[shift={(129,9)}]
\filldraw[fill=taskCard4,draw=taskCard4!65!ovMuted,line width=.45pt,rounded corners=1.5mm] (0,0) rectangle (41,42);
\node[anchor=west,font=\sffamily\bfseries\fontsize{7.2}{8}\selectfont] at (2,3){4\enspace Non-registered diff.};
\node[anchor=east,font=\sffamily\fontsize{6.5}{7.5}\selectfont,text=ovMuted] at (39,3){CT};
\node[anchor=north] at (20,10){\adjincludegraphics[trim={0.011\width} {0.705\height} {0.871\width} {0.21\height},clip,width=40mm,height=22mm,keepaspectratio]{fig2.pdf}};
\node[body,text width=38mm,font=\sffamily\fontsize{6.6}{7.7}\selectfont] at (1,28){Detect the actual difference in the second image compared to the first and report its position in coordinates.};
\end{scope}
\begin{scope}[shift={(0,53)}]
\filldraw[fill=taskCard5,draw=taskCard5!65!ovMuted,line width=.45pt,rounded corners=1.5mm] (0,0) rectangle (41,44);
\node[anchor=west,font=\sffamily\bfseries\fontsize{7.2}{8}\selectfont] at (2,3){5\enspace Registered diff.};
\node[anchor=east,font=\sffamily\fontsize{6.5}{7.5}\selectfont,text=ovMuted] at (39,3){CT};
\node[anchor=north] at (20,10){\adjincludegraphics[trim={0.854\width} {0.695\height} {0.017\width} {0.203\height},clip,width=40mm,height=22mm,keepaspectratio]{fig2.pdf}};
\node[body,text width=38mm,font=\sffamily\fontsize{6.6}{7.7}\selectfont] at (1,29){Tell me the coordinates of the difference between them.};
\end{scope}
\begin{scope}[shift={(43,53)}]
\filldraw[fill=taskCard6,draw=taskCard6!65!ovMuted,line width=.45pt,rounded corners=1.5mm] (0,0) rectangle (41,44);
\node[anchor=west,font=\sffamily\bfseries\fontsize{7.2}{8}\selectfont] at (2,3){6\enspace Referring};
\node[anchor=east,font=\sffamily\fontsize{6.5}{7.5}\selectfont,text=ovMuted] at (39,3){US / CT};
\node[anchor=north] at (20,10){\adjincludegraphics[trim={0.011\width} {0.576\height} {0.791\width} {0.322\height},clip,width=40mm,height=22mm,keepaspectratio]{fig2.pdf}};
\node[body,text width=38mm,font=\sffamily\fontsize{6.6}{7.7}\selectfont] at (1,24){Identify the bounding box of the region described by the following expression: ... Esophagus ...};
\end{scope}
\begin{scope}[shift={(86,53)}]
\filldraw[fill=taskCard7,draw=taskCard7!65!ovMuted,line width=.45pt,rounded corners=1.5mm] (0,0) rectangle (41,44);
\node[anchor=west,font=\sffamily\bfseries\fontsize{7.2}{8}\selectfont] at (2,3){7\enspace Multi-view};
\node[anchor=east,font=\sffamily\fontsize{6.5}{7.5}\selectfont,text=ovMuted] at (39,3){Mammogram};
\node[anchor=north] at (20,10){\adjincludegraphics[trim={0.338\width} {0.529\height} {0.564\width} {0.363\height},clip,width=38mm,height=22mm,keepaspectratio]{fig2.pdf}};
\node[body,text width=38mm,font=\sffamily\fontsize{6.4}{7.3}\selectfont] at (1,34){The object with a red bounding box in the first image ... is shared between these two images. Locate and recognize it in the second image.};
\end{scope}
\begin{scope}[shift={(129,53)}]
\filldraw[fill=taskCard8,draw=taskCard8!65!ovMuted,line width=.45pt,rounded corners=1.5mm] (0,0) rectangle (41,44);
\node[anchor=west,font=\sffamily\bfseries\fontsize{7.2}{8}\selectfont] at (2,3){8\enspace Patch};
\node[anchor=east,font=\sffamily\fontsize{6.5}{7.5}\selectfont,text=ovMuted] at (39,3){Fundus};
\node[anchor=north] at (10,10){\adjincludegraphics[trim={.814\width} {.522\height} {.117\width} {.363\height},clip,width=17mm,height=22mm,keepaspectratio]{fig2.pdf}};
\node[anchor=north] at (29,10){\adjincludegraphics[trim={.888\width} {.519\height} {.015\width} {.346\height},clip,width=19mm,height=22mm,keepaspectratio]{fig2.pdf}};
\node[body,text width=38mm,font=\sffamily\fontsize{6.6}{7.7}\selectfont] at (1,28){You are shown a source image and some regional cutouts. Point out where the third region comes from.};
\end{scope}

\begin{scope}[yshift=-12mm]
\draw[ovLine,dashed,line width=.65pt] (0,87)--(170,87);
\begin{scope}[yshift=4mm]
\node[anchor=north,align=center,text width=26mm,font=\sffamily\bfseries\fontsize{8.5}{10}\selectfont] at (13,150){(b) Reasoning-\\guided grounding};
\filldraw[fill=ovCream,draw=ovInk!65,rounded corners=.8mm,line width=.55pt] (28,94) rectangle (106,101);
\node[body,small,text width=74mm] at (30,95.5){\textbf{Question:} Compare these two images carefully and give me the coordinates of their difference.};
\node[anchor=north west] at (0,94){\adjincludegraphics[trim={.127\width} {.024\height} {.751\width} {.515\height},clip,width=26mm]{fig2.pdf}};
\filldraw[fill=ovSage,draw=ovInk!65,rounded corners=.6mm,line width=.55pt] (28,103) rectangle (106,108.5);
\filldraw[fill=ovPeach,draw=ovInk!65,rounded corners=.6mm,line width=.55pt] (28,110) rectangle (106,115.5);
\node[anchor=west,font=\sffamily\bfseries\fontsize{7.4}{8.5}\selectfont] at (36,105.75){Qwen3.5};
\node[anchor=east,font=\ttfamily\fontsize{7.2}{8}\selectfont] at (104,105.75){(70,183),(84,192)};
\node[anchor=west,font=\sffamily\bfseries\fontsize{7.4}{8.5}\selectfont] at (36,112.75){MedSeq-Grounder};
\node[anchor=east,font=\ttfamily\fontsize{7.2}{8}\selectfont] at (104,112.75){(80,187),(90,200)};
\fill[ovSky] (28,117) rectangle (106,123);
\draw[ovInk!75,rounded corners=.7mm,line width=.65pt] (28,117) rectangle (106,159);
\draw[ovInk!50,line width=.4pt] (28,123)--(106,123);
\node[anchor=west,font=\sffamily\bfseries\fontsize{7.8}{9}\selectfont] at (36,120){MedChain-R1 (ours)};
\node[body,text width=74mm,font=\rmfamily\fontsize{7.2}{8.5}\selectfont] at (30,125){\textcolor{ovMuted}{\texttt{\textless think\textgreater}}\enspace The initial scan of the right lung field reveals a subtle, focal opacity in the peripheral parenchyma that warrants closer inspection. Is this merely a vessel seen end-on or a true nodule? Comparing the two slices, the first image shows clear lung markings in that specific lateral region, whereas the \textcolor{ovGreen}{\textbf{second image demonstrates a new, discrete, rounded density}}. This focal lesion stands out against the aerated lung background and is distinct from the branching vascular structures nearby, \textcolor{ovGreen}{\textbf{confirming it as the primary interval change}} rather than a normal anatomical variant or artifact.\par
\textcolor{ovMuted}{\texttt{\textless/think\textgreater}}\enspace
\textcolor{ovGreen}{\texttt{\textless answer\textgreater\textbf{(82,190),(93,204)}\textless/answer\textgreater}}};

\foreach \faceY in {105.75,112.75}{
 \begin{scope}[shift={(32,\faceY)}]
 \filldraw[fill=red!12,draw=red!60!black,line width=.35pt] (0,0) circle (1.7mm);
 \draw[red!60!black,line width=.6pt,line cap=round] (-.75,-.75)--(.75,.75) (-.75,.75)--(.75,-.75);
 \end{scope}}

\begin{scope}[shift={(32,120)}]
\filldraw[fill=ovGreen!12,draw=ovGreen,line width=.35pt] (0,0) circle (1.7mm);
\draw[ovGreen,line width=.65pt,line cap=round,line join=round] (-.8,.05)--(-.25,.65)--(.85,-.6);
\end{scope}

\end{scope}
\definecolor{plotBlue}{HTML}{4D80B7}
\definecolor{plotOrange}{HTML}{B67655}
\draw[ovLine,dashed,line width=.65pt] (110,89)--(110,155);
\node[heading] at (115,91){(c) Model comparison};
\begin{scope}[shift={(0,94)},yscale=1.0701754,shift={(0,-94)}]
\filldraw[fill=white,draw=ovInk!70,rounded corners=.7mm,line width=.55pt] (113,94) rectangle (170,151);
\node[font=\sffamily\fontsize{7}{8}\selectfont] at (141.5,97){MedSeq-Grounder (7B) vs Ours};
\foreach \val in {70,75,80,85,90}{
 \pgfmathsetmacro{\yy}{137-(\val-70)*1.7}
 \draw[ovLine!45,line width=.3pt] (120,\yy)--(168,\yy);
 \node[anchor=east,font=\sffamily\fontsize{7}{8}\selectfont] at (119,\yy){\val};}
\node[rotate=90,font=\sffamily\fontsize{6.8}{8}\selectfont] at (115,120){Score (\%)};
\draw[ovInk!70,line width=.5pt] (120,101)--(120,137)--(168,137);
\fill[plotBlue!15] (122,132.665)--(130.8,128.568)--(139.6,127.956)--(148.4,125.78)--(157.2,125.27)--(166,124.692)--(166,132.665)--cycle;
\draw[plotBlue!35,dotted,line width=.45pt] (120,132.665)--(168,132.665);
\draw[plotOrange!35,dotted,line width=.45pt] (120,120.493)--(168,120.493);
\draw[plotBlue,line width=.85pt] (122,132.665)--(130.8,128.568)--(139.6,127.956)--(148.4,125.78)--(157.2,125.27)--(166,124.692);
\draw[plotOrange,dashed,line width=.85pt] (122,120.493)--(130.8,115.971)--(139.6,115.087)--(148.4,113.965)--(157.2,113.506)--(166,111.16);
\foreach \x/\y/\v in {122/132.665/72.55,130.8/128.568/74.96,139.6/127.956/75.32,148.4/125.78/76.60,157.2/125.27/76.90,166/124.692/77.24}{
\filldraw[fill=plotBlue,draw=white,line width=.35pt] (\x,\y) circle (.7mm);
\ifdim\x pt=166pt
\node[anchor=north,font=\sffamily\fontsize{6.5}{8}\selectfont,text=plotBlue!75!black] at (\x,\y+1.2){\v};
\else
\node[anchor=south,font=\sffamily\fontsize{6.5}{8}\selectfont,text=plotBlue!75!black] at (\x,\y-1.2){\v};
\fi}
\foreach \x/\y/\v in {122/120.493/79.71,130.8/115.971/82.37,139.6/115.087/82.89,148.4/113.965/83.55,157.2/113.506/83.82,166/111.16/85.20}{
\filldraw[fill=plotOrange,draw=white,line width=.3pt] (\x,\y-.8)--(\x+.8,\y)--(\x,\y+.8)--(\x-.8,\y)--cycle;
\node[anchor=north,font=\sffamily\fontsize{6.5}{8}\selectfont,text=plotOrange!75!black] at (\x,\y+1.3){\v};}
\node[align=center,font=\sffamily\fontsize{5.4}{6.6}\selectfont] at (122,140.3){MedSeq\\7B};
\node[align=center,font=\sffamily\fontsize{5.4}{6.6}\selectfont] at (130.8,140.3){4B\\dCoT-SFT};
\node[align=center,font=\sffamily\fontsize{5.4}{6.6}\selectfont] at (139.6,140.3){4B\\GRPO};
\node[align=center,font=\sffamily\fontsize{5.4}{6.6}\selectfont] at (148.4,140.3){9B\\dCoT-SFT};
\node[align=center,font=\sffamily\fontsize{5.4}{6.6}\selectfont] at (157.2,140.3){9B\\GRPO};
\node[align=center,font=\sffamily\fontsize{5.4}{6.6}\selectfont] at (166,140.3){27B\\dCoT-SFT};
\filldraw[fill=plotBlue!6,draw=plotBlue!70,rounded corners=.4mm,line width=.4pt] (141,100) rectangle (167,109);
\node[anchor=north west,align=left,font=\sffamily\fontsize{6.5}{7.5}\selectfont] at (142,101){+4.69 mIoU\\+5.49 Acc@0.5\\vs 7B baseline};
\draw[-{Latex[length=1mm,width=.8mm]},ovInk!75,line width=.5pt] (161,109)--(165.8,123.6);
\draw[plotBlue,line width=.85pt] (118,145)--(122,145);
\fill[plotBlue] (120,145) circle (.6mm);
\node[anchor=west,font=\sffamily\fontsize{6.8}{8}\selectfont] at (123,145){mIoU};
\draw[plotOrange,dashed,line width=.85pt] (143,145)--(147,145);
\fill[plotOrange] (145,144.3)--(145.7,145)--(145,145.7)--(144.3,145)--cycle;
\node[anchor=west,font=\sffamily\fontsize{6.8}{8}\selectfont] at (148,145){Acc@0.5};
\fill[plotBlue!15] (118,148) rectangle (122,149.5);
\node[anchor=west,font=\sffamily\fontsize{6.3}{7.5}\selectfont] at (123,148.75){mIoU gain over 7B baseline};
\end{scope}
\end{scope}
\end{tikzpicture}%
}
\endgroup
\end{minipage}
\caption{\textbf{Sequential medical grounding with reasoning supervision.} (a) The eight MedSG-Bench tasks, grouped into image difference grounding (IDG) and image consistency grounding (ICG). (b) A registered-difference example: zero-shot Qwen3.5 and MedSeq-Grounder return imprecise boxes, whereas \method{} first identifies the new nodule and then localizes it. (c) MedSG-Bench mIoU and Acc@0.5 of MedSeq-Grounder (7B) and \method{} at 4B and 9B (dCoT-SFT and dCoT-SFT+GRPO) and 27B (dCoT-SFT).}
\label{fig:overview}
\end{figure*}
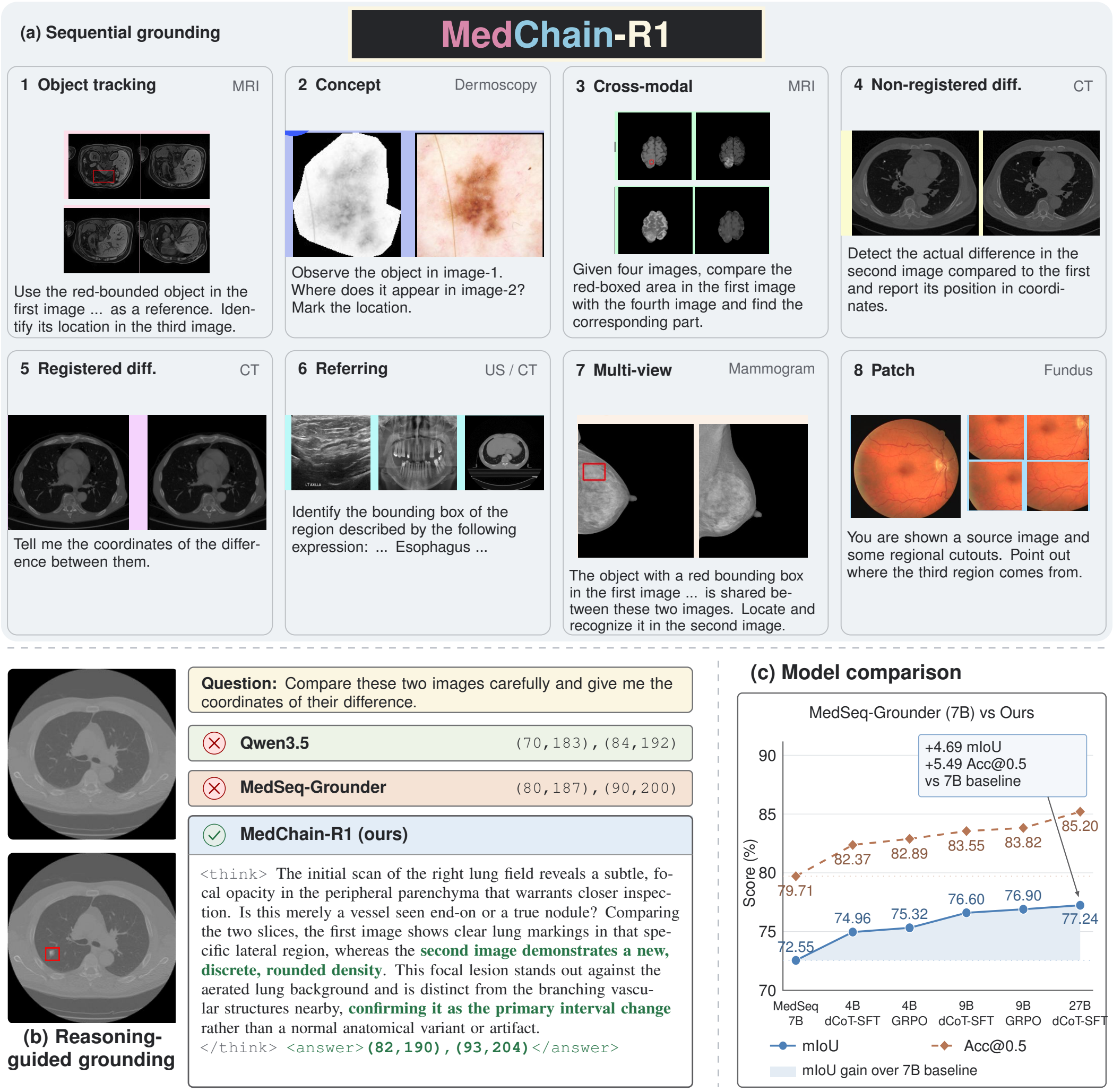

\section{Introduction}
\label{sec:intro}
A radiologist reading a follow-up chest CT (Computed Tomography) rarely cares about a nodule in isolation; the key question is whether it has changed since the prior study. The same reasoning is needed to track a lesion across slices, match regions between FLAIR and diffusion-weighted MRI, or find the same mass across mammographic views.
These tasks share one core ability: localizing a target by comparing a sequence of images. Medical visual grounding has advanced quickly~\citep{medrg, bannur2022biovil, bannur2023biovilt}, but most models still focus on single images. MedSG-Bench~\citep{medsgbench} made this gap measurable with 9,630 questions across eight tasks and ten modalities. Most general-purpose and medical multimodal LLMs score below 20 mIoU, while MedSeq-Grounder reaches 72.55 mIoU after instruction tuning on 188K sequential examples.

Box-only supervision leaves the key reasoning implicit. To answer \emph{what changed between these two scans?}, a model must match anatomy across images, compare the regions, ignore differences from windowing or positioning, and then place the box. An answer-only objective rewards the final coordinates without teaching which comparison supports them. Prior work in natural-image grounding and medical visual question answering shows that intermediate reasoning can improve accuracy beyond supervised baselines~\citep{visionr1, univgr1, medvlmr1, medr1, medgroundr1}. Sequential medical grounding has not yet used this approach because no reasoning corpus exists for the setting, and text-only reasoning models cannot inspect the images they need to explain.

\begin{figure*}[t]
\centering
\begingroup
\definecolor{mcInk}{HTML}{243247}
\definecolor{mcMuted}{HTML}{697281}
\definecolor{mcBlue}{HTML}{376B99}
\definecolor{mcPurple}{HTML}{6851B8}
\definecolor{mcTeal}{HTML}{347F70}
\definecolor{mcLine}{HTML}{B9C2CC}
\resizebox{\linewidth}{!}{%
\begin{tikzpicture}[x=1mm,y=-1mm,
 every node/.style={font=\sffamily\fontsize{8.3}{10.3}\selectfont,text=mcInk,inner sep=0pt,outer sep=0pt},
 flow/.style={-{Latex[length=2mm,width=1.5mm]},draw=mcMuted,line width=.75pt},
 body/.style={anchor=north west,align=left},
 small/.style={font=\sffamily\fontsize{7.3}{9}\selectfont},
 tag/.style={font=\ttfamily\fontsize{7.3}{9}\selectfont}]
\path[use as bounding box] (-.4,0) rectangle (170.4,116);
\draw[flow,draw=mcPurple!70,rounded corners=3mm] (17,21)--(17,5)--(140,5)--(140,21);
\node[fill=white,inner xsep=2mm,text=mcPurple,font=\sffamily\itshape\fontsize{8}{9}\selectfont] at (79,5){Original images + query + task context + annotated box};
\foreach \l/\r/\c in {0/34/mcInk,51/94/mcBlue,111/170/mcPurple}{
 \fill[\c!5] (\l+.7,22.7) rectangle (\r+.7,99.7);
 \draw[draw=mcLine,fill=white,rounded corners=1.6mm,line width=.6pt] (\l,22) rectangle (\r,99);
 \draw[\c,line width=1.2pt] (\l+3,22)--(\r-3,22);}
\node[anchor=west,font=\sffamily\bfseries\fontsize{9}{10}\selectfont] at (0,17){1\enspace Input example};
\node[anchor=west,text=mcBlue,font=\sffamily\bfseries\fontsize{9}{10}\selectfont] at (51,17){2\enspace Pseudo-CoT};
\node[anchor=west,text=mcPurple,font=\sffamily\bfseries\fontsize{9}{10}\selectfont] at (111,17){3\enspace Refined CoT};
\node[small,text=mcMuted] at (9.5,27){Image 1};
\node[small,text=mcMuted] at (24.5,27){Image 2};
\node[anchor=north west] at (2,30){\adjincludegraphics[trim={.430\width} {.611\height} {.429\width} {.149\height},clip,width=30mm]{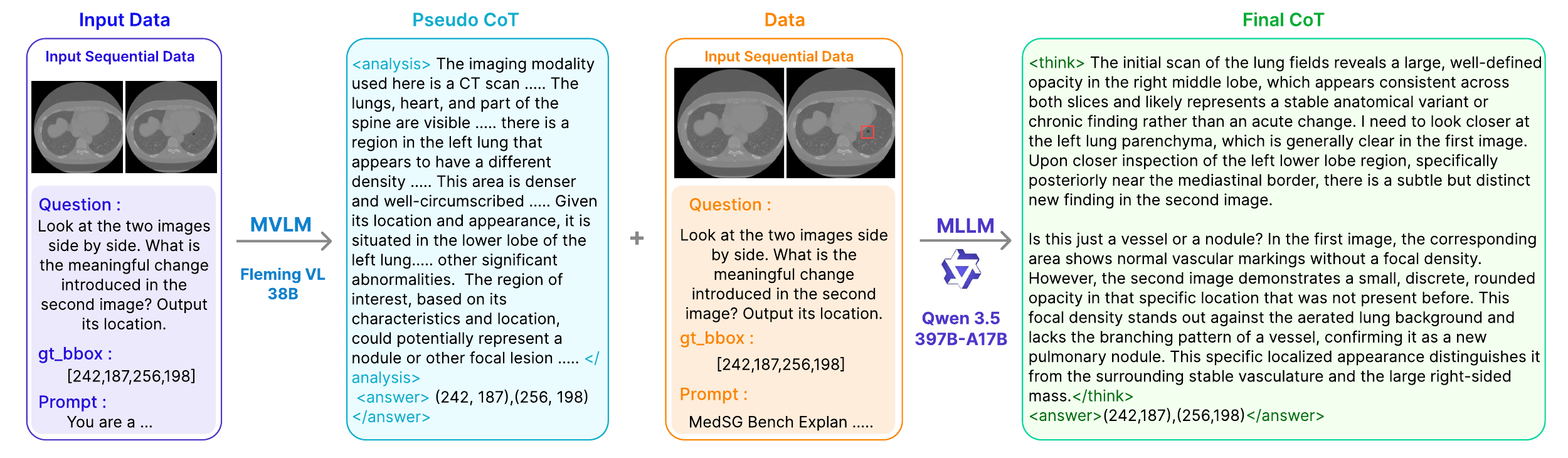}};
\node[body,font=\sffamily\bfseries\fontsize{8.3}{10.3}\selectfont] at (3,48){Question};
\node[body,text width=28mm] at (3,54){Look at the two images side by side. What is the meaningful change introduced in the second image? Output its location.};
\draw[mcLine!60,line width=.4pt] (3,82)--(31,82);
\node[body,small,text=mcMuted] at (3,85){Annotated box};
\node[body,tag] at (3,90){[242,187,256,198]};
\node[small,text=mcMuted] at (17,96){8 tasks / 10 modalities};
\begin{scope}[shift={(42.5,45)},draw=mcBlue,line width=.75pt,line cap=round]
 \filldraw[fill=mcBlue!7,rounded corners=1mm] (-3.6,-3) rectangle (3.6,2.6);
 \draw (0,-3)--(0,-4.5); \fill[mcBlue] (0,-4.5) circle (.55mm);
 \fill[mcBlue] (-1.4,-.8) circle (.5mm); \fill[mcBlue] (1.4,-.8) circle (.5mm);
 \draw (-1.2,1)--(1.2,1); \draw (-4.5,-1.3)--(-4.5,1); \draw (4.5,-1.3)--(4.5,1);
\end{scope}
\node[align=center,font=\sffamily\bfseries\fontsize{6.7}{8}\selectfont,text=mcBlue] at (42.5,54){FlemingVL\\38B};
\draw[flow,draw=mcBlue] (35,62)--(50,62);
\node[align=center,font=\sffamily\fontsize{6.5}{8}\selectfont,text=mcMuted] at (42.5,69){Medical\\teacher};
\node at (102.5,44){\includegraphics[width=8mm]{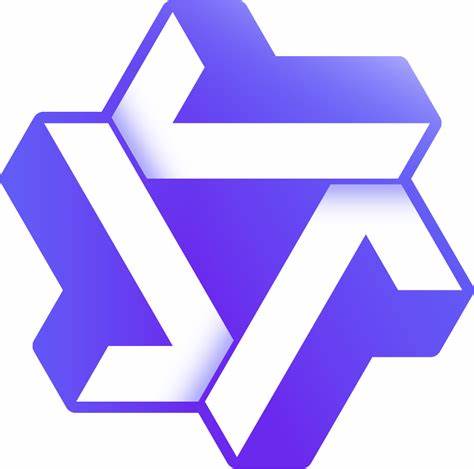}};
\node[align=center,font=\sffamily\bfseries\fontsize{6.7}{8}\selectfont,text=mcPurple] at (102.5,54){Qwen3.5\\397B-A17B};
\draw[flow,draw=mcPurple,dashed] (95,62)--(110,62);
\node[align=center,font=\sffamily\fontsize{6.5}{8}\selectfont,text=mcMuted] at (102.5,70){Optional\\sanitized\\rationale};
\node[body,tag,text=mcBlue] at (54,27){<analysis>};
\node[anchor=north east,small,text=mcMuted] at (91,27){Excerpts};
\node[body,text width=37mm] at (54,33){The imaging modality used here is a \textcolor{mcBlue}{\textbf{CT scan}} ..... The lungs, heart, and part of the spine are visible ..... there is a region in the left lung that appears to have a different density ..... This area is \textcolor{mcBlue}{denser and well-circumscribed} .....

... could potentially represent a nodule or other focal lesion .....};
\node[body,tag,text=mcBlue] at (54,85){</analysis>};
\draw[mcBlue!25,line width=.5pt] (54,90)--(91,90);
\node[tag,text=mcBlue] at (72.5,95){(242, 187),(256, 198)};
\node[body,tag,text=mcPurple] at (114,27){<think>};
\node[anchor=north east,small,text=mcMuted] at (167,27){Same example; excerpts};
\draw[mcPurple!35,line width=1.2pt] (114,34)--(114,46);
\node[body,text width=50mm] at (117,33){The initial scan of the lung fields reveals a large, well-defined opacity in the right middle lobe, which appears consistent across both slices ...};
\draw[mcBlue!40,line width=1.2pt] (114,50)--(114,60);
\node[body,text width=50mm] at (117,49){In the \textcolor{mcBlue}{\textbf{first image}}, the corresponding area shows normal vascular markings without a focal density.};
\draw[mcPurple!45,line width=1.2pt] (114,64)--(114,77);
\node[body,text width=50mm] at (117,63){However, the \textcolor{mcPurple}{\textbf{second image}} demonstrates a small, discrete, rounded opacity in that specific location that was not present before.};
\node[body,text width=53mm] at (114,80){... confirming it as a \textcolor{mcPurple}{\textbf{new pulmonary nodule}}.};
\node[body,tag,text=mcPurple] at (114,87){</think>};
\draw[mcPurple!25,line width=.5pt] (114,91)--(167,91);
\node[tag,font=\ttfamily\fontsize{7.1}{8}\selectfont,text=mcPurple] at (140.5,95){<answer>(242,187),(256,198)</answer>};
\draw[flow,draw=mcPurple,rounded corners=2mm] (160,99)--(160,108)--(151,108);
\draw[mcTeal,fill=mcTeal!8,line width=.7pt,line join=round] (141,104)--(148,104)--(145.5,107)--(145.5,110)--(143.5,111)--(143.5,107)--cycle;
\node[anchor=east,align=right,small] at (137,108){\textbf{4\enspace Quality filtering}\\and deduplication};
\draw[flow] (100,108)--(89,108);
\draw[mcBlue,fill=white,line width=.65pt] (79,104.5)--(79,110) arc[start angle=180,end angle=360,x radius=3mm,y radius=1mm]--(85,104.5);
\draw[mcBlue,fill=mcBlue!7,line width=.65pt] (82,104.5) ellipse (3mm and 1mm);
\draw[mcBlue,line width=.65pt] (79,107) arc[start angle=180,end angle=360,x radius=3mm,y radius=1mm];
\node[anchor=east,font=\sffamily\bfseries\fontsize{8.3}{10}\selectfont] at (76,108){MedChain pool};
\draw[flow] (51,108)--(40,108);
\node[align=center,small] at (18,108){\textbf{SFT\enspace /\enspace GRPO}\\Disjoint allocations};
\end{tikzpicture}%
}
\endgroup
\caption{\textbf{MedChain construction.} (1) Each training example contains an image sequence, a clinical question and an annotated box. (2) FlemingVL-38B drafts a pseudo-CoT from the images. (3) Multimodal Qwen3.5-397B-A17B receives the original images, question, task context, annotated box and an optional sanitized pseudo-CoT, and writes an image-grounded explanation in \texttt{<think>}/\texttt{<answer>} format. (4) Outputs are filtered and deduplicated into the MedChain pool, which is split into disjoint SFT and GRPO allocations.}
\label{fig:pipeline}
\end{figure*}
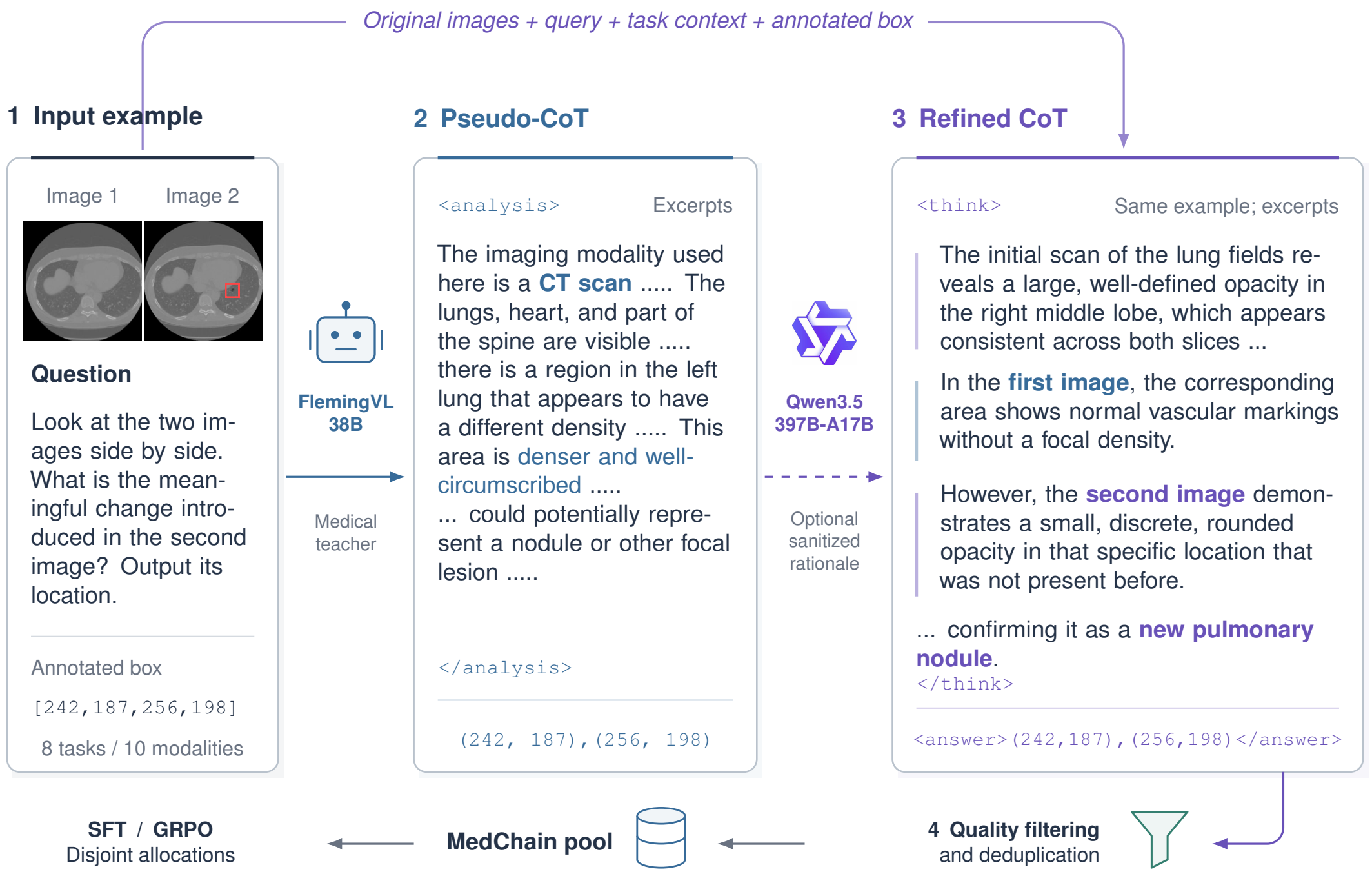

We address both problems with \method{}. We build its corpus, MedChain, with a two-stage multimodal pipeline (Figure~\ref{fig:pipeline}). First, a medical VLM drafts a preliminary rationale from the image sequence; on inspection, these drafts read like captions and rarely state the cross-image comparison the task requires. Then, a large multimodal model re-examines the original images, question, task context, and annotated box, and writes a short explanation of why the boxed region answers the question. The refiner keeps direct access to the images, which matters when the decisive evidence is a cross-image relation that a caption omits. We use the box as privileged information only during offline corpus construction. Each rationale therefore explains a known answer rather than recording an independent search process. We make this choice deliberately and directly test how the trained model uses its reasoning at inference time. We fine-tune Qwen3.5~\citep{qwen35} at 4B, 9B, and 27B parameters on this think-then-answer format, which we call dCoT-SFT. We then test whether GRPO~\citep{shao2024deepseekmath} with localization rewards provides additional gains.

Our experiments give four main findings:
\begin{enumerate}[
    leftmargin=1.2em,
    itemsep=0.3pt,
    topsep=0.3pt,
    parsep=0pt,
    partopsep=0pt
]
\item \textbf{Reasoning supervision sets a new SOTA.} dCoT-SFT outperforms MedSeq-Grounder at every model scale, reaching 74.96, 76.60, and 77.24 mIoU at 4B, 9B, and 27B, so average performance rises monotonically with model size (Table~\ref{tab:main_results}).
\item \textbf{The gains depend strongly on the task.} Compared with an answer-only 4B model, dCoT-SFT improves overall mIoU by 3.56 points (95\% CI 3.13--4.00). It gains 14.7--16.1 points on difference grounding and 6.5 points on object tracking, but loses 2.2--3.2 points on multi-view, patch, and referring grounding (Table~\ref{tab:fixed_4b}).
\item \textbf{Predictions depend on the reasoning context.} Forcing the dCoT-SFT model to answer immediately reduces mIoU by 2.9 points. Conditioning it on a rationale from another task reduces mIoU by 12.2 points.
\item \textbf{Reward optimization adds little after dCoT-SFT.} GRPO with an IoU+format reward improves benchmark mIoU by only 0.37 points and hurts concept grounding. In controlled three-seed development runs, it changes aggregate mIoU by less than 0.1 points, with a consistent gain only on multi-view grounding (Table~\ref{tab:grpo}).
\end{enumerate}

\noindent Our contributions are: (i) a two-stage multimodal rationale pipeline in which a medical VLM describes the images and a multimodal refiner re-inspects the full sequence to write the cross-image reasoning; (ii) MedChain, to our knowledge the first reasoning corpus for sequential medical grounding; (iii) dCoT-SFT models at three scales that set a new state of the art; and (iv) a controlled analysis of when reasoning helps, whether predictions depend on it, and what GRPO adds on top. We document MedChain with row- and ID-level accounting, a train--test image overlap audit, and a single-expert rationale audit. We report uncertainty throughout using paired example-level tests.

\section{Related Work}
\label{sec:related}
\noindent \textbf{Medical visual grounding.} Medical visual grounding supports interpretable vision-language models by linking text to specific image regions~\citep{medrg, bannur2022biovil, bannur2023biovilt}. Most prior work localizes findings in a single chest radiograph or pathology image~\citep{chen2024chexagent, bannur2024maira2}, while large instruction-tuned medical models extend this capability across modalities~\citep{li2023llavamed, chen2024huatuogpt, sellergren2025medgemma}. MedSG-Bench~\citep{medsgbench} introduced sequential grounding over longitudinal, multi-view, and cross-modal image sets, together with MedSeq-Grounder, which uses answer-only supervision. We use the same benchmark and study what explicit reasoning supervision adds.

\noindent \textbf{Reasoning supervision and reinforcement learning for grounding.} Chain-of-thought prompting~\citep{wei2022chain} and multimodal instruction tuning~\citep{liu2023llava, dai2023instructblip} showed that explicit intermediate reasoning can improve complex visual tasks. DeepSeek-R1~\citep{deepseekr1} popularized a cold-start reasoning SFT stage followed by GRPO~\citep{shao2024deepseekmath} with verifiable rewards, while DAPO~\citep{yu2025dapo} improved training stability. Vision-R1~\citep{visionr1} and UniVG-R1~\citep{univgr1} applied this approach to natural-image grounding. MedVLM-R1~\citep{medvlmr1}, Med-R1~\citep{medr1}, and MedGround-R1~\citep{medgroundr1} extended it to medical VQA and single-image medical grounding. We study the multi-image setting and find that reasoning SFT provides most of the gain, while IoU-rewarded GRPO adds little.

\noindent \textbf{Rationale distillation from multimodal teachers.} Prior work commonly distills rationales from large models into smaller models to scale reasoning supervision~\citep{deepseekr1, wei2022chain}. In medicine, the main challenge is keeping fluent explanations tied to visual evidence. Our refiner is multimodal: it inspects the original images rather than relying only on the teacher's text. We also condition the refiner on the annotated box. This answer conditioning improves alignment with the target, but it can also produce post-hoc rationalization. We therefore test whether the trained model actually relies on its reasoning in Section~\ref{sec:reliance}. Appendix~\ref{app:related} discusses additional related work.

\section{\method{}: Reasoning Supervision for Sequential Grounding}
\label{sec:method}
We present \method{}, which introduces explicit image-grounded reasoning supervision for sequential medical grounding through the MedChain rationale corpus, as illustrated in Figure \ref{fig:pipeline}. We use these rationales to train models with dCoT-SFT, then test whether GRPO adds further benefit beyond supervised reasoning.

\textbf{Task Description.}
Given a sequence of $N$ medical images $\mathcal{I}=\{I_1,\ldots,I_N\}$ and a clinical query $q$, we predict a box $\hat{b}=(x_1,y_1,x_2,y_2)$ that localizes the target in the correct image $I_k\in\mathcal{I}$. MedSG-Bench divides its eight tasks into two paradigms~\citep{medsgbench}. In \emph{image difference grounding} (IDG: registered and non-registered difference), the prompt gives the target image and the model must localize what changed. In \emph{image consistency grounding} (ICG: multi-view, object tracking, concept, patch, cross-modal, and referring grounding), the model must find the same structure across changes in view, slice, modality, or scale, and may also need to identify the target image. Both paradigms require cross-image comparison, while ICG additionally requires image selection.

\subsection{MedChain: image-grounded rationales}
\label{sec:medchain}

\begin{algorithm}[t]
\caption{MedChain construction via medical modality bridging.}
\label{alg:medchain}
\small
\algnewcommand{\IIf}[2]{\State \algorithmicif\ #1\ \algorithmicthen\ #2}
\begin{algorithmic}[1]
\Require examples $\mathcal{D}=\{(\mathcal{I}_i,q_i,t_i,b^*_i)\}_{i=1}^{M}$; teacher $\mathcal{F}$ (FlemingVL-38B); refiner $\mathcal{Q}$ (Qwen3.5-397B-A17B); quality threshold $\tau$
\State $\mathcal{C}\gets\emptyset$;\ \ $\mathcal{E}\gets\emptyset$ \Comment{accepted rationales; failures}
\For{each $(\mathcal{I}_i,q_i,t_i,b^*_i)\in\mathcal{D}$}
  \State $\tilde{c}_i\gets\mathcal{F}(\mathcal{I}_i,q_i,b^*_i)$ \Comment{Stage 1: medical pseudo-CoT}
  \IIf{$\tilde{c}_i$ is malformed \textbf{or} the context exceeds 8K tokens}{$\mathcal{E}\gets\mathcal{E}\cup\{i\}$; \textbf{continue}}
  \State $c^*_i\gets\mathcal{Q}(\mathcal{I}_i,q_i,t_i,b^*_i,\tilde{c}_i)$ \Comment{Stage 2: multimodal refinement}
  \IIf{$c^*_i$ is malformed \textbf{or} its answer box $\neq b^*_i$}{$\mathcal{E}\gets\mathcal{E}\cup\{i\}$; \textbf{continue}}
  \IIf{quality$(c^*_i)\geq\tau$}{$\mathcal{C}\gets\mathcal{C}\cup\{(\mathcal{I}_i,q_i,c^*_i,b^*_i)\}$}
\EndFor
\State Reprocess $\mathcal{E}$ with adjusted context splitting and merge into $\mathcal{C}$
\State Deduplicate $\mathcal{C}$ (182,573 IDs); split into disjoint $\mathcal{C}_{\text{GRPO}}$ (32,000) and $\mathcal{C}_{\text{SFT}}$ (150,573)
\State \Return $\mathcal{C}_{\text{SFT}},\ \mathcal{C}_{\text{GRPO}}$
\end{algorithmic}
\end{algorithm}

\noindent \textbf{Stage 1: Medical pseudo-CoT.} Text-only reasoning models cannot inspect CT, MRI, ultrasound or pathology images. We therefore start from a medical VLM. After comparing six candidate teachers on 100 sampled instances each (Appendix~\ref{app:A1}), we use FlemingVL-38B~\citep{flemingvl} to produce a preliminary rationale $\tilde{c}=\mathcal{F}(\mathcal{I},q,b^*)$ describing the modality, anatomy and visible content of each image, where $b^*$ is the annotated box. FlemingVL-38B acts mainly as a medical captioner: although prompted to relate the images, it describes what each image contains and rarely states how the images relate.

\begin{table*}[t]
\centering
\caption{\textbf{MedSG-Bench results.} Each cell shows mIoU (top) and \textcolor{gray}{Acc@0.5} (bottom), in percent. IDG: registered (RDG) and non-registered (NRDG) difference grounding. ICG: multi-view (MV), object tracking (OT), concept (VCG), patch (VPG), cross-modal (CMG) and referring (RG) grounding. \textbf{Bold}: our results that exceed MedSeq-Grounder, the strongest prior model, on the same metric; small numbers in parentheses give the difference from MedSeq-Grounder (\textcolor{green!55!black}{green}: higher, \textcolor{red!75!black}{red}: lower). Results for all baselines are in Table~\ref{tab:full_results}.}
\label{tab:main_results}
\scriptsize
\setlength{\tabcolsep}{3.2pt}
\renewcommand{\arraystretch}{0.95}
\resizebox{\textwidth}{!}{%
\begin{tabular}{llc|cc|cccccc|c}
\toprule
& \textbf{Model} & \textbf{Size} & \textbf{RDG} & \textbf{NRDG} & \textbf{MV} & \textbf{OT} & \textbf{VCG} & \textbf{VPG} & \textbf{CMG} & \textbf{RG} & \textbf{Avg.} \\
\midrule
\multirow{5}{*}{\rotatebox{90}{\tiny General}}
& Qwen2.5-VL & 72B & \mc{4.37}{2.60} & \mc{3.46}{0.80} & \mc{7.22}{2.78} & \mc{13.11}{7.70} & \mc{10.33}{3.50} & \mc{26.45}{6.30} & \mc{16.32}{7.00} & \mc{20.19}{14.10} & \mc{13.35}{6.12} \\
& InternVL3 & 78B & \mc{0.20}{0.00} & \mc{0.53}{0.00} & \mc{6.35}{2.43} & \mc{13.03}{8.00} & \mc{3.57}{0.90} & \mc{11.81}{2.50} & \mc{3.34}{0.85} & \mc{12.76}{8.10} & \mc{6.44}{2.90} \\
& LLaVA-OneVision & 72B & \mc{2.58}{0.80} & \mc{2.87}{0.90} & \mc{11.74}{1.39} & \mc{9.61}{2.30} & \mc{10.95}{3.30} & \mc{32.38}{20.30} & \mc{16.24}{5.40} & \mc{15.43}{6.68} & \mc{13.21}{5.18} \\
& Migician & 7B & \mc{15.26}{7.80} & \mc{14.49}{6.10} & \mc{18.16}{7.84} & \mc{21.38}{14.90} & \mc{14.23}{7.20} & \mc{28.87}{13.70} & \mc{21.41}{12.15} & \mc{25.30}{18.02} & \mc{20.29}{11.39} \\
& Qwen3.5 (zero-shot) & 27B & \mc{0.00}{0.00} & \mc{0.00}{0.00} & \mc{8.48}{4.88} & \mc{28.39}{24.50} & \mc{0.25}{0.00} & \mc{2.50}{0.20} & \mc{0.25}{0.00} & \mc{0.82}{0.00} & \mc{4.42}{3.15} \\
\midrule
\multirow{4}{*}{\rotatebox{90}{\tiny Medical}}
& HuatuoGPT-Vision & 34B & \mc{1.44}{0.00} & \mc{2.15}{0.00} & \mc{9.41}{1.65} & \mc{13.25}{8.30} & \mc{6.43}{0.70} & \mc{14.53}{1.40} & \mc{10.60}{2.60} & \mc{8.60}{1.75} & \mc{8.57}{2.09} \\
& MedGemma & 27B & \mc{0.01}{0.00} & \mc{0.06}{0.00} & \mc{5.29}{2.26} & \mc{22.26}{16.10} & \mc{2.27}{0.00} & \mc{3.11}{0.30} & \mc{2.92}{0.35} & \mc{5.35}{0.61} & \mc{4.94}{2.14} \\
& MedGRPO-7B-RL & 7B & \mc{0.09}{0.00} & \mc{0.06}{0.00} & \mc{3.68}{1.48} & \mc{17.63}{15.00} & \mc{4.57}{0.40} & \mc{18.44}{1.20} & \mc{0.78}{0.15} & \mc{9.31}{3.37} & \mc{6.27}{2.45} \\
& MedSeq-Grounder & 7B & \mc{83.29}{93.20} & \mc{83.72}{94.10} & \mc{55.03}{60.19} & \mc{62.10}{67.20} & \mc{74.11}{82.60} & \mc{85.25}{98.80} & \mc{78.77}{82.75} & \mc{60.43}{65.59} & \mc{72.55}{79.71} \\
\midrule
\multirow{5}{*}{\rotatebox{90}{\tiny Ours}}
& \cellcolor[RGB]{246,237,233}\method{} (dCoT-SFT) & 4B & \makecell{82.86\\[-3pt]\ddn{0.43}\\\textcolor{gray}{\textbf{93.30}}\\[-3pt]\dup{0.10}} & \makecell{81.30\\[-3pt]\ddn{2.42}\\\textcolor{gray}{91.20}\\[-3pt]\ddn{2.90}} & \makecell{54.01\\[-3pt]\ddn{1.02}\\\textcolor{gray}{58.97}\\[-3pt]\ddn{1.22}} & \makecell{\textbf{64.77}\\[-3pt]\dup{2.67}\\\textcolor{gray}{\textbf{71.00}}\\[-3pt]\dup{3.80}} & \makecell{70.64\\[-3pt]\ddn{3.47}\\\textcolor{gray}{78.80}\\[-3pt]\ddn{3.80}} & \makecell{\textbf{90.21}\\[-3pt]\dup{4.96}\\\textcolor{gray}{\textbf{99.70}}\\[-3pt]\dup{0.90}} & \makecell{\textbf{83.39}\\[-3pt]\dup{4.62}\\\textcolor{gray}{\textbf{87.00}}\\[-3pt]\dup{4.25}} & \makecell{\textbf{69.67}\\[-3pt]\dup{9.24}\\\textcolor{gray}{\textbf{79.28}}\\[-3pt]\dup{13.69}} & \makecell{\textbf{74.96}\\[-3pt]\dup{2.41}\\\textcolor{gray}{\textbf{82.37}}\\[-3pt]\dup{2.66}} \\
& \cellcolor[RGB]{246,237,233}\method{} (+GRPO) & 4B & \makecell{\textbf{83.82}\\[-3pt]\dup{0.53}\\\textcolor{gray}{\textbf{94.50}}\\[-3pt]\dup{1.30}} & \makecell{83.21\\[-3pt]\ddn{0.51}\\\textcolor{gray}{92.70}\\[-3pt]\ddn{1.40}} & \makecell{54.80\\[-3pt]\ddn{0.23}\\\textcolor{gray}{59.93}\\[-3pt]\ddn{0.26}} & \makecell{\textbf{67.90}\\[-3pt]\dup{5.80}\\\textcolor{gray}{\textbf{76.20}}\\[-3pt]\dup{9.00}} & \makecell{66.09\\[-3pt]\ddn{8.02}\\\textcolor{gray}{72.10}\\[-3pt]\ddn{10.50}} & \makecell{\textbf{89.43}\\[-3pt]\dup{4.18}\\\textcolor{gray}{\textbf{99.80}}\\[-3pt]\dup{1.00}} & \makecell{\textbf{83.59}\\[-3pt]\dup{4.82}\\\textcolor{gray}{\textbf{87.55}}\\[-3pt]\dup{4.80}} & \makecell{\textbf{70.72}\\[-3pt]\dup{10.29}\\\textcolor{gray}{\textbf{80.30}}\\[-3pt]\dup{14.71}} & \makecell{\textbf{75.32}\\[-3pt]\dup{2.77}\\\textcolor{gray}{\textbf{82.89}}\\[-3pt]\dup{3.18}} \\
& \cellcolor[RGB]{246,237,233}\method{} (dCoT-SFT) & 9B & \makecell{\textbf{84.00}\\[-3pt]\dup{0.71}\\\textcolor{gray}{93.10}\\[-3pt]\ddn{0.10}} & \makecell{83.02\\[-3pt]\ddn{0.70}\\\textcolor{gray}{91.60}\\[-3pt]\ddn{2.50}} & \makecell{\textbf{55.87}\\[-3pt]\dup{0.84}\\\textcolor{gray}{\textbf{60.98}}\\[-3pt]\dup{0.79}} & \makecell{\textbf{67.46}\\[-3pt]\dup{5.36}\\\textcolor{gray}{\textbf{75.00}}\\[-3pt]\dup{7.80}} & \makecell{\textbf{74.64}\\[-3pt]\dup{0.53}\\\textcolor{gray}{\textbf{83.20}}\\[-3pt]\dup{0.60}} & \makecell{\textbf{91.22}\\[-3pt]\dup{5.97}\\\textcolor{gray}{\textbf{99.90}}\\[-3pt]\dup{1.10}} & \makecell{\textbf{84.16}\\[-3pt]\dup{5.39}\\\textcolor{gray}{\textbf{87.40}}\\[-3pt]\dup{4.65}} & \makecell{\textbf{70.77}\\[-3pt]\dup{10.34}\\\textcolor{gray}{\textbf{78.95}}\\[-3pt]\dup{13.36}} & \makecell{\textbf{76.60}\\[-3pt]\dup{4.05}\\\textcolor{gray}{\textbf{83.55}}\\[-3pt]\dup{3.84}} \\
& \cellcolor[RGB]{246,237,233}\method{} (+GRPO) & 9B & \makecell{\textbf{84.16}\\[-3pt]\dup{0.87}\\\textcolor{gray}{93.20}\\[-3pt]{\tiny(0.00)}} & \makecell{83.12\\[-3pt]\ddn{0.60}\\\textcolor{gray}{91.40}\\[-3pt]\ddn{2.70}} & \makecell{\textbf{56.98}\\[-3pt]\dup{1.95}\\\textcolor{gray}{\textbf{62.02}}\\[-3pt]\dup{1.83}} & \makecell{\textbf{67.65}\\[-3pt]\dup{5.55}\\\textcolor{gray}{\textbf{74.90}}\\[-3pt]\dup{7.70}} & \makecell{\textbf{74.80}\\[-3pt]\dup{0.69}\\\textcolor{gray}{\textbf{82.80}}\\[-3pt]\dup{0.20}} & \makecell{\textbf{91.22}\\[-3pt]\dup{5.97}\\\textcolor{gray}{\textbf{99.80}}\\[-3pt]\dup{1.00}} & \makecell{\textbf{84.82}\\[-3pt]\dup{6.05}\\\textcolor{gray}{\textbf{88.25}}\\[-3pt]\dup{5.50}} & \makecell{\textbf{70.53}\\[-3pt]\dup{10.10}\\\textcolor{gray}{\textbf{79.22}}\\[-3pt]\dup{13.63}} & \makecell{\textbf{76.90}\\[-3pt]\dup{4.35}\\\textcolor{gray}{\textbf{83.82}}\\[-3pt]\dup{4.11}} \\
& \cellcolor[RGB]{246,237,233}\method{} (dCoT-SFT) & 27B & \makecell{81.87\\[-3pt]\ddn{1.42}\\\textcolor{gray}{93.00}\\[-3pt]\ddn{0.20}} & \makecell{80.97\\[-3pt]\ddn{2.75}\\\textcolor{gray}{91.60}\\[-3pt]\ddn{2.50}} & \makecell{\textbf{60.14}\\[-3pt]\dup{5.11}\\\textcolor{gray}{\textbf{66.72}}\\[-3pt]\dup{6.53}} & \makecell{\textbf{69.53}\\[-3pt]\dup{7.43}\\\textcolor{gray}{\textbf{78.30}}\\[-3pt]\dup{11.10}} & \makecell{\textbf{74.64}\\[-3pt]\dup{0.53}\\\textcolor{gray}{\textbf{83.30}}\\[-3pt]\dup{0.70}} & \makecell{\textbf{91.87}\\[-3pt]\dup{6.62}\\\textcolor{gray}{\textbf{99.90}}\\[-3pt]\dup{1.10}} & \makecell{\textbf{84.24}\\[-3pt]\dup{5.47}\\\textcolor{gray}{\textbf{88.15}}\\[-3pt]\dup{5.40}} & \makecell{\textbf{72.51}\\[-3pt]\dup{12.08}\\\textcolor{gray}{\textbf{81.98}}\\[-3pt]\dup{16.39}} & \makecell{\textbf{77.24}\\[-3pt]\dup{4.69}\\\textcolor{gray}{\textbf{85.20}}\\[-3pt]\dup{5.49}} \\
\bottomrule
\end{tabular}}
\end{table*}

\noindent \textbf{Stage 2: Multi-modal refinement.} The Stage~1 rationales capture visual content but read as captions: they describe each image without the cross-image comparison, self-checks or reasoning steps that the target requires. Stage~2 supplies this reasoning. A multimodal refiner, Qwen3.5-397B-A17B~\citep{qwen35}, receives the original images together with the question, task context $t$, annotated box and an optional sanitized $\tilde{c}$, and writes the final rationale
$c^*=\mathcal{Q}(\mathcal{I},q,t,b^*,\tilde{c})$.
The refiner is instructed to justify the boxed region from visible evidence, contrast it with nearby alternatives, include one or two brief self-checks, and avoid mentioning that the box was given (prompts in Appendix~\ref{app:B}). Because it inspects the images directly, $c^*$ does not depend only on Stage~1 text. Because it sees $b^*$, $c^*$ is an explanation of the annotated answer, not a faithful record of how any model found it.

\noindent \textbf{Filtering and accounting.} We discard outputs that are malformed, exceed the Stage~1 context limit, or produce a final box that does not match $b^*$. We reprocess failed cases using adjusted context splitting (Algorithm~\ref{alg:medchain}). After quality filtering and deduplication, MedChain contains 182,573 unique example IDs. We reserve 32,000 IDs for reinforcement learning, with 4,000 per task using seed 42. The remaining 150,573 IDs form the SFT split, released as 152,396 rows. We also audit train--test overlap using encoded and decoded-pixel hashes. We find no complete test image sequence and no source-series-plus-box pair in the SFT split. However, 147 test examples share at least one individual image with the training data (Appendix~\ref{app:A5}), so we also report results after excluding these examples.

\noindent \textbf{Rationale audit.} One medical expert reviews 2,000 rationales for clinical plausibility, consistency with the image sequence and task, and agreement with the target. The expert accepts 1,965 rationales (98.25\%; Wilson 95\% CI 97.58--98.74). The remaining 35 contain minor-to-moderate reasoning errors, although all have correct boxes. This audit uses a single rater and should not be treated as multi-specialist clinical validation (Table~\ref{tab:clinical_audit}).

\subsection{Training Protocol}
\label{sec:training}
\noindent \textbf{dCoT-SFT.} We fine-tune Qwen3.5 models at 4B, 9B, and 27B with deliberative chain-of-thought supervised fine-tuning (dCoT-SFT), i.e. full-parameter supervised learning on MedChain rationales. We use the format
\texttt{<think>}\,$c^*$\,\texttt{</think><answer>}\,$b^*$\,\texttt{</answer>}~\citep{deepseekr1},
so the model learns to state its cross-image reasoning before predicting the coordinates. We train with LLaMA-Factory~\citep{llamafactory} using AdamW, a cosine learning-rate schedule, a learning rate of $5\times10^{-6}$, and an effective batch size of 64. We select checkpoints using minimum validation loss (Appendix~\ref{app:A7}).

\noindent \textbf{GRPO on top of dCoT-SFT.} We then test whether reward optimization adds further gains by training the 4B and 9B dCoT-SFT models with GRPO~\citep{shao2024deepseekmath}. For each prompt, the policy samples $G=8$ completions. We assign each completion the reward
$r=r_{\text{IoU}}+r_{\text{fmt}}$,
where $r_{\text{IoU}}$ measures the IoU between the predicted box and $b^*$, and $r_{\text{fmt}}$ checks the output format. We compute each advantage relative to the other completions sampled for the same prompt. We train on the 32,000 held-out GRPO examples with KL coefficient $\beta=0.04$, clipping $\epsilon=0.2$, and learning rate $10^{-6}$ for up to 3,000 steps, selecting each checkpoint on a held-out development set (Appendix~\ref{app:C3}).

\section{Experiments}
\label{sec:experiments}
\noindent \textbf{Setup.} We evaluate on all 9,630 MedSG-Bench test questions~\citep{medsgbench} and report mIoU and Acc@0.5, where Acc@0.5 is the percentage of predictions with IoU $\geq 0.5$. We evaluate all baselines zero-shot using one unified prompt (Appendix~\ref{app:A6}). These include general-purpose and medical MLLMs from 1.5B to 78B parameters (full list in Appendix~\ref{app:C}), the RL-trained MedGRPO-7B-RL, and the supervised SOTA model, MedSeq-Grounder~\citep{medsgbench}.

\noindent \textbf{Statistical analysis.} Table~\ref{tab:main_results} reports representative results, while Appendix~\ref{app:C} contains the full comparison. For paired analyses, we match predictions by example ID and use 100,000 example-level bootstrap resamples (10,000 for the reasoning-context controls) together with sign-flip tests. The resulting intervals measure uncertainty from test-set sampling, not variation across training runs.

\section{Experimental Findings}
\label{sec:experimental-findings}
\subsection{Overall Findings}
\label{sec:main}

\noindent \textbf{Zero-shot models struggle with sequential grounding.} No zero-shot model exceeds 21 mIoU (Table~\ref{tab:main_results}). Larger general models do not perform better: InternVL3-78B scores below its 8B version, while the Qwen3.5 base models, including 27B, score below 5 mIoU under pixel-coordinate scoring, largely because they output degenerate or copied boxes (Appendix~\ref{app:qual}). The coordinate convention explains only part of this gap: in a separate thinking-off rerun, reading the Qwen3.5 boxes as 0--1000 normalized coordinates raises overall mIoU to at most 13.30 (27B), still far below fine-tuned models (Appendix~\ref{app:zs_coords}). Medical pretraining also provides little benefit. MedGRPO-7B-RL, which uses RL on adjacent surgical video frames, transfers poorly and reaches only 6.27 mIoU. Scale, medical knowledge and nearby video supervision alone therefore do not teach the cross-image correspondences that sequential grounding requires.

\noindent \textbf{Reasoning supervision surpasses the answer-only SOTA.} Our dCoT-SFT models outperform MedSeq-Grounder at every scale. They improve mIoU by 2.41 points at 4B, 4.05 at 9B, and 4.69 at 27B, with corresponding Acc@0.5 gains of 2.66, 3.84, and 5.49 points. At 27B, we see the largest gains on referring (+12.1 mIoU), object tracking (+7.4), patch (+6.6), and cross-modal grounding (+5.5). Part of this margin reflects the backbone: our answer-only 4B model already exceeds MedSeq-Grounder on RG, VPG and CMG (Table~\ref{tab:fixed_4b}), whereas reasoning supervision is what lifts the average above it (71.39 to 74.96 mIoU, versus 72.55). On difference grounding, our models perform similarly to MedSeq-Grounder, which remains strongest on NRDG. Since the two model families use different backbones and training data, we treat this comparison as descriptive. As a partial backbone control, we continue dCoT-SFT from the released MedSeq-Grounder checkpoint (Qwen2.5-VL-7B) and improve its mIoU/Acc@0.5 from 72.55/79.71 to 75.71/82.34 (Appendix~\ref{app:C2}). A matched answer-only continuation is not available.

\noindent \textbf{Scaling.} Average performance rises monotonically with model size (74.96, 76.60 and 77.24 mIoU), but the effect differs across tasks. From 9B to 27B, ICG performance improves or remains stable, with gains of +4.3 mIoU on MV, +2.1 on OT, and +1.7 on RG. In contrast, both IDG tasks drop by about 2 points. However, the 27B model uses a different training schedule: six epochs, a smaller warmup, and no weight decay, and it reaches a higher validation loss (Appendix~\ref{app:A7}). We therefore do not attribute this difference to model size alone.

\begin{table*}[t]
\centering
\caption{\textbf{Controlled 4B analysis.} All rows use the same 9,630 test questions. Top: answer-only SFT versus dCoT-SFT checkpoints of Qwen3.5-4B, and their difference ($\Delta$, computed from the rounded entries). Bottom: inference-time interventions on the same dCoT-SFT checkpoint. Cells show mIoU (top) and \textcolor{gray}{Acc@0.5} (bottom). Gains are shaded green and losses red.}
\label{tab:fixed_4b}
\scriptsize
\setlength{\tabcolsep}{3.5pt}
\renewcommand{\arraystretch}{0.95}
\resizebox{\textwidth}{!}{%
\begin{tabular}{l|cc|cccccc|c}
\toprule
\textbf{Qwen3.5-4B} & \textbf{RDG} & \textbf{NRDG} & \textbf{MV} & \textbf{OT} & \textbf{VCG} & \textbf{VPG} & \textbf{CMG} & \textbf{RG} & \textbf{Overall} \\
\midrule
Answer-only SFT & \mc{66.76}{81.00} & \mc{66.64}{81.40} & \mc{56.21}{63.24} & \mc{58.27}{66.40} & \mc{69.03}{78.40} & \mc{92.66}{99.70} & \mc{80.83}{85.10} & \mc{72.85}{83.47} & \mc{71.39}{80.31} \\
dCoT-SFT & \mc{82.86}{93.30} & \mc{81.30}{91.20} & \mc{54.01}{58.97} & \mc{64.77}{71.00} & \mc{70.64}{78.80} & \mc{90.21}{99.70} & \mc{83.39}{87.00} & \mc{69.67}{79.28} & \mc{74.96}{82.37} \\
$\Delta$ mIoU & \cellcolor{green!12}+16.10 & \cellcolor{green!12}+14.66 & \cellcolor{red!10}$-$2.20 & \cellcolor{green!12}+6.50 & \cellcolor{green!12}+1.61 & \cellcolor{red!10}$-$2.45 & \cellcolor{green!12}+2.56 & \cellcolor{red!10}$-$3.18 & \cellcolor{green!12}+3.56 \\
\midrule
dCoT-SFT, forced no-think & \mc{82.15}{92.20} & \mc{80.94}{90.80} & \mc{49.72}{53.57} & \mc{55.26}{58.90} & \mc{69.40}{76.40} & \mc{87.55}{99.00} & \mc{81.44}{84.85} & \mc{66.36}{73.55} & \mc{72.02}{78.66} \\
dCoT-SFT, cross-task rationale & \mc{65.19}{73.30} & \mc{80.16}{89.60} & \mc{39.55}{41.55} & \mc{46.74}{50.60} & \mc{65.12}{72.40} & \mc{86.03}{96.80} & \mc{65.78}{70.00} & \mc{56.88}{63.02} & \mc{62.77}{68.93} \\
\bottomrule
\end{tabular}}
\end{table*}

\subsection{When and How Does Reasoning Help?}
\label{sec:controlled}
\label{sec:reliance}

\noindent \textbf{Controlled comparison.} The comparison with MedSeq-Grounder mixes differences in supervision, backbone, and training data. To isolate supervision as far as our runs allow, we compare fixed 4B checkpoints trained with answer-only SFT and dCoT-SFT on the same backbone for the same number of optimization steps (Table~\ref{tab:fixed_4b}, top). All 9,630 predictions match exactly by example ID, task, and ground-truth box, with no parser failures.

\noindent \textbf{Overall effect.} dCoT-SFT improves mIoU by 3.56 points (95\% CI 3.13--4.00) and Acc@0.5 by 2.06 points (1.36--2.75), with $p<10^{-5}$ under both bootstrap and sign-flip tests. The result remains almost unchanged after excluding the 164 test examples that share an image or source series with the SFT data: $+3.61$ mIoU (3.17--4.04) and $+2.01$ Acc@0.5 (1.31--2.70) on the remaining 9,466 examples. The stable gain after this exclusion suggests that the improvement is not driven by the identified train--test overlap. The answer-only run was trained on fewer nodes, so its effective batch size, optimizer implementation and checkpoint-selection rule still differ, and neither run was repeated over seeds; we therefore treat this as a fixed-model comparison rather than a causal or multi-seed estimate. The answer-only model also scores below MedSeq-Grounder on difference grounding, which uses a different backbone, 188K training examples and its own schedule, so the absolute size of the IDG gain should be read with this in mind.

\noindent \textbf{The effect depends strongly on the task.} dCoT-SFT gains most when localization is defined by comparison: registered and non-registered difference grounding improve by 16.1 and 14.7 mIoU, while object tracking improves by 6.5. In contrast, performance drops on multi-view ($-2.2$), patch ($-2.5$), and referring grounding ($-3.2$). This split suggests that explicit reasoning is most useful when the model must identify a relation across images before it can localize the target. Patch and referring grounding may rely more on local appearance or text-to-region matching, where verbal reasoning adds less. Multi-view grounding also requires comparison, however, so its decline does not fit this explanation and remains unresolved. We therefore report the pattern as empirical rather than as a validated mechanism.

\noindent \textbf{Predictions depend on the reasoning context.} Because MedChain rationales are written with the answer in view, the model could learn to produce plausible reasoning while localizing independently of it. We test this with two inference-time interventions on the fixed 4B dCoT-SFT checkpoint (Table~\ref{tab:fixed_4b}, bottom; Appendix~\ref{app:C4}). Forcing an immediate answer (\emph{no-think}) lowers mIoU from 74.96 to 72.02 ($-2.93$, 95\% CI $-3.29$ to $-2.58$). Prefilling a deterministic rationale from another task lowers it further to 62.77 ($-12.18$, $-12.75$ to $-11.60$); all tests give $p<10^{-4}$ with 10,000 paired resamples. The much larger drop under mismatched reasoning suggests that localization is sensitive not only to the presence of reasoning, but also to its task-specific content. The effect is largest on object tracking ($-18.0$), registered difference ($-17.7$), cross-modal ($-17.6$), and multi-view grounding ($-14.5$), and smallest on non-registered difference grounding ($-1.1$). Removing reasoning hurts most on object tracking ($-9.5$) and multi-view grounding ($-4.3$). Together, these results suggest that reasoning context matters most for tasks that must track or match structures across images. These interventions do not prove that the rationale faithfully records the model's internal computation, and the cross-task rationale itself introduces a distribution shift.

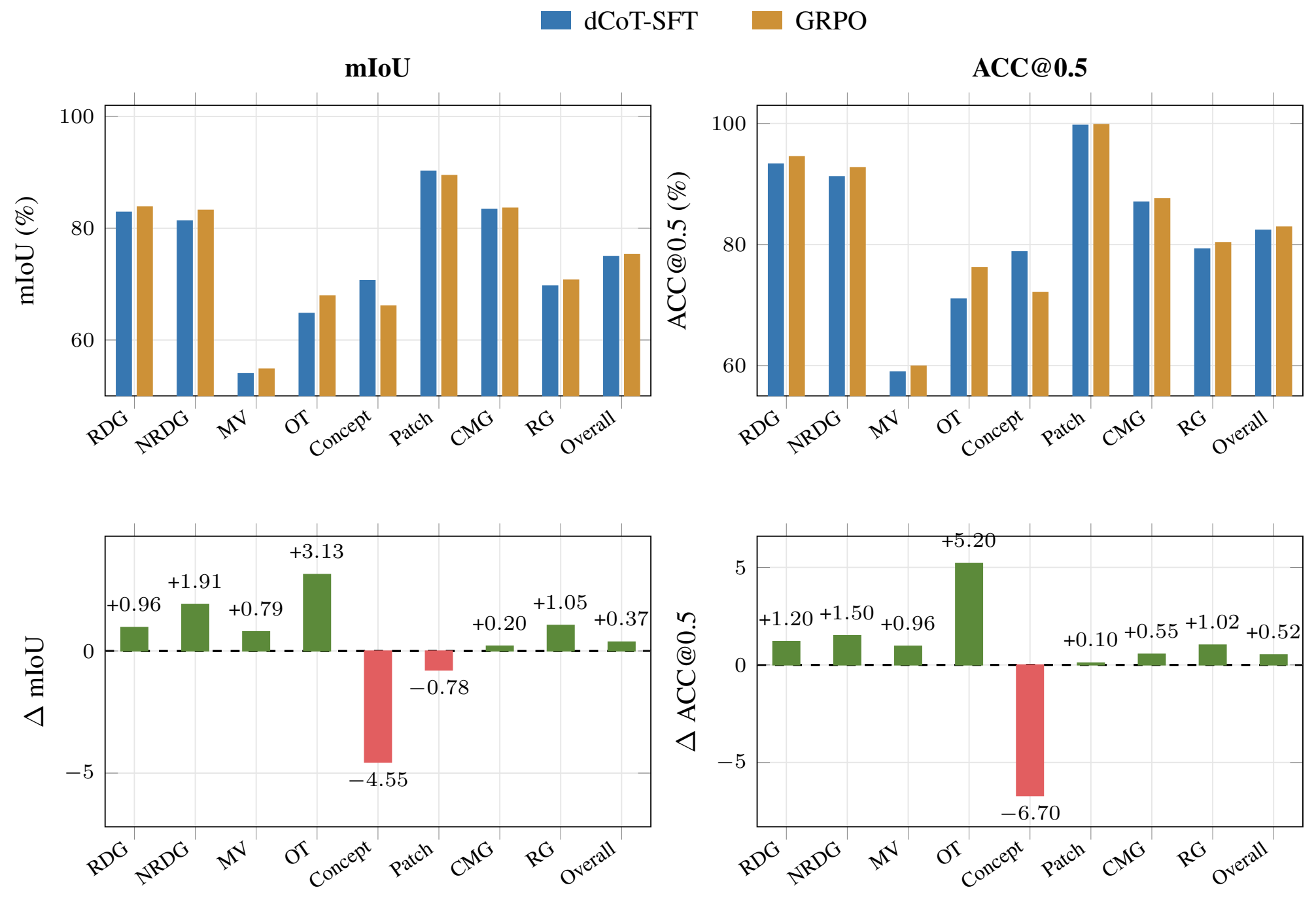
\begin{figure*}[t]
\centering
\begingroup
\definecolor{cotblue}{RGB}{55,119,177}
\definecolor{grpogold}{RGB}{205,145,55}
\definecolor{posgreen}{RGB}{92,138,58}
\definecolor{negred}{RGB}{226,94,96}
\resizebox{0.98\textwidth}{!}{%
\begin{tikzpicture}
\begin{groupplot}[
  group style={group size=2 by 2, horizontal sep=1.25cm, vertical sep=1.65cm},
  width=8.0cm,
  height=5.0cm,
  symbolic x coords={RDG,NRDG,MV,OT,Concept,Patch,CMG,RG,Overall},
  xtick={RDG,NRDG,MV,OT,Concept,Patch,CMG,RG,Overall},
  x tick label style={rotate=35,anchor=east,font=\scriptsize},
  tick label style={font=\scriptsize},
  label style={font=\small},
  title style={font=\small\bfseries},
  grid=major,
  grid style={gray!20},
  axis line style={black},
  enlarge x limits=0.06,
]

\nextgroupplot[
  title={mIoU}, ylabel={mIoU (\%)}, ymin=50, ymax=102,
  ybar, bar width=5pt,
]
\addplot[fill=cotblue,draw=cotblue] coordinates {
 (RDG,82.86) (NRDG,81.30) (MV,54.01) (OT,64.77) (Concept,70.64)
 (Patch,90.21) (CMG,83.39) (RG,69.67) (Overall,74.955)};
\addplot[fill=grpogold,draw=grpogold] coordinates {
 (RDG,83.82) (NRDG,83.21) (MV,54.80) (OT,67.90) (Concept,66.09)
 (Patch,89.43) (CMG,83.59) (RG,70.72) (Overall,75.321)};

\nextgroupplot[
  title={ACC@0.5}, ylabel={ACC@0.5 (\%)}, ymin=55, ymax=103,
  ybar, bar width=5pt,
]
\addplot[fill=cotblue,draw=cotblue] coordinates {
 (RDG,93.30) (NRDG,91.20) (MV,58.97) (OT,71.00) (Concept,78.80)
 (Patch,99.70) (CMG,87.00) (RG,79.28) (Overall,82.368)};
\addplot[fill=grpogold,draw=grpogold] coordinates {
 (RDG,94.50) (NRDG,92.70) (MV,59.93) (OT,76.20) (Concept,72.10)
 (Patch,99.80) (CMG,87.55) (RG,80.30) (Overall,82.887)};

\nextgroupplot[
  ylabel={$\Delta$ mIoU}, ymin=-7.2, ymax=4.7,
  ybar, bar width=9pt, bar shift=0pt,
  extra y ticks={0}, extra y tick labels={},
  extra y tick style={grid=major,grid style={black,dashed,line width=0.7pt}},
  nodes near coords,
  point meta=explicit symbolic,
  every node near coord/.append style={font=\scriptsize,anchor=south,yshift=2pt},
]
\addplot[fill=posgreen,draw=posgreen] coordinates {
 (RDG,0.96) [+$0.96$] (NRDG,1.91) [+$1.91$] (MV,0.79) [+$0.79$]
 (OT,3.13) [+$3.13$] (CMG,0.20) [+$0.20$] (RG,1.05) [+$1.05$]
 (Overall,0.367) [+$0.37$]};
\addplot[fill=negred,draw=negred,
  every node near coord/.append style={font=\scriptsize,anchor=north,yshift=-2pt}] coordinates {
 (Concept,-4.55) [$-4.55$] (Patch,-0.78) [$-0.78$]};

\nextgroupplot[
  ylabel={$\Delta$ ACC@0.5}, ymin=-8.3, ymax=6.6,
  ybar, bar width=9pt, bar shift=0pt,
  extra y ticks={0}, extra y tick labels={},
  extra y tick style={grid=major,grid style={black,dashed,line width=0.7pt}},
  nodes near coords,
  point meta=explicit symbolic,
  every node near coord/.append style={font=\scriptsize,anchor=south,yshift=2pt},
]
\addplot[fill=posgreen,draw=posgreen] coordinates {
 (RDG,1.20) [+$1.20$] (NRDG,1.50) [+$1.50$] (MV,0.96) [+$0.96$]
 (OT,5.20) [+$5.20$] (Patch,0.10) [+$0.10$] (CMG,0.55) [+$0.55$]
 (RG,1.02) [+$1.02$] (Overall,0.519) [+$0.52$]};
\addplot[fill=negred,draw=negred,
  every node near coord/.append style={font=\scriptsize,anchor=north,yshift=-2pt}] coordinates {
 (Concept,-6.70) [$-6.70$]};

\end{groupplot}
\node[font=\small,align=center] at ($(group c1r1.north west)!0.5!(group c2r1.north east)+(0,1.0)$)
 {\textcolor{cotblue}{\rule{10pt}{6pt}}\enspace dCoT-SFT\qquad\textcolor{grpogold}{\rule{10pt}{6pt}}\enspace GRPO};
\node[font=\scriptsize,align=center,text=negred,text width=15cm,anchor=north] at ($(group c1r2.south west)!0.5!(group c2r2.south east)+(0,-1.25)$)
  {};
\end{tikzpicture}
}
\endgroup
\caption{\textbf{Task-wise effect of GRPO after dCoT-SFT (4B, full test set).} Top: mIoU and Acc@0.5 of the fixed dCoT-SFT checkpoint and the GRPO checkpoint. Bottom: per-task differences. GRPO improves most localization-style tasks and object tracking in particular, but lowers concept grounding by 4.55 mIoU and 6.70 Acc@0.5. Only concept grounding declines substantially under GRPO.}
\label{fig:semantic_drift}
\end{figure*}

\begin{table}[t]
\centering
\caption{\textbf{GRPO after dCoT-SFT (4B; rows marked 9B use the 9B model).} Paired differences from the dCoT-SFT starting point in percentage points, with 95\% bootstrap CIs over examples or the standard deviation over training seeds. Development results use a separate 1,385-example set and are exploratory. The 9B all-task row is a point estimate from example-weighted means, without a paired interval.}
\label{tab:grpo}
\small
\setlength{\tabcolsep}{4pt}
\begin{tabular}{@{}llrr@{}}
\toprule
\textbf{Evaluation} & \textbf{GRPO variant} & \textbf{$\Delta$ mIoU} & \textbf{$\Delta$ Acc@0.5} \\
\midrule
Test, all tasks ($n$=9,630) & IoU+format & $+0.37$ [0.05, 0.69] & $+0.52$ [$-0.04$, 1.08] \\
Test, concept task & IoU+format & $-4.55$ & $-6.70$ \\
\midrule
Test 9B, all tasks & IoU+format & $+0.30$ & $+0.27$ \\
Test 9B, multi-view & IoU+format & $+1.11$ [0.39, 1.83] & $+1.05$ [$-0.44$, 2.53] \\
Test 9B, cross-modal & IoU+format & $+0.67$ [0.27, 1.09] & $+0.85$ [0.15, 1.55] \\
Test 9B, concept task & IoU+format & $+0.17$ [$-0.84$, 1.17] & $-0.40$ [$-1.70$, 0.90] \\
\midrule
Dev, all tasks & direct, step 50, 3 seeds & $+0.06$ (SD 0.08) & $+0.24$ (SD 0.15) \\
Dev, multi-view ($n$=59) & direct, step 50, 3 seeds & $+4.45$ (SD 0.63) & $+7.35$ (SD 1.96) \\
Dev, all tasks & task-balanced, threshold reward & $+0.36$ [$-0.16$, 0.89] & $+0.94$ [$-0.07$, 1.95] \\
\bottomrule
\end{tabular}
\end{table}
\subsection{Does Reward Optimization Help Beyond dCoT-SFT?}
\label{sec:grpo}

We add GRPO as a second training stage and test whether reward optimization improves on reasoning supervision.

\noindent \textbf{Benchmark effect.} GRPO improves mIoU by 0.37 points (95\% CI 0.05--0.69, $p=0.025$), but the Acc@0.5 gain is not significant ($+0.52$, $-0.04$ to 1.08, $p=0.075$) (Table~\ref{tab:grpo}). The average also hides opposite task effects: object tracking gains 3.1 mIoU and 5.2 Acc@0.5, while concept grounding loses 4.6 and 6.7, respectively (Figure~\ref{fig:semantic_drift}). This pattern suggests that an overlap-based reward can improve localization on some tasks while weakening semantic identification on others. IoU rewards overlap regardless of whether the model identifies the correct finding, which may matter most for concept grounding, where the model must recognize a lesion or structure across very different images. We do not isolate this mechanism here.

\noindent \textbf{Controlled development studies.} We next run controlled GRPO experiments on a separate 1,385-example development set (Appendix~\ref{app:grpo_followup}). An initial 3--5 point drop was an evaluation artifact: the GRPO export saved an image processor with a lower resolution limit. After restoring the original processor, a zero-learning-rate save-and-reload control exactly reproduces all SFT predictions. With matched preprocessing, direct GRPO at a fixed step changes aggregate mIoU by only $+0.06$ (SD 0.08) and Acc@0.5 by $+0.24$ (SD 0.15) across three seeds. The only consistent task-level gain is on multi-view grounding, with $+4.4$ mIoU and $+7.3$ Acc@0.5, although this task has only 59 examples. A task-balanced variant with a threshold-aware reward gives $+0.36$ mIoU and $+0.94$ Acc@0.5, but neither interval excludes zero, and the gain disappears after a prompt-formatting change.

\noindent \textbf{Headroom without gains.} We finally test whether the policy can already sample better answers than it normally returns. On 128 training prompts from the four hardest tasks, individual samples reach IoU $\geq 0.5$ in 67.7\% of cases on average, while at least one of eight samples succeeds for 84.4\% of prompts. This 16.7-point gap shows that rollout groups often contain a better localization than the policy's typical output. However, all eight samples fail on 20 of the 128 prompts, and GRPO does not convert the available headroom into benchmark gains in our configurations. Possible reasons include weak within-group contrast on the hardest examples, large differences in starting difficulty across tasks (54--90 mIoU), and a reward that measures spatial overlap without semantic correctness. These remain hypotheses to test with alternative rewards.

\section{Discussion and Conclusion}
\label{sec:discussion}
\label{sec:conclusion}
Sequential medical grounding is not simply single-image localization at scale: the main difficulty is deciding which cross-image relation matters before placing a box. Supervising that comparison with image-grounded rationales yields the best reported MedSG-Bench results at 4B, 9B and 27B, with the largest gains where the answer is defined by a cross-image comparison, while losses on appearance-matching tasks show that more reasoning is not always better. The intervention results indicate that the reasoning is not decorative, and GRPO with an overlap reward adds little once strong reasoning supervision is in place. Appendix~\ref{app:D} discusses limitations.

\subsection*{AI Use Statement}
In this work, we used generative AI tools to generate a synthetic dataset: the MedChain rationales were produced by FlemingVL-38B and Qwen3.5-397B-A17B as part of the research method (Section~\ref{sec:medchain} and Appendix~\ref{app:A}), and a sample of 2,000 rationales was audited by a medical expert. We also used generative AI tools to assist with interpreting results, in drafting parts of the discussion of per-task findings (Appendix~\ref{app:discussion}); the authors checked every interpretation against the underlying experiment outputs. Additionally, we used generative AI tools for drafting and restructuring paper sections, editing for readability, suggesting paper structure, and formatting tables and LaTeX. We have reviewed all AI-assisted work and verified all numerical statements against the experiment artifacts. We take responsibility for the final content of this work, including text, claims or artifacts produced with the aid of generative AI.

\subsection*{Ethics Statement}
This study uses the publicly distributed MedSG-Bench research benchmark and does not introduce new patient recruitment or collect additional identifiers. The generated rationales can inherit clinical errors and biases from their teacher models, are conditioned on annotated boxes, and were reviewed only in a limited single-expert audit. Neither the rationales nor the grounding models are validated for autonomous diagnosis or clinical deployment. Users must respect the privacy, license and use restrictions of each underlying data source and obtain appropriate institutional approval for any prospective clinical study. Broader impacts are discussed in Appendix~\ref{app:E}.

\subsection*{Reproducibility Statement}
Section~\ref{sec:experiments}, Section~\ref{sec:experimental-findings} and Appendix~\ref{app:A} report the data splits, evaluation protocol, optimizer settings, seeds, exact Qwen3.5 model identifiers, validation-selected checkpoint steps and the train--test overlap audit. Appendix~\ref{app:B} provides the rationale-generation prompts, and Appendix~\ref{app:C} records the additional baselines, GRPO configuration, reasoning-reliance controls and development-set studies. The supplementary material contains the training and evaluation scripts, the per-example predictions used for the paired bootstrap analyses, machine-readable overlap results and the checkpoint metadata needed to reproduce the reported analyses.

\bibliographystyle{iclr2027_conference}
\bibliography{ref}

\clearpage
\appendix

\clearpage
\begin{center}
{\Large\bf Appendix}\\[3mm]
{\large\bf Table of Content}
\end{center}
\vspace{1em}
\startcontents[sections]
{\small\printcontents[sections]{l}{1}{\setcounter{tocdepth}{3}}}

\clearpage
\newpage

\FloatBarrier
\section{MedChain-182K Dataset Construction Details}\label{app:A}


\subsection{Model Selection for Pseudo-CoT Generation}\label{app:A1}

To choose the Stage~1 teacher, we compared six candidate multimodal models (Table~\ref{tab:teacher_selection}). Each candidate generated rationales for the same 100 sampled instances under six prompt variants, and we assessed the outputs for reasoning quality, consistency with the task and image sequence, and clinical plausibility. The candidates cover three groups: general-domain MLLMs, medical MLLMs, and the task-specific sequential grounding model.

\begin{table}[H]
\centering
\caption{Candidate teachers for Stage~1 pseudo-CoT generation. All candidates were evaluated on the same 100 sampled instances with six prompt variants each. The selected teacher is highlighted.}
\label{tab:teacher_selection}
\small
\setlength{\tabcolsep}{6pt}
\begin{tabular}{@{}llc@{}}
\toprule
\textbf{Candidate model} & \textbf{Category} & \textbf{Parameters} \\
\midrule
\multicolumn{3}{@{}l}{\textit{General-domain MLLMs}} \\
\quad Qwen3-VL~\citep{qwen3vl} & General multimodal & 30B \\
\quad GPT-4o (ChatGPT API)~\citep{chatgpto3} & General multimodal & Not disclosed \\
\midrule
\multicolumn{3}{@{}l}{\textit{Medical MLLMs}} \\
\quad Lingshu~\citep{xu2025lingshu} & Medical multimodal & 32B \\
\quad QoQ-Med-VL~\citep{dai2025qoq} & Medical multimodal & 32B \\
\rowcolor[RGB]{246,237,233}\quad \textbf{FlemingVL (selected)}~\citep{flemingvl} & Medical multimodal & 38B \\
\midrule
\multicolumn{3}{@{}l}{\textit{Task-specific model}} \\
\quad MedSeq-Grounder~\citep{medsgbench} & Sequential medical grounding & 7B \\
\bottomrule
\end{tabular}
\end{table}

\noindent\textbf{Selection.} FlemingVL-38B~\citep{flemingvl} produced the most reliable, task-consistent and clinically faithful rationales across the ten imaging modalities of MedSG-Bench~\citep{medsgbench}, and we therefore use it as the Stage~1 teacher. Stage~2 refinement uses a separate multimodal model, Qwen3.5-397B-A17B (Appendix~\ref{app:A3}).

\subsection{Stage 1: Pseudo-CoT Generation (FlemingVL-38B)}\label{app:A2}

Following model selection, we developed the final
system prompt (see Appendix~\ref{app:B1}) and deployed
FlemingVL-38B~\cite{flemingvl} via
\texttt{vLLM}-based inference on NVIDIA A100
40\,GB GPUs, consuming approximately
\textbf{1,200 A100 GPU hours}. Due to the model's
8K context window limitation, samples with longer
input sequences were isolated into error files and
reprocessed in the subsequent stage.

\subsection{Stage 2: Multimodal CoT Refinement}\label{app:A3}

Qwen3.5-397B-A17B~\cite{qwen35} directly processed
the original image sequence, question, task context,
annotated box, and optional sanitized preliminary
rationale to produce the final explanation
(see Appendix~\ref{app:B2}),
executed on \textbf{NVIDIA H200 140\,GB GPUs}
and requiring approximately \textbf{1,280 H200
GPU hours}. Samples with missing pseudo-CoTs or
incorrect coordinate associations were isolated
and recursively reprocessed.

\subsection{Quality Verification}\label{app:A4}

One medical expert reviewed \textbf{2,000 rationales}
for clinical plausibility, consistency with the task
and image sequence, and correspondence to the intended
target. Of these, 1,965 were accepted (98.25\%; Wilson
95\% CI 97.58--98.74\%) and 35 contained
minor-to-moderate CoT errors; the boxes were correct in
all 35 flagged cases. Because only aggregate counts are
available, error-category counts, item-level sampling
verification, and inter-rater agreement cannot be
reported. This is a single-rater audit, not multi-expert
clinical validation.

The available audit record supports the following
breakdown and no finer subdivision. We report missing
metadata explicitly rather than reconstructing it after
the review.

\begin{table}[!ht]
\centering\small
\begin{minipage}[t]{0.47\linewidth}\centering
\caption{Clinical rationale audit record. ``Not
recorded'' denotes information absent from the retained
audit summary.}
\label{tab:clinical_audit}
\small
\begin{adjustbox}{max width=\linewidth}\begin{tabular}{p{0.56\linewidth}r}
\toprule
Audit field & Retained result \\
\midrule
Reviewer & One medical expert \\
Specialty/qualification & Not recorded \\
Clinical plausibility criterion & Assessed \\
Task/sequence consistency criterion & Assessed \\
Rationale--target correspondence & Assessed \\
Accepted & 1,965 (98.25\%) \\
Minor-to-moderate CoT error & 35 (1.75\%) \\
Incorrect box among flagged cases & 0/35 \\
Finer error subclasses/counts & Not recorded \\
Second rater / Cohen's $\kappa$ & Not available \\
\bottomrule
\end{tabular}\end{adjustbox}
\end{minipage}\hfill
\begin{minipage}[t]{0.49\linewidth}\centering
\caption{MedChain-182K construction pipeline
statistics.}
\label{tab:construction_stats}
\begin{adjustbox}{max width=\linewidth}\begin{tabular}{lc}
\toprule
\rowcolor[RGB]{242,242,242}\textbf{Component} & \textbf{Details} \\
\midrule
\multicolumn{2}{l}{\textit{Stage 1: Pseudo-CoT
Generation}} \\
\midrule
~~Model         & FlemingVL-38B \\
~~Hardware      & NVIDIA A100 40\,GB \\
~~Compute       & $\sim$1,200 A100 GPU hours \\
~~Context limit & 8K tokens \\
\midrule
\multicolumn{2}{l}{\textit{Stage 2: CoT
Refinement}} \\
\midrule
~~Model    & Qwen3.5-397B-A17B (multimodal) \\
~~Hardware & NVIDIA H200 140\,GB \\
~~Compute  & $\sim$1,280 H200 GPU hours \\
\midrule
\multicolumn{2}{l}{\textit{Dataset Statistics}} \\
\midrule
~~Filtered CoT pool & 182,573 unique IDs \\
~~GRPO allocation & 32,000 IDs (4,000/task) \\
~~Remaining SFT IDs & 150,573 \\
~~Released SFT artifact & 152,396 rows \\
~~Tasks covered & 8 clinical tasks \\
~~Modalities    & 10 modalities \\
\midrule
\multicolumn{2}{l}{\textit{Quality Verification}} \\
\midrule
~~Reviewed rationales & 2,000 \\
~~Accepted / flagged & 1,965 / 35 \\
\bottomrule
\end{tabular}\end{adjustbox}
\end{minipage}
\end{table}

Table~\ref{tab:construction_stats} summarizes the
full pipeline statistics.

\subsection{Train--Test Overlap Audit}\label{app:A5}

We hashed all referenced images in the 152,396-row
SFT artifact and the 9,630-example test set using both
encoded-file SHA-256 and decoded RGBA-pixel SHA-256.
We additionally compared complete ordered image
sequences and task-aware source series/cases. The
task-aware normalization supersedes an earlier
filename-only normalization that removed too much case
information. Table~\ref{tab:overlap_audit} reports the
verified counts.

\begin{table}[H]
\centering\small
\begin{minipage}[t]{0.52\linewidth}\centering
\caption{Detailed SFT/test overlap audit. Affected
counts refer to examples, whereas shared counts refer
to unique hashes, sequences, or source-series keys.}
\label{tab:overlap_audit}
\small
\begin{adjustbox}{max width=\linewidth}\begin{tabular}{p{0.55\linewidth}rr}
\toprule
Overlap definition & Shared & Test \\
\midrule
Encoded-file SHA-256 & 55 & 144 \\
Decoded-pixel SHA-256 & 56 & 147 \\
Complete encoded-hash sequence & 0 & 0 \\
Complete pixel-hash sequence & 0 & 0 \\
Task-aware source series/case & 12 & 17 \\
Series + normalized question & 2 & 2 \\
Series + concept & 12 & 17 \\
Series + target box & 0 & 0 \\
Series + concept + target box & 0 & 0 \\
Global target-box coordinates & 100 & 106 \\
\midrule
Image/series union excluded & --- & 164 \\
Leakage-clean paired subset & --- & 9,466 \\
\bottomrule
\end{tabular}\end{adjustbox}
\end{minipage}\hfill
\begin{minipage}[t]{0.44\linewidth}\centering
\caption{dCoT-SFT training hyperparameters
per model scale.}
\label{tab:hyperparams}
\begin{adjustbox}{max width=\linewidth}\begin{tabular}{lccc}
\toprule
\rowcolor[RGB]{242,242,242}
\textbf{Hyperparameter} & \textbf{4B}
& \textbf{9B} & \textbf{27B} \\
\midrule
Learning rate
& $5{\times}10^{-6}$
& $5{\times}10^{-6}$
& $5{\times}10^{-6}$ \\
Per-device batch size & 4 & 4 & 2 \\
Gradient accumulation & 2 & 2 & 4 \\
Effective batch size  & 64 & 64 & 64 \\
Epochs                & 2 & 2 & 6 \\
Warmup ratio          & 3\% & 3\% & 1.5\% \\
Weight decay          & 0.05 & 0.05 & 0.00 \\
Best checkpoint step  & 2000 & 2500 & 4500 \\
Best val.\ loss       & 0.2787 & 0.2633 & 0.3644 \\
\bottomrule
\end{tabular}\end{adjustbox}
\end{minipage}
\end{table}

The 147 image-content matches comprise 135 referring,
10 multi-view, one object-tracking, and one cross-modal
test example. No complete multi-image test input and no
source-series-plus-target-box pair occurs in SFT. A
global coordinate collision alone is not evidence of
leakage. Conservatively excluding the union of all
image-content and task-aware source-series matches leaves
9,466 paired examples; the dCoT-SFT minus answer-only
differences remain $+3.605$ mIoU points (95\% CI
$3.167$--$4.043$) and $+2.007$ Acc@0.5 points
($1.310$--$2.704$), with paired-bootstrap
$p<10^{-5}$ for both metrics.

\subsection{Zero-Shot Evaluation Protocol}\label{app:A6}

To ensure fair comparison across all baseline
models~\cite{bai2025qwen2, zhu2025internvl3,
li2024llava, sellergren2025medgemma}, we adopt a
unified zero-shot evaluation protocol on
MedSG-Bench~\cite{medsgbench}. Each model receives
the clinical grounding question with a
model-specific output format instruction appended,
requiring exactly one bounding box prediction with
no chain-of-thought supervision.

\vspace{6pt}

\begin{promptorange}{Zero-Shot Evaluation Prompt
(All Models)}
\ttfamily\small

\{Clinical grounding question\}\\[6pt]

\{Model-specific format suffix\}\\[6pt]

\textit{Example for Qwen3-VL:}\\
Identify the bounding box of the region
described by the following expression:
\texttt{<|object\_ref\_start|>}
\{target\}
\texttt{<|object\_ref\_end|>}\\
Output only the coordinates in integer pixel
space. Format: (x1,y1),(x2,y2). Strictly
follow this format. No additional text or
explanation.\\[6pt]

\textit{Example for generic models:}\\
Identify the bounding box of the region
described by the following expression:
\{target\}\\
Format: (x1,y1),(x2,y2). Output exactly one
bounding box and nothing else.

\end{promptorange}
\subsection{Coordinate-Convention Sensitivity of Qwen3.5 Zero-Shot Baselines}\label{app:zs_coords}

None of the 9,630 MedSG-Bench test questions specifies a coordinate space, and some Qwen vision-language models express boxes in a 0--1000 normalized frame, so near-zero Qwen3.5 scores could partly reflect a coordinate mismatch rather than a grounding failure. To test this, we score the answers of each Qwen3.5 zero-shot run under two conventions: reading the predicted box as pixel coordinates, and reading it as 0--1000 normalized coordinates rescaled by the target image width and height. The model is never asked for normalized coordinates, and the two rows of each run score an identical set of answers. Boxes are extracted with the benchmark evaluator's parser, which takes the first box in the output; answers without a parseable box score zero under both conventions. Table~\ref{tab:zs_coords} covers two sources. The first is the Qwen3.5-4B thinking-mode run reported in Table~\ref{tab:full_results}, rescored. The second is a separate thinking-off rerun of Qwen3.5-4B, 9B and 27B on all 9,630 questions, in which the model is instructed only to output the box coordinates inside answer tags. The thinking-mode outputs behind the 9B and 27B rows of Table~\ref{tab:full_results} were not retained, so they could not be rescored.

Table~\ref{tab:zs_coords} reports overall scores. In the thinking-off rerun, normalized scoring raises overall mIoU from 1.55 to 8.21 at 4B, from 1.04 to 13.05 at 9B and from 2.64 to 13.30 at 27B, so the coordinate convention explains part of the zero-shot gap for these runs. It explains almost none of the gap for the thinking-mode 4B run (4.52 pixel versus 4.55 normalized mIoU). That run returns a parseable box for only 6,368 of the 9,630 questions, so its scores mainly reflect format failures, and its object-tracking score comes largely from restating the reference box given in the question. Scaling from 9B to 27B changes normalized-coordinate mIoU only slightly (13.05 to 13.30). Under either convention, the best zero-shot result (13.30 mIoU, Qwen3.5-27B) remains far below MedSeq-Grounder (72.55) and our dCoT-SFT models (74.96 to 77.24 mIoU), so the conclusion that sequential grounding requires task-specific training is unchanged.

\begin{table}[H]
\centering
\caption{\textbf{Qwen3.5 zero-shot results under pixel and 0--1000 normalized coordinate scoring.} MedSG-Bench test set (9,630 questions), overall example-weighted scores in percent. Both columns of a row score the same set of answers; unparseable answers score zero. \emph{Parsed}: questions with a parseable box.}
\label{tab:zs_coords}
\small
\setlength{\tabcolsep}{5pt}
\begin{tabular}{ll|cc|cc|r}
\toprule
 & & \multicolumn{2}{c|}{\textbf{Pixel}} & \multicolumn{2}{c|}{\textbf{Normalized}} & \\
\textbf{Model} & \textbf{Run} & \textbf{mIoU} & \textbf{Acc@0.5} & \textbf{mIoU} & \textbf{Acc@0.5} & \textbf{Parsed} \\
\midrule
Qwen3.5-4B & thinking (Table~\ref{tab:full_results}) & 4.52 & 2.96 & 4.55 & 2.14 & 6,368 \\
\midrule
Qwen3.5-4B & thinking off & 1.55 & 0.86 & 8.21 & 2.83 & 8,139 \\
Qwen3.5-9B & thinking off & 1.04 & 0.61 & 13.05 & 6.23 & 9,630 \\
Qwen3.5-27B & thinking off & 2.64 & 1.64 & 13.30 & 7.64 & 9,466 \\
\midrule
\multicolumn{2}{l|}{\textit{Reference:} MedSeq-Grounder (7B)} & 72.55 & 79.71 & --- & --- & --- \\
\bottomrule
\end{tabular}
\end{table}

\subsection{dCoT-SFT Training Hyperparameters}\label{app:A7}

All models are fine-tuned using full-parameter
supervised fine-tuning via
LLaMA-Factory~\cite{llamafactory} with AdamW
optimizer ($\beta_1{=}0.9$, $\beta_2{=}0.999$,
$\epsilon{=}10^{-8}$) and cosine learning rate
schedule, seed 42, on
8 GPUs. Checkpoints are selected at minimum
validation loss to prevent overfitting, as
training loss continues to decrease beyond
epoch~2 while validation loss rises monotonically.
Table~\ref{tab:hyperparams} summarizes the
hyperparameters per model scale.

\noindent\textbf{Exact model identifiers.}
The multimodal base checkpoints are
\texttt{Qwen/Qwen3.5-4B},
\texttt{Qwen/Qwen3.5-9B}, and
\texttt{Qwen/Qwen3.5-27B}~\cite{qwen35}. The reported
dCoT-SFT rows use validation-selected internal
checkpoints \texttt{checkpoint-2000},
\texttt{checkpoint-2500}, and
\texttt{checkpoint-4500}, respectively. The 4B and 9B GRPO rows use checkpoints selected on a held-out development set from runs of up to 3,000 steps. Stage~2 rationale generation
uses the native-multimodal
\texttt{Qwen/Qwen3.5-397B-A17B} checkpoint, not
text-only Qwen3-235B.

\FloatBarrier
\section{MedChain Training Prompts}\label{app:B}
\subsection{Stage 1 Pseudo-CoT Generation Prompt (FlemingVL-38B)}\label{app:B1}

The following prompt is provided to
FlemingVL-38B~\cite{flemingvl} during Stage~1.
The model processes the sequential medical images
directly and generates domain-aware pseudo-CoT
descriptions, subsequently refined by multimodal
Qwen3.5-397B-A17B~\cite{qwen35} in Stage~2.

\vspace{6pt}

\begin{promptblue}{Stage 1: Pseudo-CoT Generation
Prompt (FlemingVL-38B)}

\ttfamily\small\rmfamily

You are a specialized clinical radiologist with
expertise in interpreting medical image sequences
across multiple modalities (CT, MRI, X-ray,
ultrasound, etc.) and performing precise visual
grounding.\\[5pt]

\textbf{You will be given:}
\begin{itemize}[leftmargin=1.2em,
               itemsep=0pt, topsep=2pt]
\item A sequence of medical images
\item A clinical grounding question
\item The correct bounding box of the target
      region (for verification only)
\end{itemize}

\vspace{4pt}
Your task is to generate a realistic, medically
accurate Chain-of-Thought (CoT) that follows the
reasoning process an experienced radiologist or
clinician would use, written as smooth narrative
prose rather than numbered steps.\\[5pt]

The reasoning should naturally cover the following
aspects, integrated into continuous paragraphs:\\[3pt]

\textbf{1. Identify the imaging modality and
anatomical context.}\\
Describe what type of imaging is used (CT, MRI,
X-ray, ultrasound, etc.), the imaging plane if
relevant (axial, sagittal, coronal), and the main
body region and structures that are visible.\\[3pt]

\textbf{2. Interpret the clinical question.}\\
Explain what the question is asking for: which
organ, structure, or pathology is important, and
what kind of difference, object, or region needs
to be localized. Mention any spatial cues (e.g.,
left vs right, superior vs inferior) or clinical
cues (lesion, edema, fracture, fluid,
instrument, etc.).\\[3pt]

\textbf{3. Visually analyze the images.}\\
Describe what you see in terms of medically
meaningful patterns: intensity or density changes,
shapes and borders, symmetry or asymmetry,
presence of lesions, masses, edema, or other
abnormalities. For sequence or multi-image tasks,
explain how the images relate to each other and
what changes across them.\\[3pt]

\textbf{4. Localize the target region and justify
it.}\\
Explain clearly which region best matches the
question and why. Use anatomical landmarks and
spatial relationships (e.g., ``in the lower pole
of the left kidney'', ``adjacent to the lateral
ventricle''). Connect the visual appearance and
location to the clinical description, and briefly
discuss what the finding could represent (e.g.,
nodule, infarct, hemorrhage, edema, calculus, or
instrument).\\[3pt]

\textbf{5. Conclude with the bounding box.}\\
After finishing the reasoning, indicate that the
described region corresponds to the ground-truth
bounding box and provide the coordinates.\\[5pt]

Write everything in natural, continuous sentences
and paragraphs, without explicit step numbers,
bullets, or headings. The reasoning should read
like a concise radiology teaching explanation,
not an instruction list.\\[5pt]

\textbf{Output format (STRICT):}\\[2pt]
\texttt{<analysis>}\\
\texttt{~~[Natural, flowing medical reasoning.}\\
\texttt{~~~No Step 1/2/3 and no bullet points.]}\\
\texttt{</analysis>}\\
\texttt{<answer>~(x1,~y1),(x2,~y2)~</answer>}
\\[5pt]

\textbf{Rules:}
\begin{itemize}[leftmargin=1.2em,
               itemsep=1pt, topsep=2pt]
\item Do NOT mention that the ground-truth
      bounding box was provided.
\item Do NOT copy the answer directly without
      reasoning; the analysis must stand on its
      own as a realistic clinical interpretation.
\item The explanation must clearly connect the
      image content and the clinical question to
      the localized region.
\item The final bounding box in \texttt{<answer>}
      MUST exactly match the given ground-truth
      coordinates
      \texttt{(\{x1\},\{y1\}),(\{x2\},\{y2\})}.
\end{itemize}

\end{promptblue}
\vspace{10pt}

\subsection{Stage 2 Multimodal CoT Refinement Prompt}\label{app:B2}

The following text accompanies the original image
sequence for multimodal Qwen3.5-397B-A17B~\cite{qwen35}
during Stage~2. The model directly inspects the images;
the optional FlemingVL-38B block is supplementary rather
than its only visual context. The annotated box is
privileged offline-generation information.

\vspace{6pt}

\begin{promptteal}{Stage 2: Multimodal CoT Refinement Prompt}

\ttfamily\small

\textbf{MedSG-Bench grounded explanation task}\\
Case ID: \{sample\_id\}\\
Task: \{task\}\\
\{task\_ctx\}\\
Original medical image sequence: \{images\}\\[6pt]

\textbf{Clinical question:}\\
\{question\}\\[6pt]

\textbf{Known correct bounding box:}\\
(\{x1\}, \{y1\}),(\{x2\}, \{y2\})\\[6pt]

\textbf{Instruction:}\\
The bounding box above is already correct. Explain
why this specific boxed region answers the question
but behave as if you do not know the answer yet.
Focus on what is visible inside the box, the
surrounding anatomy, and the discriminative cue
that makes this region the correct target rather
than nearby alternatives. Do not mention that the
box was given to you or that you are verifying
it.\\[4pt]

Do not search for coordinates or describe how you
would estimate the box position. Justify the box
with concise, image-grounded clinical
reasoning.\\[6pt]

\textbf{Style requirements:}\\
$\bullet$ Sound like a concise human clinician or
radiology teacher with self-reflection questions
and corrections (one or two only).\\
$\bullet$ Prefer concrete visual evidence over
generic medical textbook wording.\\
$\bullet$ Keep it short, specific, and
image-grounded.\\
$\bullet$ Do not mention that the box was given
for supervision or verification.\\
$\bullet$ End by repeating the correct box in
\texttt{<answer>}.\\[6pt]

\{prior\_block\}\\[6pt]

\textbf{Return exactly:}\\
\texttt{<think>...</think>}\\
\texttt{<answer>(\{x1\}, \{y1\}),(\{x2\},
\{y2\})</answer>}

\end{promptteal}

\vspace{10pt}

\FloatBarrier
\section{Additional Results}\label{app:C}

\begin{table*}[p] \scriptsize
    \centering
    \caption{Full MedSG-Bench results. The fixed 4B dCoT-SFT checkpoint is in Table~\ref{tab:fixed_4b}. Avg.: Average; IoU and
    \textcolor{gray}{acc@0.5} for all results are
    shown, all numbers are in percentages.}
    \begin{adjustbox}{max width=\textwidth,max totalheight=0.92\textheight}
    \begin{tabular}{lc|cc|cccccc|c}
        \toprule
        \rowcolor[RGB]{242,242,242}\multirow{3}{*}{\bf Model}
        & \multirow{3}{*}{\bf Size}
        & \multicolumn{2}{c|}{\bf IDG}
        & \multicolumn{6}{c|}{\bf ICG}
        & \multirow{3}{*}{\bf Avg.} \\
        \cmidrule(lr){3-10}
         & & RDG & NRDG & MV & OT & VCG & VPG & CMG & RG & \\
        \midrule
        \rowcolor[RGB]{246,237,233}
        \multicolumn{11}{c}{\bf General-purpose MLLMs} \\
        \midrule

\includegraphics[width=0.4cm]{Qwen.png}
Qwen2.5-VL\cite{bai2025qwen2} & 3B
& \makecell{0.59\\\textcolor{gray}{0.30}}
& \makecell{1.62\\\textcolor{gray}{1.30}}
& \makecell{7.12\\\textcolor{gray}{3.90}}
& \makecell{21.32\\\textcolor{gray}{16.80}}
& \makecell{6.98\\\textcolor{gray}{0.80}}
& \makecell{27.36\\\textcolor{gray}{3.40}}
& \makecell{10.02\\\textcolor{gray}{1.65}}
& \makecell{12.99\\\textcolor{gray}{6.82}}
& \makecell{10.94\\\textcolor{gray}{4.20}} \\
\cmidrule(lr){2-11}
\includegraphics[width=0.4cm]{Qwen.png}
Qwen2.5-VL\cite{bai2025qwen2} & 7B
& \makecell{0.88\\\textcolor{gray}{0.30}}
& \makecell{1.25\\\textcolor{gray}{0.00}}
& \makecell{8.48\\\textcolor{gray}{3.73}}
& \makecell{22.41\\\textcolor{gray}{17.80}}
& \makecell{4.22\\\textcolor{gray}{1.00}}
& \makecell{28.87\\\textcolor{gray}{5.70}}
& \makecell{16.29\\\textcolor{gray}{4.45}}
& \makecell{12.58\\\textcolor{gray}{6.21}}
& \makecell{12.31\\\textcolor{gray}{4.90}} \\
\cmidrule(lr){2-11}
\includegraphics[width=0.4cm]{Qwen.png}
Qwen2.5-VL\cite{bai2025qwen2} & 32B
& \makecell{2.69\\\textcolor{gray}{1.40}}
& \makecell{3.48\\\textcolor{gray}{1.20}}
& \makecell{7.35\\\textcolor{gray}{2.61}}
& \makecell{19.12\\\textcolor{gray}{13.40}}
& \makecell{6.53\\\textcolor{gray}{1.30}}
& \makecell{26.92\\\textcolor{gray}{7.10}}
& \makecell{12.59\\\textcolor{gray}{4.90}}
& \makecell{18.71\\\textcolor{gray}{11.67}}
& \makecell{12.47\\\textcolor{gray}{5.71}} \\
\cmidrule(lr){2-11}
\includegraphics[width=0.4cm]{Qwen.png}
Qwen2.5-VL\cite{bai2025qwen2} & 72B
& \makecell{4.37\\\textcolor{gray}{2.60}}
& \makecell{3.46\\\textcolor{gray}{0.80}}
& \makecell{7.22\\\textcolor{gray}{2.78}}
& \makecell{13.11\\\textcolor{gray}{7.70}}
& \makecell{10.33\\\textcolor{gray}{3.50}}
& \makecell{26.45\\\textcolor{gray}{6.30}}
& \makecell{16.32\\\textcolor{gray}{7.00}}
& \makecell{20.19\\\textcolor{gray}{14.10}}
& \makecell{13.35\\\textcolor{gray}{6.12}} \\
\midrule
\includegraphics[width=0.4cm]{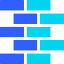}
MiniCPM-V-2\_6\cite{yao2024minicpm} & 8B
& \makecell{1.36\\\textcolor{gray}{0.00}}
& \makecell{1.50\\\textcolor{gray}{0.00}}
& \makecell{15.82\\\textcolor{gray}{5.20}}
& \makecell{24.03\\\textcolor{gray}{18.50}}
& \makecell{9.90\\\textcolor{gray}{2.10}}
& \makecell{28.65\\\textcolor{gray}{12.20}}
& \makecell{12.72\\\textcolor{gray}{3.30}}
& \makecell{12.44\\\textcolor{gray}{3.64}}
& \makecell{13.24\\\textcolor{gray}{5.27}} \\
\cmidrule(lr){2-11}
\includegraphics[width=0.4cm]{minicpm.png}
MiniCPM-O-2\_6\cite{minicpm-o} & 8B
& \makecell{1.69\\\textcolor{gray}{0.10}}
& \makecell{1.63\\\textcolor{gray}{0.00}}
& \makecell{12.11\\\textcolor{gray}{2.43}}
& \makecell{15.25\\\textcolor{gray}{9.60}}
& \makecell{9.88\\\textcolor{gray}{1.70}}
& \makecell{22.96\\\textcolor{gray}{9.20}}
& \makecell{9.53\\\textcolor{gray}{2.35}}
& \makecell{8.82\\\textcolor{gray}{2.02}}
& \makecell{10.12\\\textcolor{gray}{3.23}} \\
\midrule
\includegraphics[width=0.4cm]{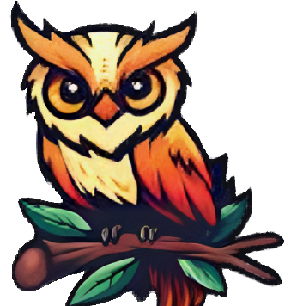}
mPLUG-Owl3\cite{ye2024mplug} & 7B
& \makecell{2.12\\\textcolor{gray}{0.00}}
& \makecell{2.55\\\textcolor{gray}{0.00}}
& \makecell{15.64\\\textcolor{gray}{3.64}}
& \makecell{15.62\\\textcolor{gray}{4.40}}
& \makecell{6.80\\\textcolor{gray}{0.80}}
& \makecell{30.42\\\textcolor{gray}{3.60}}
& \makecell{17.06\\\textcolor{gray}{4.80}}
& \makecell{11.92\\\textcolor{gray}{5.47}}
& \makecell{13.22\\\textcolor{gray}{3.19}} \\
\midrule
\includegraphics[width=0.4cm]{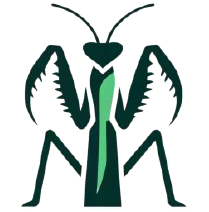}
Mantis-Idefics2\cite{jiang2024mantis} & 8B
& \makecell{0.49\\\textcolor{gray}{0.00}}
& \makecell{0.62\\\textcolor{gray}{0.00}}
& \makecell{\underline{18.69}\\\textcolor{gray}{\underline{8.59}}}
& \makecell{\underline{28.04}\\\textcolor{gray}{\underline{23.50}}}
& \makecell{6.27\\\textcolor{gray}{0.50}}
& \makecell{10.26\\\textcolor{gray}{1.10}}
& \makecell{9.59\\\textcolor{gray}{0.95}}
& \makecell{6.05\\\textcolor{gray}{0.54}}
& \makecell{9.90\\\textcolor{gray}{3.91}} \\
\midrule
\includegraphics[width=0.4cm]{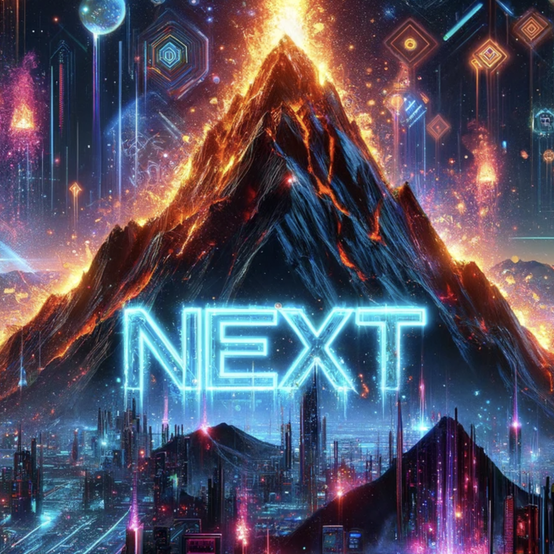}
LLaVA-OneVision\cite{li2024llava} & 7B
& \makecell{1.09\\\textcolor{gray}{0.00}}
& \makecell{0.01\\\textcolor{gray}{0.00}}
& \makecell{9.26\\\textcolor{gray}{1.13}}
& \makecell{10.50\\\textcolor{gray}{3.20}}
& \makecell{11.33\\\textcolor{gray}{1.80}}
& \makecell{22.20\\\textcolor{gray}{5.30}}
& \makecell{19.08\\\textcolor{gray}{6.70}}
& \makecell{17.11\\\textcolor{gray}{5.67}}
& \makecell{12.39\\\textcolor{gray}{3.47}} \\
\cmidrule(lr){2-11}
\includegraphics[width=0.4cm]{llava.png}
LLaVA-OneVision\cite{li2024llava} & 72B
& \makecell{2.58\\\textcolor{gray}{0.80}}
& \makecell{2.87\\\textcolor{gray}{0.90}}
& \makecell{11.74\\\textcolor{gray}{1.39}}
& \makecell{9.61\\\textcolor{gray}{2.30}}
& \makecell{10.95\\\textcolor{gray}{3.30}}
& \makecell{\underline{32.38}\\\textcolor{gray}{\underline{20.30}}}
& \makecell{16.24\\\textcolor{gray}{5.40}}
& \makecell{15.43\\\textcolor{gray}{6.68}}
& \makecell{13.21\\\textcolor{gray}{5.18}} \\
\midrule
\includegraphics[width=0.4cm]{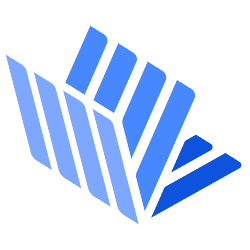}
InternVL2\cite{chen2024internvl} & 8B
& \makecell{0.18\\\textcolor{gray}{0.00}}
& \makecell{0.38\\\textcolor{gray}{0.00}}
& \makecell{17.34\\\textcolor{gray}{7.03}}
& \makecell{26.45\\\textcolor{gray}{21.20}}
& \makecell{5.56\\\textcolor{gray}{0.80}}
& \makecell{10.36\\\textcolor{gray}{0.70}}
& \makecell{6.23\\\textcolor{gray}{1.00}}
& \makecell{15.73\\\textcolor{gray}{7.69}}
& \makecell{10.24\\\textcolor{gray}{4.59}} \\
\cmidrule(lr){2-11}
\includegraphics[width=0.4cm]{internvl.png}
InternVL2\cite{chen2024internvl} & 76B
& \makecell{0.15\\\textcolor{gray}{0.00}}
& \makecell{0.15\\\textcolor{gray}{0.00}}
& \makecell{10.00\\\textcolor{gray}{3.90}}
& \makecell{15.56\\\textcolor{gray}{11.80}}
& \makecell{3.39\\\textcolor{gray}{0.40}}
& \makecell{6.64\\\textcolor{gray}{1.10}}
& \makecell{2.83\\\textcolor{gray}{0.75}}
& \makecell{15.69\\\textcolor{gray}{9.92}}
& \makecell{6.88\\\textcolor{gray}{3.53}} \\
\midrule
\includegraphics[width=0.4cm]{internvl.png}
InternVL2.5\cite{chen2024expanding} & 8B
& \makecell{0.26\\\textcolor{gray}{0.00}}
& \makecell{0.38\\\textcolor{gray}{0.00}}
& \makecell{13.52\\\textcolor{gray}{3.56}}
& \makecell{20.82\\\textcolor{gray}{13.80}}
& \makecell{1.96\\\textcolor{gray}{0.00}}
& \makecell{5.25\\\textcolor{gray}{0.00}}
& \makecell{4.70\\\textcolor{gray}{0.85}}
& \makecell{9.56\\\textcolor{gray}{3.44}}
& \makecell{7.04\\\textcolor{gray}{2.56}} \\
\cmidrule(lr){2-11}
\includegraphics[width=0.4cm]{internvl.png}
InternVL2.5\cite{chen2024expanding} & 78B
& \makecell{0.24\\\textcolor{gray}{0.10}}
& \makecell{0.32\\\textcolor{gray}{0.10}}
& \makecell{9.16\\\textcolor{gray}{2.08}}
& \makecell{16.18\\\textcolor{gray}{10.00}}
& \makecell{4.32\\\textcolor{gray}{0.50}}
& \makecell{11.86\\\textcolor{gray}{2.30}}
& \makecell{5.48\\\textcolor{gray}{1.25}}
& \makecell{10.67\\\textcolor{gray}{4.52}}
& \makecell{7.29\\\textcolor{gray}{2.55}} \\
\midrule
\includegraphics[width=0.4cm]{internvl.png}
InternVL3\cite{zhu2025internvl3} & 8B
& \makecell{1.07\\\textcolor{gray}{0.30}}
& \makecell{1.20\\\textcolor{gray}{0.00}}
& \makecell{14.36\\\textcolor{gray}{4.42}}
& \makecell{13.30\\\textcolor{gray}{6.50}}
& \makecell{6.43\\\textcolor{gray}{0.90}}
& \makecell{18.73\\\textcolor{gray}{4.60}}
& \makecell{4.73\\\textcolor{gray}{1.15}}
& \makecell{15.16\\\textcolor{gray}{7.42}}
& \makecell{9.26\\\textcolor{gray}{3.19}} \\
\cmidrule(lr){2-11}
\includegraphics[width=0.4cm]{internvl.png}
InternVL3\cite{zhu2025internvl3} & 14B
& \makecell{0.66\\\textcolor{gray}{0.00}}
& \makecell{0.71\\\textcolor{gray}{0.00}}
& \makecell{13.24\\\textcolor{gray}{5.31}}
& \makecell{19.77\\\textcolor{gray}{13.00}}
& \makecell{8.60\\\textcolor{gray}{2.10}}
& \makecell{13.17\\\textcolor{gray}{2.40}}
& \makecell{10.87\\\textcolor{gray}{3.70}}
& \makecell{14.57\\\textcolor{gray}{7.76}}
& \makecell{10.53\\\textcolor{gray}{4.41}} \\
\cmidrule(lr){2-11}
\includegraphics[width=0.4cm]{internvl.png}
InternVL3\cite{zhu2025internvl3} & 38B
& \makecell{0.98\\\textcolor{gray}{0.10}}
& \makecell{1.76\\\textcolor{gray}{0.20}}
& \makecell{12.99\\\textcolor{gray}{4.79}}
& \makecell{19.27\\\textcolor{gray}{13.60}}
& \makecell{7.63\\\textcolor{gray}{2.10}}
& \makecell{17.76\\\textcolor{gray}{2.90}}
& \makecell{6.47\\\textcolor{gray}{1.75}}
& \makecell{16.59\\\textcolor{gray}{10.05}}
& \makecell{10.37\\\textcolor{gray}{4.44}} \\
\cmidrule(lr){2-11}
\includegraphics[width=0.4cm]{internvl.png}
InternVL3\cite{zhu2025internvl3} & 78B
& \makecell{0.20\\\textcolor{gray}{0.00}}
& \makecell{0.53\\\textcolor{gray}{0.00}}
& \makecell{6.35\\\textcolor{gray}{2.43}}
& \makecell{13.03\\\textcolor{gray}{8.00}}
& \makecell{3.57\\\textcolor{gray}{0.90}}
& \makecell{11.81\\\textcolor{gray}{2.50}}
& \makecell{3.34\\\textcolor{gray}{0.85}}
& \makecell{12.76\\\textcolor{gray}{8.10}}
& \makecell{6.44\\\textcolor{gray}{2.90}} \\
\midrule
\includegraphics[width=0.4cm]{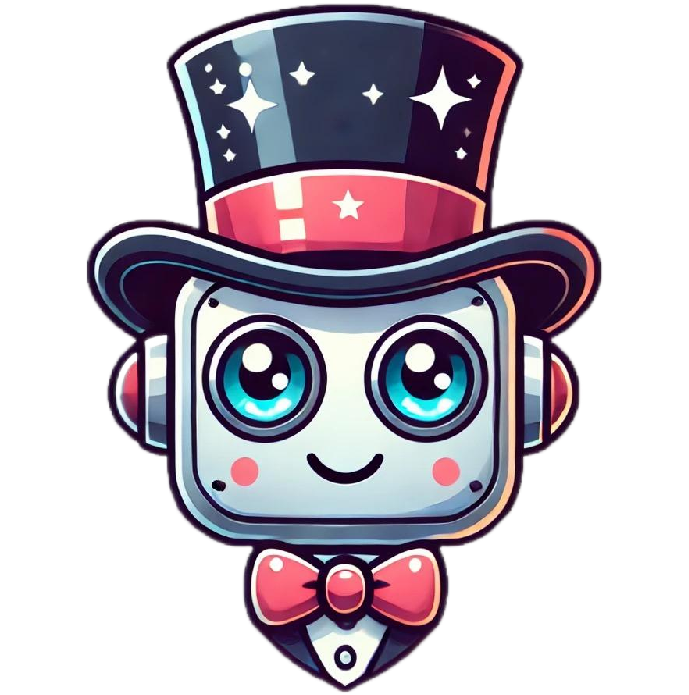}
Migician\cite{li2025migician} & 7B
& \makecell{\underline{15.26}\\\textcolor{gray}{\underline{7.80}}}
& \makecell{\underline{14.49}\\\textcolor{gray}{\underline{6.10}}}
& \makecell{18.16\\\textcolor{gray}{7.84}}
& \makecell{21.38\\\textcolor{gray}{14.90}}
& \makecell{\underline{14.23}\\\textcolor{gray}{\underline{7.20}}}
& \makecell{28.87\\\textcolor{gray}{13.70}}
& \makecell{\underline{21.41}\\\textcolor{gray}{\underline{12.15}}}
& \makecell{\underline{25.30}\\\textcolor{gray}{\underline{18.02}}}
& \makecell{\underline{20.29}\\\textcolor{gray}{\underline{11.39}}} \\
        \midrule
        \rowcolor[RGB]{246,237,233}
        \multicolumn{11}{c}{\bf Medical-domain specialized MLLMs} \\
        \midrule
\includegraphics[width=0.4cm]{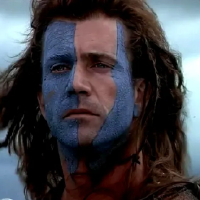}
HuatuoGPT-Vision\cite{chen2024huatuogpt} & 7B
& \makecell{1.35\\\textcolor{gray}{0.00}}
& \makecell{1.84\\\textcolor{gray}{0.20}}
& \makecell{10.42\\\textcolor{gray}{2.78}}
& \makecell{14.57\\\textcolor{gray}{9.20}}
& \makecell{7.99\\\textcolor{gray}{0.80}}
& \makecell{15.52\\\textcolor{gray}{2.30}}
& \makecell{9.46\\\textcolor{gray}{2.15}}
& \makecell{9.60\\\textcolor{gray}{1.82}}
& \makecell{8.97\\\textcolor{gray}{2.36}} \\
\cmidrule(lr){2-11}
\includegraphics[width=0.4cm]{huatuo.png}
HuatuoGPT-Vision\cite{chen2024huatuogpt} & 34B
& \makecell{1.44\\\textcolor{gray}{0.00}}
& \makecell{2.15\\\textcolor{gray}{0.00}}
& \makecell{9.41\\\textcolor{gray}{1.65}}
& \makecell{13.25\\\textcolor{gray}{8.30}}
& \makecell{6.43\\\textcolor{gray}{0.70}}
& \makecell{14.53\\\textcolor{gray}{1.40}}
& \makecell{10.60\\\textcolor{gray}{2.60}}
& \makecell{8.60\\\textcolor{gray}{1.75}}
& \makecell{8.57\\\textcolor{gray}{2.09}} \\
\midrule
\includegraphics[width=0.4cm]{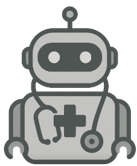}
MedSeq-Grounder & 7B
& \makecell{83.29\\\textcolor{gray}{93.20}}
& \makecell{83.72\\\textcolor{gray}{94.10}}
& \makecell{55.03\\\textcolor{gray}{60.19}}
& \makecell{62.10\\\textcolor{gray}{67.20}}
& \makecell{74.11\\\textcolor{gray}{82.60}}
& \makecell{85.25\\\textcolor{gray}{98.80}}
& \makecell{78.77\\\textcolor{gray}{82.75}}
& \makecell{60.43\\\textcolor{gray}{65.59}}
& \makecell{72.55\\\textcolor{gray}{79.71}} \\
\midrule
\rowcolor[RGB]{246,237,233}
\includegraphics[width=0.4cm]{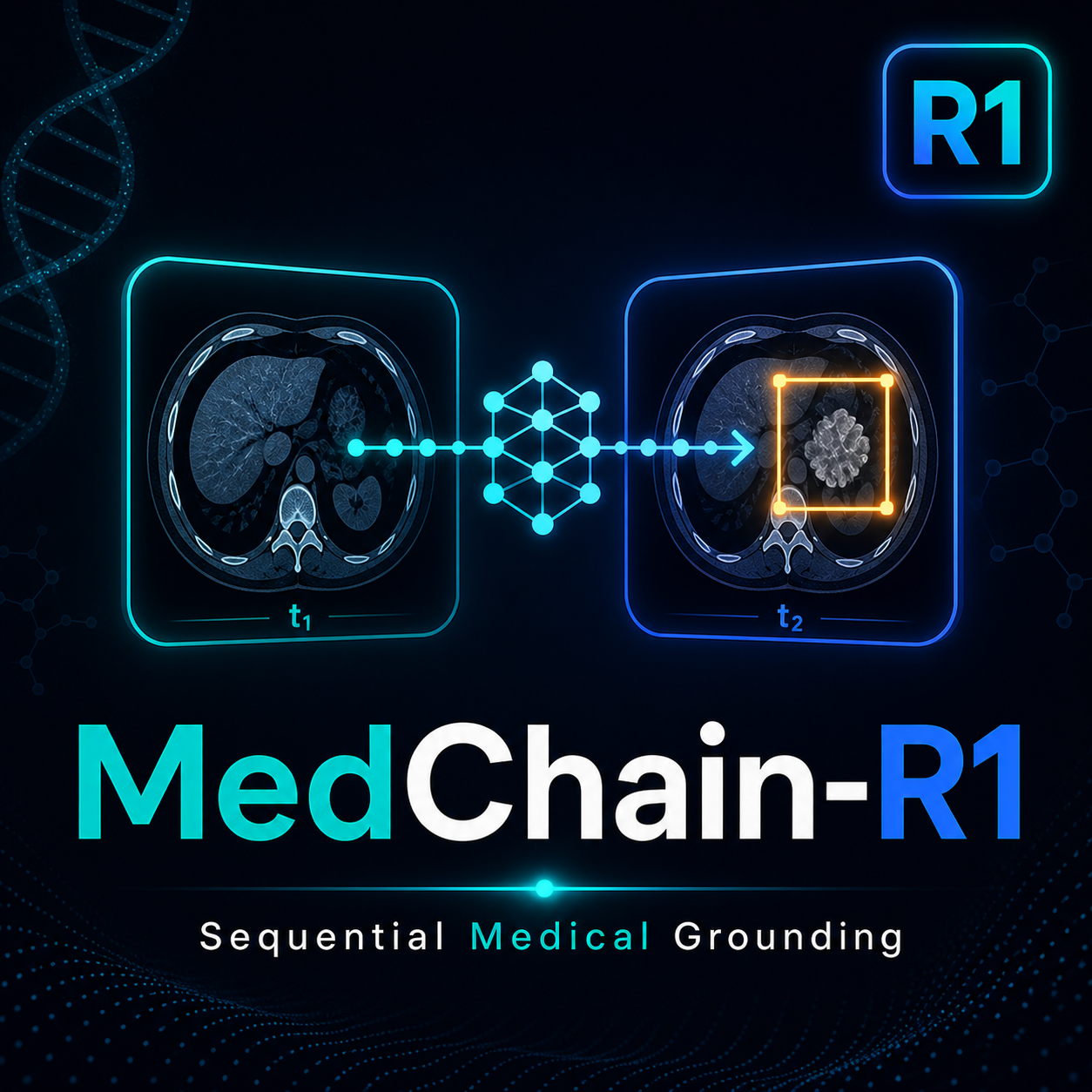}
\textbf{Ours-dCoT+GRPO (4B)} & 4B
& \makecell{\textbf{83.82}\\\textcolor{gray}{94.50}}
& \makecell{83.21\\\textcolor{gray}{92.70}}
& \makecell{54.80\\\textcolor{gray}{59.93}}
& \makecell{\textbf{67.90}\\\textcolor{gray}{\textbf{76.20}}}
& \makecell{66.09\\\textcolor{gray}{72.10}}
& \makecell{\textbf{89.43}\\\textcolor{gray}{\textbf{99.80}}}
& \makecell{\textbf{83.59}\\\textcolor{gray}{\textbf{87.55}}}
& \makecell{\textbf{70.72}\\\textcolor{gray}{\textbf{80.30}}}
& \makecell{\textbf{75.32}\\\textcolor{gray}{\textbf{82.89}}} \\
\midrule
\rowcolor[RGB]{246,237,233}
\includegraphics[width=0.4cm]{MedChain-R1.png}
\textbf{Ours-CoT-SFT (9B)} & 9B
& \makecell{\textbf{84.00}\\\textcolor{gray}{\textbf{93.10}}}
& \makecell{83.02\\\textcolor{gray}{91.60}}
& \makecell{\textbf{55.87}\\\textcolor{gray}{\textbf{60.98}}}
& \makecell{\textbf{67.46}\\\textcolor{gray}{\textbf{75.00}}}
& \makecell{\textbf{74.64}\\\textcolor{gray}{\textbf{83.20}}}
& \makecell{\textbf{91.22}\\\textcolor{gray}{\textbf{99.90}}}
& \makecell{\textbf{84.16}\\\textcolor{gray}{\textbf{87.40}}}
& \makecell{\textbf{70.77}\\\textcolor{gray}{\textbf{78.95}}}
& \makecell{\textbf{76.60}\\\textcolor{gray}{\textbf{83.55}}} \\
\midrule
\rowcolor[RGB]{246,237,233}
\includegraphics[width=0.4cm]{MedChain-R1.png}
\textbf{Ours-dCoT+GRPO (9B)} & 9B
& \makecell{\textbf{84.16}\\\textcolor{gray}{93.20}}
& \makecell{83.12\\\textcolor{gray}{91.40}}
& \makecell{\textbf{56.98}\\\textcolor{gray}{\textbf{62.02}}}
& \makecell{\textbf{67.65}\\\textcolor{gray}{\textbf{74.90}}}
& \makecell{\textbf{74.80}\\\textcolor{gray}{\textbf{82.80}}}
& \makecell{\textbf{91.22}\\\textcolor{gray}{\textbf{99.80}}}
& \makecell{\textbf{84.82}\\\textcolor{gray}{\textbf{88.25}}}
& \makecell{\textbf{70.53}\\\textcolor{gray}{\textbf{79.22}}}
& \makecell{\textbf{76.90}\\\textcolor{gray}{\textbf{83.82}}} \\
\midrule
\rowcolor[RGB]{246,237,233}
\includegraphics[width=0.4cm]{MedChain-R1.png}
\textbf{Ours-CoT-SFT (27B)} & 27B
& \makecell{81.87\\\textcolor{gray}{93.00}}
& \makecell{80.97\\\textcolor{gray}{91.60}}
& \makecell{\textbf{60.14}\\\textcolor{gray}{\textbf{66.72}}}
& \makecell{\textbf{69.53}\\\textcolor{gray}{\textbf{78.30}}}
& \makecell{\textbf{74.64}\\\textcolor{gray}{\textbf{83.30}}}
& \makecell{\textbf{91.87}\\\textcolor{gray}{\textbf{99.90}}}
& \makecell{\textbf{84.24}\\\textcolor{gray}{\textbf{88.15}}}
& \makecell{\textbf{72.51}\\\textcolor{gray}{\textbf{81.98}}}
& \makecell{\textbf{77.24}\\\textcolor{gray}{\textbf{85.20}}} \\
        \bottomrule
    \end{tabular}%
    \end{adjustbox}
    \label{tab:full_results}
\end{table*}

\begin{table*}[!htbp]
\centering
\caption{Ablation of training stages for
\method{} on MedSG-Bench across model scales.
IoU and \textcolor{gray}{\texttt{acc@0.5}}
are reported, all numbers in percentages.}
\label{tab:ablation_stages}
\resizebox{\textwidth}{!}{%
\begin{tabular}{lc | cc | cccccc | c}
\toprule
\rowcolor[RGB]{242,242,242}\multirow{2}{*}{\textbf{Model}}
& \multirow{2}{*}{\textbf{Size}}
& \multicolumn{2}{c|}{\textbf{IDG}}
& \multicolumn{6}{c|}{\textbf{ICG}}
& \multirow{2}{*}{\textbf{Avg.}} \\
\cmidrule(lr){3-4} \cmidrule(lr){5-10}
& & \textbf{RDG} & \textbf{NRDG}
& \textbf{MV} & \textbf{OT}
& \textbf{VCG} & \textbf{VPG}
& \textbf{CMG} & \textbf{RG} & \\
\midrule
\multicolumn{11}{l}{\textit{Qwen3.5~---~4B}} \\
\midrule
\includegraphics[width=0.4cm]{MedChain-R1.png}
~~Zero-shot & 4B
& \makecell{0.00\\\textcolor{gray}{0.00}}
& \makecell{0.00\\\textcolor{gray}{0.00}}
& \makecell{8.27\\\textcolor{gray}{4.88}}
& \makecell{26.66\\\textcolor{gray}{22.60}}
& \makecell{1.50\\\textcolor{gray}{0.00}}
& \makecell{2.65\\\textcolor{gray}{0.10}}
& \makecell{0.35\\\textcolor{gray}{0.00}}
& \makecell{1.69\\\textcolor{gray}{0.13}}
& \makecell{4.52\\\textcolor{gray}{2.96}} \\
\cmidrule(lr){2-11}
\rowcolor[RGB]{246,237,233}\includegraphics[width=0.4cm]{MedChain-R1.png}
~~+~dCoT-SFT+GRPO & 4B
& \makecell{\textbf{83.82}\\\textcolor{gray}{94.50}}
& \makecell{83.21\\\textcolor{gray}{92.70}}
& \makecell{54.80\\\textcolor{gray}{59.93}}
& \makecell{\textbf{67.90}\\\textcolor{gray}{76.20}}
& \makecell{66.09\\\textcolor{gray}{72.10}}
& \makecell{\textbf{89.43}\\\textcolor{gray}{99.80}}
& \makecell{\textbf{83.59}\\\textcolor{gray}{87.55}}
& \makecell{\textbf{70.72}\\\textcolor{gray}{80.30}}
& \makecell{\textbf{75.32}\\\textcolor{gray}{82.89}} \\
\midrule
\multicolumn{11}{l}{\textit{Qwen3.5~---~9B}} \\
\midrule
\includegraphics[width=0.4cm]{MedChain-R1.png}
~~Zero-shot & 9B
& \makecell{0.00\\\textcolor{gray}{0.00}}
& \makecell{0.00\\\textcolor{gray}{0.00}}
& \makecell{8.57\\\textcolor{gray}{4.97}}
& \makecell{28.30\\\textcolor{gray}{24.50}}
& \makecell{0.52\\\textcolor{gray}{0.00}}
& \makecell{1.62\\\textcolor{gray}{0.20}}
& \makecell{0.25\\\textcolor{gray}{0.00}}
& \makecell{1.26\\\textcolor{gray}{0.13}}
& \makecell{4.43\\\textcolor{gray}{3.18}} \\
\cmidrule(lr){2-11}
\rowcolor[RGB]{246,237,233}\includegraphics[width=0.4cm]{MedChain-R1.png}
~~+~CoT-SFT & 9B
& \makecell{\textbf{84.00}\\\textcolor{gray}{93.10}}
& \makecell{83.02\\\textcolor{gray}{91.60}}
& \makecell{\textbf{55.87}\\\textcolor{gray}{60.98}}
& \makecell{\textbf{67.46}\\\textcolor{gray}{75.00}}
& \makecell{\textbf{74.64}\\\textcolor{gray}{83.20}}
& \makecell{\textbf{91.22}\\\textcolor{gray}{99.90}}
& \makecell{\textbf{84.16}\\\textcolor{gray}{87.40}}
& \makecell{\textbf{70.77}\\\textcolor{gray}{78.95}}
& \makecell{\textbf{76.60}\\\textcolor{gray}{83.55}} \\
\cmidrule(lr){2-11}
\rowcolor[RGB]{246,237,233}\includegraphics[width=0.4cm]{MedChain-R1.png}
~~+~dCoT-SFT+GRPO & 9B
& \makecell{\textbf{84.16}\\\textcolor{gray}{93.20}}
& \makecell{83.12\\\textcolor{gray}{91.40}}
& \makecell{\textbf{56.98}\\\textcolor{gray}{62.02}}
& \makecell{\textbf{67.65}\\\textcolor{gray}{74.90}}
& \makecell{\textbf{74.80}\\\textcolor{gray}{82.80}}
& \makecell{\textbf{91.22}\\\textcolor{gray}{99.80}}
& \makecell{\textbf{84.82}\\\textcolor{gray}{88.25}}
& \makecell{\textbf{70.53}\\\textcolor{gray}{79.22}}
& \makecell{\textbf{76.90}\\\textcolor{gray}{83.82}} \\
\midrule
\multicolumn{11}{l}{\textit{Qwen3.5~---~27B}} \\
\midrule
\includegraphics[width=0.4cm]{MedChain-R1.png}
~~Zero-shot & 27B
& \makecell{0.00\\\textcolor{gray}{0.00}}
& \makecell{0.00\\\textcolor{gray}{0.00}}
& \makecell{8.48\\\textcolor{gray}{4.88}}
& \makecell{28.39\\\textcolor{gray}{24.50}}
& \makecell{0.25\\\textcolor{gray}{0.00}}
& \makecell{2.50\\\textcolor{gray}{0.20}}
& \makecell{0.25\\\textcolor{gray}{0.00}}
& \makecell{0.82\\\textcolor{gray}{0.00}}
& \makecell{4.42\\\textcolor{gray}{3.15}} \\
\cmidrule(lr){2-11}
\rowcolor[RGB]{246,237,233}\includegraphics[width=0.4cm]{MedChain-R1.png}
~~+~CoT-SFT & 27B
& \makecell{81.87\\\textcolor{gray}{93.00}}
& \makecell{80.97\\\textcolor{gray}{91.60}}
& \makecell{\textbf{60.14}\\\textcolor{gray}{66.72}}
& \makecell{\textbf{69.53}\\\textcolor{gray}{78.30}}
& \makecell{\textbf{74.64}\\\textcolor{gray}{83.30}}
& \makecell{\textbf{91.87}\\\textcolor{gray}{99.90}}
& \makecell{\textbf{84.24}\\\textcolor{gray}{88.15}}
& \makecell{\textbf{72.51}\\\textcolor{gray}{81.98}}
& \makecell{\textbf{77.24}\\\textcolor{gray}{85.20}} \\
\bottomrule
\end{tabular}%
}
\vspace{2pt}
{\small \textbf{Bold} = best in column per scale.
Top = IoU, \textcolor{gray}{bottom = acc@0.5}. Table~\ref{tab:fixed_4b}
separates the verified dCoT-SFT and GRPO checkpoints.}
\end{table*}


Table~\ref{tab:appendix_medical_mllms} reports
performance of additional general-domain and
medical-specialized MLLMs on
MedSG-Bench~\cite{medsgbench} under the zero-shot
evaluation protocol described in
Appendix~\ref{app:A6}. These results complement Table~\ref{tab:full_results} and further confirm the substantial
sequential grounding gap across all model families,
including recent models such as
Qwen3-VL~\cite{qwen3vl} and
MedGemma~\cite{sellergren2025medgemma}.

\begin{table*}[!ht]
\centering
\caption{Performance of additional MLLMs on
MedSG-Bench. IoU and
\textcolor{gray}{\texttt{acc@0.5}} are reported,
all numbers in percentages.}
\label{tab:appendix_medical_mllms}
\resizebox{\textwidth}{!}{%
\begin{tabular}{lc | cc | cccccc | c}
\rowcolor[RGB]{242,242,242}
\toprule
\multirow{2}{*}{\textbf{Model}}
& \multirow{2}{*}{\textbf{Size}}
& \multicolumn{2}{c|}{\textbf{IDG}}
& \multicolumn{6}{c|}{\textbf{ICG}}
& \multirow{2}{*}{\textbf{Avg.}} \\
\cmidrule(lr){3-4} \cmidrule(lr){5-10}
& & \textbf{RDG} & \textbf{NRDG}
& \textbf{MV} & \textbf{OT}
& \textbf{VCG} & \textbf{VPG}
& \textbf{CMG} & \textbf{RG} & \\
\midrule
\rowcolor[RGB]{246,237,233}
\multicolumn{11}{c}{\textbf{General-domain MLLMs}} \\
\midrule
Qwen3-VL\cite{qwen3vl} & 4B
& \makecell{0.01\\\textcolor{gray}{0.00}}
& \makecell{0.00\\\textcolor{gray}{0.00}}
& \makecell{7.27\\\textcolor{gray}{3.83}}
& \makecell{25.08\\\textcolor{gray}{21.30}}
& \makecell{0.62\\\textcolor{gray}{0.00}}
& \makecell{2.67\\\textcolor{gray}{0.00}}
& \makecell{0.20\\\textcolor{gray}{0.00}}
& \makecell{0.94\\\textcolor{gray}{0.00}}
& \makecell{4.00\\\textcolor{gray}{2.67}} \\
\cmidrule(lr){2-11}
Qwen3-VL\cite{qwen3vl} & 8B
& \makecell{0.04\\\textcolor{gray}{0.00}}
& \makecell{0.00\\\textcolor{gray}{0.00}}
& \makecell{7.43\\\textcolor{gray}{3.83}}
& \makecell{24.73\\\textcolor{gray}{20.60}}
& \makecell{0.74\\\textcolor{gray}{0.00}}
& \makecell{1.87\\\textcolor{gray}{0.00}}
& \makecell{0.46\\\textcolor{gray}{0.05}}
& \makecell{1.12\\\textcolor{gray}{0.00}}
& \makecell{4.00\\\textcolor{gray}{2.61}} \\
\cmidrule(lr){2-11}
Qwen3-VL\cite{qwen3vl} & 30B
& \makecell{0.01\\\textcolor{gray}{0.00}}
& \makecell{0.00\\\textcolor{gray}{0.00}}
& \makecell{3.12\\\textcolor{gray}{1.66}}
& \makecell{0.65\\\textcolor{gray}{0.40}}
& \makecell{0.82\\\textcolor{gray}{0.00}}
& \makecell{2.50\\\textcolor{gray}{0.00}}
& \makecell{0.10\\\textcolor{gray}{0.00}}
& \makecell{1.05\\\textcolor{gray}{0.00}}
& \makecell{0.97\\\textcolor{gray}{0.24}} \\
\midrule
Gemma\cite{gemma3} & 4B
& \makecell{0.03\\\textcolor{gray}{0.00}}
& \makecell{0.00\\\textcolor{gray}{0.00}}
& \makecell{0.24\\\textcolor{gray}{0.09}}
& \makecell{0.95\\\textcolor{gray}{0.80}}
& \makecell{0.79\\\textcolor{gray}{0.00}}
& \makecell{3.12\\\textcolor{gray}{0.00}}
& \makecell{0.11\\\textcolor{gray}{0.05}}
& \makecell{1.09\\\textcolor{gray}{0.00}}
& \makecell{0.73\\\textcolor{gray}{0.10}} \\
\cmidrule(lr){2-11}
Gemma\cite{gemma3} & 26B
& \makecell{0.01\\\textcolor{gray}{0.00}}
& \makecell{0.00\\\textcolor{gray}{0.00}}
& \makecell{0.59\\\textcolor{gray}{0.26}}
& \makecell{2.83\\\textcolor{gray}{2.70}}
& \makecell{0.17\\\textcolor{gray}{0.00}}
& \makecell{1.26\\\textcolor{gray}{0.00}}
& \makecell{0.02\\\textcolor{gray}{0.00}}
& \makecell{0.59\\\textcolor{gray}{0.00}}
& \makecell{0.61\\\textcolor{gray}{0.31}} \\
\midrule
\rowcolor[RGB]{246,237,233}
\multicolumn{11}{c}{\textbf{Medical-domain
specialized MLLMs}} \\
\midrule
Med-VLM-R1\cite{medvlmr1} & 2B
& \makecell{0.01\\\textcolor{gray}{0.00}}
& \makecell{0.01\\\textcolor{gray}{0.00}}
& \makecell{4.16\\\textcolor{gray}{2.53}}
& \makecell{8.98\\\textcolor{gray}{7.00}}
& \makecell{0.06\\\textcolor{gray}{0.00}}
& \makecell{0.68\\\textcolor{gray}{0.00}}
& \makecell{0.05\\\textcolor{gray}{0.00}}
& \makecell{0.35\\\textcolor{gray}{0.00}}
& \makecell{1.57\\\textcolor{gray}{1.03}} \\
\midrule
MedGemma\cite{sellergren2025medgemma} & 1.5B
& \makecell{0.03\\\textcolor{gray}{0.00}}
& \makecell{0.01\\\textcolor{gray}{0.00}}
& \makecell{8.53\\\textcolor{gray}{4.97}}
& \makecell{28.46\\\textcolor{gray}{24.40}}
& \makecell{0.91\\\textcolor{gray}{0.00}}
& \makecell{2.08\\\textcolor{gray}{0.10}}
& \makecell{0.46\\\textcolor{gray}{0.00}}
& \makecell{1.01\\\textcolor{gray}{0.00}}
& \makecell{4.54\\\textcolor{gray}{3.14}} \\
\cmidrule(lr){2-11}
MedGemma\cite{sellergren2025medgemma} & 4B
& \makecell{0.58\\\textcolor{gray}{0.00}}
& \makecell{0.26\\\textcolor{gray}{0.00}}
& \makecell{5.67\\\textcolor{gray}{2.44}}
& \makecell{18.86\\\textcolor{gray}{11.70}}
& \makecell{9.42\\\textcolor{gray}{1.30}}
& \makecell{7.06\\\textcolor{gray}{1.30}}
& \makecell{3.75\\\textcolor{gray}{0.95}}
& \makecell{8.49\\\textcolor{gray}{1.28}}
& \makecell{6.52\\\textcolor{gray}{2.17}} \\
\cmidrule(lr){2-11}
MedGemma\cite{sellergren2025medgemma} & 27B
& \makecell{0.01\\\textcolor{gray}{0.00}}
& \makecell{0.06\\\textcolor{gray}{0.00}}
& \makecell{5.29\\\textcolor{gray}{2.26}}
& \makecell{22.26\\\textcolor{gray}{16.10}}
& \makecell{2.27\\\textcolor{gray}{0.00}}
& \makecell{3.11\\\textcolor{gray}{0.30}}
& \makecell{2.92\\\textcolor{gray}{0.35}}
& \makecell{5.35\\\textcolor{gray}{0.61}}
& \makecell{4.94\\\textcolor{gray}{2.14}} \\
\midrule
Fleming VL\cite{flemingvl} & 8B
& \makecell{0.05\\\textcolor{gray}{0.00}}
& \makecell{0.03\\\textcolor{gray}{0.00}}
& \makecell{6.50\\\textcolor{gray}{3.57}}
& \makecell{22.57\\\textcolor{gray}{18.20}}
& \makecell{1.48\\\textcolor{gray}{0.10}}
& \makecell{2.92\\\textcolor{gray}{0.10}}
& \makecell{0.89\\\textcolor{gray}{0.10}}
& \makecell{0.00\\\textcolor{gray}{0.00}}
& \makecell{3.77\\\textcolor{gray}{2.36}} \\
\midrule
Lingshu \cite{xu2025lingshu} & 7B
& \makecell{0.20\\\textcolor{gray}{0.00}}
& \makecell{0.03\\\textcolor{gray}{0.00}}
& \makecell{4.94\\\textcolor{gray}{1.83}}
& \makecell{19.53\\\textcolor{gray}{16.00}}
& \makecell{3.47\\\textcolor{gray}{0.10}}
& \makecell{18.30\\\textcolor{gray}{1.00}}
& \makecell{3.48\\\textcolor{gray}{0.10}}
& \makecell{4.22\\\textcolor{gray}{1.69}}
& \makecell{6.27\\\textcolor{gray}{2.27}} \\
\cmidrule(lr){2-11}
Lingshu \cite{xu2025lingshu}& 32B
& \makecell{0.04\\\textcolor{gray}{0.00}}
& \makecell{0.02\\\textcolor{gray}{0.00}}
& \makecell{4.54\\\textcolor{gray}{1.74}}
& \makecell{17.54\\\textcolor{gray}{13.20}}
& \makecell{1.34\\\textcolor{gray}{0.20}}
& \makecell{0.03\\\textcolor{gray}{0.00}}
& \makecell{0.59\\\textcolor{gray}{0.00}}
& \makecell{0.00\\\textcolor{gray}{0.00}}
& \makecell{2.63\\\textcolor{gray}{1.60}} \\
\midrule
QoQ-Med3 \cite{dai2025qoq} & 8B
& \makecell{0.04\\\textcolor{gray}{0.00}}
& \makecell{0.00\\\textcolor{gray}{0.00}}
& \makecell{7.35\\\textcolor{gray}{3.75}}
& \makecell{24.92\\\textcolor{gray}{21.00}}
& \makecell{0.66\\\textcolor{gray}{0.00}}
& \makecell{1.64\\\textcolor{gray}{0.00}}
& \makecell{0.50\\\textcolor{gray}{0.10}}
& \makecell{1.12\\\textcolor{gray}{0.00}}
& \makecell{3.98\\\textcolor{gray}{2.65}} \\
\bottomrule
\end{tabular}%
}
\vspace{2pt}
{\small Top = IoU,
\textcolor{gray}{bottom = acc@0.5}.}
\end{table*}

\subsection{Task-Specific RL Baseline}\label{app:C1}

We evaluated MedGRPO-7B-RL with direct box generation
on all 9,630 test examples without parser failures.
Table~\ref{tab:medgrpo} shows that training for adjacent
surgical video frames transfers poorly to longitudinal,
cross-view, and cross-modal medical sequences.

\begin{table}[!ht]
\centering\small
\caption{MedGRPO-7B-RL transfer to MedSG-Bench;
mIoU/Acc@0.5 in percent.}
\label{tab:medgrpo}
\begin{tabular}{lc@{\qquad}lc}
\toprule
Task & Score & Task & Score \\
\midrule
RDG & 0.09/0.00 & NRDG & 0.06/0.00 \\
Multi-view & 3.68/1.48 & Object Tracking & 17.63/15.00 \\
Concept & 4.57/0.40 & Patch & 18.44/1.20 \\
Crossmodal & 0.78/0.15 & Referring & 9.31/3.37 \\
\midrule
\multicolumn{3}{r}{Overall} & 6.27/2.45 \\
\bottomrule
\end{tabular}
\end{table}

\subsection{Qwen2.5-VL-7B Continued Training}\label{app:C2}

Full-parameter CoT-SFT continued from the published
MedSeq-Grounder checkpoint reaches 75.71 mIoU and
82.34 Acc@0.5, compared with the published checkpoint's
72.55/79.71. This is useful continued-training evidence,
but not a matched answer-only-versus-CoT experiment: the
initialization is already MedSeq-Grounder-trained and no
equally trained answer-only continuation is available.

\subsection{GRPO Configuration}\label{app:C3}

The GRPO run uses 32,000 examples
(eight tasks $\times$ 4,000; seed 42), eight rollouts per
prompt, one policy iteration, equal implicitly summed IoU
and binary-format rewards, KL coefficient $\beta=0.04$,
clipping $\epsilon=0.20$, initial learning rate $10^{-6}$
with linear decay, AdamW, per-device batch size 8,
gradient accumulation 2, bfloat16, and a 1,200-token
completion limit. The effective sampling temperature is
1.0.

\subsection{Reasoning-Reliance Inference Controls}\label{app:C4}

We evaluated two inference-time interventions on the
same 4B dCoT-SFT validation-selected checkpoint-2000
used for the 74.955 mIoU / 82.368 Acc@0.5 baseline.
Both controls use the identical 9,630-example test set
and deterministic decoding. In the \emph{no-think}
condition, the assistant response is prefixed with
\texttt{<answer>} so the model must immediately generate
a box. In the \emph{corrupted-reasoning} condition, each
example is deterministically assigned a rationale from a
different task, followed by an \texttt{<answer>} prefix.
There were no inference failures.

Assistant prefilling occasionally removes only the outer
parentheses from an otherwise unambiguous coordinate
string (for example,
\texttt{<answer>90,142),(106,161</answer>}). We therefore
apply one delimiter-tolerant answer-tag parser to the
baseline and both interventions; this leaves the normal
baseline score unchanged and prevents valid coordinates
from being mapped to the zero box.

\begin{table}[H]
\centering\small
\begin{minipage}[t]{0.36\linewidth}\centering
\caption{Reasoning-reliance controls on the fixed 4B
dCoT-SFT checkpoint; all values are percentages.}
\label{tab:reasoning_controls}
\begin{adjustbox}{max width=\linewidth}\begin{tabular}{lrr}
\toprule
Inference condition & mIoU & Acc@0.5 \\
\midrule
Normal dCoT inference & 74.955 & 82.368 \\
Forced no-think & 72.025 & 78.660 \\
Corrupted cross-task reasoning & 62.773 & 68.930 \\
\bottomrule
\end{tabular}\end{adjustbox}
\end{minipage}\hfill
\begin{minipage}[t]{0.6\linewidth}\centering
\caption{Paired intervention-minus-normal differences
from 10,000 example-level bootstrap resamples. Differences
and confidence intervals are percentage points.}
\label{tab:reasoning_controls_paired}
\small
\begin{adjustbox}{max width=\linewidth}\begin{tabular}{lrr}
\toprule
Intervention & $\Delta$ mIoU [95\% CI] & $\Delta$ Acc@0.5 [95\% CI] \\
\midrule
No-think & $-2.930$ [$-3.293,-2.580$] & $-3.707$ [$-4.309,-3.126$] \\
Corrupted & $-12.182$ [$-12.745,-11.603$] & $-13.437$ [$-14.216,-12.648$] \\
\bottomrule
\end{tabular}\end{adjustbox}
\end{minipage}
\end{table}

\begin{table}[H]
\centering
\caption{Task-wise reasoning-control results on the fixed
4B dCoT-SFT checkpoint. Each cell reports mIoU/Acc@0.5
in percent.}
\label{tab:reasoning_controls_taskwise}
\footnotesize
\begin{tabular}{lrrr}
\toprule
Task & Normal dCoT & Forced no-think & Corrupted reasoning \\
\midrule
Registered Difference & 82.86/93.30 & 82.15/92.20 & 65.19/73.30 \\
Non-Registered Difference & 81.30/91.20 & 80.94/90.80 & 80.16/89.60 \\
Multi-view & 54.01/58.97 & 49.72/53.57 & 39.55/41.55 \\
Object Tracking & 64.77/71.00 & 55.26/58.90 & 46.74/50.60 \\
Concept & 70.64/78.80 & 69.40/76.40 & 65.12/72.40 \\
Patch & 90.21/99.70 & 87.55/99.00 & 86.03/96.80 \\
Crossmodal & 83.39/87.00 & 81.44/84.85 & 65.78/70.00 \\
Referring & 69.67/79.28 & 66.36/73.55 & 56.88/63.02 \\
\midrule
Overall & 74.96/82.37 & 72.02/78.66 & 62.77/68.93 \\
\bottomrule
\end{tabular}
\end{table}

All paired bootstrap and sign-flip tests reach
$p<10^{-4}$. Removing generated reasoning causes a
moderate but significant decline, while an unrelated
reasoning prefix causes a substantially larger decline.
This supports sensitivity to reasoning context, not the
stronger claim that the displayed rationale is a faithful
causal trace. Corrupted prefilling also constitutes a
deliberate distribution shift. These controls use the
SFT checkpoint rather than the GRPO checkpoint because
the retained local GRPO weights are incomplete; they
must therefore be compared with the 74.955/82.368 SFT
baseline, not the 75.321/82.887 GRPO result.



\subsection{Task-wise Paired Bootstrap Comparisons}\label{app:taskwise_boot}

Tables~\ref{tab:taskwise_boot_dcot} and~\ref{tab:taskwise_boot_grpo} compare the fixed 4B dCoT-SFT checkpoint-2000 and the GRPO checkpoint with the fixed 4B answer-only SFT checkpoint on each task of the 9,630-example test set. Predictions are paired by example ID, and each task uses 100,000 paired example-level bootstrap resamples. Differences are the comparison model minus answer-only SFT, in percentage points. Each task contains 1,000 examples except multi-view (1,148), cross-modal (2,000) and referring (1,482); the overall scores in Tables~\ref{tab:main_results} and~\ref{tab:fixed_4b} are example-weighted means over these tasks. With 16 tests per table, a Bonferroni-adjusted threshold is $0.05/16\approx0.003$. Reported $p<10^{-5}$ indicates that no resample reached the null.

The dCoT-SFT difference on concept grounding is not significant ($p=0.063$ for mIoU, $p=0.78$ for Acc@0.5). On patch grounding, mIoU decreases while Acc@0.5 is unchanged (99.7 for both models), so the loss reflects box tightness rather than missed targets. On multi-view and referring grounding, the Acc@0.5 decreases ($-4.27$ and $-4.18$) are larger than the corresponding mIoU decreases. GRPO remains below answer-only SFT on concept, multi-view, patch and referring grounding.

\begin{table}[!ht]
\centering
\caption{\textbf{Task-wise paired comparison: dCoT-SFT versus answer-only SFT (4B).} Paired example-level bootstrap on the 9,630 MedSG-Bench test questions (100,000 resamples per task). Cells show mIoU (top) and \textcolor{gray}{Acc@0.5} (bottom) in percent; $\Delta$ rows give the comparison minus answer-only SFT in percentage points, computed from unrounded scores, with 95\% CIs and $p$-values. Significant gains ($p<0.05$) are shaded green and significant losses red. The Overall column repeats the paired analysis of Section~\ref{sec:controlled}.}
\label{tab:taskwise_boot_dcot}
\scriptsize
\setlength{\tabcolsep}{3.5pt}
\renewcommand{\arraystretch}{0.95}
\resizebox{\textwidth}{!}{%
\begin{tabular}{l|cc|cccccc|c}
\toprule
\textbf{Qwen3.5-4B} & \textbf{RDG} & \textbf{NRDG} & \textbf{MV} & \textbf{OT} & \textbf{VCG} & \textbf{VPG} & \textbf{CMG} & \textbf{RG} & \textbf{Overall} \\
$N$ & 1,000 & 1,000 & 1,148 & 1,000 & 1,000 & 1,000 & 2,000 & 1,482 & 9,630 \\
\midrule
Answer-only SFT & \mc{66.76}{81.00} & \mc{66.64}{81.40} & \mc{56.21}{63.24} & \mc{58.27}{66.40} & \mc{69.03}{78.40} & \mc{92.66}{99.70} & \mc{80.83}{85.10} & \mc{72.85}{83.47} & \mc{71.39}{80.31} \\
dCoT-SFT (ckpt.\ 2000) & \mc{82.86}{93.30} & \mc{81.30}{91.20} & \mc{54.01}{58.97} & \mc{64.77}{71.00} & \mc{70.64}{78.80} & \mc{90.21}{99.70} & \mc{83.39}{87.00} & \mc{69.67}{79.28} & \mc{74.96}{82.37} \\
\midrule
$\Delta$ mIoU & \cellcolor{green!12}+16.10 & \cellcolor{green!12}+14.67 & \cellcolor{red!10}$-$2.20 & \cellcolor{green!12}+6.50 & +1.61 & \cellcolor{red!10}$-$2.45 & \cellcolor{green!12}+2.56 & \cellcolor{red!10}$-$3.18 & \cellcolor{green!12}+3.56 \\
\quad 95\% CI & {\tiny[14.94, 17.28]} & {\tiny[13.54, 15.80]} & {\tiny[$-$3.53, $-$0.87]} & {\tiny[4.72, 8.29]} & {\tiny[$-$0.10, 3.30]} & {\tiny[$-$2.81, $-$2.11]} & {\tiny[1.81, 3.31]} & {\tiny[$-$4.25, $-$2.13]} & {\tiny[3.13, 4.00]} \\
\quad $p$ & {\tiny $<10^{-5}$} & {\tiny $<10^{-5}$} & {\tiny 0.0010} & {\tiny $<10^{-5}$} & {\tiny 0.0634} & {\tiny $<10^{-5}$} & {\tiny $<10^{-5}$} & {\tiny $<10^{-5}$} & {\tiny $<10^{-5}$} \\
\midrule
$\Delta$ Acc@0.5 & \cellcolor{green!12}+12.30 & \cellcolor{green!12}+9.80 & \cellcolor{red!10}$-$4.27 & \cellcolor{green!12}+4.60 & +0.40 & +0.00 & \cellcolor{green!12}+1.90 & \cellcolor{red!10}$-$4.18 & \cellcolor{green!12}+2.06 \\
\quad 95\% CI & {\tiny[10.20, 14.40]} & {\tiny[7.90, 11.80]} & {\tiny[$-$6.62, $-$1.92]} & {\tiny[1.80, 7.40]} & {\tiny[$-$2.00, 2.80]} & {\tiny[$-$0.40, 0.40]} & {\tiny[0.60, 3.20]} & {\tiny[$-$6.07, $-$2.36]} & {\tiny[1.36, 2.75]} \\
\quad $p$ & {\tiny $<10^{-5}$} & {\tiny $<10^{-5}$} & {\tiny 0.0004} & {\tiny 0.0013} & {\tiny 0.7772} & {\tiny 1.0000} & {\tiny 0.0041} & {\tiny $<10^{-5}$} & {\tiny $<10^{-5}$} \\
\bottomrule
\end{tabular}}
\end{table}

\begin{table}[!ht]
\centering
\caption{\textbf{Task-wise paired comparison: GRPO versus answer-only SFT (4B).} Paired example-level bootstrap on the 9,630 MedSG-Bench test questions (100,000 resamples per task). Cells show mIoU (top) and \textcolor{gray}{Acc@0.5} (bottom) in percent; $\Delta$ rows give the comparison minus answer-only SFT in percentage points, computed from unrounded scores, with 95\% CIs and $p$-values. Significant gains ($p<0.05$) are shaded green and significant losses red. An overall paired interval was not computed for this comparison (---).}
\label{tab:taskwise_boot_grpo}
\scriptsize
\setlength{\tabcolsep}{3.5pt}
\renewcommand{\arraystretch}{0.95}
\resizebox{\textwidth}{!}{%
\begin{tabular}{l|cc|cccccc|c}
\toprule
\textbf{Qwen3.5-4B} & \textbf{RDG} & \textbf{NRDG} & \textbf{MV} & \textbf{OT} & \textbf{VCG} & \textbf{VPG} & \textbf{CMG} & \textbf{RG} & \textbf{Overall} \\
$N$ & 1,000 & 1,000 & 1,148 & 1,000 & 1,000 & 1,000 & 2,000 & 1,482 & 9,630 \\
\midrule
Answer-only SFT & \mc{66.76}{81.00} & \mc{66.64}{81.40} & \mc{56.21}{63.24} & \mc{58.27}{66.40} & \mc{69.03}{78.40} & \mc{92.66}{99.70} & \mc{80.83}{85.10} & \mc{72.85}{83.47} & \mc{71.39}{80.31} \\
GRPO & \mc{83.82}{94.50} & \mc{83.21}{92.70} & \mc{54.80}{59.93} & \mc{67.90}{76.20} & \mc{66.09}{72.10} & \mc{89.43}{99.80} & \mc{83.59}{87.55} & \mc{70.72}{80.30} & \mc{75.32}{82.89} \\
\midrule
$\Delta$ mIoU & \cellcolor{green!12}+17.05 & \cellcolor{green!12}+16.57 & \cellcolor{red!10}$-$1.41 & \cellcolor{green!12}+9.63 & \cellcolor{red!10}$-$2.93 & \cellcolor{red!10}$-$3.24 & \cellcolor{green!12}+2.76 & \cellcolor{red!10}$-$2.12 & --- \\
\quad 95\% CI & {\tiny[15.88, 18.25]} & {\tiny[15.39, 17.77]} & {\tiny[$-$2.71, $-$0.10]} & {\tiny[7.93, 11.37]} & {\tiny[$-$4.65, $-$1.24]} & {\tiny[$-$3.57, $-$2.91]} & {\tiny[2.00, 3.51]} & {\tiny[$-$3.15, $-$1.11]} & --- \\
\quad $p$ & {\tiny $<10^{-5}$} & {\tiny $<10^{-5}$} & {\tiny 0.0341} & {\tiny $<10^{-5}$} & {\tiny 0.0007} & {\tiny $<10^{-5}$} & {\tiny $<10^{-5}$} & {\tiny $<10^{-5}$} & --- \\
\midrule
$\Delta$ Acc@0.5 & \cellcolor{green!12}+13.50 & \cellcolor{green!12}+11.30 & \cellcolor{red!10}$-$3.31 & \cellcolor{green!12}+9.80 & \cellcolor{red!10}$-$6.30 & +0.10 & \cellcolor{green!12}+2.45 & \cellcolor{red!10}$-$3.17 & --- \\
\quad 95\% CI & {\tiny[11.40, 15.70]} & {\tiny[9.30, 13.40]} & {\tiny[$-$5.75, $-$0.87]} & {\tiny[7.10, 12.50]} & {\tiny[$-$8.80, $-$3.90]} & {\tiny[$-$0.20, 0.50]} & {\tiny[1.20, 3.70]} & {\tiny[$-$5.06, $-$1.35]} & --- \\
\quad $p$ & {\tiny $<10^{-5}$} & {\tiny $<10^{-5}$} & {\tiny 0.0098} & {\tiny $<10^{-5}$} & {\tiny $<10^{-5}$} & {\tiny 0.7615} & {\tiny 0.0002} & {\tiny 0.0009} & --- \\
\bottomrule
\end{tabular}}
\end{table}

Table~\ref{tab:taskwise_boot_grpo9b} applies the same paired analysis to GRPO at 9B, comparing the GRPO checkpoint with its 9B dCoT-SFT starting point (checkpoint-2500). GRPO significantly improves multi-view ($+1.11$ mIoU, $p=0.0025$) and cross-modal grounding ($+0.67$ mIoU, $p=0.0018$), and no task shows a significant decrease. Concept grounding is unchanged ($+0.17$ mIoU, $p=0.74$), in contrast to the 4B decrease. The overall change is $+0.30$ mIoU and $+0.27$ Acc@0.5 (example-weighted means; no paired overall interval was computed).

\begin{table}[!ht]
\centering
\caption{\textbf{Task-wise paired comparison: GRPO versus dCoT-SFT (9B).} Paired example-level bootstrap on the 9,630 MedSG-Bench test questions (100,000 resamples per task). Cells show mIoU (top) and \textcolor{gray}{Acc@0.5} (bottom) in percent; $\Delta$ rows give GRPO minus the dCoT-SFT checkpoint-2500 in percentage points, computed from unrounded scores, with 95\% CIs and $p$-values. Significant gains ($p<0.05$) are shaded green and significant losses red. The Overall $\Delta$ is a point estimate from example-weighted means; a paired overall interval was not computed (---).}
\label{tab:taskwise_boot_grpo9b}
\scriptsize
\setlength{\tabcolsep}{3.5pt}
\renewcommand{\arraystretch}{0.95}
\resizebox{\textwidth}{!}{%
\begin{tabular}{l|cc|cccccc|c}
\toprule
\textbf{Qwen3.5-9B} & \textbf{RDG} & \textbf{NRDG} & \textbf{MV} & \textbf{OT} & \textbf{VCG} & \textbf{VPG} & \textbf{CMG} & \textbf{RG} & \textbf{Overall} \\
$N$ & 1,000 & 1,000 & 1,148 & 1,000 & 1,000 & 1,000 & 2,000 & 1,482 & 9,630 \\
\midrule
dCoT-SFT (ckpt.\ 2500) & \mc{84.00}{93.10} & \mc{83.02}{91.60} & \mc{55.87}{60.98} & \mc{67.46}{75.00} & \mc{74.64}{83.20} & \mc{91.22}{99.90} & \mc{84.16}{87.40} & \mc{70.77}{78.95} & \mc{76.60}{83.55} \\
GRPO & \mc{84.16}{93.20} & \mc{83.12}{91.40} & \mc{56.98}{62.02} & \mc{67.65}{74.90} & \mc{74.80}{82.80} & \mc{91.22}{99.80} & \mc{84.82}{88.25} & \mc{70.53}{79.22} & \mc{76.90}{83.82} \\
\midrule
$\Delta$ mIoU & +0.16 & +0.10 & \cellcolor{green!12}+1.11 & +0.19 & +0.17 & $-$0.00 & \cellcolor{green!12}+0.67 & $-$0.24 & +0.30 \\
\quad 95\% CI & {\tiny[$-$0.21, 0.51]} & {\tiny[$-$0.23, 0.40]} & {\tiny[0.39, 1.83]} & {\tiny[$-$0.49, 0.89]} & {\tiny[$-$0.84, 1.17]} & {\tiny[$-$0.07, 0.07]} & {\tiny[0.27, 1.09]} & {\tiny[$-$0.75, 0.26]} & --- \\
\quad $p$ & {\tiny 0.3642} & {\tiny 0.5353} & {\tiny 0.0025} & {\tiny 0.5806} & {\tiny 0.7443} & {\tiny 0.9362} & {\tiny 0.0018} & {\tiny 0.3490} & --- \\
\midrule
$\Delta$ Acc@0.5 & +0.10 & $-$0.20 & +1.05 & $-$0.10 & $-$0.40 & $-$0.10 & \cellcolor{green!12}+0.85 & +0.27 & +0.27 \\
\quad 95\% CI & {\tiny[$-$0.70, 0.90]} & {\tiny[$-$0.70, 0.30]} & {\tiny[$-$0.44, 2.53]} & {\tiny[$-$1.60, 1.50]} & {\tiny[$-$1.70, 0.90]} & {\tiny[$-$0.30, 0.00]} & {\tiny[0.15, 1.55]} & {\tiny[$-$0.81, 1.35]} & --- \\
\quad $p$ & {\tiny 0.8960} & {\tiny 0.5348} & {\tiny 0.1907} & {\tiny 0.9505} & {\tiny 0.6053} & {\tiny 0.6356} & {\tiny 0.0185} & {\tiny 0.6592} & --- \\
\bottomrule
\end{tabular}}
\end{table}

\subsection{GRPO on the Full Benchmark: Task-wise Effects and Hypotheses}
\label{app:grpo_main}

Figure~\ref{fig:semantic_drift} compares the fixed 4B dCoT-SFT checkpoint with the GRPO checkpoint on every task. The paired aggregate gains are $+0.367$ mIoU points (95\% CI $0.049$--$0.686$, $p=0.0247$) and $+0.519$ Acc@0.5 points ($-0.042$--$1.080$, $p=0.0746$). We therefore claim a small mIoU improvement, not a significant Acc@0.5 gain, and not a general failure of GRPO.

\noindent \textbf{Reward-variance hypothesis.} dCoT-SFT can produce low-diversity rollout groups, which reduces within-group reward variance and hence the GRPO advantage signal~\citep{shao2024deepseekmath, yu2025dapo}. The sampling diagnostic in Appendix~\ref{app:grpo_followup} shows that diversity is not absent on most training prompts, so this is at most a partial explanation.

\noindent \textbf{Cross-task imbalance hypothesis.} Starting mIoU spans 54 to 90 points across the eight tasks. Such heterogeneity motivates task-balanced sampling, but score differences alone do not show that easy tasks dominate the gradients. Our task-balanced follow-up did not produce a significant aggregate gain.

\noindent \textbf{Semantic drift.} The IoU+format reward improves most localization-style tasks but lowers concept grounding by $4.55$ mIoU and $6.70$ Acc@0.5. Spatial overlap rewards do not directly preserve the clinical identification that concept grounding requires~\citep{medrg, bannur2022biovil, medgroundr1}. These observations motivate hypotheses about exploration, task imbalance and semantic preservation but do not isolate their causal contributions.

\subsection{Controlled GRPO Development Studies}\label{app:C5}
\phantomsection\label{app:grpo_followup}

\noindent \textbf{Scope and preprocessing control.}
These experiments use a 1,385-example development set,
not the 9,630-example benchmark, and initialize from the
4B dCoT-SFT checkpoint-2000. An early GRPO export saved a
processor with a reduced image-resolution limit. A zero-learning-rate
save/reload control reproduced an apparent regression with that
processor, but exactly reproduced all 1,385 SFT predictions
with the original processor. The early 3--5-point regressions
therefore cannot be attributed to learned policy degradation.
Native-resolution retraining uses the SFT bounds of 65,536
minimum and 16,777,216 maximum pixels. Historical batch-8
and later batch-1 results are kept separate.

\noindent \textbf{Fixed-checkpoint replication across seeds.}
Native-resolution direct GRPO was evaluated at step 50,
fixed before evaluating the two additional seeds.
Under the historical batch-8 protocol, the three-seed
mean improves over SFT by only $+0.064$ mIoU and
$+0.241$ Acc@0.5 points (unrounded differences).

\begin{table}[H]
\centering\small
\begin{minipage}[t]{0.4\linewidth}\centering
\small
\caption{Fixed step-50 direct GRPO on the development set,
historical batch-8 evaluation. Values are percentages;
SD is the sample standard deviation across training seeds.}
\begin{adjustbox}{max width=\linewidth}\begin{tabular}{lrr}
\toprule
Run & mIoU & Acc@0.5 \\
\midrule
CoT-SFT & 78.038 & 86.065 \\
GRPO, seed 42 & 78.189 & 86.426 \\
GRPO, seed 43 & 78.088 & 86.137 \\
GRPO, seed 44 & 78.028 & 86.354 \\
Mean $\pm$ SD & $78.101\pm0.081$ & $86.306\pm0.150$ \\
\bottomrule
\end{tabular}\end{adjustbox}
\end{minipage}\hfill
\begin{minipage}[t]{0.56\linewidth}\centering
\small
\caption{Positive development point estimates with original
prompts. All rows use identical examples, the SFT processor,
and batch-1 evaluation. Step 100 is a secondary checkpoint;
step 200 was the preprocessing ablation's primary endpoint.}
\begin{adjustbox}{max width=\linewidth}\begin{tabular}{lrrrr}
\toprule
Model & mIoU & Acc@0.5 & $\Delta$ mIoU & $\Delta$ Acc \\
\midrule
CoT-SFT & 77.877 & 85.560 & --- & --- \\
No-DIoU, step 100 & 78.235 & 86.498 & +0.357 & +0.939 \\
No-DIoU, step 200 & 78.202 & 86.209 & +0.325 & +0.650 \\
\bottomrule
\end{tabular}\end{adjustbox}
\end{minipage}
\end{table}

Multi-view is the clearest positive task-level pattern:
mean mIoU increases from 53.589 to $58.038\pm0.625$
($+4.449$ points), and mean Acc@0.5 from 57.627 to
$64.972\pm1.957$ ($+7.345$ points). This exploratory
result covers only 59 development examples; other tasks
do not uniformly improve. It does not establish an overall
4--5-point gain or eliminate the historical batch-8 caveat.

\noindent \textbf{Task-balanced, threshold-aware GRPO.}
A follow-up uses 15,000 RL examples, 3,750 each from Concept,
Multi-view, Object Tracking, and Referring. The reward is IoU
plus $0.2$ times a sigmoid threshold reward centered at
IoU $0.5$ (temperature $0.05$), plus $0.01$ times the format
reward; the DIoU coefficient is zero. Settings are eight
rollouts, two policy iterations, centered within-group
advantages, task-balanced sampling, no replay, all-token
policy loss, frozen vision encoder, learning rate
$5\!\times\!10^{-7}$, cosine decay over 200 updates,
warmup ratio $0.05$, $\beta=0.04$, $\epsilon=0.2$, and
seed 42. Eight GPUs use per-device batch 2, accumulation 4,
and a 768-token completion limit. Evaluation uses batch 1.

For step 100 versus SFT, 100,000 paired example-bootstrap
resamples yield $+0.357$ mIoU points (95\% CI
$[-0.164,0.886]$, $p=0.182$) and $+0.939$ Acc@0.5
points ($[-0.072,1.949]$, $p=0.085$). Neither aggregate
gain is significant at $0.05$. These intervals quantify
example sampling, not repeated-training uncertainty.

\noindent \textbf{Prompt sensitivity and a negative control.}
Removing literal image-marker text, while retaining image
inputs, changes SFT to 78.212/86.570 and step-200 GRPO to
78.196/86.282. The step-200 advantage therefore does not
persist under this condition. A cleaned-prompt step-100
comparison is not yet available. A separate 16-update
mixed-success-weighting pilot on 507 fixed development
examples also does not beat SFT (75.434/84.221 versus
75.667/85.010). These qualifications prevent selecting
only favorable checkpoints or evaluation conventions.

\noindent \textbf{Sampling headroom, not deployable accuracy.}
On 128 training-only diagnostic prompts (32 per hard task),
eight stochastic completions yield sampled pass@1 67.676\%
and pass@8 84.375\%, with success defined as IoU $\geq0.5$.
Multi-view yields 46.484\%/62.500\%, and Object Tracking
67.969\%/93.750\%. Sampled pass@1 averages the eight
completions, rather than greedy decoding. These are neither
held-out benchmark results nor a deployable oracle selector.

\noindent \textbf{Inference repeatability.}
On 54 outcome-selected diagnostic examples (38 changed
cases and 16 unchanged controls), repeated greedy inference
with one fixed GRPO checkpoint yields zero text, box, or
Acc@0.5 changes, both on the same node and across two nodes.
This supports repeatability only under those settings,
not clinical correctness or training-seed reproducibility.

\noindent \textbf{Full-pass experiment status.}
The later 30,000-example, eight-task no-DIoU run is not
reported as a completed result. No performance claim is
made before its training and benchmark evaluation finish.

\FloatBarrier
\section{Extended Discussion of Findings}
\label{app:discussion}

This section collects clarifications and analyses that support the main text but did not fit within the page limit. All numbers are taken from results reported elsewhere in the paper; no new experiments are introduced here.

\subsection{Design of the two-stage pipeline}
\label{app:disc_bridging}

MedChain separates \emph{seeing} from \emph{reasoning}. In Stage~1, FlemingVL-38B, a medical vision-language model, describes the modality, anatomy and visible findings of each image. These descriptions are medically accurate but read like captions: each image is described on its own, and the relation between images, which is what the grounding question asks about, is rarely stated. In Stage~2, Qwen3.5-397B-A17B receives the original image sequence together with the question, task context, annotated box and the Stage~1 description, and writes the final rationale. Because the refiner inspects the images directly, it can state the cross-image comparison that decides the answer, for example a density present in one scan and absent in its prior, or a structure that keeps its position relative to the mediastinum across slices, and it can add brief self-checks such as ruling out a windowing artifact (Figure~\ref{fig:qual_case1}). Every statement in the final rationale therefore remains tied to visible evidence, while the reasoning itself comes from the stage that can see the whole sequence.

\subsection{Reading the task-level pattern}
\label{app:disc_tasks}

Table~\ref{tab:task_synthesis} brings together, for each task, the effect of reasoning supervision (dCoT-SFT minus answer-only SFT), the two inference-time interventions on the dCoT-SFT checkpoint, and the effect of GRPO on the full benchmark. All values are mIoU differences in percentage points on the 9,630-example test set, computed from the rounded entries in Tables~\ref{tab:fixed_4b} and~\ref{tab:reasoning_controls_taskwise} and Figure~\ref{fig:semantic_drift}.

\begin{table}[H]
\centering
\caption{Per-task synthesis of the 4B analyses (mIoU, percentage points). \emph{Training effect}: dCoT-SFT minus answer-only SFT. \emph{No-think} and \emph{Cross-task}: change when the dCoT-SFT checkpoint is forced to answer immediately or is given a rationale from another task. \emph{GRPO}: GRPO minus dCoT-SFT.}
\label{tab:task_synthesis}
\small
\setlength{\tabcolsep}{5pt}
\begin{tabular}{@{}lrrrr@{}}
\toprule
\textbf{Task} & \textbf{Training effect} & \textbf{No-think} & \textbf{Cross-task} & \textbf{GRPO} \\
\midrule
Registered difference (RDG) & $+16.10$ & $-0.71$ & $-17.67$ & $+0.96$ \\
Non-registered difference (NRDG) & $+14.66$ & $-0.36$ & $-1.14$ & $+1.91$ \\
Multi-view (MV) & $-2.20$ & $-4.29$ & $-14.46$ & $+0.79$ \\
Object tracking (OT) & $+6.50$ & $-9.51$ & $-18.03$ & $+3.13$ \\
Concept (VCG) & $+1.61$ & $-1.24$ & $-5.52$ & $-4.55$ \\
Patch (VPG) & $-2.45$ & $-2.66$ & $-4.18$ & $-0.78$ \\
Cross-modal (CMG) & $+2.56$ & $-1.95$ & $-17.61$ & $+0.20$ \\
Referring (RG) & $-3.18$ & $-3.31$ & $-12.79$ & $+1.05$ \\
\midrule
Overall & $+3.56$ & $-2.93$ & $-12.18$ & $+0.37$ \\
\bottomrule
\end{tabular}
\end{table}

Several observations follow from this view. We state them as hypotheses, since each rests on single training runs.

\textbf{Object tracking depends on reasoning at every stage.} It gains from reasoning supervision, is the task most harmed by removing the reasoning ($-9.51$) or replacing it ($-18.03$), and gains most from GRPO ($+3.13$). Tracking a structure across slices requires explicitly carrying anatomical context from one image to the next, which is what the rationales describe.

\textbf{Training benefit and inference-time reliance can dissociate.} Non-registered difference grounding shows one of the largest training effects ($+14.66$) but is almost unaffected by removing ($-0.36$) or replacing ($-1.14$) the rationale. Registered difference grounding, by contrast, is strongly affected by a mismatched rationale ($-17.67$). One reading is that for non-registered pairs, reasoning supervision acts mainly as a training signal that shapes the model's visual representation, rather than as a computation the model performs at inference time. A second, not exclusive, reading is that the answer-only baseline is particularly weak on difference grounding (Appendix~\ref{app:disc_baseline}), which inflates the measured training effect. Distinguishing these explanations requires matched multi-seed runs.

\textbf{Some tasks rely on reasoning that does not help them.} Multi-view and referring grounding are both hurt by reasoning supervision ($-2.20$ and $-3.18$), yet the trained model depends heavily on its rationale for these tasks (cross-task replacement costs $-14.46$ and $-12.79$). The model uses the reasoning, but the reasoning it learned for these tasks appears less useful than direct localization. This is consistent with the rationales being less informative for these tasks, for example because correspondences between mammographic views or between a phrase and a small region are harder to verbalize than a change between two scans. Multi-view is also the task where development-set GRPO produced its only consistent gain (Appendix~\ref{app:grpo_followup}), which suggests that reward optimization can partly correct reasoning that supervised rationales teach imperfectly.

\textbf{Concept grounding is where the overlap reward fails.} Concept grounding gains little from reasoning supervision, depends only moderately on its rationale, and is the one task that GRPO clearly harms ($-4.55$). This task requires recognizing the same lesion or structure across very different images; an IoU reward can be increased by boxes that overlap the target without identifying it, which is the failure mode discussed in Section~\ref{sec:grpo}.

\subsection{The answer-only baseline and MedSeq-Grounder}
\label{app:disc_baseline}

The answer-only 4B model scores 66.76 and 66.64 mIoU on registered and non-registered difference grounding, well below MedSeq-Grounder (83.29 and 83.72), which is also trained without rationales. The two answer-only models differ in backbone (Qwen3.5-4B versus Qwen2.5-VL-7B), training data (the MedChain SFT split versus MedSG-188K) and training schedule, so their gap is not attributable to any single factor. The practical consequence is that the $+14.7$ to $+16.1$ mIoU training effect on difference grounding in Table~\ref{tab:fixed_4b} measures the benefit of reasoning supervision \emph{relative to our answer-only run}, and may overstate the benefit relative to a stronger answer-only model. Two results bound this concern. First, the 4B dCoT-SFT model comes within 2.5 mIoU of MedSeq-Grounder on difference grounding despite the smaller backbone, and exceeds it on four of the eight tasks and on average. Second, continuing dCoT-SFT from the MedSeq-Grounder checkpoint itself raises it from 72.55 to 75.71 mIoU (Appendix~\ref{app:C2}), so reasoning supervision still helps when the starting point is the strongest answer-only model.

\subsection{Model scale}
\label{app:disc_scale}

Average performance increases monotonically with model size (74.96, 76.60 and 77.24 mIoU at 4B, 9B and 27B). The per-task picture is less uniform: from 9B to 27B, consistency tasks improve or stay level while both difference tasks decline by about 2 mIoU. Because the 27B model was trained for six epochs with a smaller warmup and no weight decay, and reached a higher validation loss than the 9B model (Table~\ref{tab:hyperparams}), this pattern should not be read as a property of scale alone. A matched schedule across sizes would be needed to separate the effect of capacity from the effect of optimization.

\subsection{Why GRPO adds little on top of dCoT-SFT}
\label{app:disc_grpo}

GRPO learns from differences in reward among completions sampled for the same prompt. After dCoT-SFT, the policy already solves most prompts on most samples, so many groups are likely to contain several completions with similar IoU and therefore provide weak contrast. At the same time, the sampling diagnostic shows real headroom: at least one of eight samples succeeds on 84.4\% of hard training prompts while the average sample succeeds on 67.7\%. That this headroom does not translate into benchmark gains suggests that the bottleneck is not the absence of good samples but the signal used to prefer them. An IoU reward cannot distinguish a box that overlaps the target by chance from one that identifies the correct finding, which is consistent with the concept-grounding decline. Rewards that also score semantic correctness, or that reward the image-selection step separately from localization, are natural next steps.

The development studies also carry a methodological lesson. An apparent 3 to 5 point regression after GRPO was caused by an exported image processor with a lower resolution limit, not by learning. A zero-learning-rate save-and-reload control exposed the problem immediately. We recommend such a control whenever a fine-tuned vision-language model is exported and re-evaluated, since preprocessing mismatches can easily be mistaken for training effects of the same size as the effects being studied.

\subsection{What box conditioning implies}
\label{app:disc_conditioning}

Both stages of the pipeline see the annotated box. This makes the rationales explanations of a known answer: they are aligned with the correct region by construction, but they do not record how a model would search for it. We chose this design so that every rationale is aligned with the correct region; a rationale written without the answer could justify an incorrect region and teach the student to reason fluently toward wrong answers. The cost is a risk of post-hoc rationalization. The inference-time controls address the part of this concern that matters for the trained model: its localization changes substantially when the rationale is removed or replaced, so the rationale is not ignored. They do not show that the rationale faithfully reflects the model's internal computation, and they cannot rule out that some of the text is decorative. Rationales generated without the answer and filtered by correctness, and multi-rater clinical review, would strengthen the corpus in future versions.

\subsection{Practical takeaways}
\label{app:disc_takeaways}

For practitioners building grounding models for multi-image clinical data, our results suggest four points. First, reasoning supervision is most valuable when the target is defined by a relation between images, such as change or correspondence across slices. Second, refinement should keep access to the images rather than relying on a caption. Third, gains should be checked per task, since reasoning supervision can hurt appearance-matching tasks while improving the average. Fourth, reward optimization after reasoning supervision should be evaluated with matched preprocessing, multiple seeds and rewards that credit correct identification, not only overlap.

\FloatBarrier
\section{Limitations}\label{app:D}

First, both generation stages inherit model blind spots,
and the annotated box conditions the final explanation;
the rationales may therefore be post-hoc and should not
be interpreted as faithful internal reasoning. The
single-rater audit covers only 2,000 examples and lacks
item-level labels, specialty metadata, error-category
counts, and inter-rater agreement. Second, the 4B
answer-only and dCoT-SFT runs differ in training stage,
effective batch, optimizer implementation, and checkpoint
selection. Their paired test analysis is a fixed-model
comparison, not a causal or multi-seed estimate. Third,
the SFT and test sets share image content in 147 test
examples and a task-aware source series in 17; although
no complete sequence or series-plus-box duplicates and a
leakage-clean sensitivity analysis are reported, strict
image- or study-level disjointness is not established.
Fourth, evaluation is limited to MedSG-Bench and the CoT
benefit is negative on several tasks. Finally, the GRPO
findings apply only to the verified IoU+format reward and
optimization configuration; they do not establish that
GRPO is generally ineffective. The no-think and corrupted-
reasoning controls are inference-time distribution shifts
on one SFT checkpoint and do not by themselves establish
rationale faithfulness or training-run causality. The tasks where reasoning supervision does not yet help behave very differently from one another, and a single IoU-based reward does not supervise them well. Future
work should use
matched multi-seed controls, patient/study-level splits,
multi-specialist audits, and richer rewards.

\FloatBarrier
\section{Broader Impacts}\label{app:E}

\method{} and MedChain-182K are intended for
research use in medical AI. Positive impacts
include improved interpretability of medical image
grounding through explicit clinical reasoning
chains~\cite{wei2022chain}, which could support
radiologist workflows in longitudinal scan review,
multi-view analysis, and cross-modal examination.
Negative risks include potential over-reliance on
model predictions in clinical settings without
appropriate human oversight, and the possibility
that errors in pseudo-CoT generation
(Appendix~\ref{app:A2}) propagate into model behavior.
We strongly caution against deployment without
rigorous clinical validation. MedChain-182K is
constructed from publicly available benchmark
data~\cite{medsgbench} and does not contain
identifiable patient information.

\FloatBarrier
\makeatletter\setlength{\@fptop}{0pt plus 1fil}\setlength{\@fpbot}{0pt plus 1fil}\makeatother
\section{Qualitative Case Studies}
\label{app:qual}
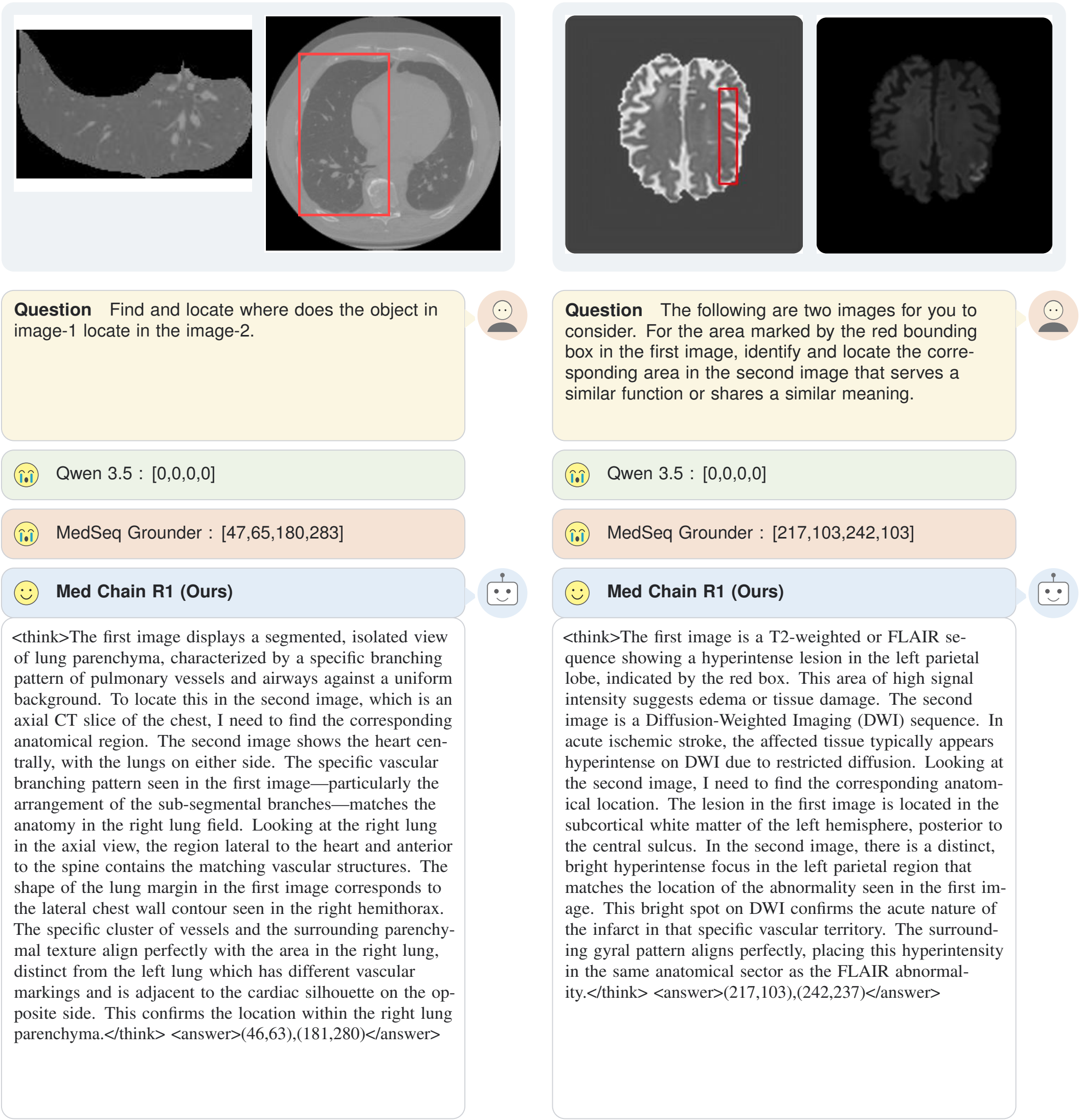
\begin{figure*}[!ht]
    \centering
    \begingroup
\definecolor{qInk}{HTML}{252629}\definecolor{qBlue}{HTML}{3B73A7}
\definecolor{qLine}{HTML}{B9BEC5}
\definecolor{qCream}{HTML}{FBF6E2}
\definecolor{qSage}{HTML}{EDF3E7}
\definecolor{qPeach}{HTML}{F5E3D5}
\definecolor{qSky}{HTML}{E3EDF7}
\definecolor{qBack}{HTML}{EEF2F5}
\resizebox{\textwidth}{!}{%
\begin{tikzpicture}[x=1mm,y=-1mm,every node/.style={font=\sffamily\fontsize{8}{9.5}\selectfont,text=qInk},box/.style={anchor=north west,align=left,inner sep=2mm,outer sep=0pt,rounded corners=2mm,draw=qLine!70,line width=.45pt,text width=70mm}]
\begin{scope}[shift={(0,0)}]
\fill[qBack,rounded corners=2mm] (0,0) rectangle (82,43);
\node[anchor=north,inner sep=0pt] at (21,2){\adjincludegraphics[trim={0.01\width} {0.695\height} {0.8784\width} {0.01\height},clip,width=38mm,height=38mm,keepaspectratio]{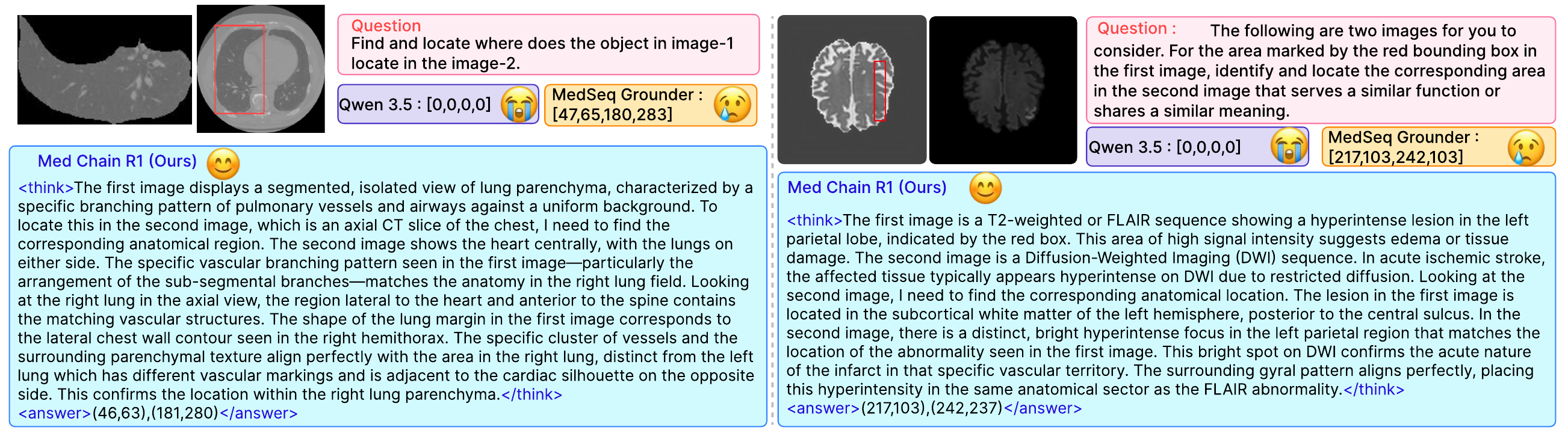}};
\node[anchor=north,inner sep=0pt] at (61,2){\adjincludegraphics[trim={0.125\width} {0.695\height} {0.792\width} {0.01\height},clip,width=38mm,height=38mm,keepaspectratio]{sample.pdf}};
\node[box,fill=qCream] (question0) at (0,46){\parbox[t][20mm][t]{70mm}{\textbf{Question}\quad Find and locate where does the object in image-1 locate in the image-2.}};
\node[box,fill=qSage] (qwen0) at ([yshift=-1.5mm]question0.south west){\tikz[x=1mm,y=-1mm,baseline=-.7mm]{\filldraw[fill=yellow!55!white,draw=qInk!80,line width=.35pt] (0,0) circle (1.9mm);\draw[qInk,line width=.35pt,line cap=round] (-1.15,-.35) .. controls (-.95,-.8) and (-.5,-.8) .. (-.3,-.35) (.3,-.35) .. controls (.5,-.8) and (.95,-.8) .. (1.15,-.35);\fill[qBlue!65!cyan,rounded corners=.12mm] (-1,-.2) rectangle (-.65,1.25);\fill[qBlue!65!cyan,rounded corners=.12mm] (.65,-.2) rectangle (1,1.25);\fill[qInk!85] (0,.8) ellipse (.3mm and .42mm);}\quad Qwen 3.5 : [0,0,0,0]};
\node[box,fill=qPeach] (med0) at ([yshift=-1.5mm]qwen0.south west){\tikz[x=1mm,y=-1mm,baseline=-.7mm]{\filldraw[fill=yellow!55!white,draw=qInk!80,line width=.35pt] (0,0) circle (1.9mm);\draw[qInk,line width=.35pt,line cap=round] (-1.15,-.35) .. controls (-.95,-.8) and (-.5,-.8) .. (-.3,-.35) (.3,-.35) .. controls (.5,-.8) and (.95,-.8) .. (1.15,-.35);\fill[qBlue!65!cyan,rounded corners=.12mm] (-1,-.2) rectangle (-.65,1.25);\fill[qBlue!65!cyan,rounded corners=.12mm] (.65,-.2) rectangle (1,1.25);\fill[qInk!85] (0,.8) ellipse (.3mm and .42mm);}\quad MedSeq Grounder : [47,65,180,283]};
\node[box,fill=qSky] (ours0) at ([yshift=-1.5mm]med0.south west){\tikz[x=1mm,y=-1mm,baseline=-.7mm]{\filldraw[fill=yellow!55!white,draw=qInk!80,line width=.35pt] (0,0) circle (1.9mm);\fill[qInk] (-.65,-.45) circle (.18mm);\fill[qInk] (.65,-.45) circle (.18mm);\draw[qInk,line width=.45pt,line cap=round] (-.85,.35) .. controls (-.45,1.2) and (.45,1.2) .. (.85,.35);}\quad\textbf{Med Chain R1 (Ours)}};
\node[box,fill=white,font=\rmfamily\fontsize{8}{9.5}\selectfont] at ([yshift=-0mm]ours0.south west){\parbox[t][76mm][t]{70mm}{\textless{}think\textgreater{}The first image displays a segmented, isolated view of lung parenchyma, characterized by a specific branching pattern of pulmonary vessels and airways against a uniform background. To locate this in the second image, which is an axial CT slice of the chest, I need to find the corresponding anatomical region. The second image shows the heart centrally, with the lungs on either side. The specific vascular branching pattern seen in the first image---particularly the arrangement of the sub-segmental branches---matches the anatomy in the right lung field. Looking at the right lung in the axial view, the region lateral to the heart and anterior to the spine contains the matching vascular structures. The shape of the lung margin in the first image corresponds to the lateral chest wall contour seen in the right hemithorax. The specific cluster of vessels and the surrounding parenchymal texture align perfectly with the area in the right lung, distinct from the left lung which has different vascular markings and is adjacent to the cardiac silhouette on the opposite side. This confirms the location within the right lung parenchyma.\textless{}/think\textgreater{} \textless{}answer\textgreater{}(46,63),(181,280)\textless{}/answer\textgreater{}}};

\fill[qCream] ([yshift=-3mm]question0.north east)--++(2mm,-1.5mm)--++(-2mm,-1.5mm)--cycle;
\fill[qSky] ([yshift=-3mm]ours0.north east)--++(2mm,-1.5mm)--++(-2mm,-1.5mm)--cycle;
\begin{scope}[shift={([xshift=6mm,yshift=-4mm]question0.north east)},scale=1.2]
\filldraw[fill=qPeach,draw=white,line width=1.2pt] (0,0) circle (3.5mm);
\fill[qInk!65] (-2,2.2) .. controls (-1.7,.5) and (1.7,.5) .. (2,2.2)--cycle;
\filldraw[fill=qCream,draw=qInk!65,line width=.35pt] (0,-.7) circle (1.25mm);
\draw[qInk!65,line width=.45pt] (-1.15,-1.15) .. controls (-.7,-2.25) and (.7,-2.25) .. (1.15,-1.15);
\fill[qInk] (-.4,-.7) circle (.12mm); \fill[qInk] (.4,-.7) circle (.12mm);
\end{scope}
\begin{scope}[shift={([xshift=6mm,yshift=-4mm]ours0.north east)},scale=1.2]
\filldraw[fill=qSky,draw=white,line width=1.2pt] (0,0) circle (3.5mm);
\draw[qInk!65,line width=.45pt] (0,-1.5)--(0,-2.25);
\fill[qInk!65] (0,-2.4) circle (.3mm);
\filldraw[fill=white,draw=qInk!65,rounded corners=.65mm,line width=.45pt] (-2,-1.45) rectangle (2,1.55);
\fill[qInk!75] (-.8,-.25) circle (.3mm); \fill[qInk!75] (.8,-.25) circle (.3mm);
\draw[qInk!65,line width=.4pt] (-.75,.65) .. controls (-.25,1.2) and (.25,1.2) .. (.75,.65);
\end{scope}
\end{scope}
\begin{scope}[shift={(88,0)}]
\fill[qBack,rounded corners=2mm] (0,0) rectangle (82,43);
\node[anchor=north,inner sep=0pt] at (21,2){\adjincludegraphics[trim={0.496\width} {0.626\height} {0.409\width} {0.034\height},clip,width=38mm,height=38mm,keepaspectratio]{sample.pdf}};
\node[anchor=north,inner sep=0pt] at (61,2){\adjincludegraphics[trim={0.593\width} {0.626\height} {0.313\width} {0.034\height},clip,width=38mm,height=38mm,keepaspectratio]{sample.pdf}};
\node[box,fill=qCream] (question1) at (0,46){\parbox[t][20mm][t]{70mm}{\textbf{Question}\quad The following are two images for you to consider. For the area marked by the red bounding box in the first image, identify and locate the corresponding area in the second image that serves a similar function or shares a similar meaning.}};
\node[box,fill=qSage] (qwen1) at ([yshift=-1.5mm]question1.south west){\tikz[x=1mm,y=-1mm,baseline=-.7mm]{\filldraw[fill=yellow!55!white,draw=qInk!80,line width=.35pt] (0,0) circle (1.9mm);\draw[qInk,line width=.35pt,line cap=round] (-1.15,-.35) .. controls (-.95,-.8) and (-.5,-.8) .. (-.3,-.35) (.3,-.35) .. controls (.5,-.8) and (.95,-.8) .. (1.15,-.35);\fill[qBlue!65!cyan,rounded corners=.12mm] (-1,-.2) rectangle (-.65,1.25);\fill[qBlue!65!cyan,rounded corners=.12mm] (.65,-.2) rectangle (1,1.25);\fill[qInk!85] (0,.8) ellipse (.3mm and .42mm);}\quad Qwen 3.5 : [0,0,0,0]};
\node[box,fill=qPeach] (med1) at ([yshift=-1.5mm]qwen1.south west){\tikz[x=1mm,y=-1mm,baseline=-.7mm]{\filldraw[fill=yellow!55!white,draw=qInk!80,line width=.35pt] (0,0) circle (1.9mm);\draw[qInk,line width=.35pt,line cap=round] (-1.15,-.35) .. controls (-.95,-.8) and (-.5,-.8) .. (-.3,-.35) (.3,-.35) .. controls (.5,-.8) and (.95,-.8) .. (1.15,-.35);\fill[qBlue!65!cyan,rounded corners=.12mm] (-1,-.2) rectangle (-.65,1.25);\fill[qBlue!65!cyan,rounded corners=.12mm] (.65,-.2) rectangle (1,1.25);\fill[qInk!85] (0,.8) ellipse (.3mm and .42mm);}\quad MedSeq Grounder : [217,103,242,103]};
\node[box,fill=qSky] (ours1) at ([yshift=-1.5mm]med1.south west){\tikz[x=1mm,y=-1mm,baseline=-.7mm]{\filldraw[fill=yellow!55!white,draw=qInk!80,line width=.35pt] (0,0) circle (1.9mm);\fill[qInk] (-.65,-.45) circle (.18mm);\fill[qInk] (.65,-.45) circle (.18mm);\draw[qInk,line width=.45pt,line cap=round] (-.85,.35) .. controls (-.45,1.2) and (.45,1.2) .. (.85,.35);}\quad\textbf{Med Chain R1 (Ours)}};
\node[box,fill=white,font=\rmfamily\fontsize{8}{9.5}\selectfont] at ([yshift=-0mm]ours1.south west){\parbox[t][76mm][t]{70mm}{\textless{}think\textgreater{}The first image is a T2-weighted or FLAIR sequence showing a hyperintense lesion in the left parietal lobe, indicated by the red box. This area of high signal intensity suggests edema or tissue damage. The second image is a Diffusion-Weighted Imaging (DWI) sequence. In acute ischemic stroke, the affected tissue typically appears hyperintense on DWI due to restricted diffusion. Looking at the second image, I need to find the corresponding anatomical location. The lesion in the first image is located in the subcortical white matter of the left hemisphere, posterior to the central sulcus. In the second image, there is a distinct, bright hyperintense focus in the left parietal region that matches the location of the abnormality seen in the first image. This bright spot on DWI confirms the acute nature of the infarct in that specific vascular territory. The surrounding gyral pattern aligns perfectly, placing this hyperintensity in the same anatomical sector as the FLAIR abnormality.\textless{}/think\textgreater{} \textless{}answer\textgreater{}(217,103),(242,237)\textless{}/answer\textgreater{}}};

\fill[qCream] ([yshift=-3mm]question1.north east)--++(2mm,-1.5mm)--++(-2mm,-1.5mm)--cycle;
\fill[qSky] ([yshift=-3mm]ours1.north east)--++(2mm,-1.5mm)--++(-2mm,-1.5mm)--cycle;
\begin{scope}[shift={([xshift=6mm,yshift=-4mm]question1.north east)},scale=1.2]
\filldraw[fill=qPeach,draw=white,line width=1.2pt] (0,0) circle (3.5mm);
\fill[qInk!65] (-2,2.2) .. controls (-1.7,.5) and (1.7,.5) .. (2,2.2)--cycle;
\filldraw[fill=qCream,draw=qInk!65,line width=.35pt] (0,-.7) circle (1.25mm);
\draw[qInk!65,line width=.45pt] (-1.15,-1.15) .. controls (-.7,-2.25) and (.7,-2.25) .. (1.15,-1.15);
\fill[qInk] (-.4,-.7) circle (.12mm); \fill[qInk] (.4,-.7) circle (.12mm);
\end{scope}
\begin{scope}[shift={([xshift=6mm,yshift=-4mm]ours1.north east)},scale=1.2]
\filldraw[fill=qSky,draw=white,line width=1.2pt] (0,0) circle (3.5mm);
\draw[qInk!65,line width=.45pt] (0,-1.5)--(0,-2.25);
\fill[qInk!65] (0,-2.4) circle (.3mm);
\filldraw[fill=white,draw=qInk!65,rounded corners=.65mm,line width=.45pt] (-2,-1.45) rectangle (2,1.55);
\fill[qInk!75] (-.8,-.25) circle (.3mm); \fill[qInk!75] (.8,-.25) circle (.3mm);
\draw[qInk!65,line width=.4pt] (-.75,.65) .. controls (-.25,1.2) and (.25,1.2) .. (.75,.65);
\end{scope}
\end{scope}
\end{tikzpicture}%
}
\endgroup
    \caption{
        \textbf{Qualitative results of \method{} on sequential medical image grounding.}
        Two representative examples from MedSG-Bench: object tracking across CT slices
        (\textit{left}) and cross-modal consistency grounding between FLAIR and DWI MRI
        sequences (\textit{right}). For each example, zero-shot Qwen3.5 produces degenerate
        outputs ($[0,0,0,0]$), MedSeq-Grounder predicts plausible but imprecise boxes, and
        \method{} generates structured clinical reasoning chains (\texttt{<think>}) that
        identify the correct anatomical correspondence before predicting precise bounding
        box coordinates (\texttt{<answer>}). The reasoning traces demonstrate cross-image
        analysis vascular pattern matching in CT and ischemic lesion localization via
        DWI hyperintensity in MRI  that is absent in both baselines.
    }
    \label{fig:sample}
\end{figure*}

\begin{figure*}[p]
\centering
\begingroup
\definecolor{qInk}{HTML}{252629}\definecolor{qBlue}{HTML}{3B73A7}
\definecolor{qLine}{HTML}{B9BEC5}
\definecolor{qCream}{HTML}{FBF6E2}
\definecolor{qSage}{HTML}{EDF3E7}
\definecolor{qPeach}{HTML}{F5E3D5}
\definecolor{qSky}{HTML}{E3EDF7}
\definecolor{qBack}{HTML}{EEF2F5}
\resizebox{\textwidth}{!}{%
\begin{tikzpicture}[x=1mm,y=-1mm,every node/.style={font=\sffamily\fontsize{8.5}{10.2}\selectfont,text=qInk},box/.style={anchor=north west,align=left,inner sep=2mm,outer sep=0pt,rounded corners=2mm,draw=qLine!70,line width=.45pt,text width=164mm}]

\fill[qBack,rounded corners=2mm] (0,0) rectangle (170,86);
\node[anchor=north,inner sep=0pt] at (43,3){\adjincludegraphics[trim={0.018\width} {0.03\height} {0.755\width} {0.03\height},clip,width=80mm,height=80mm,keepaspectratio]{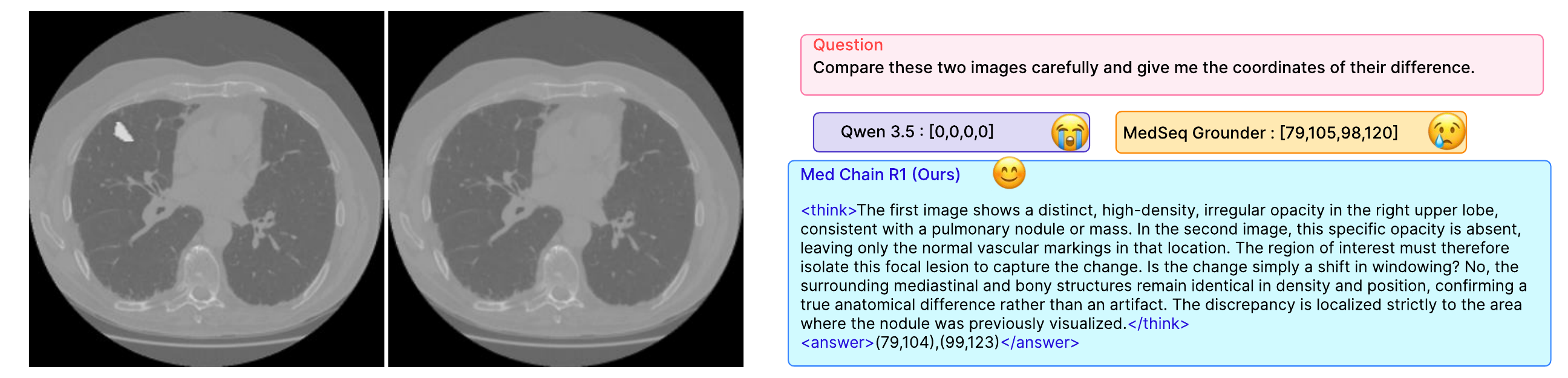}};
\node[anchor=north,inner sep=0pt] at (127,3){\adjincludegraphics[trim={0.248\width} {0.03\height} {0.525\width} {0.03\height},clip,width=80mm,height=80mm,keepaspectratio]{qual_case1.pdf}};
\node[box,fill=qCream] (question) at (0,90){\textbf{Question}\quad Compare these two images carefully and give me the coordinates of their difference.};
\node[box,fill=qSage] (qwen) at ([yshift=-2mm]question.south west){\tikz[x=1mm,y=-1mm,baseline=-.7mm]{\filldraw[fill=yellow!55!white,draw=qInk!80,line width=.35pt] (0,0) circle (1.9mm);\draw[qInk,line width=.35pt,line cap=round] (-1.15,-.35) .. controls (-.95,-.8) and (-.5,-.8) .. (-.3,-.35) (.3,-.35) .. controls (.5,-.8) and (.95,-.8) .. (1.15,-.35);\fill[qBlue!65!cyan,rounded corners=.12mm] (-1,-.2) rectangle (-.65,1.25);\fill[qBlue!65!cyan,rounded corners=.12mm] (.65,-.2) rectangle (1,1.25);\fill[qInk!85] (0,.8) ellipse (.3mm and .42mm);}\quad Qwen 3.5 : [0,0,0,0]};
\node[box,fill=qPeach] (medseq) at ([yshift=-1.5mm]qwen.south west){\tikz[x=1mm,y=-1mm,baseline=-.7mm]{\filldraw[fill=yellow!55!white,draw=qInk!80,line width=.35pt] (0,0) circle (1.9mm);\draw[qInk,line width=.35pt,line cap=round] (-1.15,-.35) .. controls (-.95,-.8) and (-.5,-.8) .. (-.3,-.35) (.3,-.35) .. controls (.5,-.8) and (.95,-.8) .. (1.15,-.35);\fill[qBlue!65!cyan,rounded corners=.12mm] (-1,-.2) rectangle (-.65,1.25);\fill[qBlue!65!cyan,rounded corners=.12mm] (.65,-.2) rectangle (1,1.25);\fill[qInk!85] (0,.8) ellipse (.3mm and .42mm);}\quad MedSeq Grounder : [79,105,98,120]};
\node[box,fill=qSky] (ours) at ([yshift=-2mm]medseq.south west){\tikz[x=1mm,y=-1mm,baseline=-.7mm]{\filldraw[fill=yellow!55!white,draw=qInk!80,line width=.35pt] (0,0) circle (1.9mm);\fill[qInk] (-.65,-.45) circle (.18mm);\fill[qInk] (.65,-.45) circle (.18mm);\draw[qInk,line width=.45pt,line cap=round] (-.85,.35) .. controls (-.45,1.2) and (.45,1.2) .. (.85,.35);}\quad\textbf{Med Chain R1 (Ours)}};
\node[box,fill=white,rounded corners=2mm,font=\rmfamily\fontsize{9.2}{11}\selectfont] (trace) at ([yshift=-0mm]ours.south west){\textless{}think\textgreater{}The first image shows a distinct, high-density, irregular opacity in the right upper lobe, consistent with a pulmonary nodule or mass. In the second image, this specific opacity is absent, leaving only the normal vascular markings in that location. The region of interest must therefore isolate this focal lesion to capture the change. Is the change simply a shift in windowing? No, the surrounding mediastinal and bony structures remain identical in density and position, confirming a true anatomical difference rather than an artifact. The discrepancy is localized strictly to the area where the nodule was previously visualized.\textless{}/think\textgreater{} \textless{}answer\textgreater{}(79,104),(99,123)\textless{}/answer\textgreater{}};

\fill[qCream] ([yshift=-3mm]question.north east)--++(2mm,-1.5mm)--++(-2mm,-1.5mm)--cycle;
\fill[qSky] ([yshift=-3mm]ours.north east)--++(2mm,-1.5mm)--++(-2mm,-1.5mm)--cycle;
\begin{scope}[shift={([xshift=6mm,yshift=-4mm]question.north east)},scale=1.2]
\filldraw[fill=qPeach,draw=white,line width=1.2pt] (0,0) circle (3.5mm);
\fill[qInk!65] (-2,2.2) .. controls (-1.7,.5) and (1.7,.5) .. (2,2.2)--cycle;
\filldraw[fill=qCream,draw=qInk!65,line width=.35pt] (0,-.7) circle (1.25mm);
\draw[qInk!65,line width=.45pt] (-1.15,-1.15) .. controls (-.7,-2.25) and (.7,-2.25) .. (1.15,-1.15);
\fill[qInk] (-.4,-.7) circle (.12mm); \fill[qInk] (.4,-.7) circle (.12mm);
\end{scope}
\begin{scope}[shift={([xshift=6mm,yshift=-4mm]ours.north east)},scale=1.2]
\filldraw[fill=qSky,draw=white,line width=1.2pt] (0,0) circle (3.5mm);
\draw[qInk!65,line width=.45pt] (0,-1.5)--(0,-2.25);
\fill[qInk!65] (0,-2.4) circle (.3mm);
\filldraw[fill=white,draw=qInk!65,rounded corners=.65mm,line width=.45pt] (-2,-1.45) rectangle (2,1.55);
\fill[qInk!75] (-.8,-.25) circle (.3mm); \fill[qInk!75] (.8,-.25) circle (.3mm);
\draw[qInk!65,line width=.4pt] (-.75,.65) .. controls (-.25,1.2) and (.25,1.2) .. (.75,.65);
\end{scope}
\end{tikzpicture}%
}
\endgroup
\caption{
    \textbf{Case Study 1: Registered
    Difference Grounding (CT).}
    Given two registered axial CT chest
    scans, the task is to localize the
    interval change between them.
    \textbf{Qwen3.5} completely fails,
    predicting \texttt{[0,0,0,0]}
    unable to detect any difference.
    \textbf{MedSeq-Grounder} predicts
    \texttt{[79,105,98,120]}, identifying
    the approximate region but with
    imprecise boundaries.
    \textbf{\method{} (Ours)} reasons
    that the first image shows a distinct
    high-density irregular opacity in the
    right upper lobe consistent with a
    pulmonary nodule, absent in the second
    image, rules out windowing artifact by
    confirming stable mediastinal and bony
    structures, and correctly localizes
    the nodule at
    \texttt{(79,104),(99,123)}.
    This case illustrates that correct
    difference grounding requires clinical
    concept identification
    distinguishing a true focal lesion
    from normal vascular markings
    rather than naive pixel comparison.
}
\label{fig:qual_case1}
\end{figure*}

\begin{figure*}[p]
\centering
\begingroup
\definecolor{qInk}{HTML}{252629}\definecolor{qBlue}{HTML}{3B73A7}
\definecolor{qLine}{HTML}{B9BEC5}
\definecolor{qCream}{HTML}{FBF6E2}
\definecolor{qSage}{HTML}{EDF3E7}
\definecolor{qPeach}{HTML}{F5E3D5}
\definecolor{qSky}{HTML}{E3EDF7}
\definecolor{qBack}{HTML}{EEF2F5}
\resizebox{\textwidth}{!}{%
\begin{tikzpicture}[x=1mm,y=-1mm,every node/.style={font=\sffamily\fontsize{8.5}{10.2}\selectfont,text=qInk},box/.style={anchor=north west,align=left,inner sep=2mm,outer sep=0pt,rounded corners=2mm,draw=qLine!70,line width=.45pt,text width=164mm}]

\fill[qBack,rounded corners=2mm] (0,0) rectangle (170,86);
\node[anchor=north,inner sep=0pt] at (43,3){\adjincludegraphics[trim={0.022\width} {0.053\height} {0.756\width} {0.036\height},clip,width=80mm,height=80mm,keepaspectratio]{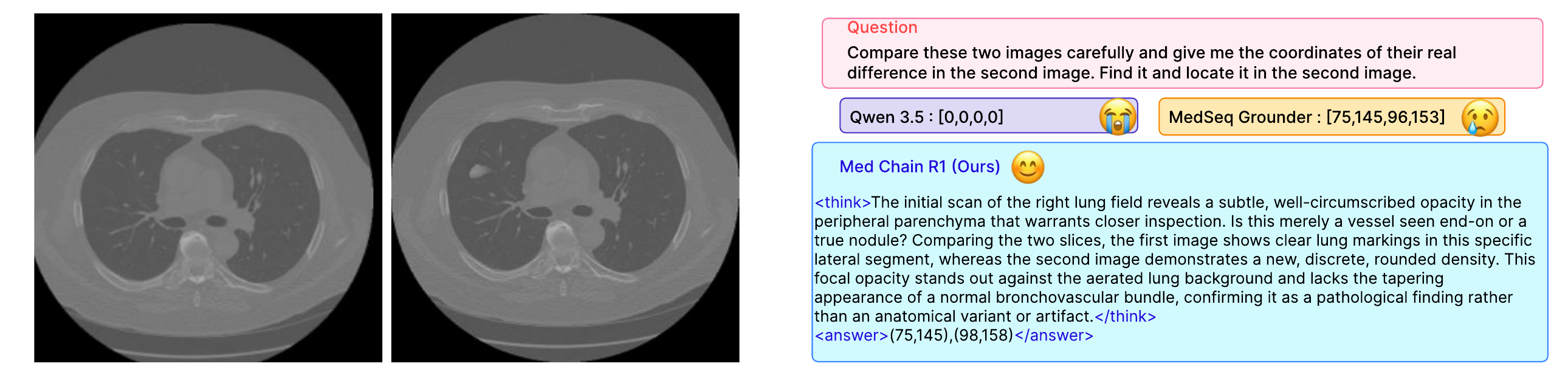}};
\node[anchor=north,inner sep=0pt] at (127,3){\adjincludegraphics[trim={0.25\width} {0.053\height} {0.528\width} {0.036\height},clip,width=80mm,height=80mm,keepaspectratio]{qual_case2.pdf}};
\node[box,fill=qCream] (question) at (0,90){\textbf{Question}\quad Compare these two images carefully and give me the coordinates of their real difference in the second image. Find it and locate it in the second image.};
\node[box,fill=qSage] (qwen) at ([yshift=-2mm]question.south west){\tikz[x=1mm,y=-1mm,baseline=-.7mm]{\filldraw[fill=yellow!55!white,draw=qInk!80,line width=.35pt] (0,0) circle (1.9mm);\draw[qInk,line width=.35pt,line cap=round] (-1.15,-.35) .. controls (-.95,-.8) and (-.5,-.8) .. (-.3,-.35) (.3,-.35) .. controls (.5,-.8) and (.95,-.8) .. (1.15,-.35);\fill[qBlue!65!cyan,rounded corners=.12mm] (-1,-.2) rectangle (-.65,1.25);\fill[qBlue!65!cyan,rounded corners=.12mm] (.65,-.2) rectangle (1,1.25);\fill[qInk!85] (0,.8) ellipse (.3mm and .42mm);}\quad Qwen 3.5 : [0,0,0,0]};
\node[box,fill=qPeach] (medseq) at ([yshift=-1.5mm]qwen.south west){\tikz[x=1mm,y=-1mm,baseline=-.7mm]{\filldraw[fill=yellow!55!white,draw=qInk!80,line width=.35pt] (0,0) circle (1.9mm);\draw[qInk,line width=.35pt,line cap=round] (-1.15,-.35) .. controls (-.95,-.8) and (-.5,-.8) .. (-.3,-.35) (.3,-.35) .. controls (.5,-.8) and (.95,-.8) .. (1.15,-.35);\fill[qBlue!65!cyan,rounded corners=.12mm] (-1,-.2) rectangle (-.65,1.25);\fill[qBlue!65!cyan,rounded corners=.12mm] (.65,-.2) rectangle (1,1.25);\fill[qInk!85] (0,.8) ellipse (.3mm and .42mm);}\quad MedSeq Grounder : [75,145,96,153]};
\node[box,fill=qSky] (ours) at ([yshift=-2mm]medseq.south west){\tikz[x=1mm,y=-1mm,baseline=-.7mm]{\filldraw[fill=yellow!55!white,draw=qInk!80,line width=.35pt] (0,0) circle (1.9mm);\fill[qInk] (-.65,-.45) circle (.18mm);\fill[qInk] (.65,-.45) circle (.18mm);\draw[qInk,line width=.45pt,line cap=round] (-.85,.35) .. controls (-.45,1.2) and (.45,1.2) .. (.85,.35);}\quad\textbf{Med Chain R1 (Ours)}};
\node[box,fill=white,rounded corners=2mm,font=\rmfamily\fontsize{9.2}{11}\selectfont] (trace) at ([yshift=-0mm]ours.south west){\textless{}think\textgreater{}The initial scan of the right lung field reveals a subtle, well-circumscribed opacity in the peripheral parenchyma that warrants closer inspection. Is this merely a vessel seen end-on or a true nodule? Comparing the two slices, the first image shows clear lung markings in this specific lateral segment, whereas the second image demonstrates a new, discrete, rounded density. This focal opacity stands out against the aerated lung background and lacks the tapering appearance of a normal bronchovascular bundle, confirming it as a pathological finding rather than an anatomical variant or artifact.\textless{}/think\textgreater{} \textless{}answer\textgreater{}(75,145),(98,158)\textless{}/answer\textgreater{}};

\fill[qCream] ([yshift=-3mm]question.north east)--++(2mm,-1.5mm)--++(-2mm,-1.5mm)--cycle;
\fill[qSky] ([yshift=-3mm]ours.north east)--++(2mm,-1.5mm)--++(-2mm,-1.5mm)--cycle;
\begin{scope}[shift={([xshift=6mm,yshift=-4mm]question.north east)},scale=1.2]
\filldraw[fill=qPeach,draw=white,line width=1.2pt] (0,0) circle (3.5mm);
\fill[qInk!65] (-2,2.2) .. controls (-1.7,.5) and (1.7,.5) .. (2,2.2)--cycle;
\filldraw[fill=qCream,draw=qInk!65,line width=.35pt] (0,-.7) circle (1.25mm);
\draw[qInk!65,line width=.45pt] (-1.15,-1.15) .. controls (-.7,-2.25) and (.7,-2.25) .. (1.15,-1.15);
\fill[qInk] (-.4,-.7) circle (.12mm); \fill[qInk] (.4,-.7) circle (.12mm);
\end{scope}
\begin{scope}[shift={([xshift=6mm,yshift=-4mm]ours.north east)},scale=1.2]
\filldraw[fill=qSky,draw=white,line width=1.2pt] (0,0) circle (3.5mm);
\draw[qInk!65,line width=.45pt] (0,-1.5)--(0,-2.25);
\fill[qInk!65] (0,-2.4) circle (.3mm);
\filldraw[fill=white,draw=qInk!65,rounded corners=.65mm,line width=.45pt] (-2,-1.45) rectangle (2,1.55);
\fill[qInk!75] (-.8,-.25) circle (.3mm); \fill[qInk!75] (.8,-.25) circle (.3mm);
\draw[qInk!65,line width=.4pt] (-.75,.65) .. controls (-.25,1.2) and (.25,1.2) .. (.75,.65);
\end{scope}
\end{tikzpicture}%
}
\endgroup
\caption{
    \textbf{Case Study 2: Non-Registered
    Difference Grounding (CT).}
    Given two non-registered axial CT chest
    scans, the task is to localize the real
    difference in the second image.
    \textbf{Qwen3.5} again predicts
    \texttt{[0,0,0,0]}, failing completely
    to detect any change.
    \textbf{MedSeq-Grounder} predicts
    \texttt{[75,145,96,153]}, localizing
    the correct general region but with
    imprecise boundaries.
    \textbf{\method{} (Ours)} reasons
    that the right lung field shows a
    subtle well-circumscribed opacity in
    the peripheral parenchyma, questions
    whether it is a vessel seen end-on or
    a true nodule, confirms it as a new
    pathological finding by noting the
    absence of the tapering appearance of
    a normal bronchovascular bundle, and
    correctly localizes it at
    \texttt{(75,145),(98,158)}.
    Compared to Case~1 (registered CT),
    this harder non-registered setting
    requires the model to establish spatial
    correspondence without alignment cues,
    yet \method{}'s clinical reasoning
    correctly identifies and localizes the
    interval finding where both baselines
    fail.
}
\label{fig:qual_case2}
\end{figure*}

\begin{figure*}[p]
\centering
\begingroup
\definecolor{qInk}{HTML}{252629}\definecolor{qBlue}{HTML}{3B73A7}
\definecolor{qLine}{HTML}{B9BEC5}
\definecolor{qCream}{HTML}{FBF6E2}
\definecolor{qSage}{HTML}{EDF3E7}
\definecolor{qPeach}{HTML}{F5E3D5}
\definecolor{qSky}{HTML}{E3EDF7}
\definecolor{qBack}{HTML}{EEF2F5}
\resizebox{\textwidth}{!}{%
\begin{tikzpicture}[x=1mm,y=-1mm,every node/.style={font=\sffamily\fontsize{8.5}{10.2}\selectfont,text=qInk},box/.style={anchor=north west,align=left,inner sep=2mm,outer sep=0pt,rounded corners=2mm,draw=qLine!70,line width=.45pt,text width=164mm}]

\fill[qBack,rounded corners=2mm] (0,0) rectangle (170,86);
\node[anchor=north,inner sep=0pt] at (43,3){\adjincludegraphics[trim={0.01\width} {0.063\height} {0.72\width} {0.067\height},clip,width=80mm,height=80mm,keepaspectratio]{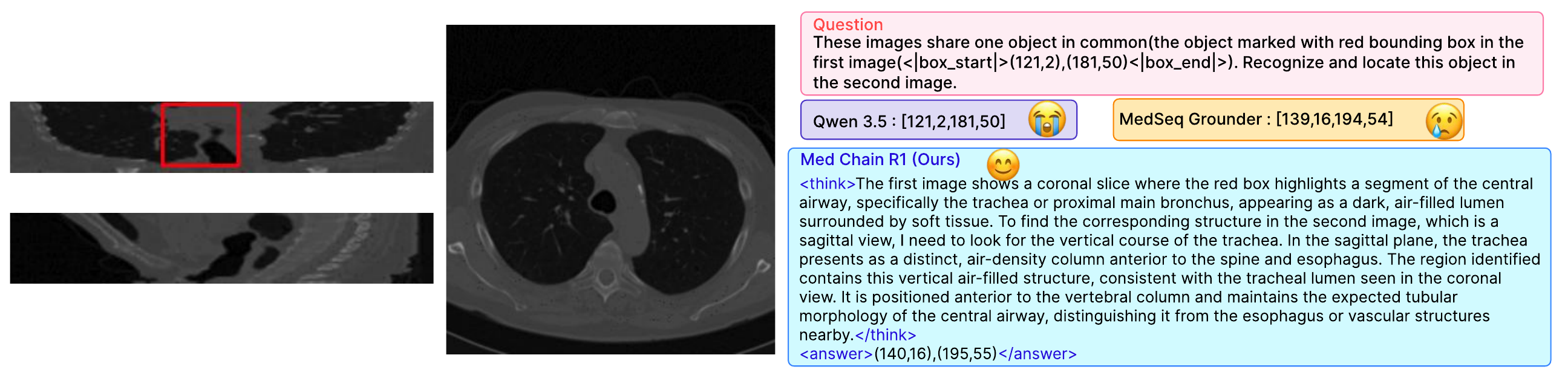}};
\node[anchor=north,inner sep=0pt] at (127,3){\adjincludegraphics[trim={0.285\width} {0.063\height} {0.505\width} {0.067\height},clip,width=80mm,height=80mm,keepaspectratio]{qual_case3.pdf}};
\node[box,fill=qCream] (question) at (0,90){\textbf{Question}\quad These images share one object in common(the object marked with red bounding box in the first image(\textless{}|box\_start|\textgreater{}(121,2),(181,50)\textless{}|box\_end|\textgreater{}). Recognize and locate this object in the second image.};
\node[box,fill=qSage] (qwen) at ([yshift=-2mm]question.south west){\tikz[x=1mm,y=-1mm,baseline=-.7mm]{\filldraw[fill=yellow!55!white,draw=qInk!80,line width=.35pt] (0,0) circle (1.9mm);\draw[qInk,line width=.35pt,line cap=round] (-1.15,-.35) .. controls (-.95,-.8) and (-.5,-.8) .. (-.3,-.35) (.3,-.35) .. controls (.5,-.8) and (.95,-.8) .. (1.15,-.35);\fill[qBlue!65!cyan,rounded corners=.12mm] (-1,-.2) rectangle (-.65,1.25);\fill[qBlue!65!cyan,rounded corners=.12mm] (.65,-.2) rectangle (1,1.25);\fill[qInk!85] (0,.8) ellipse (.3mm and .42mm);}\quad Qwen 3.5 : [121,2,181,50]};
\node[box,fill=qPeach] (medseq) at ([yshift=-1.5mm]qwen.south west){\tikz[x=1mm,y=-1mm,baseline=-.7mm]{\filldraw[fill=yellow!55!white,draw=qInk!80,line width=.35pt] (0,0) circle (1.9mm);\draw[qInk,line width=.35pt,line cap=round] (-1.15,-.35) .. controls (-.95,-.8) and (-.5,-.8) .. (-.3,-.35) (.3,-.35) .. controls (.5,-.8) and (.95,-.8) .. (1.15,-.35);\fill[qBlue!65!cyan,rounded corners=.12mm] (-1,-.2) rectangle (-.65,1.25);\fill[qBlue!65!cyan,rounded corners=.12mm] (.65,-.2) rectangle (1,1.25);\fill[qInk!85] (0,.8) ellipse (.3mm and .42mm);}\quad MedSeq Grounder : [139,16,194,54]};
\node[box,fill=qSky] (ours) at ([yshift=-2mm]medseq.south west){\tikz[x=1mm,y=-1mm,baseline=-.7mm]{\filldraw[fill=yellow!55!white,draw=qInk!80,line width=.35pt] (0,0) circle (1.9mm);\fill[qInk] (-.65,-.45) circle (.18mm);\fill[qInk] (.65,-.45) circle (.18mm);\draw[qInk,line width=.45pt,line cap=round] (-.85,.35) .. controls (-.45,1.2) and (.45,1.2) .. (.85,.35);}\quad\textbf{Med Chain R1 (Ours)}};
\node[box,fill=white,rounded corners=2mm,font=\rmfamily\fontsize{9.2}{11}\selectfont] (trace) at ([yshift=-0mm]ours.south west){\textless{}think\textgreater{}The first image shows a coronal slice where the red box highlights a segment of the central airway, specifically the trachea or proximal main bronchus, appearing as a dark, air-filled lumen surrounded by soft tissue. To find the corresponding structure in the second image, which is a sagittal view, I need to look for the vertical course of the trachea. In the sagittal plane, the trachea presents as a distinct, air-density column anterior to the spine and esophagus. The region identified contains this vertical air-filled structure, consistent with the tracheal lumen seen in the coronal view. It is positioned anterior to the vertebral column and maintains the expected tubular morphology of the central airway, distinguishing it from the esophagus or vascular structures nearby.\textless{}/think\textgreater{} \textless{}answer\textgreater{}(140,16),(195,55)\textless{}/answer\textgreater{}};

\fill[qCream] ([yshift=-3mm]question.north east)--++(2mm,-1.5mm)--++(-2mm,-1.5mm)--cycle;
\fill[qSky] ([yshift=-3mm]ours.north east)--++(2mm,-1.5mm)--++(-2mm,-1.5mm)--cycle;
\begin{scope}[shift={([xshift=6mm,yshift=-4mm]question.north east)},scale=1.2]
\filldraw[fill=qPeach,draw=white,line width=1.2pt] (0,0) circle (3.5mm);
\fill[qInk!65] (-2,2.2) .. controls (-1.7,.5) and (1.7,.5) .. (2,2.2)--cycle;
\filldraw[fill=qCream,draw=qInk!65,line width=.35pt] (0,-.7) circle (1.25mm);
\draw[qInk!65,line width=.45pt] (-1.15,-1.15) .. controls (-.7,-2.25) and (.7,-2.25) .. (1.15,-1.15);
\fill[qInk] (-.4,-.7) circle (.12mm); \fill[qInk] (.4,-.7) circle (.12mm);
\end{scope}
\begin{scope}[shift={([xshift=6mm,yshift=-4mm]ours.north east)},scale=1.2]
\filldraw[fill=qSky,draw=white,line width=1.2pt] (0,0) circle (3.5mm);
\draw[qInk!65,line width=.45pt] (0,-1.5)--(0,-2.25);
\fill[qInk!65] (0,-2.4) circle (.3mm);
\filldraw[fill=white,draw=qInk!65,rounded corners=.65mm,line width=.45pt] (-2,-1.45) rectangle (2,1.55);
\fill[qInk!75] (-.8,-.25) circle (.3mm); \fill[qInk!75] (.8,-.25) circle (.3mm);
\draw[qInk!65,line width=.4pt] (-.75,.65) .. controls (-.25,1.2) and (.25,1.2) .. (.75,.65);
\end{scope}
\end{tikzpicture}%
}
\endgroup
\caption{
    \textbf{Case Study 3: Crossmodal
    Grounding (CT).}
    Given a coronal CT slice with a
    red-boxed reference region and an
    axial CT slice, the task is to
    localize the same anatomical structure
    in the second image across views.
    \textbf{Qwen3.5} simply copies the
    reference box coordinates
    \texttt{[121,2,181,50]}, failing to
    perform any cross-view reasoning.
    \textbf{MedSeq-Grounder} predicts
    \texttt{[139,16,194,54]}, identifying
    the approximate region but without
    view-consistent reasoning.
    \textbf{\method{} (Ours)} identifies
    the red-boxed region in the coronal
    view as the trachea or proximal main
    bronchus  a dark air-filled lumen
    surrounded by soft tissue  reasons
    that in the sagittal axial view the
    trachea presents as a vertical
    air-density column anterior to the
    spine and esophagus, distinguishes it
    from nearby vascular structures by its
    tubular morphology, and correctly
    localizes it at
    \texttt{(140,16),(195,55)}.
    This case demonstrates that crossmodal
    grounding requires explicit anatomical
    concept transfer across imaging planes,
    which \method{}'s clinical reasoning
    chain uniquely provides.
}
\label{fig:qual_case3}
\end{figure*}

\begin{figure*}[p]
\centering
\begingroup
\definecolor{qInk}{HTML}{252629}\definecolor{qBlue}{HTML}{3B73A7}
\definecolor{qLine}{HTML}{B9BEC5}
\definecolor{qCream}{HTML}{FBF6E2}
\definecolor{qSage}{HTML}{EDF3E7}
\definecolor{qPeach}{HTML}{F5E3D5}
\definecolor{qSky}{HTML}{E3EDF7}
\definecolor{qBack}{HTML}{EEF2F5}
\resizebox{\textwidth}{!}{%
\begin{tikzpicture}[x=1mm,y=-1mm,every node/.style={font=\sffamily\fontsize{9}{11}\selectfont,text=qInk},box/.style={anchor=north west,align=left,inner sep=2mm,outer sep=0pt,rounded corners=2mm,draw=qLine!70,line width=.45pt,text width=156mm}]
\fill[qBack,rounded corners=2mm] (0,0) rectangle (170,44);
\node[anchor=north,inner sep=0pt] at (21,2){\adjincludegraphics[trim={0.008\width} {0.477\height} {0.871\width} {0.023\height},clip,width=40mm,height=40mm,keepaspectratio]{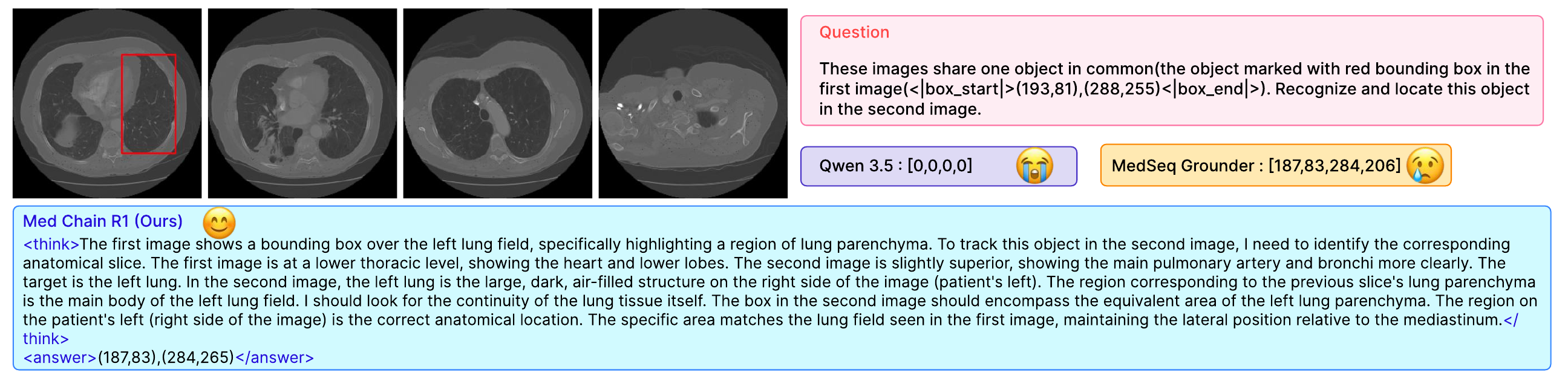}};
\node[anchor=north,inner sep=0pt] at (63.5,2){\adjincludegraphics[trim={0.133\width} {0.477\height} {0.747\width} {0.023\height},clip,width=40mm,height=40mm,keepaspectratio]{qual_case4.pdf}};
\node[anchor=north,inner sep=0pt] at (106,2){\adjincludegraphics[trim={0.257\width} {0.477\height} {0.622\width} {0.023\height},clip,width=40mm,height=40mm,keepaspectratio]{qual_case4.pdf}};
\node[anchor=north,inner sep=0pt] at (148.5,2){\adjincludegraphics[trim={0.382\width} {0.477\height} {0.497\width} {0.023\height},clip,width=40mm,height=40mm,keepaspectratio]{qual_case4.pdf}};
\node[box,fill=qCream] (question) at (0,47){\textbf{Question}\quad These images share one object in common(the object marked with red bounding box in the first image(\textless{}|box\_start|\textgreater{}(193,81),(288,255)\textless{}|box\_end|\textgreater{}). Recognize and locate this object in the second image.};
\node[box,fill=qSage,text width=73mm] (qwen) at ([yshift=-2mm]question.south west){\tikz[x=1mm,y=-1mm,baseline=-.7mm]{\filldraw[fill=yellow!55!white,draw=qInk!80,line width=.35pt] (0,0) circle (1.9mm);\draw[qInk,line width=.35pt,line cap=round] (-1.15,-.35) .. controls (-.95,-.8) and (-.5,-.8) .. (-.3,-.35) (.3,-.35) .. controls (.5,-.8) and (.95,-.8) .. (1.15,-.35);\fill[qBlue!65!cyan,rounded corners=.12mm] (-1,-.2) rectangle (-.65,1.25);\fill[qBlue!65!cyan,rounded corners=.12mm] (.65,-.2) rectangle (1,1.25);\fill[qInk!85] (0,.8) ellipse (.3mm and .42mm);}\quad Qwen 3.5 : [0,0,0,0]};
\node[box,fill=qPeach,text width=73mm] (medseq) at ([xshift=83mm,yshift=-2mm]question.south west){\tikz[x=1mm,y=-1mm,baseline=-.7mm]{\filldraw[fill=yellow!55!white,draw=qInk!80,line width=.35pt] (0,0) circle (1.9mm);\draw[qInk,line width=.35pt,line cap=round] (-1.15,-.35) .. controls (-.95,-.8) and (-.5,-.8) .. (-.3,-.35) (.3,-.35) .. controls (.5,-.8) and (.95,-.8) .. (1.15,-.35);\fill[qBlue!65!cyan,rounded corners=.12mm] (-1,-.2) rectangle (-.65,1.25);\fill[qBlue!65!cyan,rounded corners=.12mm] (.65,-.2) rectangle (1,1.25);\fill[qInk!85] (0,.8) ellipse (.3mm and .42mm);}\quad MedSeq Grounder : [187,83,284,206]};
\node[box,fill=qSky] (ours) at ([yshift=-2mm]qwen.south west){\tikz[x=1mm,y=-1mm,baseline=-.7mm]{\filldraw[fill=yellow!55!white,draw=qInk!80,line width=.35pt] (0,0) circle (1.9mm);\fill[qInk] (-.65,-.45) circle (.18mm);\fill[qInk] (.65,-.45) circle (.18mm);\draw[qInk,line width=.45pt,line cap=round] (-.85,.35) .. controls (-.45,1.2) and (.45,1.2) .. (.85,.35);}\quad\textbf{Med Chain R1 (Ours)}};
\node[box,fill=white,font=\rmfamily\fontsize{9.2}{11}\selectfont] at (ours.south west){\textless{}think\textgreater{}The first image shows a bounding box over the left lung field, specifically highlighting a region of lung parenchyma. To track this object in the second image, I need to identify the corresponding anatomical slice. The first image is at a lower thoracic level, showing the heart and lower lobes. The second image is slightly superior, showing the main pulmonary artery and bronchi more clearly. The target is the left lung. In the second image, the left lung is the large, dark, air-filled structure on the right side of the image (patient's left). The region corresponding to the previous slice's lung parenchyma is the main body of the left lung field. I should look for the continuity of the lung tissue itself. The box in the second image should encompass the equivalent area of the left lung parenchyma. The region on the patient's left (right side of the image) is the correct anatomical location. The specific area matches the lung field seen in the first image, maintaining the lateral position relative to the mediastinum.\textless{}/ think\textgreater{} \textless{}answer\textgreater{}(187,83),(284,265)\textless{}/answer\textgreater{}};

\fill[qCream] ([yshift=-3mm]question.north east)--++(2mm,-1.5mm)--++(-2mm,-1.5mm)--cycle;
\fill[qSky] ([yshift=-3mm]ours.north east)--++(2mm,-1.5mm)--++(-2mm,-1.5mm)--cycle;
\begin{scope}[shift={([xshift=6mm,yshift=-4mm]question.north east)},scale=1.2]
\filldraw[fill=qPeach,draw=white,line width=1.2pt] (0,0) circle (3.5mm);
\fill[qInk!65] (-2,2.2) .. controls (-1.7,.5) and (1.7,.5) .. (2,2.2)--cycle;
\filldraw[fill=qCream,draw=qInk!65,line width=.35pt] (0,-.7) circle (1.25mm);
\draw[qInk!65,line width=.45pt] (-1.15,-1.15) .. controls (-.7,-2.25) and (.7,-2.25) .. (1.15,-1.15);
\fill[qInk] (-.4,-.7) circle (.12mm); \fill[qInk] (.4,-.7) circle (.12mm);
\end{scope}
\begin{scope}[shift={([xshift=6mm,yshift=-4mm]ours.north east)},scale=1.2]
\filldraw[fill=qSky,draw=white,line width=1.2pt] (0,0) circle (3.5mm);
\draw[qInk!65,line width=.45pt] (0,-1.5)--(0,-2.25);
\fill[qInk!65] (0,-2.4) circle (.3mm);
\filldraw[fill=white,draw=qInk!65,rounded corners=.65mm,line width=.45pt] (-2,-1.45) rectangle (2,1.55);
\fill[qInk!75] (-.8,-.25) circle (.3mm); \fill[qInk!75] (.8,-.25) circle (.3mm);
\draw[qInk!65,line width=.4pt] (-.75,.65) .. controls (-.25,1.2) and (.25,1.2) .. (.75,.65);
\end{scope}
\end{tikzpicture}%
}
\endgroup
\caption{
    \textbf{Case Study 4: Object Tracking
    (CT).}
    Given four sequential axial CT slices
    with a red-boxed reference region in
    the first image, the task is to
    localize the same object in the second
    image across slice levels.
    \textbf{Qwen3.5} predicts
    \texttt{[0,0,0,0]}, failing entirely.
    \textbf{MedSeq-Grounder} predicts
    \texttt{[187,83,284,206]}, identifying
    the correct region but with imprecise
    inferior boundary.
    \textbf{\method{} (Ours)} reasons
    that the red box in the first image
    highlights left lung parenchyma at a
    lower thoracic level showing the heart
    and lower lobes, identifies that the
    second image is a slightly superior
    slice showing the main pulmonary artery
    and bronchi, tracks the left lung as
    the large air-filled structure on the
    patient's left, and correctly localizes
    the corresponding parenchymal region
    at \texttt{(187,83),(284,265)} by
    maintaining lateral position relative
    to the mediastinum across slice levels.
    This case highlights that object
    tracking across CT slices requires
    anatomical continuity reasoning
    understanding how structures evolve
    across imaging planes  which neither
    baseline can perform.
}
\label{fig:qual_case4}
\end{figure*}

\begin{figure*}[p]
\centering
\begingroup
\definecolor{qInk}{HTML}{252629}\definecolor{qBlue}{HTML}{3B73A7}
\definecolor{qLine}{HTML}{B9BEC5}
\definecolor{qCream}{HTML}{FBF6E2}
\definecolor{qSage}{HTML}{EDF3E7}
\definecolor{qPeach}{HTML}{F5E3D5}
\definecolor{qSky}{HTML}{E3EDF7}
\definecolor{qBack}{HTML}{EEF2F5}
\resizebox{\textwidth}{!}{%
\begin{tikzpicture}[x=1mm,y=-1mm,every node/.style={font=\sffamily\fontsize{8.5}{10.2}\selectfont,text=qInk},box/.style={anchor=north west,align=left,inner sep=2mm,outer sep=0pt,rounded corners=2mm,draw=qLine!70,line width=.45pt,text width=164mm}]

\fill[qBack,rounded corners=2mm] (0,0) rectangle (170,86);
\node[anchor=north,inner sep=0pt] at (43,3){\adjincludegraphics[trim={0.01\width} {0.03\height} {0.792\width} {0.03\height},clip,width=80mm,height=80mm,keepaspectratio]{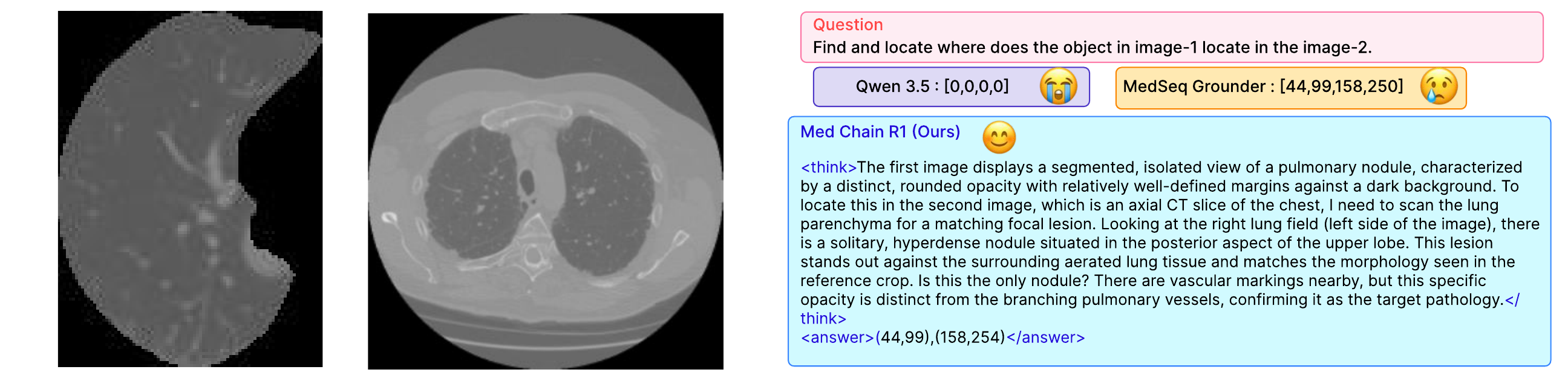}};
\node[anchor=north,inner sep=0pt] at (127,3){\adjincludegraphics[trim={0.235\width} {0.025\height} {0.538\width} {0.035\height},clip,width=80mm,height=80mm,keepaspectratio]{qual_case5.pdf}};
\node[box,fill=qCream] (question) at (0,90){\textbf{Question}\quad Find and locate where does the object in image-1 locate in the image-2.};
\node[box,fill=qSage] (qwen) at ([yshift=-2mm]question.south west){\tikz[x=1mm,y=-1mm,baseline=-.7mm]{\filldraw[fill=yellow!55!white,draw=qInk!80,line width=.35pt] (0,0) circle (1.9mm);\draw[qInk,line width=.35pt,line cap=round] (-1.15,-.35) .. controls (-.95,-.8) and (-.5,-.8) .. (-.3,-.35) (.3,-.35) .. controls (.5,-.8) and (.95,-.8) .. (1.15,-.35);\fill[qBlue!65!cyan,rounded corners=.12mm] (-1,-.2) rectangle (-.65,1.25);\fill[qBlue!65!cyan,rounded corners=.12mm] (.65,-.2) rectangle (1,1.25);\fill[qInk!85] (0,.8) ellipse (.3mm and .42mm);}\quad Qwen 3.5 : [0,0,0,0]};
\node[box,fill=qPeach] (medseq) at ([yshift=-1.5mm]qwen.south west){\tikz[x=1mm,y=-1mm,baseline=-.7mm]{\filldraw[fill=yellow!55!white,draw=qInk!80,line width=.35pt] (0,0) circle (1.9mm);\draw[qInk,line width=.35pt,line cap=round] (-1.15,-.35) .. controls (-.95,-.8) and (-.5,-.8) .. (-.3,-.35) (.3,-.35) .. controls (.5,-.8) and (.95,-.8) .. (1.15,-.35);\fill[qBlue!65!cyan,rounded corners=.12mm] (-1,-.2) rectangle (-.65,1.25);\fill[qBlue!65!cyan,rounded corners=.12mm] (.65,-.2) rectangle (1,1.25);\fill[qInk!85] (0,.8) ellipse (.3mm and .42mm);}\quad MedSeq Grounder : [44,99,158,250]};
\node[box,fill=qSky] (ours) at ([yshift=-2mm]medseq.south west){\tikz[x=1mm,y=-1mm,baseline=-.7mm]{\filldraw[fill=yellow!55!white,draw=qInk!80,line width=.35pt] (0,0) circle (1.9mm);\fill[qInk] (-.65,-.45) circle (.18mm);\fill[qInk] (.65,-.45) circle (.18mm);\draw[qInk,line width=.45pt,line cap=round] (-.85,.35) .. controls (-.45,1.2) and (.45,1.2) .. (.85,.35);}\quad\textbf{Med Chain R1 (Ours)}};
\node[box,fill=white,rounded corners=2mm,font=\rmfamily\fontsize{9.2}{11}\selectfont] (trace) at ([yshift=-0mm]ours.south west){\textless{}think\textgreater{}The first image displays a segmented, isolated view of a pulmonary nodule, characterized by a distinct, rounded opacity with relatively well-defined margins against a dark background. To locate this in the second image, which is an axial CT slice of the chest, I need to scan the lung parenchyma for a matching focal lesion. Looking at the right lung field (left side of the image), there is a solitary, hyperdense nodule situated in the posterior aspect of the upper lobe. This lesion stands out against the surrounding aerated lung tissue and matches the morphology seen in the reference crop. Is this the only nodule? There are vascular markings nearby, but this specific opacity is distinct from the branching pulmonary vessels, confirming it as the target pathology.\textless{}/ think\textgreater{} \textless{}answer\textgreater{}(44,99),(158,254)\textless{}/answer\textgreater{}};

\fill[qCream] ([yshift=-3mm]question.north east)--++(2mm,-1.5mm)--++(-2mm,-1.5mm)--cycle;
\fill[qSky] ([yshift=-3mm]ours.north east)--++(2mm,-1.5mm)--++(-2mm,-1.5mm)--cycle;
\begin{scope}[shift={([xshift=6mm,yshift=-4mm]question.north east)},scale=1.2]
\filldraw[fill=qPeach,draw=white,line width=1.2pt] (0,0) circle (3.5mm);
\fill[qInk!65] (-2,2.2) .. controls (-1.7,.5) and (1.7,.5) .. (2,2.2)--cycle;
\filldraw[fill=qCream,draw=qInk!65,line width=.35pt] (0,-.7) circle (1.25mm);
\draw[qInk!65,line width=.45pt] (-1.15,-1.15) .. controls (-.7,-2.25) and (.7,-2.25) .. (1.15,-1.15);
\fill[qInk] (-.4,-.7) circle (.12mm); \fill[qInk] (.4,-.7) circle (.12mm);
\end{scope}
\begin{scope}[shift={([xshift=6mm,yshift=-4mm]ours.north east)},scale=1.2]
\filldraw[fill=qSky,draw=white,line width=1.2pt] (0,0) circle (3.5mm);
\draw[qInk!65,line width=.45pt] (0,-1.5)--(0,-2.25);
\fill[qInk!65] (0,-2.4) circle (.3mm);
\filldraw[fill=white,draw=qInk!65,rounded corners=.65mm,line width=.45pt] (-2,-1.45) rectangle (2,1.55);
\fill[qInk!75] (-.8,-.25) circle (.3mm); \fill[qInk!75] (.8,-.25) circle (.3mm);
\draw[qInk!65,line width=.4pt] (-.75,.65) .. controls (-.25,1.2) and (.25,1.2) .. (.75,.65);
\end{scope}
\end{tikzpicture}%
}
\endgroup
\caption{
    \textbf{Case Study 5: Concept
    Localization (CT).}
    Given a segmented isolated crop of a
    pulmonary nodule (image~1) and a full
    axial CT chest slice (image~2), the
    task is to localize the same nodule
    in the full scan.
    \textbf{Qwen3.5} predicts
    \texttt{[0,0,0,0]}, failing completely.
    \textbf{MedSeq-Grounder} predicts
    \texttt{[44,99,158,250]}, spanning an
    overly large region that includes
    surrounding lung tissue.
    \textbf{\method{} (Ours)} identifies
    the reference crop as a rounded
    well-defined opacity consistent with
    a pulmonary nodule, scans the right
    lung field for a matching focal lesion,
    locates a solitary hyperdense nodule
    in the posterior aspect of the upper
    lobe, distinguishes it from nearby
    branching pulmonary vessels by its
    distinct morphology, and correctly
    tightens the localization to
    \texttt{(44,99),(158,254)}.
    This case is the hardest task type for
    simple RL  concept localization
    requires matching a pathological
    finding across heterogeneous imaging
    contexts, explaining the severe
    semantic drift ($-6.70$ Acc@0.5)
    observed under IoU-only GRPO.
}
\label{fig:qual_case5}
\end{figure*}

Figure~\ref{fig:sample} and Figures~\ref{fig:qual_case1}--\ref{fig:qual_case5} show representative predictions of zero-shot Qwen3.5, MedSeq-Grounder and \method{}. The examples were selected for illustration and are not a random sample.


\section{Extended Related Work}
\label{app:related}

\noindent \textbf{Medical vision-language models.} Large-scale medical VLMs such as LLaVA-Med~\citep{li2023llavamed}, HuatuoGPT-Vision~\citep{chen2024huatuogpt} and MedGemma~\citep{sellergren2025medgemma} perform well on medical VQA and report generation. FlemingVL~\citep{flemingvl} provides a unified framework for heterogeneous medical visual understanding across 2D, 3D and video modalities and serves as the Stage~1 teacher in our pipeline. General-purpose MLLMs including Qwen2.5-VL~\citep{bai2025qwen2}, InternVL3~\citep{zhu2025internvl3} and the multi-image grounding model Migician~\citep{li2025migician} serve as zero-shot baselines. None of these models is trained for sequential medical grounding, and none receives cross-image reasoning supervision.

\noindent \textbf{Chain-of-thought distillation.} Distilling reasoning from large models into smaller deployable ones~\citep{deepseekr1, wei2022chain} is a practical way to scale chain-of-thought supervision. In medical settings the central difficulty is aligning fluent clinical explanations with visual evidence across a sequence. In our pipeline FlemingVL-38B~\citep{flemingvl} supplies a preliminary visual rationale, and multimodal Qwen3.5-397B-A17B~\citep{qwen35} re-examines the original images before writing the final box-conditioned explanation. This differs from text-only distillation because the refiner keeps direct visual access.

\noindent \textbf{Visual grounding architectures.} Classical grounding models such as TransVG~\citep{transvg}, MDETR~\citep{mdetr} and Grounding DINO~\citep{groundingdino} predict boxes from a single image and text query. Sequential grounding adds image selection and cross-image correspondence, which motivates generating an explicit comparison before the box.

\begin{figure*}[p]
\centering
\begingroup
\definecolor{qInk}{HTML}{252629}\definecolor{qBlue}{HTML}{3B73A7}
\definecolor{qLine}{HTML}{B9BEC5}
\definecolor{qCream}{HTML}{FBF6E2}
\definecolor{qSage}{HTML}{EDF3E7}
\definecolor{qPeach}{HTML}{F5E3D5}
\definecolor{qSky}{HTML}{E3EDF7}
\definecolor{qBack}{HTML}{EEF2F5}
\resizebox{\textwidth}{!}{%
\begin{tikzpicture}[x=1mm,y=-1mm,every node/.style={font=\sffamily\fontsize{8.5}{10.2}\selectfont,text=qInk},box/.style={anchor=north west,align=left,inner sep=2mm,outer sep=0pt,rounded corners=2mm,draw=qLine!70,line width=.45pt,text width=164mm}]

\fill[qBack,rounded corners=2mm] (0,0) rectangle (170,86);
\node[anchor=north,inner sep=0pt] at (43,3){\adjincludegraphics[trim={0.003\width} {0.018\height} {0.763\width} {0.016\height},clip,width=80mm,height=80mm,keepaspectratio]{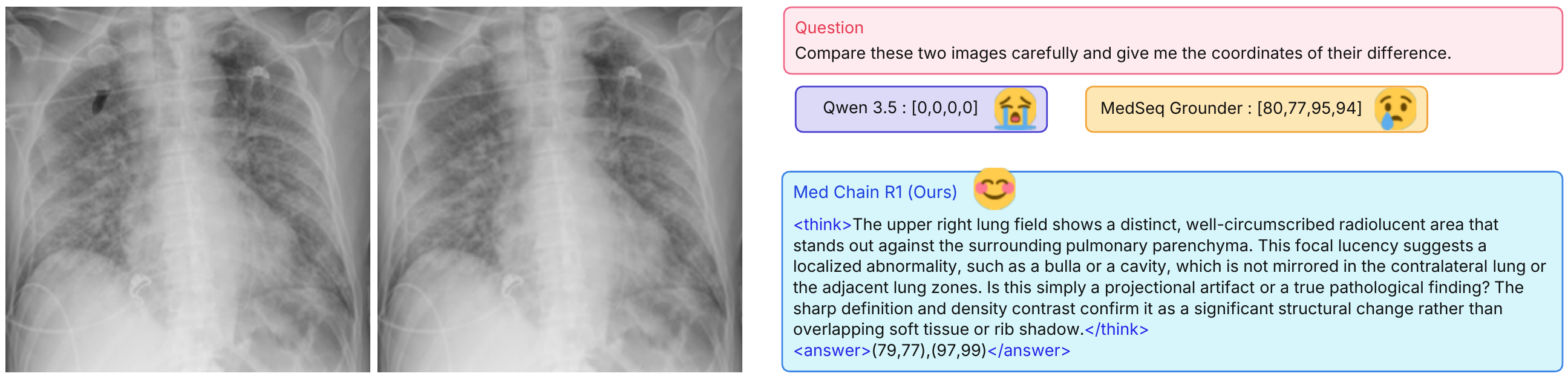}};
\node[anchor=north,inner sep=0pt] at (127,3){\adjincludegraphics[trim={0.241\width} {0.018\height} {0.526\width} {0.016\height},clip,width=80mm,height=80mm,keepaspectratio]{qual_case6.pdf}};
\node[box,fill=qCream] (question) at (0,90){\textbf{Question}\quad Compare these two images carefully and give me the coordinates of their difference.};
\node[box,fill=qSage] (qwen) at ([yshift=-2mm]question.south west){\tikz[x=1mm,y=-1mm,baseline=-.7mm]{\filldraw[fill=yellow!55!white,draw=qInk!80,line width=.35pt] (0,0) circle (1.9mm);\draw[qInk,line width=.35pt,line cap=round] (-1.15,-.35) .. controls (-.95,-.8) and (-.5,-.8) .. (-.3,-.35) (.3,-.35) .. controls (.5,-.8) and (.95,-.8) .. (1.15,-.35);\fill[qBlue!65!cyan,rounded corners=.12mm] (-1,-.2) rectangle (-.65,1.25);\fill[qBlue!65!cyan,rounded corners=.12mm] (.65,-.2) rectangle (1,1.25);\fill[qInk!85] (0,.8) ellipse (.3mm and .42mm);}\quad Qwen 3.5 : [0,0,0,0]};
\node[box,fill=qPeach] (medseq) at ([yshift=-1.5mm]qwen.south west){\tikz[x=1mm,y=-1mm,baseline=-.7mm]{\filldraw[fill=yellow!55!white,draw=qInk!80,line width=.35pt] (0,0) circle (1.9mm);\draw[qInk,line width=.35pt,line cap=round] (-1.15,-.35) .. controls (-.95,-.8) and (-.5,-.8) .. (-.3,-.35) (.3,-.35) .. controls (.5,-.8) and (.95,-.8) .. (1.15,-.35);\fill[qBlue!65!cyan,rounded corners=.12mm] (-1,-.2) rectangle (-.65,1.25);\fill[qBlue!65!cyan,rounded corners=.12mm] (.65,-.2) rectangle (1,1.25);\fill[qInk!85] (0,.8) ellipse (.3mm and .42mm);}\quad MedSeq Grounder : [80,77,95,94]};
\node[box,fill=qSky] (ours) at ([yshift=-2mm]medseq.south west){\tikz[x=1mm,y=-1mm,baseline=-.7mm]{\filldraw[fill=yellow!55!white,draw=qInk!80,line width=.35pt] (0,0) circle (1.9mm);\fill[qInk] (-.65,-.45) circle (.18mm);\fill[qInk] (.65,-.45) circle (.18mm);\draw[qInk,line width=.45pt,line cap=round] (-.85,.35) .. controls (-.45,1.2) and (.45,1.2) .. (.85,.35);}\quad\textbf{Med Chain R1 (Ours)}};
\node[box,fill=white,rounded corners=2mm,font=\rmfamily\fontsize{9.2}{11}\selectfont] (trace) at ([yshift=-0mm]ours.south west){\textless{}think\textgreater{}The upper right lung field shows a distinct, well-circumscribed radiolucent area that stands out against the surrounding pulmonary parenchyma. This focal lucency suggests a localized abnormality, such as a bulla or a cavity, which is not mirrored in the contralateral lung or the adjacent lung zones. Is this simply a projectional artifact or a true pathological finding? The sharp definition and density contrast confirm it as a significant structural change rather than overlapping soft tissue or rib shadow.\textless{}/think\textgreater{} \textless{}answer\textgreater{}(79,77),(97,99)\textless{}/answer\textgreater{}};

\fill[qCream] ([yshift=-3mm]question.north east)--++(2mm,-1.5mm)--++(-2mm,-1.5mm)--cycle;
\fill[qSky] ([yshift=-3mm]ours.north east)--++(2mm,-1.5mm)--++(-2mm,-1.5mm)--cycle;
\begin{scope}[shift={([xshift=6mm,yshift=-4mm]question.north east)},scale=1.2]
\filldraw[fill=qPeach,draw=white,line width=1.2pt] (0,0) circle (3.5mm);
\fill[qInk!65] (-2,2.2) .. controls (-1.7,.5) and (1.7,.5) .. (2,2.2)--cycle;
\filldraw[fill=qCream,draw=qInk!65,line width=.35pt] (0,-.7) circle (1.25mm);
\draw[qInk!65,line width=.45pt] (-1.15,-1.15) .. controls (-.7,-2.25) and (.7,-2.25) .. (1.15,-1.15);
\fill[qInk] (-.4,-.7) circle (.12mm); \fill[qInk] (.4,-.7) circle (.12mm);
\end{scope}
\begin{scope}[shift={([xshift=6mm,yshift=-4mm]ours.north east)},scale=1.2]
\filldraw[fill=qSky,draw=white,line width=1.2pt] (0,0) circle (3.5mm);
\draw[qInk!65,line width=.45pt] (0,-1.5)--(0,-2.25);
\fill[qInk!65] (0,-2.4) circle (.3mm);
\filldraw[fill=white,draw=qInk!65,rounded corners=.65mm,line width=.45pt] (-2,-1.45) rectangle (2,1.55);
\fill[qInk!75] (-.8,-.25) circle (.3mm); \fill[qInk!75] (.8,-.25) circle (.3mm);
\draw[qInk!65,line width=.4pt] (-.75,.65) .. controls (-.25,1.2) and (.25,1.2) .. (.75,.65);
\end{scope}
\end{tikzpicture}%
}
\endgroup
\caption{
    \textbf{Case Study 6: Registered Difference Grounding (Chest X-ray).}
    Given two registered chest radiographs, the task is to localize the difference between them.
    \textbf{Qwen3.5} predicts \texttt{[0,0,0,0]}, an empty box.
    \textbf{MedSeq-Grounder} predicts \texttt{[80,77,95,94]}.
    \textbf{\method{} (Ours)} identifies a well-circumscribed radiolucent area in the upper right lung field, checks whether it could be a projectional artifact or overlapping rib shadow, and predicts
    \texttt{(79,77),(97,99)}.
}
\label{fig:qual_case6}
\end{figure*}

\begin{figure*}[p]
\centering
\begingroup
\definecolor{qInk}{HTML}{252629}\definecolor{qBlue}{HTML}{3B73A7}
\definecolor{qLine}{HTML}{B9BEC5}
\definecolor{qCream}{HTML}{FBF6E2}
\definecolor{qSage}{HTML}{EDF3E7}
\definecolor{qPeach}{HTML}{F5E3D5}
\definecolor{qSky}{HTML}{E3EDF7}
\definecolor{qBack}{HTML}{EEF2F5}
\resizebox{\textwidth}{!}{%
\begin{tikzpicture}[x=1mm,y=-1mm,every node/.style={font=\sffamily\fontsize{8.5}{10.2}\selectfont,text=qInk},box/.style={anchor=north west,align=left,inner sep=2mm,outer sep=0pt,rounded corners=2mm,draw=qLine!70,line width=.45pt,text width=164mm}]

\fill[qBack,rounded corners=2mm] (0,0) rectangle (170,86);
\node[anchor=north,inner sep=0pt] at (43,3){\adjincludegraphics[trim={0.003\width} {0.121\height} {0.763\width} {0.119\height},clip,width=80mm,height=80mm,keepaspectratio]{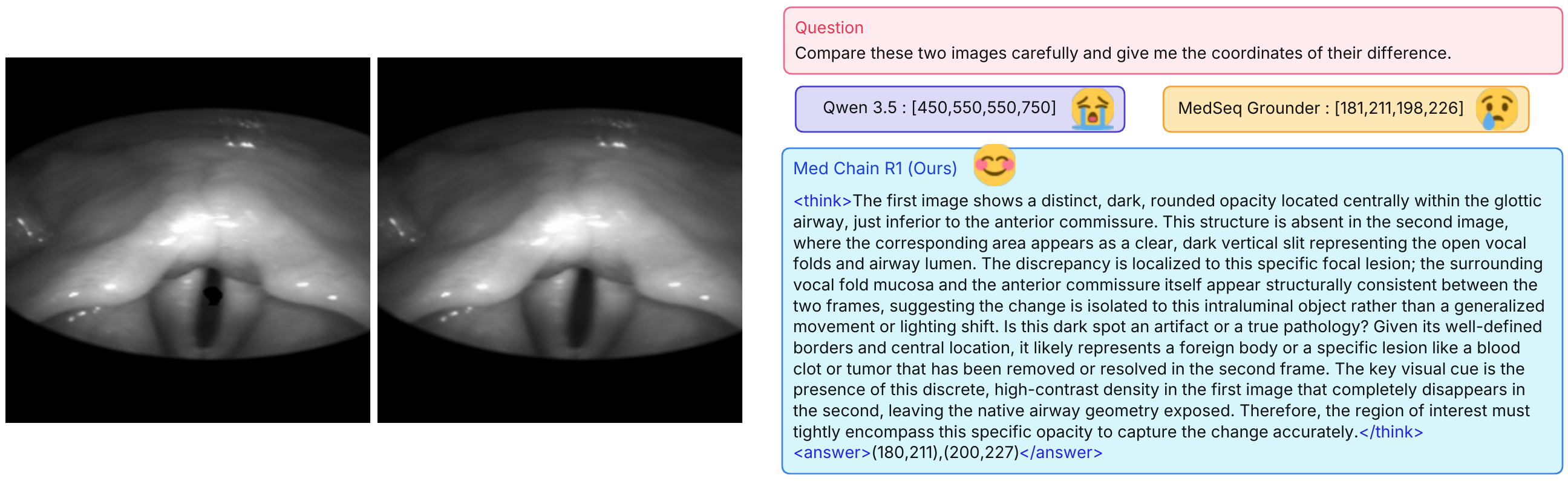}};
\node[anchor=north,inner sep=0pt] at (127,3){\adjincludegraphics[trim={0.241\width} {0.121\height} {0.526\width} {0.119\height},clip,width=80mm,height=80mm,keepaspectratio]{qual_case7.pdf}};
\node[box,fill=qCream] (question) at (0,90){\textbf{Question}\quad Compare these two images carefully and give me the coordinates of their difference.};
\node[box,fill=qSage] (qwen) at ([yshift=-2mm]question.south west){\tikz[x=1mm,y=-1mm,baseline=-.7mm]{\filldraw[fill=yellow!55!white,draw=qInk!80,line width=.35pt] (0,0) circle (1.9mm);\draw[qInk,line width=.35pt,line cap=round] (-1.15,-.35) .. controls (-.95,-.8) and (-.5,-.8) .. (-.3,-.35) (.3,-.35) .. controls (.5,-.8) and (.95,-.8) .. (1.15,-.35);\fill[qBlue!65!cyan,rounded corners=.12mm] (-1,-.2) rectangle (-.65,1.25);\fill[qBlue!65!cyan,rounded corners=.12mm] (.65,-.2) rectangle (1,1.25);\fill[qInk!85] (0,.8) ellipse (.3mm and .42mm);}\quad Qwen 3.5 : [450,550,550,750]};
\node[box,fill=qPeach] (medseq) at ([yshift=-1.5mm]qwen.south west){\tikz[x=1mm,y=-1mm,baseline=-.7mm]{\filldraw[fill=yellow!55!white,draw=qInk!80,line width=.35pt] (0,0) circle (1.9mm);\draw[qInk,line width=.35pt,line cap=round] (-1.15,-.35) .. controls (-.95,-.8) and (-.5,-.8) .. (-.3,-.35) (.3,-.35) .. controls (.5,-.8) and (.95,-.8) .. (1.15,-.35);\fill[qBlue!65!cyan,rounded corners=.12mm] (-1,-.2) rectangle (-.65,1.25);\fill[qBlue!65!cyan,rounded corners=.12mm] (.65,-.2) rectangle (1,1.25);\fill[qInk!85] (0,.8) ellipse (.3mm and .42mm);}\quad MedSeq Grounder : [181,211,198,226]};
\node[box,fill=qSky] (ours) at ([yshift=-2mm]medseq.south west){\tikz[x=1mm,y=-1mm,baseline=-.7mm]{\filldraw[fill=yellow!55!white,draw=qInk!80,line width=.35pt] (0,0) circle (1.9mm);\fill[qInk] (-.65,-.45) circle (.18mm);\fill[qInk] (.65,-.45) circle (.18mm);\draw[qInk,line width=.45pt,line cap=round] (-.85,.35) .. controls (-.45,1.2) and (.45,1.2) .. (.85,.35);}\quad\textbf{Med Chain R1 (Ours)}};
\node[box,fill=white,rounded corners=2mm,font=\rmfamily\fontsize{9.2}{11}\selectfont] (trace) at ([yshift=-0mm]ours.south west){\textless{}think\textgreater{}The first image shows a distinct, dark, rounded opacity located centrally within the glottic airway, just inferior to the anterior commissure. This structure is absent in the second image, where the corresponding area appears as a clear, dark vertical slit representing the open vocal folds and airway lumen. The discrepancy is localized to this specific focal lesion; the surrounding vocal fold mucosa and the anterior commissure itself appear structurally consistent between the two frames, suggesting the change is isolated to this intraluminal object rather than a generalized movement or lighting shift. Is this dark spot an artifact or a true pathology? Given its well-defined borders and central location, it likely represents a foreign body or a specific lesion like a blood clot or tumor that has been removed or resolved in the second frame. The key visual cue is the presence of this discrete, high-contrast density in the first image that completely disappears in the second, leaving the native airway geometry exposed. Therefore, the region of interest must tightly encompass this specific opacity to capture the change accurately.\textless{}/think\textgreater{} \textless{}answer\textgreater{}(180,211),(200,227)\textless{}/answer\textgreater{}};

\fill[qCream] ([yshift=-3mm]question.north east)--++(2mm,-1.5mm)--++(-2mm,-1.5mm)--cycle;
\fill[qSky] ([yshift=-3mm]ours.north east)--++(2mm,-1.5mm)--++(-2mm,-1.5mm)--cycle;
\begin{scope}[shift={([xshift=6mm,yshift=-4mm]question.north east)},scale=1.2]
\filldraw[fill=qPeach,draw=white,line width=1.2pt] (0,0) circle (3.5mm);
\fill[qInk!65] (-2,2.2) .. controls (-1.7,.5) and (1.7,.5) .. (2,2.2)--cycle;
\filldraw[fill=qCream,draw=qInk!65,line width=.35pt] (0,-.7) circle (1.25mm);
\draw[qInk!65,line width=.45pt] (-1.15,-1.15) .. controls (-.7,-2.25) and (.7,-2.25) .. (1.15,-1.15);
\fill[qInk] (-.4,-.7) circle (.12mm); \fill[qInk] (.4,-.7) circle (.12mm);
\end{scope}
\begin{scope}[shift={([xshift=6mm,yshift=-4mm]ours.north east)},scale=1.2]
\filldraw[fill=qSky,draw=white,line width=1.2pt] (0,0) circle (3.5mm);
\draw[qInk!65,line width=.45pt] (0,-1.5)--(0,-2.25);
\fill[qInk!65] (0,-2.4) circle (.3mm);
\filldraw[fill=white,draw=qInk!65,rounded corners=.65mm,line width=.45pt] (-2,-1.45) rectangle (2,1.55);
\fill[qInk!75] (-.8,-.25) circle (.3mm); \fill[qInk!75] (.8,-.25) circle (.3mm);
\draw[qInk!65,line width=.4pt] (-.75,.65) .. controls (-.25,1.2) and (.25,1.2) .. (.75,.65);
\end{scope}
\end{tikzpicture}%
}
\endgroup
\caption{
    \textbf{Case Study 7: Registered Difference Grounding (Laryngoscopy).}
    Given two registered laryngoscopic frames, the task is to localize the difference between them.
    \textbf{Qwen3.5} predicts \texttt{[450,550,550,750]}.
    \textbf{MedSeq-Grounder} predicts \texttt{[181,211,198,226]}.
    \textbf{\method{} (Ours)} describes a dark, rounded intraluminal opacity in the glottic airway that is present in the first frame and absent in the second, notes that the surrounding vocal fold mucosa is unchanged, and predicts
    \texttt{(180,211),(200,227)}.
}
\label{fig:qual_case7}
\end{figure*}

\clearpage
\begin{figure*}[p]
\centering
\begingroup
\definecolor{qInk}{HTML}{252629}\definecolor{qBlue}{HTML}{3B73A7}
\definecolor{qLine}{HTML}{B9BEC5}
\definecolor{qCream}{HTML}{FBF6E2}
\definecolor{qSage}{HTML}{EDF3E7}
\definecolor{qPeach}{HTML}{F5E3D5}
\definecolor{qSky}{HTML}{E3EDF7}
\definecolor{qBack}{HTML}{EEF2F5}
\resizebox{\textwidth}{!}{%
\begin{tikzpicture}[x=1mm,y=-1mm,every node/.style={font=\sffamily\fontsize{8.5}{10.2}\selectfont,text=qInk},box/.style={anchor=north west,align=left,inner sep=2mm,outer sep=0pt,rounded corners=2mm,draw=qLine!70,line width=.45pt,text width=164mm}]

\fill[qBack,rounded corners=2mm] (0,0) rectangle (170,86);
\node[anchor=north,inner sep=0pt] at (43,3){\adjincludegraphics[trim={0.003\width} {0.103\height} {0.763\width} {0.102\height},clip,width=80mm,height=80mm,keepaspectratio]{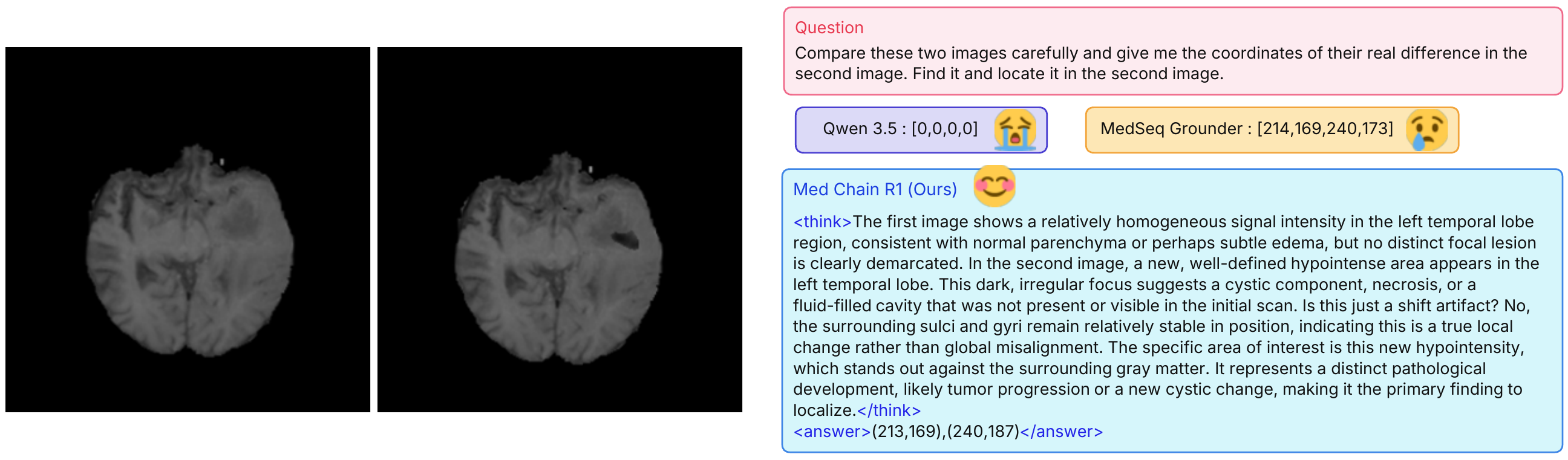}};
\node[anchor=north,inner sep=0pt] at (127,3){\adjincludegraphics[trim={0.241\width} {0.103\height} {0.526\width} {0.102\height},clip,width=80mm,height=80mm,keepaspectratio]{qual_case8.pdf}};
\node[box,fill=qCream] (question) at (0,90){\textbf{Question}\quad Compare these two images carefully and give me the coordinates of their real difference in the second image. Find it and locate it in the second image.};
\node[box,fill=qSage] (qwen) at ([yshift=-2mm]question.south west){\tikz[x=1mm,y=-1mm,baseline=-.7mm]{\filldraw[fill=yellow!55!white,draw=qInk!80,line width=.35pt] (0,0) circle (1.9mm);\draw[qInk,line width=.35pt,line cap=round] (-1.15,-.35) .. controls (-.95,-.8) and (-.5,-.8) .. (-.3,-.35) (.3,-.35) .. controls (.5,-.8) and (.95,-.8) .. (1.15,-.35);\fill[qBlue!65!cyan,rounded corners=.12mm] (-1,-.2) rectangle (-.65,1.25);\fill[qBlue!65!cyan,rounded corners=.12mm] (.65,-.2) rectangle (1,1.25);\fill[qInk!85] (0,.8) ellipse (.3mm and .42mm);}\quad Qwen 3.5 : [0,0,0,0]};
\node[box,fill=qPeach] (medseq) at ([yshift=-1.5mm]qwen.south west){\tikz[x=1mm,y=-1mm,baseline=-.7mm]{\filldraw[fill=yellow!55!white,draw=qInk!80,line width=.35pt] (0,0) circle (1.9mm);\draw[qInk,line width=.35pt,line cap=round] (-1.15,-.35) .. controls (-.95,-.8) and (-.5,-.8) .. (-.3,-.35) (.3,-.35) .. controls (.5,-.8) and (.95,-.8) .. (1.15,-.35);\fill[qBlue!65!cyan,rounded corners=.12mm] (-1,-.2) rectangle (-.65,1.25);\fill[qBlue!65!cyan,rounded corners=.12mm] (.65,-.2) rectangle (1,1.25);\fill[qInk!85] (0,.8) ellipse (.3mm and .42mm);}\quad MedSeq Grounder : [214,169,240,173]};
\node[box,fill=qSky] (ours) at ([yshift=-2mm]medseq.south west){\tikz[x=1mm,y=-1mm,baseline=-.7mm]{\filldraw[fill=yellow!55!white,draw=qInk!80,line width=.35pt] (0,0) circle (1.9mm);\fill[qInk] (-.65,-.45) circle (.18mm);\fill[qInk] (.65,-.45) circle (.18mm);\draw[qInk,line width=.45pt,line cap=round] (-.85,.35) .. controls (-.45,1.2) and (.45,1.2) .. (.85,.35);}\quad\textbf{Med Chain R1 (Ours)}};
\node[box,fill=white,rounded corners=2mm,font=\rmfamily\fontsize{9.2}{11}\selectfont] (trace) at ([yshift=-0mm]ours.south west){\textless{}think\textgreater{}The first image shows a relatively homogeneous signal intensity in the left temporal lobe region, consistent with normal parenchyma or perhaps subtle edema, but no distinct focal lesion is clearly demarcated. In the second image, a new, well-defined hypointense area appears in the left temporal lobe. This dark, irregular focus suggests a cystic component, necrosis, or a fluid-filled cavity that was not present or visible in the initial scan. Is this just a shift artifact? No, the surrounding sulci and gyri remain relatively stable in position, indicating this is a true local change rather than global misalignment. The specific area of interest is this new hypointensity, which stands out against the surrounding gray matter. It represents a distinct pathological development, likely tumor progression or a new cystic change, making it the primary finding to localize.\textless{}/think\textgreater{} \textless{}answer\textgreater{}(213,169),(240,187)\textless{}/answer\textgreater{}};

\fill[qCream] ([yshift=-3mm]question.north east)--++(2mm,-1.5mm)--++(-2mm,-1.5mm)--cycle;
\fill[qSky] ([yshift=-3mm]ours.north east)--++(2mm,-1.5mm)--++(-2mm,-1.5mm)--cycle;
\begin{scope}[shift={([xshift=6mm,yshift=-4mm]question.north east)},scale=1.2]
\filldraw[fill=qPeach,draw=white,line width=1.2pt] (0,0) circle (3.5mm);
\fill[qInk!65] (-2,2.2) .. controls (-1.7,.5) and (1.7,.5) .. (2,2.2)--cycle;
\filldraw[fill=qCream,draw=qInk!65,line width=.35pt] (0,-.7) circle (1.25mm);
\draw[qInk!65,line width=.45pt] (-1.15,-1.15) .. controls (-.7,-2.25) and (.7,-2.25) .. (1.15,-1.15);
\fill[qInk] (-.4,-.7) circle (.12mm); \fill[qInk] (.4,-.7) circle (.12mm);
\end{scope}
\begin{scope}[shift={([xshift=6mm,yshift=-4mm]ours.north east)},scale=1.2]
\filldraw[fill=qSky,draw=white,line width=1.2pt] (0,0) circle (3.5mm);
\draw[qInk!65,line width=.45pt] (0,-1.5)--(0,-2.25);
\fill[qInk!65] (0,-2.4) circle (.3mm);
\filldraw[fill=white,draw=qInk!65,rounded corners=.65mm,line width=.45pt] (-2,-1.45) rectangle (2,1.55);
\fill[qInk!75] (-.8,-.25) circle (.3mm); \fill[qInk!75] (.8,-.25) circle (.3mm);
\draw[qInk!65,line width=.4pt] (-.75,.65) .. controls (-.25,1.2) and (.25,1.2) .. (.75,.65);
\end{scope}
\end{tikzpicture}%
}
\endgroup
\caption{
    \textbf{Case Study 8: Non-Registered Difference Grounding (Brain MRI).}
    Given two non-registered brain MRI slices, the task is to localize the real difference in the second image.
    \textbf{Qwen3.5} predicts \texttt{[0,0,0,0]}, an empty box.
    \textbf{MedSeq-Grounder} predicts \texttt{[214,169,240,173]}.
    \textbf{\method{} (Ours)} identifies a new hypointense focus in the left temporal lobe, rules out a global shift by noting that the surrounding sulci and gyri are stable, and predicts
    \texttt{(213,169),(240,187)}.
}
\label{fig:qual_case8}
\end{figure*}

\begin{figure*}[p]
\centering
\begingroup
\definecolor{qInk}{HTML}{252629}\definecolor{qBlue}{HTML}{3B73A7}
\definecolor{qLine}{HTML}{B9BEC5}
\definecolor{qCream}{HTML}{FBF6E2}
\definecolor{qSage}{HTML}{EDF3E7}
\definecolor{qPeach}{HTML}{F5E3D5}
\definecolor{qSky}{HTML}{E3EDF7}
\definecolor{qBack}{HTML}{EEF2F5}
\resizebox{\textwidth}{!}{%
\begin{tikzpicture}[x=1mm,y=-1mm,every node/.style={font=\sffamily\fontsize{8.5}{10.2}\selectfont,text=qInk},box/.style={anchor=north west,align=left,inner sep=2mm,outer sep=0pt,rounded corners=2mm,draw=qLine!70,line width=.45pt,text width=164mm}]

\fill[qBack,rounded corners=2mm] (0,0) rectangle (170,86);
\node[anchor=north,inner sep=0pt] at (43,3){\adjincludegraphics[trim={0.003\width} {0.121\height} {0.763\width} {0.119\height},clip,width=80mm,height=80mm,keepaspectratio]{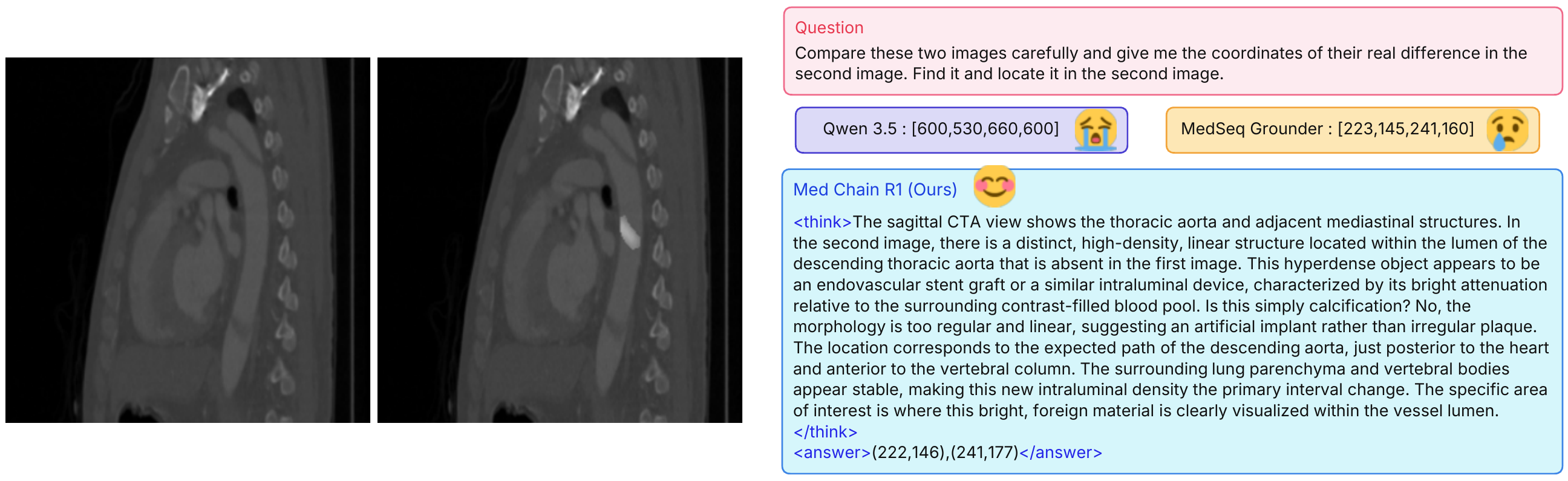}};
\node[anchor=north,inner sep=0pt] at (127,3){\adjincludegraphics[trim={0.241\width} {0.121\height} {0.526\width} {0.119\height},clip,width=80mm,height=80mm,keepaspectratio]{qual_case9.pdf}};
\node[box,fill=qCream] (question) at (0,90){\textbf{Question}\quad Compare these two images carefully and give me the coordinates of their real difference in the second image. Find it and locate it in the second image.};
\node[box,fill=qSage] (qwen) at ([yshift=-2mm]question.south west){\tikz[x=1mm,y=-1mm,baseline=-.7mm]{\filldraw[fill=yellow!55!white,draw=qInk!80,line width=.35pt] (0,0) circle (1.9mm);\draw[qInk,line width=.35pt,line cap=round] (-1.15,-.35) .. controls (-.95,-.8) and (-.5,-.8) .. (-.3,-.35) (.3,-.35) .. controls (.5,-.8) and (.95,-.8) .. (1.15,-.35);\fill[qBlue!65!cyan,rounded corners=.12mm] (-1,-.2) rectangle (-.65,1.25);\fill[qBlue!65!cyan,rounded corners=.12mm] (.65,-.2) rectangle (1,1.25);\fill[qInk!85] (0,.8) ellipse (.3mm and .42mm);}\quad Qwen 3.5 : [600,530,660,600]};
\node[box,fill=qPeach] (medseq) at ([yshift=-1.5mm]qwen.south west){\tikz[x=1mm,y=-1mm,baseline=-.7mm]{\filldraw[fill=yellow!55!white,draw=qInk!80,line width=.35pt] (0,0) circle (1.9mm);\draw[qInk,line width=.35pt,line cap=round] (-1.15,-.35) .. controls (-.95,-.8) and (-.5,-.8) .. (-.3,-.35) (.3,-.35) .. controls (.5,-.8) and (.95,-.8) .. (1.15,-.35);\fill[qBlue!65!cyan,rounded corners=.12mm] (-1,-.2) rectangle (-.65,1.25);\fill[qBlue!65!cyan,rounded corners=.12mm] (.65,-.2) rectangle (1,1.25);\fill[qInk!85] (0,.8) ellipse (.3mm and .42mm);}\quad MedSeq Grounder : [223,145,241,160]};
\node[box,fill=qSky] (ours) at ([yshift=-2mm]medseq.south west){\tikz[x=1mm,y=-1mm,baseline=-.7mm]{\filldraw[fill=yellow!55!white,draw=qInk!80,line width=.35pt] (0,0) circle (1.9mm);\fill[qInk] (-.65,-.45) circle (.18mm);\fill[qInk] (.65,-.45) circle (.18mm);\draw[qInk,line width=.45pt,line cap=round] (-.85,.35) .. controls (-.45,1.2) and (.45,1.2) .. (.85,.35);}\quad\textbf{Med Chain R1 (Ours)}};
\node[box,fill=white,rounded corners=2mm,font=\rmfamily\fontsize{9.2}{11}\selectfont] (trace) at ([yshift=-0mm]ours.south west){\textless{}think\textgreater{}The sagittal CTA view shows the thoracic aorta and adjacent mediastinal structures. In the second image, there is a distinct, high-density, linear structure located within the lumen of the descending thoracic aorta that is absent in the first image. This hyperdense object appears to be an endovascular stent graft or a similar intraluminal device, characterized by its bright attenuation relative to the surrounding contrast-filled blood pool. Is this simply calcification? No, the morphology is too regular and linear, suggesting an artificial implant rather than irregular plaque. The location corresponds to the expected path of the descending aorta, just posterior to the heart and anterior to the vertebral column. The surrounding lung parenchyma and vertebral bodies appear stable, making this new intraluminal density the primary interval change. The specific area of interest is where this bright, foreign material is clearly visualized within the vessel lumen.\textless{}/think\textgreater{} \textless{}answer\textgreater{}(222,146),(241,177)\textless{}/answer\textgreater{}};

\fill[qCream] ([yshift=-3mm]question.north east)--++(2mm,-1.5mm)--++(-2mm,-1.5mm)--cycle;
\fill[qSky] ([yshift=-3mm]ours.north east)--++(2mm,-1.5mm)--++(-2mm,-1.5mm)--cycle;
\begin{scope}[shift={([xshift=6mm,yshift=-4mm]question.north east)},scale=1.2]
\filldraw[fill=qPeach,draw=white,line width=1.2pt] (0,0) circle (3.5mm);
\fill[qInk!65] (-2,2.2) .. controls (-1.7,.5) and (1.7,.5) .. (2,2.2)--cycle;
\filldraw[fill=qCream,draw=qInk!65,line width=.35pt] (0,-.7) circle (1.25mm);
\draw[qInk!65,line width=.45pt] (-1.15,-1.15) .. controls (-.7,-2.25) and (.7,-2.25) .. (1.15,-1.15);
\fill[qInk] (-.4,-.7) circle (.12mm); \fill[qInk] (.4,-.7) circle (.12mm);
\end{scope}
\begin{scope}[shift={([xshift=6mm,yshift=-4mm]ours.north east)},scale=1.2]
\filldraw[fill=qSky,draw=white,line width=1.2pt] (0,0) circle (3.5mm);
\draw[qInk!65,line width=.45pt] (0,-1.5)--(0,-2.25);
\fill[qInk!65] (0,-2.4) circle (.3mm);
\filldraw[fill=white,draw=qInk!65,rounded corners=.65mm,line width=.45pt] (-2,-1.45) rectangle (2,1.55);
\fill[qInk!75] (-.8,-.25) circle (.3mm); \fill[qInk!75] (.8,-.25) circle (.3mm);
\draw[qInk!65,line width=.4pt] (-.75,.65) .. controls (-.25,1.2) and (.25,1.2) .. (.75,.65);
\end{scope}
\end{tikzpicture}%
}
\endgroup
\caption{
    \textbf{Case Study 9: Non-Registered Difference Grounding (CT Angiography).}
    Given two non-registered sagittal CT angiography images, the task is to localize the real difference in the second image.
    \textbf{Qwen3.5} predicts \texttt{[600,530,660,600]}.
    \textbf{MedSeq-Grounder} predicts \texttt{[223,145,241,160]}.
    \textbf{\method{} (Ours)} identifies a new linear hyperdense structure within the lumen of the descending thoracic aorta, distinguishes it from calcification by its regular morphology, and predicts
    \texttt{(222,146),(241,177)}.
}
\label{fig:qual_case9}
\end{figure*}

\clearpage
\begin{figure*}[p]
\centering
\begingroup
\definecolor{qInk}{HTML}{252629}\definecolor{qBlue}{HTML}{3B73A7}
\definecolor{qLine}{HTML}{B9BEC5}
\definecolor{qCream}{HTML}{FBF6E2}
\definecolor{qSage}{HTML}{EDF3E7}
\definecolor{qPeach}{HTML}{F5E3D5}
\definecolor{qSky}{HTML}{E3EDF7}
\definecolor{qBack}{HTML}{EEF2F5}
\resizebox{\textwidth}{!}{%
\begin{tikzpicture}[x=1mm,y=-1mm,every node/.style={font=\sffamily\fontsize{8.5}{10.2}\selectfont,text=qInk},box/.style={anchor=north west,align=left,inner sep=2mm,outer sep=0pt,rounded corners=2mm,draw=qLine!70,line width=.45pt,text width=164mm}]

\fill[qBack,rounded corners=2mm] (0,0) rectangle (170,86);
\node[anchor=north,inner sep=0pt] at (43,3){\adjincludegraphics[trim={0.003\width} {0.121\height} {0.763\width} {0.119\height},clip,width=80mm,height=80mm,keepaspectratio]{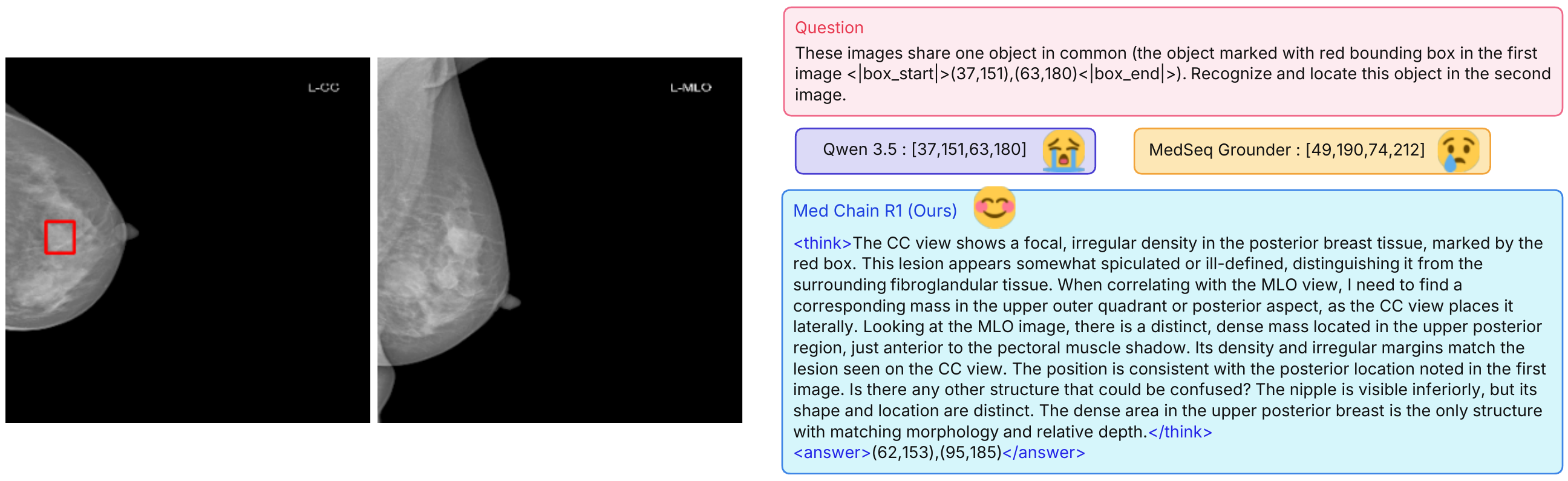}};
\node[anchor=north,inner sep=0pt] at (127,3){\adjincludegraphics[trim={0.241\width} {0.121\height} {0.526\width} {0.119\height},clip,width=80mm,height=80mm,keepaspectratio]{qual_case10.pdf}};
\node[box,fill=qCream] (question) at (0,90){\textbf{Question}\quad These images share one object in common (the object marked with red bounding box in the first image \textless{}|box\_start|\textgreater{}(37,151),(63,180)\textless{}|box\_end|\textgreater{}). Recognize and locate this object in the second image.};
\node[box,fill=qSage] (qwen) at ([yshift=-2mm]question.south west){\tikz[x=1mm,y=-1mm,baseline=-.7mm]{\filldraw[fill=yellow!55!white,draw=qInk!80,line width=.35pt] (0,0) circle (1.9mm);\draw[qInk,line width=.35pt,line cap=round] (-1.15,-.35) .. controls (-.95,-.8) and (-.5,-.8) .. (-.3,-.35) (.3,-.35) .. controls (.5,-.8) and (.95,-.8) .. (1.15,-.35);\fill[qBlue!65!cyan,rounded corners=.12mm] (-1,-.2) rectangle (-.65,1.25);\fill[qBlue!65!cyan,rounded corners=.12mm] (.65,-.2) rectangle (1,1.25);\fill[qInk!85] (0,.8) ellipse (.3mm and .42mm);}\quad Qwen 3.5 : [37,151,63,180]};
\node[box,fill=qPeach] (medseq) at ([yshift=-1.5mm]qwen.south west){\tikz[x=1mm,y=-1mm,baseline=-.7mm]{\filldraw[fill=yellow!55!white,draw=qInk!80,line width=.35pt] (0,0) circle (1.9mm);\draw[qInk,line width=.35pt,line cap=round] (-1.15,-.35) .. controls (-.95,-.8) and (-.5,-.8) .. (-.3,-.35) (.3,-.35) .. controls (.5,-.8) and (.95,-.8) .. (1.15,-.35);\fill[qBlue!65!cyan,rounded corners=.12mm] (-1,-.2) rectangle (-.65,1.25);\fill[qBlue!65!cyan,rounded corners=.12mm] (.65,-.2) rectangle (1,1.25);\fill[qInk!85] (0,.8) ellipse (.3mm and .42mm);}\quad MedSeq Grounder : [49,190,74,212]};
\node[box,fill=qSky] (ours) at ([yshift=-2mm]medseq.south west){\tikz[x=1mm,y=-1mm,baseline=-.7mm]{\filldraw[fill=yellow!55!white,draw=qInk!80,line width=.35pt] (0,0) circle (1.9mm);\fill[qInk] (-.65,-.45) circle (.18mm);\fill[qInk] (.65,-.45) circle (.18mm);\draw[qInk,line width=.45pt,line cap=round] (-.85,.35) .. controls (-.45,1.2) and (.45,1.2) .. (.85,.35);}\quad\textbf{Med Chain R1 (Ours)}};
\node[box,fill=white,rounded corners=2mm,font=\rmfamily\fontsize{9.2}{11}\selectfont] (trace) at ([yshift=-0mm]ours.south west){\textless{}think\textgreater{}The CC view shows a focal, irregular density in the posterior breast tissue, marked by the red box. This lesion appears somewhat spiculated or ill-defined, distinguishing it from the surrounding fibroglandular tissue. When correlating with the MLO view, I need to find a corresponding mass in the upper outer quadrant or posterior aspect, as the CC view places it laterally. Looking at the MLO image, there is a distinct, dense mass located in the upper posterior region, just anterior to the pectoral muscle shadow. Its density and irregular margins match the lesion seen on the CC view. The position is consistent with the posterior location noted in the first image. Is there any other structure that could be confused? The nipple is visible inferiorly, but its shape and location are distinct. The dense area in the upper posterior breast is the only structure with matching morphology and relative depth.\textless{}/think\textgreater{} \textless{}answer\textgreater{}(62,153),(95,185)\textless{}/answer\textgreater{}};

\fill[qCream] ([yshift=-3mm]question.north east)--++(2mm,-1.5mm)--++(-2mm,-1.5mm)--cycle;
\fill[qSky] ([yshift=-3mm]ours.north east)--++(2mm,-1.5mm)--++(-2mm,-1.5mm)--cycle;
\begin{scope}[shift={([xshift=6mm,yshift=-4mm]question.north east)},scale=1.2]
\filldraw[fill=qPeach,draw=white,line width=1.2pt] (0,0) circle (3.5mm);
\fill[qInk!65] (-2,2.2) .. controls (-1.7,.5) and (1.7,.5) .. (2,2.2)--cycle;
\filldraw[fill=qCream,draw=qInk!65,line width=.35pt] (0,-.7) circle (1.25mm);
\draw[qInk!65,line width=.45pt] (-1.15,-1.15) .. controls (-.7,-2.25) and (.7,-2.25) .. (1.15,-1.15);
\fill[qInk] (-.4,-.7) circle (.12mm); \fill[qInk] (.4,-.7) circle (.12mm);
\end{scope}
\begin{scope}[shift={([xshift=6mm,yshift=-4mm]ours.north east)},scale=1.2]
\filldraw[fill=qSky,draw=white,line width=1.2pt] (0,0) circle (3.5mm);
\draw[qInk!65,line width=.45pt] (0,-1.5)--(0,-2.25);
\fill[qInk!65] (0,-2.4) circle (.3mm);
\filldraw[fill=white,draw=qInk!65,rounded corners=.65mm,line width=.45pt] (-2,-1.45) rectangle (2,1.55);
\fill[qInk!75] (-.8,-.25) circle (.3mm); \fill[qInk!75] (.8,-.25) circle (.3mm);
\draw[qInk!65,line width=.4pt] (-.75,.65) .. controls (-.25,1.2) and (.25,1.2) .. (.75,.65);
\end{scope}
\end{tikzpicture}%
}
\endgroup
\caption{
    \textbf{Case Study 10: Multi-view Grounding (Mammography).}
    Given a craniocaudal (CC) view with a red-boxed lesion and a mediolateral oblique (MLO) view, the task is to localize the same lesion in the MLO view.
    \textbf{Qwen3.5} predicts \texttt{[37,151,63,180]}, copying the reference box.
    \textbf{MedSeq-Grounder} predicts \texttt{[49,190,74,212]}.
    \textbf{\method{} (Ours)} relates the posterior position of the lesion on the CC view to the MLO projection, locates a dense, irregular mass anterior to the pectoral muscle shadow, and predicts
    \texttt{(62,153),(95,185)}.
}
\label{fig:qual_case10}
\end{figure*}

\begin{figure*}[p]
\centering
\begingroup
\definecolor{qInk}{HTML}{252629}\definecolor{qBlue}{HTML}{3B73A7}
\definecolor{qLine}{HTML}{B9BEC5}
\definecolor{qCream}{HTML}{FBF6E2}
\definecolor{qSage}{HTML}{EDF3E7}
\definecolor{qPeach}{HTML}{F5E3D5}
\definecolor{qSky}{HTML}{E3EDF7}
\definecolor{qBack}{HTML}{EEF2F5}
\resizebox{\textwidth}{!}{%
\begin{tikzpicture}[x=1mm,y=-1mm,every node/.style={font=\sffamily\fontsize{8.5}{10.2}\selectfont,text=qInk},box/.style={anchor=north west,align=left,inner sep=2mm,outer sep=0pt,rounded corners=2mm,draw=qLine!70,line width=.45pt,text width=164mm}]

\fill[qBack,rounded corners=2mm] (0,0) rectangle (170,86);
\node[anchor=north,inner sep=0pt] at (85,3){\adjincludegraphics[trim={0.13\width} {0.011\height} {0.652\width} {0.009\height},clip,width=164mm,height=80mm,keepaspectratio]{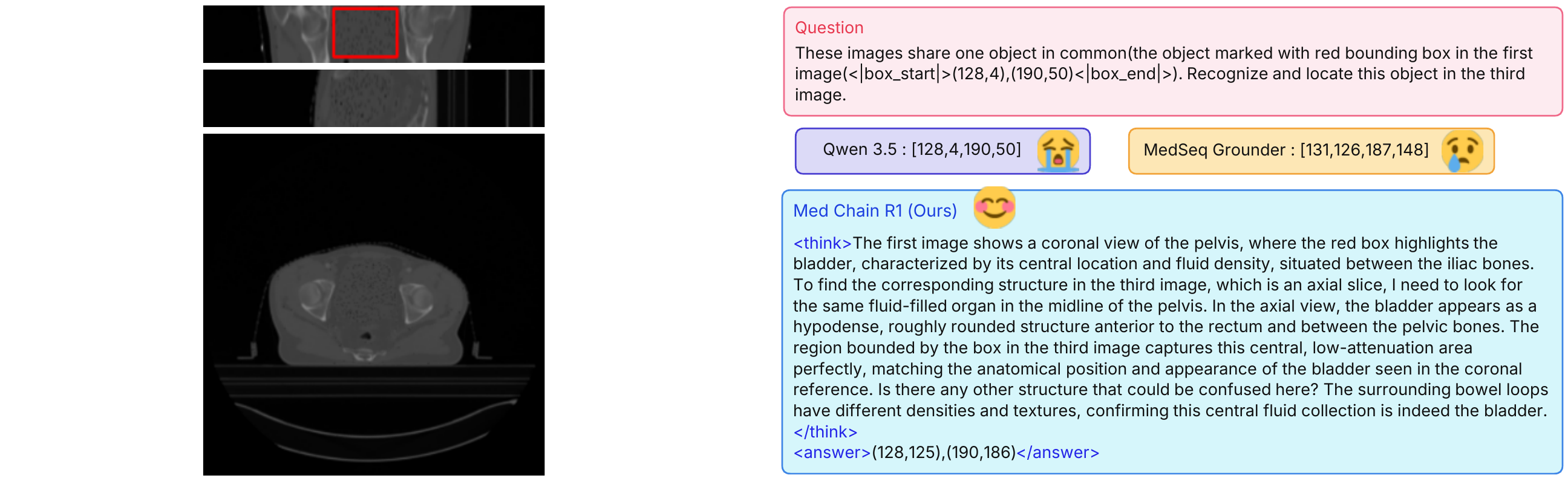}};
\node[box,fill=qCream] (question) at (0,90){\textbf{Question}\quad These images share one object in common(the object marked with red bounding box in the first image(\textless{}|box\_start|\textgreater{}(128,4),(190,50)\textless{}|box\_end|\textgreater{}). Recognize and locate this object in the third image.};
\node[box,fill=qSage] (qwen) at ([yshift=-2mm]question.south west){\tikz[x=1mm,y=-1mm,baseline=-.7mm]{\filldraw[fill=yellow!55!white,draw=qInk!80,line width=.35pt] (0,0) circle (1.9mm);\draw[qInk,line width=.35pt,line cap=round] (-1.15,-.35) .. controls (-.95,-.8) and (-.5,-.8) .. (-.3,-.35) (.3,-.35) .. controls (.5,-.8) and (.95,-.8) .. (1.15,-.35);\fill[qBlue!65!cyan,rounded corners=.12mm] (-1,-.2) rectangle (-.65,1.25);\fill[qBlue!65!cyan,rounded corners=.12mm] (.65,-.2) rectangle (1,1.25);\fill[qInk!85] (0,.8) ellipse (.3mm and .42mm);}\quad Qwen 3.5 : [128,4,190,50]};
\node[box,fill=qPeach] (medseq) at ([yshift=-1.5mm]qwen.south west){\tikz[x=1mm,y=-1mm,baseline=-.7mm]{\filldraw[fill=yellow!55!white,draw=qInk!80,line width=.35pt] (0,0) circle (1.9mm);\draw[qInk,line width=.35pt,line cap=round] (-1.15,-.35) .. controls (-.95,-.8) and (-.5,-.8) .. (-.3,-.35) (.3,-.35) .. controls (.5,-.8) and (.95,-.8) .. (1.15,-.35);\fill[qBlue!65!cyan,rounded corners=.12mm] (-1,-.2) rectangle (-.65,1.25);\fill[qBlue!65!cyan,rounded corners=.12mm] (.65,-.2) rectangle (1,1.25);\fill[qInk!85] (0,.8) ellipse (.3mm and .42mm);}\quad MedSeq Grounder : [131,126,187,148]};
\node[box,fill=qSky] (ours) at ([yshift=-2mm]medseq.south west){\tikz[x=1mm,y=-1mm,baseline=-.7mm]{\filldraw[fill=yellow!55!white,draw=qInk!80,line width=.35pt] (0,0) circle (1.9mm);\fill[qInk] (-.65,-.45) circle (.18mm);\fill[qInk] (.65,-.45) circle (.18mm);\draw[qInk,line width=.45pt,line cap=round] (-.85,.35) .. controls (-.45,1.2) and (.45,1.2) .. (.85,.35);}\quad\textbf{Med Chain R1 (Ours)}};
\node[box,fill=white,rounded corners=2mm,font=\rmfamily\fontsize{9.2}{11}\selectfont] (trace) at ([yshift=-0mm]ours.south west){\textless{}think\textgreater{}The first image shows a coronal view of the pelvis, where the red box highlights the bladder, characterized by its central location and fluid density, situated between the iliac bones. To find the corresponding structure in the third image, which is an axial slice, I need to look for the same fluid-filled organ in the midline of the pelvis. In the axial view, the bladder appears as a hypodense, roughly rounded structure anterior to the rectum and between the pelvic bones. The region bounded by the box in the third image captures this central, low-attenuation area perfectly, matching the anatomical position and appearance of the bladder seen in the coronal reference. Is there any other structure that could be confused here? The surrounding bowel loops have different densities and textures, confirming this central fluid collection is indeed the bladder.\textless{}/think\textgreater{} \textless{}answer\textgreater{}(128,125),(190,186)\textless{}/answer\textgreater{}};

\fill[qCream] ([yshift=-3mm]question.north east)--++(2mm,-1.5mm)--++(-2mm,-1.5mm)--cycle;
\fill[qSky] ([yshift=-3mm]ours.north east)--++(2mm,-1.5mm)--++(-2mm,-1.5mm)--cycle;
\begin{scope}[shift={([xshift=6mm,yshift=-4mm]question.north east)},scale=1.2]
\filldraw[fill=qPeach,draw=white,line width=1.2pt] (0,0) circle (3.5mm);
\fill[qInk!65] (-2,2.2) .. controls (-1.7,.5) and (1.7,.5) .. (2,2.2)--cycle;
\filldraw[fill=qCream,draw=qInk!65,line width=.35pt] (0,-.7) circle (1.25mm);
\draw[qInk!65,line width=.45pt] (-1.15,-1.15) .. controls (-.7,-2.25) and (.7,-2.25) .. (1.15,-1.15);
\fill[qInk] (-.4,-.7) circle (.12mm); \fill[qInk] (.4,-.7) circle (.12mm);
\end{scope}
\begin{scope}[shift={([xshift=6mm,yshift=-4mm]ours.north east)},scale=1.2]
\filldraw[fill=qSky,draw=white,line width=1.2pt] (0,0) circle (3.5mm);
\draw[qInk!65,line width=.45pt] (0,-1.5)--(0,-2.25);
\fill[qInk!65] (0,-2.4) circle (.3mm);
\filldraw[fill=white,draw=qInk!65,rounded corners=.65mm,line width=.45pt] (-2,-1.45) rectangle (2,1.55);
\fill[qInk!75] (-.8,-.25) circle (.3mm); \fill[qInk!75] (.8,-.25) circle (.3mm);
\draw[qInk!65,line width=.4pt] (-.75,.65) .. controls (-.25,1.2) and (.25,1.2) .. (.75,.65);
\end{scope}
\end{tikzpicture}%
}
\endgroup
\caption{
    \textbf{Case Study 11: Multi-view Grounding (CT).}
    Given a coronal pelvic CT image with the bladder marked by a red box, the task is to localize the same structure in the third image, an axial slice.
    \textbf{Qwen3.5} predicts \texttt{[128,4,190,50]}, copying the reference box.
    \textbf{MedSeq-Grounder} predicts \texttt{[131,126,187,148]}.
    \textbf{\method{} (Ours)} identifies the boxed structure as the bladder, searches the axial slice for the midline fluid-density organ anterior to the rectum, and predicts
    \texttt{(128,125),(190,186)}.
}
\label{fig:qual_case11}
\end{figure*}

\clearpage
\begin{figure*}[p]
\centering
\begingroup
\definecolor{qInk}{HTML}{252629}\definecolor{qBlue}{HTML}{3B73A7}
\definecolor{qLine}{HTML}{B9BEC5}
\definecolor{qCream}{HTML}{FBF6E2}
\definecolor{qSage}{HTML}{EDF3E7}
\definecolor{qPeach}{HTML}{F5E3D5}
\definecolor{qSky}{HTML}{E3EDF7}
\definecolor{qBack}{HTML}{EEF2F5}
\resizebox{\textwidth}{!}{%
\begin{tikzpicture}[x=1mm,y=-1mm,every node/.style={font=\sffamily\fontsize{9}{11}\selectfont,text=qInk},box/.style={anchor=north west,align=left,inner sep=2mm,outer sep=0pt,rounded corners=2mm,draw=qLine!70,line width=.45pt,text width=156mm}]
\fill[qBack,rounded corners=2mm] (0,0) rectangle (170,44);
\node[anchor=north,inner sep=0pt] at (41,2){\adjincludegraphics[trim={0.023\width} {0.508\height} {0.836\width} {0.01\height},clip,width=40.0mm,height=40mm,keepaspectratio]{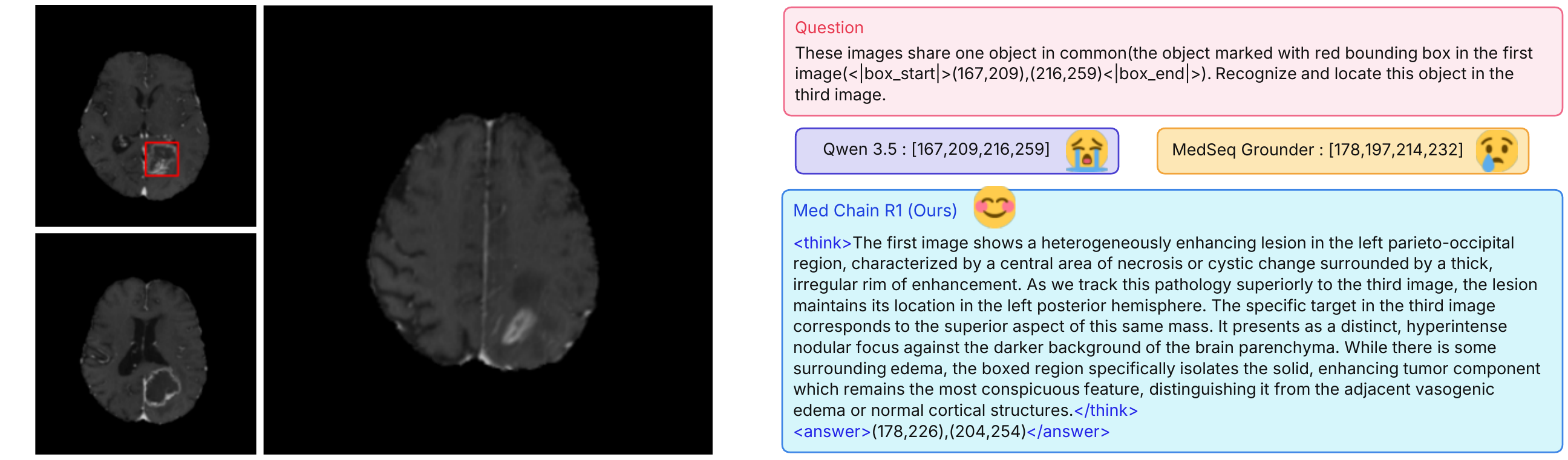}};
\node[anchor=north,inner sep=0pt] at (85,2){\adjincludegraphics[trim={0.023\width} {0.012\height} {0.836\width} {0.506\height},clip,width=40.0mm,height=40mm,keepaspectratio]{qual_case12.pdf}};
\node[anchor=north,inner sep=0pt] at (129,2){\adjincludegraphics[trim={0.168\width} {0.012\height} {0.545\width} {0.01\height},clip,width=40.0mm,height=40mm,keepaspectratio]{qual_case12.pdf}};
\node[box,fill=qCream] (question) at (0,47){\textbf{Question}\quad These images share one object in common(the object marked with red bounding box in the first image(\textless{}|box\_start|\textgreater{}(167,209),(216,259)\textless{}|box\_end|\textgreater{}). Recognize and locate this object in the third image.};
\node[box,fill=qSage,text width=73mm] (qwen) at ([yshift=-2mm]question.south west){\tikz[x=1mm,y=-1mm,baseline=-.7mm]{\filldraw[fill=yellow!55!white,draw=qInk!80,line width=.35pt] (0,0) circle (1.9mm);\draw[qInk,line width=.35pt,line cap=round] (-1.15,-.35) .. controls (-.95,-.8) and (-.5,-.8) .. (-.3,-.35) (.3,-.35) .. controls (.5,-.8) and (.95,-.8) .. (1.15,-.35);\fill[qBlue!65!cyan,rounded corners=.12mm] (-1,-.2) rectangle (-.65,1.25);\fill[qBlue!65!cyan,rounded corners=.12mm] (.65,-.2) rectangle (1,1.25);\fill[qInk!85] (0,.8) ellipse (.3mm and .42mm);}\quad Qwen 3.5 : [167,209,216,259]};
\node[box,fill=qPeach,text width=73mm] (medseq) at ([xshift=83mm,yshift=-2mm]question.south west){\tikz[x=1mm,y=-1mm,baseline=-.7mm]{\filldraw[fill=yellow!55!white,draw=qInk!80,line width=.35pt] (0,0) circle (1.9mm);\draw[qInk,line width=.35pt,line cap=round] (-1.15,-.35) .. controls (-.95,-.8) and (-.5,-.8) .. (-.3,-.35) (.3,-.35) .. controls (.5,-.8) and (.95,-.8) .. (1.15,-.35);\fill[qBlue!65!cyan,rounded corners=.12mm] (-1,-.2) rectangle (-.65,1.25);\fill[qBlue!65!cyan,rounded corners=.12mm] (.65,-.2) rectangle (1,1.25);\fill[qInk!85] (0,.8) ellipse (.3mm and .42mm);}\quad MedSeq Grounder : [178,197,214,232]};
\node[box,fill=qSky] (ours) at ([yshift=-2mm]qwen.south west){\tikz[x=1mm,y=-1mm,baseline=-.7mm]{\filldraw[fill=yellow!55!white,draw=qInk!80,line width=.35pt] (0,0) circle (1.9mm);\fill[qInk] (-.65,-.45) circle (.18mm);\fill[qInk] (.65,-.45) circle (.18mm);\draw[qInk,line width=.45pt,line cap=round] (-.85,.35) .. controls (-.45,1.2) and (.45,1.2) .. (.85,.35);}\quad\textbf{Med Chain R1 (Ours)}};
\node[box,fill=white,font=\rmfamily\fontsize{9.2}{11}\selectfont] at (ours.south west){\textless{}think\textgreater{}The first image shows a heterogeneously enhancing lesion in the left parieto-occipital region, characterized by a central area of necrosis or cystic change surrounded by a thick, irregular rim of enhancement. As we track this pathology superiorly to the third image, the lesion maintains its location in the left posterior hemisphere. The specific target in the third image corresponds to the superior aspect of this same mass. It presents as a distinct, hyperintense nodular focus against the darker background of the brain parenchyma. While there is some surrounding edema, the boxed region specifically isolates the solid, enhancing tumor component which remains the most conspicuous feature, distinguishing it from the adjacent vasogenic edema or normal cortical structures.\textless{}/think\textgreater{} \textless{}answer\textgreater{}(178,226),(204,254)\textless{}/answer\textgreater{}};

\fill[qCream] ([yshift=-3mm]question.north east)--++(2mm,-1.5mm)--++(-2mm,-1.5mm)--cycle;
\fill[qSky] ([yshift=-3mm]ours.north east)--++(2mm,-1.5mm)--++(-2mm,-1.5mm)--cycle;
\begin{scope}[shift={([xshift=6mm,yshift=-4mm]question.north east)},scale=1.2]
\filldraw[fill=qPeach,draw=white,line width=1.2pt] (0,0) circle (3.5mm);
\fill[qInk!65] (-2,2.2) .. controls (-1.7,.5) and (1.7,.5) .. (2,2.2)--cycle;
\filldraw[fill=qCream,draw=qInk!65,line width=.35pt] (0,-.7) circle (1.25mm);
\draw[qInk!65,line width=.45pt] (-1.15,-1.15) .. controls (-.7,-2.25) and (.7,-2.25) .. (1.15,-1.15);
\fill[qInk] (-.4,-.7) circle (.12mm); \fill[qInk] (.4,-.7) circle (.12mm);
\end{scope}
\begin{scope}[shift={([xshift=6mm,yshift=-4mm]ours.north east)},scale=1.2]
\filldraw[fill=qSky,draw=white,line width=1.2pt] (0,0) circle (3.5mm);
\draw[qInk!65,line width=.45pt] (0,-1.5)--(0,-2.25);
\fill[qInk!65] (0,-2.4) circle (.3mm);
\filldraw[fill=white,draw=qInk!65,rounded corners=.65mm,line width=.45pt] (-2,-1.45) rectangle (2,1.55);
\fill[qInk!75] (-.8,-.25) circle (.3mm); \fill[qInk!75] (.8,-.25) circle (.3mm);
\draw[qInk!65,line width=.4pt] (-.75,.65) .. controls (-.25,1.2) and (.25,1.2) .. (.75,.65);
\end{scope}
\end{tikzpicture}%
}
\endgroup
\caption{
    \textbf{Case Study 12: Object Tracking (Contrast-enhanced MRI).}
    Given three brain MRI slices with an enhancing lesion marked by a red box in the first image, the task is to localize the same lesion in the third image.
    \textbf{Qwen3.5} predicts \texttt{[167,209,216,259]}, copying the reference box.
    \textbf{MedSeq-Grounder} predicts \texttt{[178,197,214,232]}.
    \textbf{\method{} (Ours)} follows the left posterior lesion superiorly, separates the solid enhancing component from the surrounding vasogenic edema, and predicts
    \texttt{(178,226),(204,254)}.
}
\label{fig:qual_case12}
\end{figure*}

\begin{figure*}[p]
\centering
\begingroup
\definecolor{qInk}{HTML}{252629}\definecolor{qBlue}{HTML}{3B73A7}
\definecolor{qLine}{HTML}{B9BEC5}
\definecolor{qCream}{HTML}{FBF6E2}
\definecolor{qSage}{HTML}{EDF3E7}
\definecolor{qPeach}{HTML}{F5E3D5}
\definecolor{qSky}{HTML}{E3EDF7}
\definecolor{qBack}{HTML}{EEF2F5}
\resizebox{\textwidth}{!}{%
\begin{tikzpicture}[x=1mm,y=-1mm,every node/.style={font=\sffamily\fontsize{9}{11}\selectfont,text=qInk},box/.style={anchor=north west,align=left,inner sep=2mm,outer sep=0pt,rounded corners=2mm,draw=qLine!70,line width=.45pt,text width=156mm}]
\fill[qBack,rounded corners=2mm] (0,0) rectangle (170,28);
\node[anchor=north,inner sep=0pt] at (15,2){\adjincludegraphics[trim={0.003\width} {0.565\height} {0.917\width} {0.082\height},clip,width=24.0mm,height=24.0mm,keepaspectratio]{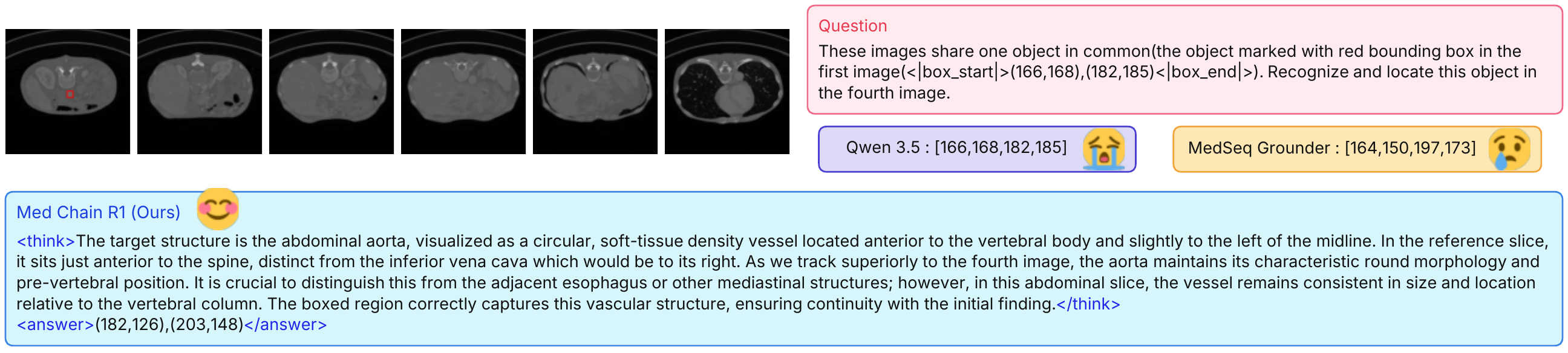}};
\node[anchor=north,inner sep=0pt] at (43,2){\adjincludegraphics[trim={0.088\width} {0.565\height} {0.833\width} {0.082\height},clip,width=24.0mm,height=24.0mm,keepaspectratio]{qual_case13.pdf}};
\node[anchor=north,inner sep=0pt] at (71,2){\adjincludegraphics[trim={0.172\width} {0.565\height} {0.749\width} {0.082\height},clip,width=24.0mm,height=24.0mm,keepaspectratio]{qual_case13.pdf}};
\node[anchor=north,inner sep=0pt] at (99,2){\adjincludegraphics[trim={0.256\width} {0.565\height} {0.665\width} {0.082\height},clip,width=24.0mm,height=24.0mm,keepaspectratio]{qual_case13.pdf}};
\node[anchor=north,inner sep=0pt] at (127,2){\adjincludegraphics[trim={0.34\width} {0.565\height} {0.581\width} {0.082\height},clip,width=24.0mm,height=24.0mm,keepaspectratio]{qual_case13.pdf}};
\node[anchor=north,inner sep=0pt] at (155,2){\adjincludegraphics[trim={0.424\width} {0.565\height} {0.496\width} {0.082\height},clip,width=24.0mm,height=24.0mm,keepaspectratio]{qual_case13.pdf}};
\node[box,fill=qCream] (question) at (0,31){\textbf{Question}\quad These images share one object in common(the object marked with red bounding box in the first image(\textless{}|box\_start|\textgreater{}(166,168),(182,185)\textless{}|box\_end|\textgreater{}). Recognize and locate this object in the fourth image.};
\node[box,fill=qSage,text width=73mm] (qwen) at ([yshift=-2mm]question.south west){\tikz[x=1mm,y=-1mm,baseline=-.7mm]{\filldraw[fill=yellow!55!white,draw=qInk!80,line width=.35pt] (0,0) circle (1.9mm);\draw[qInk,line width=.35pt,line cap=round] (-1.15,-.35) .. controls (-.95,-.8) and (-.5,-.8) .. (-.3,-.35) (.3,-.35) .. controls (.5,-.8) and (.95,-.8) .. (1.15,-.35);\fill[qBlue!65!cyan,rounded corners=.12mm] (-1,-.2) rectangle (-.65,1.25);\fill[qBlue!65!cyan,rounded corners=.12mm] (.65,-.2) rectangle (1,1.25);\fill[qInk!85] (0,.8) ellipse (.3mm and .42mm);}\quad Qwen 3.5 : [166,168,182,185]};
\node[box,fill=qPeach,text width=73mm] (medseq) at ([xshift=83mm,yshift=-2mm]question.south west){\tikz[x=1mm,y=-1mm,baseline=-.7mm]{\filldraw[fill=yellow!55!white,draw=qInk!80,line width=.35pt] (0,0) circle (1.9mm);\draw[qInk,line width=.35pt,line cap=round] (-1.15,-.35) .. controls (-.95,-.8) and (-.5,-.8) .. (-.3,-.35) (.3,-.35) .. controls (.5,-.8) and (.95,-.8) .. (1.15,-.35);\fill[qBlue!65!cyan,rounded corners=.12mm] (-1,-.2) rectangle (-.65,1.25);\fill[qBlue!65!cyan,rounded corners=.12mm] (.65,-.2) rectangle (1,1.25);\fill[qInk!85] (0,.8) ellipse (.3mm and .42mm);}\quad MedSeq Grounder : [164,150,197,173]};
\node[box,fill=qSky] (ours) at ([yshift=-2mm]qwen.south west){\tikz[x=1mm,y=-1mm,baseline=-.7mm]{\filldraw[fill=yellow!55!white,draw=qInk!80,line width=.35pt] (0,0) circle (1.9mm);\fill[qInk] (-.65,-.45) circle (.18mm);\fill[qInk] (.65,-.45) circle (.18mm);\draw[qInk,line width=.45pt,line cap=round] (-.85,.35) .. controls (-.45,1.2) and (.45,1.2) .. (.85,.35);}\quad\textbf{Med Chain R1 (Ours)}};
\node[box,fill=white,font=\rmfamily\fontsize{9.2}{11}\selectfont] at (ours.south west){\textless{}think\textgreater{}The target structure is the abdominal aorta, visualized as a circular, soft-tissue density vessel located anterior to the vertebral body and slightly to the left of the midline. In the reference slice, it sits just anterior to the spine, distinct from the inferior vena cava which would be to its right. As we track superiorly to the fourth image, the aorta maintains its characteristic round morphology and pre-vertebral position. It is crucial to distinguish this from the adjacent esophagus or other mediastinal structures; however, in this abdominal slice, the vessel remains consistent in size and location relative to the vertebral column. The boxed region correctly captures this vascular structure, ensuring continuity with the initial finding.\textless{}/think\textgreater{} \textless{}answer\textgreater{}(182,126),(203,148)\textless{}/answer\textgreater{}};

\fill[qCream] ([yshift=-3mm]question.north east)--++(2mm,-1.5mm)--++(-2mm,-1.5mm)--cycle;
\fill[qSky] ([yshift=-3mm]ours.north east)--++(2mm,-1.5mm)--++(-2mm,-1.5mm)--cycle;
\begin{scope}[shift={([xshift=6mm,yshift=-4mm]question.north east)},scale=1.2]
\filldraw[fill=qPeach,draw=white,line width=1.2pt] (0,0) circle (3.5mm);
\fill[qInk!65] (-2,2.2) .. controls (-1.7,.5) and (1.7,.5) .. (2,2.2)--cycle;
\filldraw[fill=qCream,draw=qInk!65,line width=.35pt] (0,-.7) circle (1.25mm);
\draw[qInk!65,line width=.45pt] (-1.15,-1.15) .. controls (-.7,-2.25) and (.7,-2.25) .. (1.15,-1.15);
\fill[qInk] (-.4,-.7) circle (.12mm); \fill[qInk] (.4,-.7) circle (.12mm);
\end{scope}
\begin{scope}[shift={([xshift=6mm,yshift=-4mm]ours.north east)},scale=1.2]
\filldraw[fill=qSky,draw=white,line width=1.2pt] (0,0) circle (3.5mm);
\draw[qInk!65,line width=.45pt] (0,-1.5)--(0,-2.25);
\fill[qInk!65] (0,-2.4) circle (.3mm);
\filldraw[fill=white,draw=qInk!65,rounded corners=.65mm,line width=.45pt] (-2,-1.45) rectangle (2,1.55);
\fill[qInk!75] (-.8,-.25) circle (.3mm); \fill[qInk!75] (.8,-.25) circle (.3mm);
\draw[qInk!65,line width=.4pt] (-.75,.65) .. controls (-.25,1.2) and (.25,1.2) .. (.75,.65);
\end{scope}
\end{tikzpicture}%
}
\endgroup
\caption{
    \textbf{Case Study 13: Object Tracking (Abdominal CT).}
    Given a sequence of axial CT slices with the abdominal aorta marked by a red box in the first image, the task is to localize the same structure in the fourth image.
    \textbf{Qwen3.5} predicts \texttt{[166,168,182,185]}, copying the reference box.
    \textbf{MedSeq-Grounder} predicts \texttt{[164,150,197,173]}.
    \textbf{\method{} (Ours)} tracks the round prevertebral vessel across slices, distinguishes it from the inferior vena cava and the esophagus, and predicts
    \texttt{(182,126),(203,148)}.
}
\label{fig:qual_case13}
\end{figure*}

\clearpage
\begin{figure*}[p]
\centering
\begingroup
\definecolor{qInk}{HTML}{252629}\definecolor{qBlue}{HTML}{3B73A7}
\definecolor{qLine}{HTML}{B9BEC5}
\definecolor{qCream}{HTML}{FBF6E2}
\definecolor{qSage}{HTML}{EDF3E7}
\definecolor{qPeach}{HTML}{F5E3D5}
\definecolor{qSky}{HTML}{E3EDF7}
\definecolor{qBack}{HTML}{EEF2F5}
\resizebox{\textwidth}{!}{%
\begin{tikzpicture}[x=1mm,y=-1mm,every node/.style={font=\sffamily\fontsize{8.5}{10.2}\selectfont,text=qInk},box/.style={anchor=north west,align=left,inner sep=2mm,outer sep=0pt,rounded corners=2mm,draw=qLine!70,line width=.45pt,text width=164mm}]

\fill[qBack,rounded corners=2mm] (0,0) rectangle (170,86);
\node[anchor=north,inner sep=0pt] at (43,3){\adjincludegraphics[trim={0.01\width} {0.011\height} {0.85\width} {0.01\height},clip,width=80mm,height=80mm,keepaspectratio]{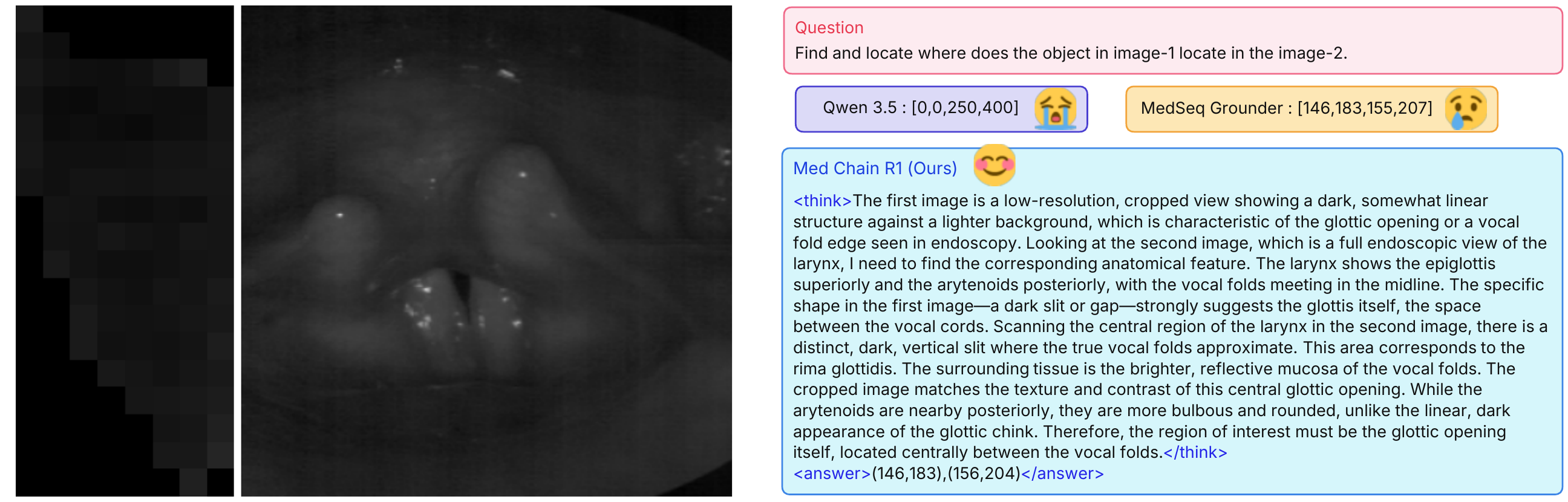}};
\node[anchor=north,inner sep=0pt] at (127,3){\adjincludegraphics[trim={0.154\width} {0.011\height} {0.533\width} {0.01\height},clip,width=80mm,height=80mm,keepaspectratio]{qual_case14.pdf}};
\node[box,fill=qCream] (question) at (0,90){\textbf{Question}\quad Find and locate where does the object in image-1 locate in the image-2.};
\node[box,fill=qSage] (qwen) at ([yshift=-2mm]question.south west){\tikz[x=1mm,y=-1mm,baseline=-.7mm]{\filldraw[fill=yellow!55!white,draw=qInk!80,line width=.35pt] (0,0) circle (1.9mm);\draw[qInk,line width=.35pt,line cap=round] (-1.15,-.35) .. controls (-.95,-.8) and (-.5,-.8) .. (-.3,-.35) (.3,-.35) .. controls (.5,-.8) and (.95,-.8) .. (1.15,-.35);\fill[qBlue!65!cyan,rounded corners=.12mm] (-1,-.2) rectangle (-.65,1.25);\fill[qBlue!65!cyan,rounded corners=.12mm] (.65,-.2) rectangle (1,1.25);\fill[qInk!85] (0,.8) ellipse (.3mm and .42mm);}\quad Qwen 3.5 : [0,0,250,400]};
\node[box,fill=qPeach] (medseq) at ([yshift=-1.5mm]qwen.south west){\tikz[x=1mm,y=-1mm,baseline=-.7mm]{\filldraw[fill=yellow!55!white,draw=qInk!80,line width=.35pt] (0,0) circle (1.9mm);\draw[qInk,line width=.35pt,line cap=round] (-1.15,-.35) .. controls (-.95,-.8) and (-.5,-.8) .. (-.3,-.35) (.3,-.35) .. controls (.5,-.8) and (.95,-.8) .. (1.15,-.35);\fill[qBlue!65!cyan,rounded corners=.12mm] (-1,-.2) rectangle (-.65,1.25);\fill[qBlue!65!cyan,rounded corners=.12mm] (.65,-.2) rectangle (1,1.25);\fill[qInk!85] (0,.8) ellipse (.3mm and .42mm);}\quad MedSeq Grounder : [146,183,155,207]};
\node[box,fill=qSky] (ours) at ([yshift=-2mm]medseq.south west){\tikz[x=1mm,y=-1mm,baseline=-.7mm]{\filldraw[fill=yellow!55!white,draw=qInk!80,line width=.35pt] (0,0) circle (1.9mm);\fill[qInk] (-.65,-.45) circle (.18mm);\fill[qInk] (.65,-.45) circle (.18mm);\draw[qInk,line width=.45pt,line cap=round] (-.85,.35) .. controls (-.45,1.2) and (.45,1.2) .. (.85,.35);}\quad\textbf{Med Chain R1 (Ours)}};
\node[box,fill=white,rounded corners=2mm,font=\rmfamily\fontsize{9.2}{11}\selectfont] (trace) at ([yshift=-0mm]ours.south west){\textless{}think\textgreater{}The first image is a low-resolution, cropped view showing a dark, somewhat linear structure against a lighter background, which is characteristic of the glottic opening or a vocal fold edge seen in endoscopy. Looking at the second image, which is a full endoscopic view of the larynx, I need to find the corresponding anatomical feature. The larynx shows the epiglottis superiorly and the arytenoids posteriorly, with the vocal folds meeting in the midline. The specific shape in the first image---a dark slit or gap---strongly suggests the glottis itself, the space between the vocal cords. Scanning the central region of the larynx in the second image, there is a distinct, dark, vertical slit where the true vocal folds approximate. This area corresponds to the rima glottidis. The surrounding tissue is the brighter, reflective mucosa of the vocal folds. The cropped image matches the texture and contrast of this central glottic opening. While the arytenoids are nearby posteriorly, they are more bulbous and rounded, unlike the linear, dark appearance of the glottic chink. Therefore, the region of interest must be the glottic opening itself, located centrally between the vocal folds.\textless{}/think\textgreater{} \textless{}answer\textgreater{}(146,183),(156,204)\textless{}/answer\textgreater{}};

\fill[qCream] ([yshift=-3mm]question.north east)--++(2mm,-1.5mm)--++(-2mm,-1.5mm)--cycle;
\fill[qSky] ([yshift=-3mm]ours.north east)--++(2mm,-1.5mm)--++(-2mm,-1.5mm)--cycle;
\begin{scope}[shift={([xshift=6mm,yshift=-4mm]question.north east)},scale=1.2]
\filldraw[fill=qPeach,draw=white,line width=1.2pt] (0,0) circle (3.5mm);
\fill[qInk!65] (-2,2.2) .. controls (-1.7,.5) and (1.7,.5) .. (2,2.2)--cycle;
\filldraw[fill=qCream,draw=qInk!65,line width=.35pt] (0,-.7) circle (1.25mm);
\draw[qInk!65,line width=.45pt] (-1.15,-1.15) .. controls (-.7,-2.25) and (.7,-2.25) .. (1.15,-1.15);
\fill[qInk] (-.4,-.7) circle (.12mm); \fill[qInk] (.4,-.7) circle (.12mm);
\end{scope}
\begin{scope}[shift={([xshift=6mm,yshift=-4mm]ours.north east)},scale=1.2]
\filldraw[fill=qSky,draw=white,line width=1.2pt] (0,0) circle (3.5mm);
\draw[qInk!65,line width=.45pt] (0,-1.5)--(0,-2.25);
\fill[qInk!65] (0,-2.4) circle (.3mm);
\filldraw[fill=white,draw=qInk!65,rounded corners=.65mm,line width=.45pt] (-2,-1.45) rectangle (2,1.55);
\fill[qInk!75] (-.8,-.25) circle (.3mm); \fill[qInk!75] (.8,-.25) circle (.3mm);
\draw[qInk!65,line width=.4pt] (-.75,.65) .. controls (-.25,1.2) and (.25,1.2) .. (.75,.65);
\end{scope}
\end{tikzpicture}%
}
\endgroup
\caption{
    \textbf{Case Study 14: Concept Grounding (Laryngoscopy).}
    Given a cropped endoscopic region (image~1) and a full laryngoscopic view (image~2), the task is to localize the cropped content in the full view.
    \textbf{Qwen3.5} predicts \texttt{[0,0,250,400]}.
    \textbf{MedSeq-Grounder} predicts \texttt{[146,183,155,207]}.
    \textbf{\method{} (Ours)} matches the dark linear structure in the crop to the glottic opening between the vocal folds, distinguishes it from the more rounded arytenoids, and predicts
    \texttt{(146,183),(156,204)}.
}
\label{fig:qual_case14}
\end{figure*}

\begin{figure*}[p]
\centering
\begingroup
\definecolor{qInk}{HTML}{252629}\definecolor{qBlue}{HTML}{3B73A7}
\definecolor{qLine}{HTML}{B9BEC5}
\definecolor{qCream}{HTML}{FBF6E2}
\definecolor{qSage}{HTML}{EDF3E7}
\definecolor{qPeach}{HTML}{F5E3D5}
\definecolor{qSky}{HTML}{E3EDF7}
\definecolor{qBack}{HTML}{EEF2F5}
\resizebox{\textwidth}{!}{%
\begin{tikzpicture}[x=1mm,y=-1mm,every node/.style={font=\sffamily\fontsize{8.5}{10.2}\selectfont,text=qInk},box/.style={anchor=north west,align=left,inner sep=2mm,outer sep=0pt,rounded corners=2mm,draw=qLine!70,line width=.45pt,text width=164mm}]

\fill[qBack,rounded corners=2mm] (0,0) rectangle (170,86);
\node[anchor=north,inner sep=0pt] at (43,3){\adjincludegraphics[trim={0.003\width} {0.178\height} {0.737\width} {0.176\height},clip,width=80mm,height=80mm,keepaspectratio]{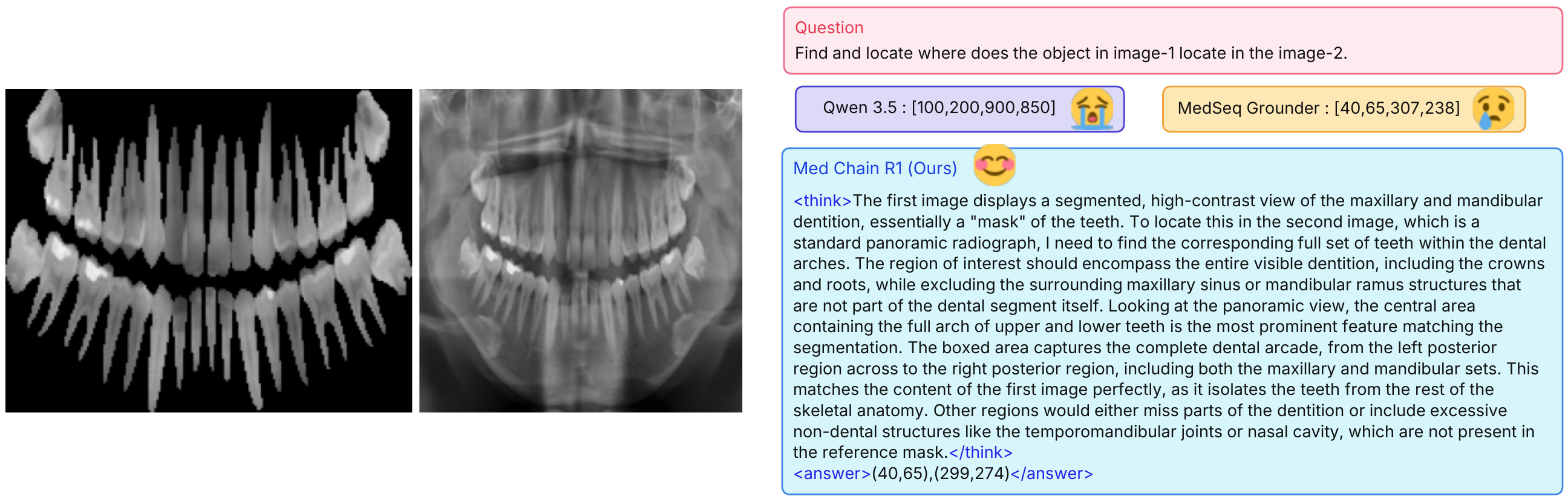}};
\node[anchor=north,inner sep=0pt] at (127,3){\adjincludegraphics[trim={0.268\width} {0.178\height} {0.526\width} {0.176\height},clip,width=80mm,height=80mm,keepaspectratio]{qual_case15.pdf}};
\node[box,fill=qCream] (question) at (0,90){\textbf{Question}\quad Find and locate where does the object in image-1 locate in the image-2.};
\node[box,fill=qSage] (qwen) at ([yshift=-2mm]question.south west){\tikz[x=1mm,y=-1mm,baseline=-.7mm]{\filldraw[fill=yellow!55!white,draw=qInk!80,line width=.35pt] (0,0) circle (1.9mm);\draw[qInk,line width=.35pt,line cap=round] (-1.15,-.35) .. controls (-.95,-.8) and (-.5,-.8) .. (-.3,-.35) (.3,-.35) .. controls (.5,-.8) and (.95,-.8) .. (1.15,-.35);\fill[qBlue!65!cyan,rounded corners=.12mm] (-1,-.2) rectangle (-.65,1.25);\fill[qBlue!65!cyan,rounded corners=.12mm] (.65,-.2) rectangle (1,1.25);\fill[qInk!85] (0,.8) ellipse (.3mm and .42mm);}\quad Qwen 3.5 : [100,200,900,850]};
\node[box,fill=qPeach] (medseq) at ([yshift=-1.5mm]qwen.south west){\tikz[x=1mm,y=-1mm,baseline=-.7mm]{\filldraw[fill=yellow!55!white,draw=qInk!80,line width=.35pt] (0,0) circle (1.9mm);\draw[qInk,line width=.35pt,line cap=round] (-1.15,-.35) .. controls (-.95,-.8) and (-.5,-.8) .. (-.3,-.35) (.3,-.35) .. controls (.5,-.8) and (.95,-.8) .. (1.15,-.35);\fill[qBlue!65!cyan,rounded corners=.12mm] (-1,-.2) rectangle (-.65,1.25);\fill[qBlue!65!cyan,rounded corners=.12mm] (.65,-.2) rectangle (1,1.25);\fill[qInk!85] (0,.8) ellipse (.3mm and .42mm);}\quad MedSeq Grounder : [40,65,307,238]};
\node[box,fill=qSky] (ours) at ([yshift=-2mm]medseq.south west){\tikz[x=1mm,y=-1mm,baseline=-.7mm]{\filldraw[fill=yellow!55!white,draw=qInk!80,line width=.35pt] (0,0) circle (1.9mm);\fill[qInk] (-.65,-.45) circle (.18mm);\fill[qInk] (.65,-.45) circle (.18mm);\draw[qInk,line width=.45pt,line cap=round] (-.85,.35) .. controls (-.45,1.2) and (.45,1.2) .. (.85,.35);}\quad\textbf{Med Chain R1 (Ours)}};
\node[box,fill=white,rounded corners=2mm,font=\rmfamily\fontsize{9.2}{11}\selectfont] (trace) at ([yshift=-0mm]ours.south west){\textless{}think\textgreater{}The first image displays a segmented, high-contrast view of the maxillary and mandibular dentition, essentially a ``mask'' of the teeth. To locate this in the second image, which is a standard panoramic radiograph, I need to find the corresponding full set of teeth within the dental arches. The region of interest should encompass the entire visible dentition, including the crowns and roots, while excluding the surrounding maxillary sinus or mandibular ramus structures that are not part of the dental segment itself. Looking at the panoramic view, the central area containing the full arch of upper and lower teeth is the most prominent feature matching the segmentation. The boxed area captures the complete dental arcade, from the left posterior region across to the right posterior region, including both the maxillary and mandibular sets. This matches the content of the first image perfectly, as it isolates the teeth from the rest of the skeletal anatomy. Other regions would either miss parts of the dentition or include excessive non-dental structures like the temporomandibular joints or nasal cavity, which are not present in the reference mask.\textless{}/think\textgreater{} \textless{}answer\textgreater{}(40,65),(299,274)\textless{}/answer\textgreater{}};

\fill[qCream] ([yshift=-3mm]question.north east)--++(2mm,-1.5mm)--++(-2mm,-1.5mm)--cycle;
\fill[qSky] ([yshift=-3mm]ours.north east)--++(2mm,-1.5mm)--++(-2mm,-1.5mm)--cycle;
\begin{scope}[shift={([xshift=6mm,yshift=-4mm]question.north east)},scale=1.2]
\filldraw[fill=qPeach,draw=white,line width=1.2pt] (0,0) circle (3.5mm);
\fill[qInk!65] (-2,2.2) .. controls (-1.7,.5) and (1.7,.5) .. (2,2.2)--cycle;
\filldraw[fill=qCream,draw=qInk!65,line width=.35pt] (0,-.7) circle (1.25mm);
\draw[qInk!65,line width=.45pt] (-1.15,-1.15) .. controls (-.7,-2.25) and (.7,-2.25) .. (1.15,-1.15);
\fill[qInk] (-.4,-.7) circle (.12mm); \fill[qInk] (.4,-.7) circle (.12mm);
\end{scope}
\begin{scope}[shift={([xshift=6mm,yshift=-4mm]ours.north east)},scale=1.2]
\filldraw[fill=qSky,draw=white,line width=1.2pt] (0,0) circle (3.5mm);
\draw[qInk!65,line width=.45pt] (0,-1.5)--(0,-2.25);
\fill[qInk!65] (0,-2.4) circle (.3mm);
\filldraw[fill=white,draw=qInk!65,rounded corners=.65mm,line width=.45pt] (-2,-1.45) rectangle (2,1.55);
\fill[qInk!75] (-.8,-.25) circle (.3mm); \fill[qInk!75] (.8,-.25) circle (.3mm);
\draw[qInk!65,line width=.4pt] (-.75,.65) .. controls (-.25,1.2) and (.25,1.2) .. (.75,.65);
\end{scope}
\end{tikzpicture}%
}
\endgroup
\caption{
    \textbf{Case Study 15: Concept Grounding (Dental Panoramic Radiograph).}
    Given a segmentation-style mask of the dentition (image~1) and a panoramic radiograph (image~2), the task is to localize the masked content in the radiograph.
    \textbf{Qwen3.5} predicts \texttt{[100,200,900,850]}.
    \textbf{MedSeq-Grounder} predicts \texttt{[40,65,307,238]}.
    \textbf{\method{} (Ours)} matches the mask to the full maxillary and mandibular dental arcade, excludes the sinus, mandibular ramus and temporomandibular joints, and predicts
    \texttt{(40,65),(299,274)}.
}
\label{fig:qual_case15}
\end{figure*}

\clearpage
\begin{figure*}[p]
\centering
\begingroup
\definecolor{qInk}{HTML}{252629}\definecolor{qBlue}{HTML}{3B73A7}
\definecolor{qLine}{HTML}{B9BEC5}
\definecolor{qCream}{HTML}{FBF6E2}
\definecolor{qSage}{HTML}{EDF3E7}
\definecolor{qPeach}{HTML}{F5E3D5}
\definecolor{qSky}{HTML}{E3EDF7}
\definecolor{qBack}{HTML}{EEF2F5}
\resizebox{\textwidth}{!}{%
\begin{tikzpicture}[x=1mm,y=-1mm,every node/.style={font=\sffamily\fontsize{9}{11}\selectfont,text=qInk},box/.style={anchor=north west,align=left,inner sep=2mm,outer sep=0pt,rounded corners=2mm,draw=qLine!70,line width=.45pt,text width=156mm}]
\fill[qBack,rounded corners=2mm] (0,0) rectangle (170,42);
\node[anchor=north,inner sep=0pt] at (22,2){\adjincludegraphics[trim={0.027\width} {0.551\height} {0.863\width} {0.012\height},clip,width=38.0mm,height=38.0mm,keepaspectratio]{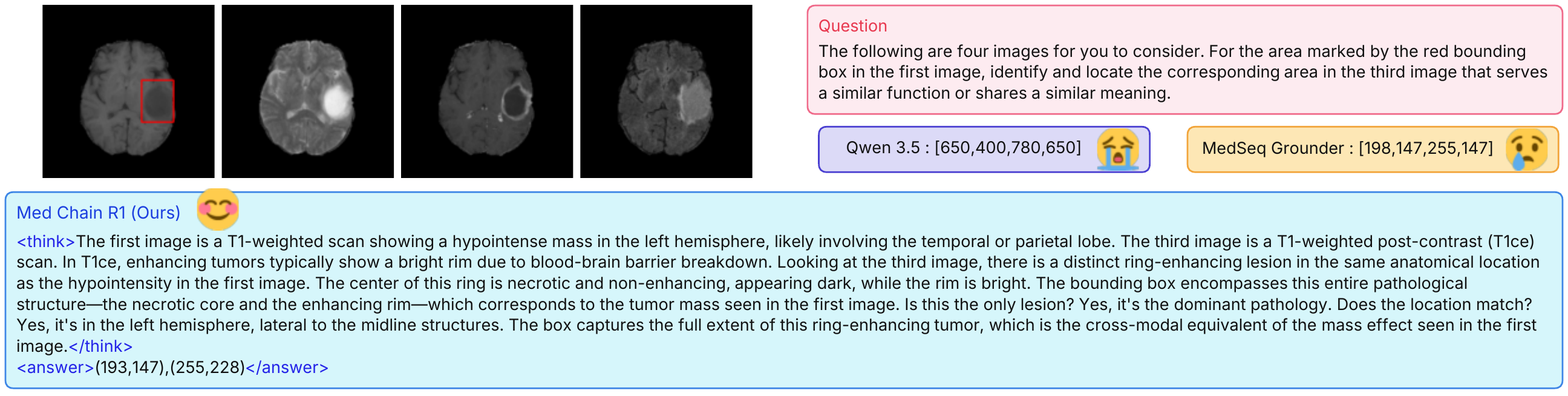}};
\node[anchor=north,inner sep=0pt] at (64,2){\adjincludegraphics[trim={0.141\width} {0.551\height} {0.748\width} {0.012\height},clip,width=38.0mm,height=38.0mm,keepaspectratio]{qual_case16.pdf}};
\node[anchor=north,inner sep=0pt] at (106,2){\adjincludegraphics[trim={0.256\width} {0.551\height} {0.634\width} {0.012\height},clip,width=38.0mm,height=38.0mm,keepaspectratio]{qual_case16.pdf}};
\node[anchor=north,inner sep=0pt] at (148,2){\adjincludegraphics[trim={0.37\width} {0.551\height} {0.52\width} {0.012\height},clip,width=38.0mm,height=38.0mm,keepaspectratio]{qual_case16.pdf}};
\node[box,fill=qCream] (question) at (0,45){\textbf{Question}\quad The following are four images for you to consider. For the area marked by the red bounding box in the first image, identify and locate the corresponding area in the third image that serves a similar function or shares a similar meaning.};
\node[box,fill=qSage,text width=73mm] (qwen) at ([yshift=-2mm]question.south west){\tikz[x=1mm,y=-1mm,baseline=-.7mm]{\filldraw[fill=yellow!55!white,draw=qInk!80,line width=.35pt] (0,0) circle (1.9mm);\draw[qInk,line width=.35pt,line cap=round] (-1.15,-.35) .. controls (-.95,-.8) and (-.5,-.8) .. (-.3,-.35) (.3,-.35) .. controls (.5,-.8) and (.95,-.8) .. (1.15,-.35);\fill[qBlue!65!cyan,rounded corners=.12mm] (-1,-.2) rectangle (-.65,1.25);\fill[qBlue!65!cyan,rounded corners=.12mm] (.65,-.2) rectangle (1,1.25);\fill[qInk!85] (0,.8) ellipse (.3mm and .42mm);}\quad Qwen 3.5 : [650,400,780,650]};
\node[box,fill=qPeach,text width=73mm] (medseq) at ([xshift=83mm,yshift=-2mm]question.south west){\tikz[x=1mm,y=-1mm,baseline=-.7mm]{\filldraw[fill=yellow!55!white,draw=qInk!80,line width=.35pt] (0,0) circle (1.9mm);\draw[qInk,line width=.35pt,line cap=round] (-1.15,-.35) .. controls (-.95,-.8) and (-.5,-.8) .. (-.3,-.35) (.3,-.35) .. controls (.5,-.8) and (.95,-.8) .. (1.15,-.35);\fill[qBlue!65!cyan,rounded corners=.12mm] (-1,-.2) rectangle (-.65,1.25);\fill[qBlue!65!cyan,rounded corners=.12mm] (.65,-.2) rectangle (1,1.25);\fill[qInk!85] (0,.8) ellipse (.3mm and .42mm);}\quad MedSeq Grounder : [198,147,255,147]};
\node[box,fill=qSky] (ours) at ([yshift=-2mm]qwen.south west){\tikz[x=1mm,y=-1mm,baseline=-.7mm]{\filldraw[fill=yellow!55!white,draw=qInk!80,line width=.35pt] (0,0) circle (1.9mm);\fill[qInk] (-.65,-.45) circle (.18mm);\fill[qInk] (.65,-.45) circle (.18mm);\draw[qInk,line width=.45pt,line cap=round] (-.85,.35) .. controls (-.45,1.2) and (.45,1.2) .. (.85,.35);}\quad\textbf{Med Chain R1 (Ours)}};
\node[box,fill=white,font=\rmfamily\fontsize{9.2}{11}\selectfont] at (ours.south west){\textless{}think\textgreater{}The first image is a T1-weighted scan showing a hypointense mass in the left hemisphere, likely involving the temporal or parietal lobe. The third image is a T1-weighted post-contrast (T1ce) scan. In T1ce, enhancing tumors typically show a bright rim due to blood-brain barrier breakdown. Looking at the third image, there is a distinct ring-enhancing lesion in the same anatomical location as the hypointensity in the first image. The center of this ring is necrotic and non-enhancing, appearing dark, while the rim is bright. The bounding box encompasses this entire pathological structure---the necrotic core and the enhancing rim---which corresponds to the tumor mass seen in the first image. Is this the only lesion? Yes, it's the dominant pathology. Does the location match? Yes, it's in the left hemisphere, lateral to the midline structures. The box captures the full extent of this ring-enhancing tumor, which is the cross-modal equivalent of the mass effect seen in the first image.\textless{}/think\textgreater{} \textless{}answer\textgreater{}(193,147),(255,228)\textless{}/answer\textgreater{}};

\fill[qCream] ([yshift=-3mm]question.north east)--++(2mm,-1.5mm)--++(-2mm,-1.5mm)--cycle;
\fill[qSky] ([yshift=-3mm]ours.north east)--++(2mm,-1.5mm)--++(-2mm,-1.5mm)--cycle;
\begin{scope}[shift={([xshift=6mm,yshift=-4mm]question.north east)},scale=1.2]
\filldraw[fill=qPeach,draw=white,line width=1.2pt] (0,0) circle (3.5mm);
\fill[qInk!65] (-2,2.2) .. controls (-1.7,.5) and (1.7,.5) .. (2,2.2)--cycle;
\filldraw[fill=qCream,draw=qInk!65,line width=.35pt] (0,-.7) circle (1.25mm);
\draw[qInk!65,line width=.45pt] (-1.15,-1.15) .. controls (-.7,-2.25) and (.7,-2.25) .. (1.15,-1.15);
\fill[qInk] (-.4,-.7) circle (.12mm); \fill[qInk] (.4,-.7) circle (.12mm);
\end{scope}
\begin{scope}[shift={([xshift=6mm,yshift=-4mm]ours.north east)},scale=1.2]
\filldraw[fill=qSky,draw=white,line width=1.2pt] (0,0) circle (3.5mm);
\draw[qInk!65,line width=.45pt] (0,-1.5)--(0,-2.25);
\fill[qInk!65] (0,-2.4) circle (.3mm);
\filldraw[fill=white,draw=qInk!65,rounded corners=.65mm,line width=.45pt] (-2,-1.45) rectangle (2,1.55);
\fill[qInk!75] (-.8,-.25) circle (.3mm); \fill[qInk!75] (.8,-.25) circle (.3mm);
\draw[qInk!65,line width=.4pt] (-.75,.65) .. controls (-.25,1.2) and (.25,1.2) .. (.75,.65);
\end{scope}
\end{tikzpicture}%
}
\endgroup
\caption{
    \textbf{Case Study 16: Cross-modal Grounding (Brain MRI, T1 to T1ce).}
    Given four MRI sequences with a hypointense mass marked by a red box on the T1-weighted image, the task is to localize the corresponding region in the third image, a post-contrast T1 (T1ce) scan.
    \textbf{Qwen3.5} predicts \texttt{[650,400,780,650]}.
    \textbf{MedSeq-Grounder} predicts \texttt{[198,147,255,147]}, a zero-height box.
    \textbf{\method{} (Ours)} relates the T1 hypointensity to a ring-enhancing lesion with a necrotic core at the same location on T1ce, and predicts
    \texttt{(193,147),(255,228)}.
}
\label{fig:qual_case16}
\end{figure*}

\begin{figure*}[p]
\centering
\begingroup
\definecolor{qInk}{HTML}{252629}\definecolor{qBlue}{HTML}{3B73A7}
\definecolor{qLine}{HTML}{B9BEC5}
\definecolor{qCream}{HTML}{FBF6E2}
\definecolor{qSage}{HTML}{EDF3E7}
\definecolor{qPeach}{HTML}{F5E3D5}
\definecolor{qSky}{HTML}{E3EDF7}
\definecolor{qBack}{HTML}{EEF2F5}
\resizebox{\textwidth}{!}{%
\begin{tikzpicture}[x=1mm,y=-1mm,every node/.style={font=\sffamily\fontsize{8.5}{10.2}\selectfont,text=qInk},box/.style={anchor=north west,align=left,inner sep=2mm,outer sep=0pt,rounded corners=2mm,draw=qLine!70,line width=.45pt,text width=164mm}]

\fill[qBack,rounded corners=2mm] (0,0) rectangle (170,86);
\node[anchor=north,inner sep=0pt] at (43,3){\adjincludegraphics[trim={0.003\width} {0.103\height} {0.763\width} {0.102\height},clip,width=80mm,height=80mm,keepaspectratio]{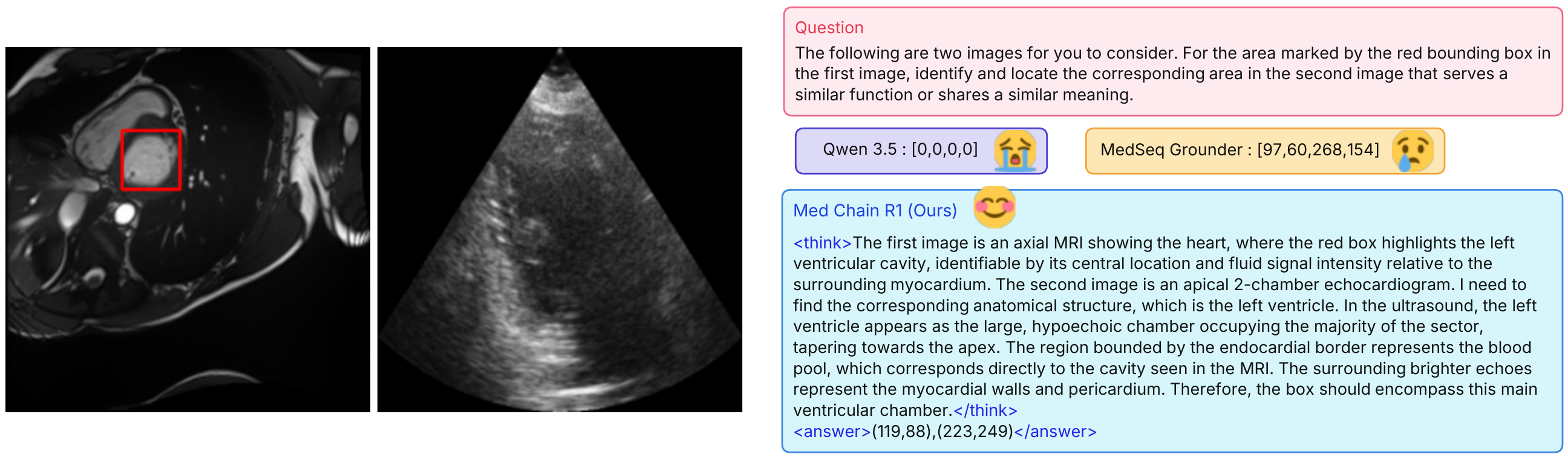}};
\node[anchor=north,inner sep=0pt] at (127,3){\adjincludegraphics[trim={0.241\width} {0.103\height} {0.526\width} {0.102\height},clip,width=80mm,height=80mm,keepaspectratio]{qual_case17.pdf}};
\node[box,fill=qCream] (question) at (0,90){\textbf{Question}\quad The following are two images for you to consider. For the area marked by the red bounding box in the first image, identify and locate the corresponding area in the second image that serves a similar function or shares a similar meaning.};
\node[box,fill=qSage] (qwen) at ([yshift=-2mm]question.south west){\tikz[x=1mm,y=-1mm,baseline=-.7mm]{\filldraw[fill=yellow!55!white,draw=qInk!80,line width=.35pt] (0,0) circle (1.9mm);\draw[qInk,line width=.35pt,line cap=round] (-1.15,-.35) .. controls (-.95,-.8) and (-.5,-.8) .. (-.3,-.35) (.3,-.35) .. controls (.5,-.8) and (.95,-.8) .. (1.15,-.35);\fill[qBlue!65!cyan,rounded corners=.12mm] (-1,-.2) rectangle (-.65,1.25);\fill[qBlue!65!cyan,rounded corners=.12mm] (.65,-.2) rectangle (1,1.25);\fill[qInk!85] (0,.8) ellipse (.3mm and .42mm);}\quad Qwen 3.5 : [0,0,0,0]};
\node[box,fill=qPeach] (medseq) at ([yshift=-1.5mm]qwen.south west){\tikz[x=1mm,y=-1mm,baseline=-.7mm]{\filldraw[fill=yellow!55!white,draw=qInk!80,line width=.35pt] (0,0) circle (1.9mm);\draw[qInk,line width=.35pt,line cap=round] (-1.15,-.35) .. controls (-.95,-.8) and (-.5,-.8) .. (-.3,-.35) (.3,-.35) .. controls (.5,-.8) and (.95,-.8) .. (1.15,-.35);\fill[qBlue!65!cyan,rounded corners=.12mm] (-1,-.2) rectangle (-.65,1.25);\fill[qBlue!65!cyan,rounded corners=.12mm] (.65,-.2) rectangle (1,1.25);\fill[qInk!85] (0,.8) ellipse (.3mm and .42mm);}\quad MedSeq Grounder : [97,60,268,154]};
\node[box,fill=qSky] (ours) at ([yshift=-2mm]medseq.south west){\tikz[x=1mm,y=-1mm,baseline=-.7mm]{\filldraw[fill=yellow!55!white,draw=qInk!80,line width=.35pt] (0,0) circle (1.9mm);\fill[qInk] (-.65,-.45) circle (.18mm);\fill[qInk] (.65,-.45) circle (.18mm);\draw[qInk,line width=.45pt,line cap=round] (-.85,.35) .. controls (-.45,1.2) and (.45,1.2) .. (.85,.35);}\quad\textbf{Med Chain R1 (Ours)}};
\node[box,fill=white,rounded corners=2mm,font=\rmfamily\fontsize{9.2}{11}\selectfont] (trace) at ([yshift=-0mm]ours.south west){\textless{}think\textgreater{}The first image is an axial MRI showing the heart, where the red box highlights the left ventricular cavity, identifiable by its central location and fluid signal intensity relative to the surrounding myocardium. The second image is an apical 2-chamber echocardiogram. I need to find the corresponding anatomical structure, which is the left ventricle. In the ultrasound, the left ventricle appears as the large, hypoechoic chamber occupying the majority of the sector, tapering towards the apex. The region bounded by the endocardial border represents the blood pool, which corresponds directly to the cavity seen in the MRI. The surrounding brighter echoes represent the myocardial walls and pericardium. Therefore, the box should encompass this main ventricular chamber.\textless{}/think\textgreater{} \textless{}answer\textgreater{}(119,88),(223,249)\textless{}/answer\textgreater{}};

\fill[qCream] ([yshift=-3mm]question.north east)--++(2mm,-1.5mm)--++(-2mm,-1.5mm)--cycle;
\fill[qSky] ([yshift=-3mm]ours.north east)--++(2mm,-1.5mm)--++(-2mm,-1.5mm)--cycle;
\begin{scope}[shift={([xshift=6mm,yshift=-4mm]question.north east)},scale=1.2]
\filldraw[fill=qPeach,draw=white,line width=1.2pt] (0,0) circle (3.5mm);
\fill[qInk!65] (-2,2.2) .. controls (-1.7,.5) and (1.7,.5) .. (2,2.2)--cycle;
\filldraw[fill=qCream,draw=qInk!65,line width=.35pt] (0,-.7) circle (1.25mm);
\draw[qInk!65,line width=.45pt] (-1.15,-1.15) .. controls (-.7,-2.25) and (.7,-2.25) .. (1.15,-1.15);
\fill[qInk] (-.4,-.7) circle (.12mm); \fill[qInk] (.4,-.7) circle (.12mm);
\end{scope}
\begin{scope}[shift={([xshift=6mm,yshift=-4mm]ours.north east)},scale=1.2]
\filldraw[fill=qSky,draw=white,line width=1.2pt] (0,0) circle (3.5mm);
\draw[qInk!65,line width=.45pt] (0,-1.5)--(0,-2.25);
\fill[qInk!65] (0,-2.4) circle (.3mm);
\filldraw[fill=white,draw=qInk!65,rounded corners=.65mm,line width=.45pt] (-2,-1.45) rectangle (2,1.55);
\fill[qInk!75] (-.8,-.25) circle (.3mm); \fill[qInk!75] (.8,-.25) circle (.3mm);
\draw[qInk!65,line width=.4pt] (-.75,.65) .. controls (-.25,1.2) and (.25,1.2) .. (.75,.65);
\end{scope}
\end{tikzpicture}%
}
\endgroup
\caption{
    \textbf{Case Study 17: Cross-modal Grounding (Cardiac MRI to Echocardiography).}
    Given an axial cardiac MRI image with the left ventricular cavity marked by a red box and an apical two-chamber echocardiogram, the task is to localize the corresponding region in the echocardiogram.
    \textbf{Qwen3.5} predicts \texttt{[0,0,0,0]}, an empty box.
    \textbf{MedSeq-Grounder} predicts \texttt{[97,60,268,154]}.
    \textbf{\method{} (Ours)} identifies the boxed region as the left ventricular cavity, locates the corresponding hypoechoic chamber bounded by the endocardial border in the ultrasound, and predicts
    \texttt{(119,88),(223,249)}.
}
\label{fig:qual_case17}
\end{figure*}

\clearpage
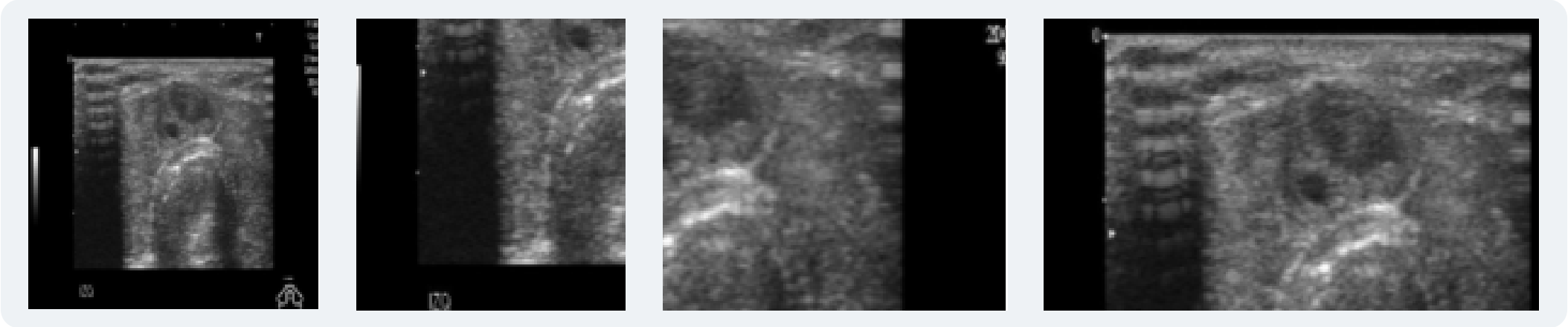
\begin{figure*}[p]
\centering
\begingroup
\definecolor{qInk}{HTML}{252629}\definecolor{qBlue}{HTML}{3B73A7}
\definecolor{qLine}{HTML}{B9BEC5}
\definecolor{qCream}{HTML}{FBF6E2}
\definecolor{qSage}{HTML}{EDF3E7}
\definecolor{qPeach}{HTML}{F5E3D5}
\definecolor{qSky}{HTML}{E3EDF7}
\definecolor{qBack}{HTML}{EEF2F5}
\resizebox{\textwidth}{!}{%
\begin{tikzpicture}[x=1mm,y=-1mm,every node/.style={font=\sffamily\fontsize{9}{11}\selectfont,text=qInk},box/.style={anchor=north west,align=left,inner sep=2mm,outer sep=0pt,rounded corners=2mm,draw=qLine!70,line width=.45pt,text width=156mm}]
\fill[qBack,rounded corners=2mm] (0,0) rectangle (170,29.3);
\node[anchor=north,inner sep=0pt] at (15.6,2){\adjincludegraphics[trim={0.003\width} {0.536\height} {0.915\width} {0.055\height},clip,width=25.3mm,height=25.3mm,keepaspectratio]{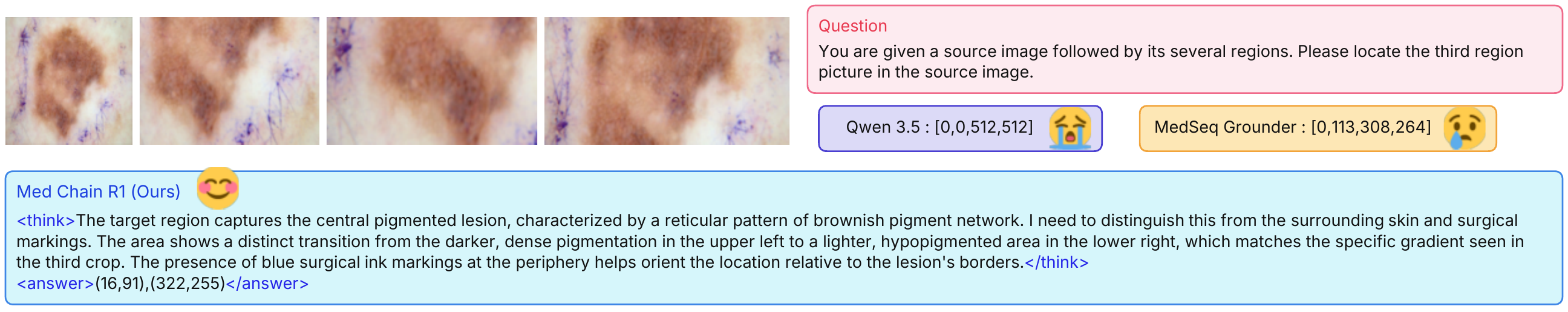}};
\node[anchor=north,inner sep=0pt] at (50.2,2){\adjincludegraphics[trim={0.089\width} {0.536\height} {0.796\width} {0.055\height},clip,width=35.8mm,height=25.3mm,keepaspectratio]{qual_case18.pdf}};
\node[anchor=north,inner sep=0pt] at (93.1,2){\adjincludegraphics[trim={0.208\width} {0.536\height} {0.657\width} {0.055\height},clip,width=42.1mm,height=25.3mm,keepaspectratio]{qual_case18.pdf}};
\node[anchor=north,inner sep=0pt] at (142.6,2){\adjincludegraphics[trim={0.347\width} {0.536\height} {0.496\width} {0.055\height},clip,width=48.8mm,height=25.3mm,keepaspectratio]{qual_case18.pdf}};
\node[box,fill=qCream] (question) at (0,32.3){\textbf{Question}\quad You are given a source image followed by its several regions. Please locate the third region picture in the source image.};
\node[box,fill=qSage,text width=73mm] (qwen) at ([yshift=-2mm]question.south west){\tikz[x=1mm,y=-1mm,baseline=-.7mm]{\filldraw[fill=yellow!55!white,draw=qInk!80,line width=.35pt] (0,0) circle (1.9mm);\draw[qInk,line width=.35pt,line cap=round] (-1.15,-.35) .. controls (-.95,-.8) and (-.5,-.8) .. (-.3,-.35) (.3,-.35) .. controls (.5,-.8) and (.95,-.8) .. (1.15,-.35);\fill[qBlue!65!cyan,rounded corners=.12mm] (-1,-.2) rectangle (-.65,1.25);\fill[qBlue!65!cyan,rounded corners=.12mm] (.65,-.2) rectangle (1,1.25);\fill[qInk!85] (0,.8) ellipse (.3mm and .42mm);}\quad Qwen 3.5 : [0,0,512,512]};
\node[box,fill=qPeach,text width=73mm] (medseq) at ([xshift=83mm,yshift=-2mm]question.south west){\tikz[x=1mm,y=-1mm,baseline=-.7mm]{\filldraw[fill=yellow!55!white,draw=qInk!80,line width=.35pt] (0,0) circle (1.9mm);\draw[qInk,line width=.35pt,line cap=round] (-1.15,-.35) .. controls (-.95,-.8) and (-.5,-.8) .. (-.3,-.35) (.3,-.35) .. controls (.5,-.8) and (.95,-.8) .. (1.15,-.35);\fill[qBlue!65!cyan,rounded corners=.12mm] (-1,-.2) rectangle (-.65,1.25);\fill[qBlue!65!cyan,rounded corners=.12mm] (.65,-.2) rectangle (1,1.25);\fill[qInk!85] (0,.8) ellipse (.3mm and .42mm);}\quad MedSeq Grounder : [0,113,308,264]};
\node[box,fill=qSky] (ours) at ([yshift=-2mm]qwen.south west){\tikz[x=1mm,y=-1mm,baseline=-.7mm]{\filldraw[fill=yellow!55!white,draw=qInk!80,line width=.35pt] (0,0) circle (1.9mm);\fill[qInk] (-.65,-.45) circle (.18mm);\fill[qInk] (.65,-.45) circle (.18mm);\draw[qInk,line width=.45pt,line cap=round] (-.85,.35) .. controls (-.45,1.2) and (.45,1.2) .. (.85,.35);}\quad\textbf{Med Chain R1 (Ours)}};
\node[box,fill=white,font=\rmfamily\fontsize{9.2}{11}\selectfont] at (ours.south west){\textless{}think\textgreater{}The target region captures the central pigmented lesion, characterized by a reticular pattern of brownish pigment network. I need to distinguish this from the surrounding skin and surgical markings. The area shows a distinct transition from the darker, dense pigmentation in the upper left to a lighter, hypopigmented area in the lower right, which matches the specific gradient seen in the third crop. The presence of blue surgical ink markings at the periphery helps orient the location relative to the lesion's borders.\textless{}/think\textgreater{} \textless{}answer\textgreater{}(16,91),(322,255)\textless{}/answer\textgreater{}};

\fill[qCream] ([yshift=-3mm]question.north east)--++(2mm,-1.5mm)--++(-2mm,-1.5mm)--cycle;
\fill[qSky] ([yshift=-3mm]ours.north east)--++(2mm,-1.5mm)--++(-2mm,-1.5mm)--cycle;
\begin{scope}[shift={([xshift=6mm,yshift=-4mm]question.north east)},scale=1.2]
\filldraw[fill=qPeach,draw=white,line width=1.2pt] (0,0) circle (3.5mm);
\fill[qInk!65] (-2,2.2) .. controls (-1.7,.5) and (1.7,.5) .. (2,2.2)--cycle;
\filldraw[fill=qCream,draw=qInk!65,line width=.35pt] (0,-.7) circle (1.25mm);
\draw[qInk!65,line width=.45pt] (-1.15,-1.15) .. controls (-.7,-2.25) and (.7,-2.25) .. (1.15,-1.15);
\fill[qInk] (-.4,-.7) circle (.12mm); \fill[qInk] (.4,-.7) circle (.12mm);
\end{scope}
\begin{scope}[shift={([xshift=6mm,yshift=-4mm]ours.north east)},scale=1.2]
\filldraw[fill=qSky,draw=white,line width=1.2pt] (0,0) circle (3.5mm);
\draw[qInk!65,line width=.45pt] (0,-1.5)--(0,-2.25);
\fill[qInk!65] (0,-2.4) circle (.3mm);
\filldraw[fill=white,draw=qInk!65,rounded corners=.65mm,line width=.45pt] (-2,-1.45) rectangle (2,1.55);
\fill[qInk!75] (-.8,-.25) circle (.3mm); \fill[qInk!75] (.8,-.25) circle (.3mm);
\draw[qInk!65,line width=.4pt] (-.75,.65) .. controls (-.25,1.2) and (.25,1.2) .. (.75,.65);
\end{scope}
\end{tikzpicture}%
}
\endgroup
\caption{
    \textbf{Case Study 18: Patch Grounding (Dermoscopy).}
    Given a dermoscopic source image followed by several of its regions, the task is to localize the third region in the source image.
    \textbf{Qwen3.5} predicts \texttt{[0,0,512,512]}, the whole image.
    \textbf{MedSeq-Grounder} predicts \texttt{[0,113,308,264]}.
    \textbf{\method{} (Ours)} uses the gradient from dense pigment network to a lighter hypopigmented area, together with the peripheral surgical ink markings, to place the region, and predicts
    \texttt{(16,91),(322,255)}.
}
\label{fig:qual_case18}
\end{figure*}

\begin{figure*}[p]
\centering
\begingroup
\definecolor{qInk}{HTML}{252629}\definecolor{qBlue}{HTML}{3B73A7}
\definecolor{qLine}{HTML}{B9BEC5}
\definecolor{qCream}{HTML}{FBF6E2}
\definecolor{qSage}{HTML}{EDF3E7}
\definecolor{qPeach}{HTML}{F5E3D5}
\definecolor{qSky}{HTML}{E3EDF7}
\definecolor{qBack}{HTML}{EEF2F5}
\resizebox{\textwidth}{!}{%
\begin{tikzpicture}[x=1mm,y=-1mm,every node/.style={font=\sffamily\fontsize{9}{11}\selectfont,text=qInk},box/.style={anchor=north west,align=left,inner sep=2mm,outer sep=0pt,rounded corners=2mm,draw=qLine!70,line width=.45pt,text width=156mm}]
\fill[qBack,rounded corners=2mm] (0,0) rectangle (170,35.6);
\node[anchor=north,inner sep=0pt] at (18.8,2){\adjincludegraphics[trim={0.015\width} {0.557\height} {0.888\width} {0.015\height},clip,width=31.6mm,height=31.6mm,keepaspectratio]{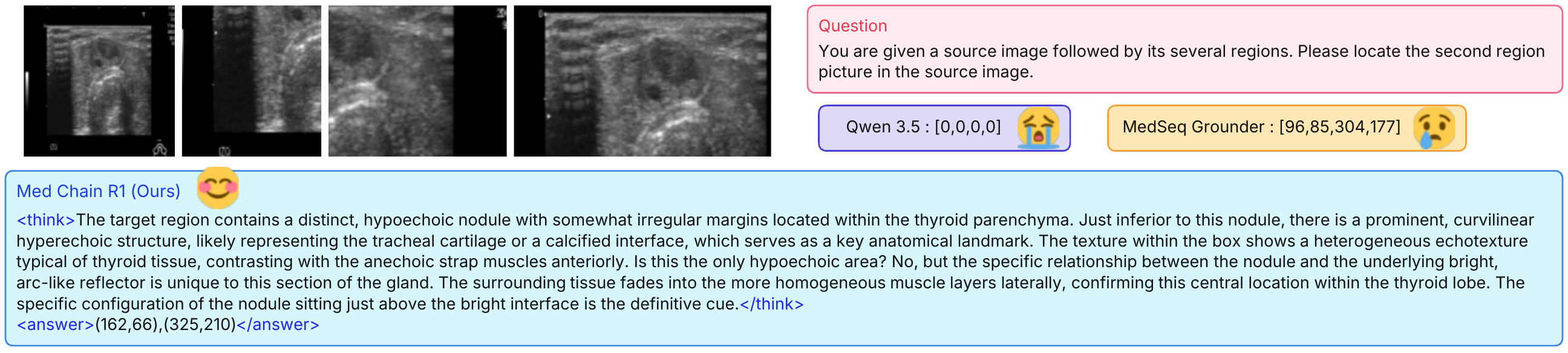}};
\node[anchor=north,inner sep=0pt] at (53.2,2){\adjincludegraphics[trim={0.116\width} {0.557\height} {0.795\width} {0.015\height},clip,width=29.2mm,height=31.6mm,keepaspectratio]{qual_case19.pdf}};
\node[anchor=north,inner sep=0pt] at (90.6,2){\adjincludegraphics[trim={0.21\width} {0.557\height} {0.676\width} {0.015\height},clip,width=37.4mm,height=31.6mm,keepaspectratio]{qual_case19.pdf}};
\node[anchor=north,inner sep=0pt] at (140.1,2){\adjincludegraphics[trim={0.328\width} {0.557\height} {0.508\width} {0.015\height},clip,width=53.7mm,height=31.6mm,keepaspectratio]{qual_case19.pdf}};
\node[box,fill=qCream] (question) at (0,38.6){\textbf{Question}\quad You are given a source image followed by its several regions. Please locate the second region picture in the source image.};
\node[box,fill=qSage,text width=73mm] (qwen) at ([yshift=-2mm]question.south west){\tikz[x=1mm,y=-1mm,baseline=-.7mm]{\filldraw[fill=yellow!55!white,draw=qInk!80,line width=.35pt] (0,0) circle (1.9mm);\draw[qInk,line width=.35pt,line cap=round] (-1.15,-.35) .. controls (-.95,-.8) and (-.5,-.8) .. (-.3,-.35) (.3,-.35) .. controls (.5,-.8) and (.95,-.8) .. (1.15,-.35);\fill[qBlue!65!cyan,rounded corners=.12mm] (-1,-.2) rectangle (-.65,1.25);\fill[qBlue!65!cyan,rounded corners=.12mm] (.65,-.2) rectangle (1,1.25);\fill[qInk!85] (0,.8) ellipse (.3mm and .42mm);}\quad Qwen 3.5 : [0,0,0,0]};
\node[box,fill=qPeach,text width=73mm] (medseq) at ([xshift=83mm,yshift=-2mm]question.south west){\tikz[x=1mm,y=-1mm,baseline=-.7mm]{\filldraw[fill=yellow!55!white,draw=qInk!80,line width=.35pt] (0,0) circle (1.9mm);\draw[qInk,line width=.35pt,line cap=round] (-1.15,-.35) .. controls (-.95,-.8) and (-.5,-.8) .. (-.3,-.35) (.3,-.35) .. controls (.5,-.8) and (.95,-.8) .. (1.15,-.35);\fill[qBlue!65!cyan,rounded corners=.12mm] (-1,-.2) rectangle (-.65,1.25);\fill[qBlue!65!cyan,rounded corners=.12mm] (.65,-.2) rectangle (1,1.25);\fill[qInk!85] (0,.8) ellipse (.3mm and .42mm);}\quad MedSeq Grounder : [96,85,304,177]};
\node[box,fill=qSky] (ours) at ([yshift=-2mm]qwen.south west){\tikz[x=1mm,y=-1mm,baseline=-.7mm]{\filldraw[fill=yellow!55!white,draw=qInk!80,line width=.35pt] (0,0) circle (1.9mm);\fill[qInk] (-.65,-.45) circle (.18mm);\fill[qInk] (.65,-.45) circle (.18mm);\draw[qInk,line width=.45pt,line cap=round] (-.85,.35) .. controls (-.45,1.2) and (.45,1.2) .. (.85,.35);}\quad\textbf{Med Chain R1 (Ours)}};
\node[box,fill=white,font=\rmfamily\fontsize{9.2}{11}\selectfont] at (ours.south west){\textless{}think\textgreater{}The target region contains a distinct, hypoechoic nodule with somewhat irregular margins located within the thyroid parenchyma. Just inferior to this nodule, there is a prominent, curvilinear hyperechoic structure, likely representing the tracheal cartilage or a calcified interface, which serves as a key anatomical landmark. The texture within the box shows a heterogeneous echotexture typical of thyroid tissue, contrasting with the anechoic strap muscles anteriorly. Is this the only hypoechoic area? No, but the specific relationship between the nodule and the underlying bright, arc-like reflector is unique to this section of the gland. The surrounding tissue fades into the more homogeneous muscle layers laterally, confirming this central location within the thyroid lobe. The specific configuration of the nodule sitting just above the bright interface is the definitive cue.\textless{}/think\textgreater{} \textless{}answer\textgreater{}(162,66),(325,210)\textless{}/answer\textgreater{}};

\fill[qCream] ([yshift=-3mm]question.north east)--++(2mm,-1.5mm)--++(-2mm,-1.5mm)--cycle;
\fill[qSky] ([yshift=-3mm]ours.north east)--++(2mm,-1.5mm)--++(-2mm,-1.5mm)--cycle;
\begin{scope}[shift={([xshift=6mm,yshift=-4mm]question.north east)},scale=1.2]
\filldraw[fill=qPeach,draw=white,line width=1.2pt] (0,0) circle (3.5mm);
\fill[qInk!65] (-2,2.2) .. controls (-1.7,.5) and (1.7,.5) .. (2,2.2)--cycle;
\filldraw[fill=qCream,draw=qInk!65,line width=.35pt] (0,-.7) circle (1.25mm);
\draw[qInk!65,line width=.45pt] (-1.15,-1.15) .. controls (-.7,-2.25) and (.7,-2.25) .. (1.15,-1.15);
\fill[qInk] (-.4,-.7) circle (.12mm); \fill[qInk] (.4,-.7) circle (.12mm);
\end{scope}
\begin{scope}[shift={([xshift=6mm,yshift=-4mm]ours.north east)},scale=1.2]
\filldraw[fill=qSky,draw=white,line width=1.2pt] (0,0) circle (3.5mm);
\draw[qInk!65,line width=.45pt] (0,-1.5)--(0,-2.25);
\fill[qInk!65] (0,-2.4) circle (.3mm);
\filldraw[fill=white,draw=qInk!65,rounded corners=.65mm,line width=.45pt] (-2,-1.45) rectangle (2,1.55);
\fill[qInk!75] (-.8,-.25) circle (.3mm); \fill[qInk!75] (.8,-.25) circle (.3mm);
\draw[qInk!65,line width=.4pt] (-.75,.65) .. controls (-.25,1.2) and (.25,1.2) .. (.75,.65);
\end{scope}
\end{tikzpicture}%
}
\endgroup
\caption{
    \textbf{Case Study 19: Patch Grounding (Thyroid Ultrasound).}
    Given a thyroid ultrasound source image followed by several of its regions, the task is to localize the second region in the source image.
    \textbf{Qwen3.5} predicts \texttt{[0,0,0,0]}, an empty box.
    \textbf{MedSeq-Grounder} predicts \texttt{[96,85,304,177]}.
    \textbf{\method{} (Ours)} anchors the region on a hypoechoic nodule sitting above a curvilinear hyperechoic interface, and predicts
    \texttt{(162,66),(325,210)}.
}
\label{fig:qual_case19}
\end{figure*}

\clearpage
\begin{figure*}[p]
\centering
\begingroup
\definecolor{qInk}{HTML}{252629}\definecolor{qBlue}{HTML}{3B73A7}
\definecolor{qLine}{HTML}{B9BEC5}
\definecolor{qCream}{HTML}{FBF6E2}
\definecolor{qSage}{HTML}{EDF3E7}
\definecolor{qPeach}{HTML}{F5E3D5}
\definecolor{qSky}{HTML}{E3EDF7}
\definecolor{qBack}{HTML}{EEF2F5}
\resizebox{\textwidth}{!}{%
\begin{tikzpicture}[x=1mm,y=-1mm,every node/.style={font=\sffamily\fontsize{9}{11}\selectfont,text=qInk},box/.style={anchor=north west,align=left,inner sep=2mm,outer sep=0pt,rounded corners=2mm,draw=qLine!70,line width=.45pt,text width=156mm}]
\fill[qBack,rounded corners=2mm] (0,0) rectangle (170,44);
\node[anchor=north,inner sep=0pt] at (41,2){\adjincludegraphics[trim={0.003\width} {0.197\height} {0.842\width} {0.194\height},clip,width=40.0mm,height=40mm,keepaspectratio]{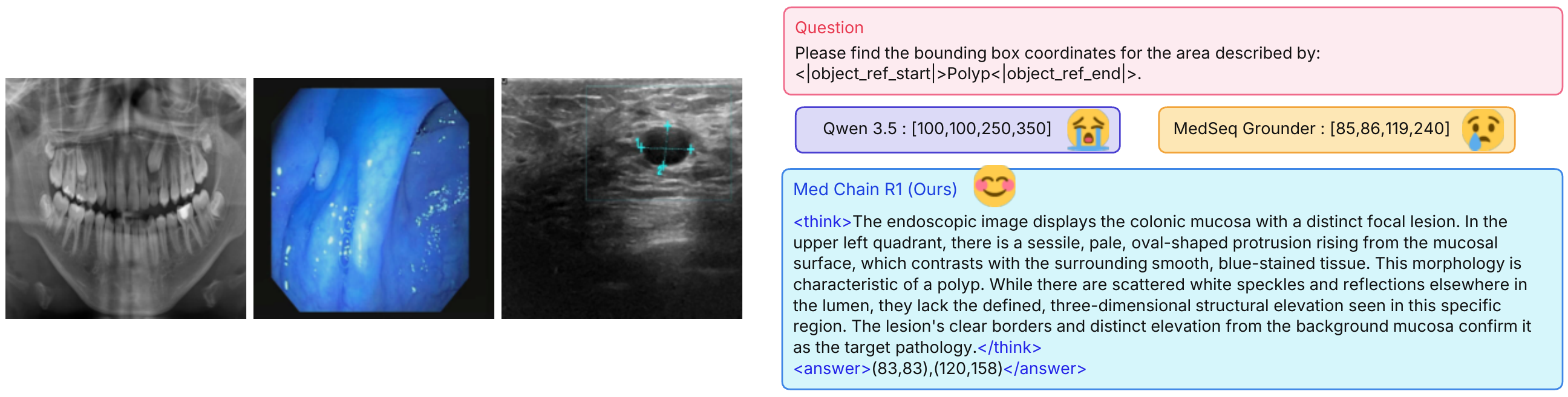}};
\node[anchor=north,inner sep=0pt] at (85,2){\adjincludegraphics[trim={0.162\width} {0.197\height} {0.684\width} {0.194\height},clip,width=40.0mm,height=40mm,keepaspectratio]{qual_case20.pdf}};
\node[anchor=north,inner sep=0pt] at (129,2){\adjincludegraphics[trim={0.32\width} {0.197\height} {0.526\width} {0.194\height},clip,width=40.0mm,height=40mm,keepaspectratio]{qual_case20.pdf}};
\node[box,fill=qCream] (question) at (0,47){\textbf{Question}\quad Please find the bounding box coordinates for the area described by: \textless{}|object\_ref\_start|\textgreater{}Polyp\textless{}|object\_ref\_end|\textgreater{}.};
\node[box,fill=qSage,text width=73mm] (qwen) at ([yshift=-2mm]question.south west){\tikz[x=1mm,y=-1mm,baseline=-.7mm]{\filldraw[fill=yellow!55!white,draw=qInk!80,line width=.35pt] (0,0) circle (1.9mm);\draw[qInk,line width=.35pt,line cap=round] (-1.15,-.35) .. controls (-.95,-.8) and (-.5,-.8) .. (-.3,-.35) (.3,-.35) .. controls (.5,-.8) and (.95,-.8) .. (1.15,-.35);\fill[qBlue!65!cyan,rounded corners=.12mm] (-1,-.2) rectangle (-.65,1.25);\fill[qBlue!65!cyan,rounded corners=.12mm] (.65,-.2) rectangle (1,1.25);\fill[qInk!85] (0,.8) ellipse (.3mm and .42mm);}\quad Qwen 3.5 : [100,100,250,350]};
\node[box,fill=qPeach,text width=73mm] (medseq) at ([xshift=83mm,yshift=-2mm]question.south west){\tikz[x=1mm,y=-1mm,baseline=-.7mm]{\filldraw[fill=yellow!55!white,draw=qInk!80,line width=.35pt] (0,0) circle (1.9mm);\draw[qInk,line width=.35pt,line cap=round] (-1.15,-.35) .. controls (-.95,-.8) and (-.5,-.8) .. (-.3,-.35) (.3,-.35) .. controls (.5,-.8) and (.95,-.8) .. (1.15,-.35);\fill[qBlue!65!cyan,rounded corners=.12mm] (-1,-.2) rectangle (-.65,1.25);\fill[qBlue!65!cyan,rounded corners=.12mm] (.65,-.2) rectangle (1,1.25);\fill[qInk!85] (0,.8) ellipse (.3mm and .42mm);}\quad MedSeq Grounder : [85,86,119,240]};
\node[box,fill=qSky] (ours) at ([yshift=-2mm]qwen.south west){\tikz[x=1mm,y=-1mm,baseline=-.7mm]{\filldraw[fill=yellow!55!white,draw=qInk!80,line width=.35pt] (0,0) circle (1.9mm);\fill[qInk] (-.65,-.45) circle (.18mm);\fill[qInk] (.65,-.45) circle (.18mm);\draw[qInk,line width=.45pt,line cap=round] (-.85,.35) .. controls (-.45,1.2) and (.45,1.2) .. (.85,.35);}\quad\textbf{Med Chain R1 (Ours)}};
\node[box,fill=white,font=\rmfamily\fontsize{9.2}{11}\selectfont] at (ours.south west){\textless{}think\textgreater{}The endoscopic image displays the colonic mucosa with a distinct focal lesion. In the upper left quadrant, there is a sessile, pale, oval-shaped protrusion rising from the mucosal surface, which contrasts with the surrounding smooth, blue-stained tissue. This morphology is characteristic of a polyp. While there are scattered white speckles and reflections elsewhere in the lumen, they lack the defined, three-dimensional structural elevation seen in this specific region. The lesion's clear borders and distinct elevation from the background mucosa confirm it as the target pathology.\textless{}/think\textgreater{} \textless{}answer\textgreater{}(83,83),(120,158)\textless{}/answer\textgreater{}};

\fill[qCream] ([yshift=-3mm]question.north east)--++(2mm,-1.5mm)--++(-2mm,-1.5mm)--cycle;
\fill[qSky] ([yshift=-3mm]ours.north east)--++(2mm,-1.5mm)--++(-2mm,-1.5mm)--cycle;
\begin{scope}[shift={([xshift=6mm,yshift=-4mm]question.north east)},scale=1.2]
\filldraw[fill=qPeach,draw=white,line width=1.2pt] (0,0) circle (3.5mm);
\fill[qInk!65] (-2,2.2) .. controls (-1.7,.5) and (1.7,.5) .. (2,2.2)--cycle;
\filldraw[fill=qCream,draw=qInk!65,line width=.35pt] (0,-.7) circle (1.25mm);
\draw[qInk!65,line width=.45pt] (-1.15,-1.15) .. controls (-.7,-2.25) and (.7,-2.25) .. (1.15,-1.15);
\fill[qInk] (-.4,-.7) circle (.12mm); \fill[qInk] (.4,-.7) circle (.12mm);
\end{scope}
\begin{scope}[shift={([xshift=6mm,yshift=-4mm]ours.north east)},scale=1.2]
\filldraw[fill=qSky,draw=white,line width=1.2pt] (0,0) circle (3.5mm);
\draw[qInk!65,line width=.45pt] (0,-1.5)--(0,-2.25);
\fill[qInk!65] (0,-2.4) circle (.3mm);
\filldraw[fill=white,draw=qInk!65,rounded corners=.65mm,line width=.45pt] (-2,-1.45) rectangle (2,1.55);
\fill[qInk!75] (-.8,-.25) circle (.3mm); \fill[qInk!75] (.8,-.25) circle (.3mm);
\draw[qInk!65,line width=.4pt] (-.75,.65) .. controls (-.25,1.2) and (.25,1.2) .. (.75,.65);
\end{scope}
\end{tikzpicture}%
}
\endgroup
\caption{
    \textbf{Case Study 20: Referring Grounding (Colonoscopy).}
    Given the referring expression \emph{Polyp}, the task is to localize the described region.
    \textbf{Qwen3.5} predicts \texttt{[100,100,250,350]}.
    \textbf{MedSeq-Grounder} predicts \texttt{[85,86,119,240]}.
    \textbf{\method{} (Ours)} describes a sessile, pale, oval protrusion in the upper left quadrant of the colonoscopic image, distinguishes it from scattered specular reflections, and predicts
    \texttt{(83,83),(120,158)}.
}
\label{fig:qual_case20}
\end{figure*}

\begin{figure*}[p]
\centering
\begingroup
\definecolor{qInk}{HTML}{252629}\definecolor{qBlue}{HTML}{3B73A7}
\definecolor{qLine}{HTML}{B9BEC5}
\definecolor{qCream}{HTML}{FBF6E2}
\definecolor{qSage}{HTML}{EDF3E7}
\definecolor{qPeach}{HTML}{F5E3D5}
\definecolor{qSky}{HTML}{E3EDF7}
\definecolor{qBack}{HTML}{EEF2F5}
\resizebox{\textwidth}{!}{%
\begin{tikzpicture}[x=1mm,y=-1mm,every node/.style={font=\sffamily\fontsize{9}{11}\selectfont,text=qInk},box/.style={anchor=north west,align=left,inner sep=2mm,outer sep=0pt,rounded corners=2mm,draw=qLine!70,line width=.45pt,text width=156mm}]
\fill[qBack,rounded corners=2mm] (0,0) rectangle (170,44);
\node[anchor=north,inner sep=0pt] at (41,2){\adjincludegraphics[trim={0.003\width} {0.238\height} {0.842\width} {0.236\height},clip,width=40.0mm,height=40mm,keepaspectratio]{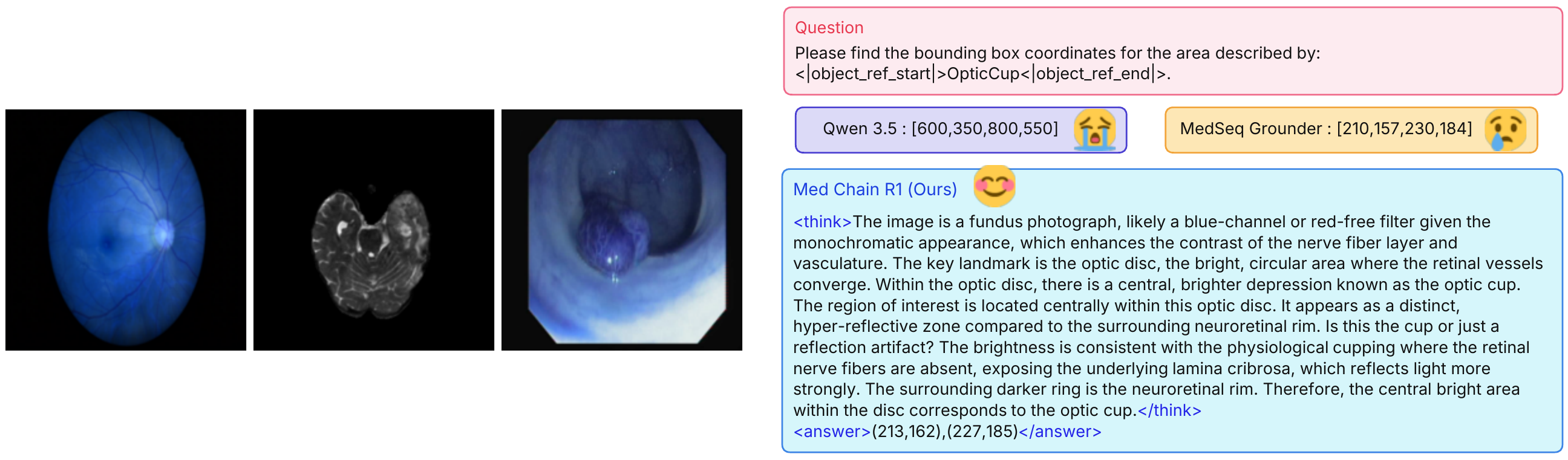}};
\node[anchor=north,inner sep=0pt] at (85,2){\adjincludegraphics[trim={0.162\width} {0.238\height} {0.684\width} {0.236\height},clip,width=40.0mm,height=40mm,keepaspectratio]{qual_case21.pdf}};
\node[anchor=north,inner sep=0pt] at (129,2){\adjincludegraphics[trim={0.32\width} {0.238\height} {0.526\width} {0.236\height},clip,width=40.0mm,height=40mm,keepaspectratio]{qual_case21.pdf}};
\node[box,fill=qCream] (question) at (0,47){\textbf{Question}\quad Please find the bounding box coordinates for the area described by: \textless{}|object\_ref\_start|\textgreater{}OpticCup\textless{}|object\_ref\_end|\textgreater{}.};
\node[box,fill=qSage,text width=73mm] (qwen) at ([yshift=-2mm]question.south west){\tikz[x=1mm,y=-1mm,baseline=-.7mm]{\filldraw[fill=yellow!55!white,draw=qInk!80,line width=.35pt] (0,0) circle (1.9mm);\draw[qInk,line width=.35pt,line cap=round] (-1.15,-.35) .. controls (-.95,-.8) and (-.5,-.8) .. (-.3,-.35) (.3,-.35) .. controls (.5,-.8) and (.95,-.8) .. (1.15,-.35);\fill[qBlue!65!cyan,rounded corners=.12mm] (-1,-.2) rectangle (-.65,1.25);\fill[qBlue!65!cyan,rounded corners=.12mm] (.65,-.2) rectangle (1,1.25);\fill[qInk!85] (0,.8) ellipse (.3mm and .42mm);}\quad Qwen 3.5 : [600,350,800,550]};
\node[box,fill=qPeach,text width=73mm] (medseq) at ([xshift=83mm,yshift=-2mm]question.south west){\tikz[x=1mm,y=-1mm,baseline=-.7mm]{\filldraw[fill=yellow!55!white,draw=qInk!80,line width=.35pt] (0,0) circle (1.9mm);\draw[qInk,line width=.35pt,line cap=round] (-1.15,-.35) .. controls (-.95,-.8) and (-.5,-.8) .. (-.3,-.35) (.3,-.35) .. controls (.5,-.8) and (.95,-.8) .. (1.15,-.35);\fill[qBlue!65!cyan,rounded corners=.12mm] (-1,-.2) rectangle (-.65,1.25);\fill[qBlue!65!cyan,rounded corners=.12mm] (.65,-.2) rectangle (1,1.25);\fill[qInk!85] (0,.8) ellipse (.3mm and .42mm);}\quad MedSeq Grounder : [210,157,230,184]};
\node[box,fill=qSky] (ours) at ([yshift=-2mm]qwen.south west){\tikz[x=1mm,y=-1mm,baseline=-.7mm]{\filldraw[fill=yellow!55!white,draw=qInk!80,line width=.35pt] (0,0) circle (1.9mm);\fill[qInk] (-.65,-.45) circle (.18mm);\fill[qInk] (.65,-.45) circle (.18mm);\draw[qInk,line width=.45pt,line cap=round] (-.85,.35) .. controls (-.45,1.2) and (.45,1.2) .. (.85,.35);}\quad\textbf{Med Chain R1 (Ours)}};
\node[box,fill=white,font=\rmfamily\fontsize{9.2}{11}\selectfont] at (ours.south west){\textless{}think\textgreater{}The image is a fundus photograph, likely a blue-channel or red-free filter given the monochromatic appearance, which enhances the contrast of the nerve fiber layer and vasculature. The key landmark is the optic disc, the bright, circular area where the retinal vessels converge. Within the optic disc, there is a central, brighter depression known as the optic cup. The region of interest is located centrally within this optic disc. It appears as a distinct, hyper-reflective zone compared to the surrounding neuroretinal rim. Is this the cup or just a reflection artifact? The brightness is consistent with the physiological cupping where the retinal nerve fibers are absent, exposing the underlying lamina cribrosa, which reflects light more strongly. The surrounding darker ring is the neuroretinal rim. Therefore, the central bright area within the disc corresponds to the optic cup.\textless{}/think\textgreater{} \textless{}answer\textgreater{}(213,162),(227,185)\textless{}/answer\textgreater{}};

\fill[qCream] ([yshift=-3mm]question.north east)--++(2mm,-1.5mm)--++(-2mm,-1.5mm)--cycle;
\fill[qSky] ([yshift=-3mm]ours.north east)--++(2mm,-1.5mm)--++(-2mm,-1.5mm)--cycle;
\begin{scope}[shift={([xshift=6mm,yshift=-4mm]question.north east)},scale=1.2]
\filldraw[fill=qPeach,draw=white,line width=1.2pt] (0,0) circle (3.5mm);
\fill[qInk!65] (-2,2.2) .. controls (-1.7,.5) and (1.7,.5) .. (2,2.2)--cycle;
\filldraw[fill=qCream,draw=qInk!65,line width=.35pt] (0,-.7) circle (1.25mm);
\draw[qInk!65,line width=.45pt] (-1.15,-1.15) .. controls (-.7,-2.25) and (.7,-2.25) .. (1.15,-1.15);
\fill[qInk] (-.4,-.7) circle (.12mm); \fill[qInk] (.4,-.7) circle (.12mm);
\end{scope}
\begin{scope}[shift={([xshift=6mm,yshift=-4mm]ours.north east)},scale=1.2]
\filldraw[fill=qSky,draw=white,line width=1.2pt] (0,0) circle (3.5mm);
\draw[qInk!65,line width=.45pt] (0,-1.5)--(0,-2.25);
\fill[qInk!65] (0,-2.4) circle (.3mm);
\filldraw[fill=white,draw=qInk!65,rounded corners=.65mm,line width=.45pt] (-2,-1.45) rectangle (2,1.55);
\fill[qInk!75] (-.8,-.25) circle (.3mm); \fill[qInk!75] (.8,-.25) circle (.3mm);
\draw[qInk!65,line width=.4pt] (-.75,.65) .. controls (-.25,1.2) and (.25,1.2) .. (.75,.65);
\end{scope}
\end{tikzpicture}%
}
\endgroup
\caption{
    \textbf{Case Study 21: Referring Grounding (Fundus Photography).}
    Given the referring expression \emph{OpticCup}, the task is to localize the described region.
    \textbf{Qwen3.5} predicts \texttt{[600,350,800,550]}.
    \textbf{MedSeq-Grounder} predicts \texttt{[210,157,230,184]}.
    \textbf{\method{} (Ours)} first locates the optic disc where the retinal vessels converge, then the central bright depression within it, and predicts
    \texttt{(213,162),(227,185)}.
}
\label{fig:qual_case21}
\end{figure*}

\clearpage

\end{document}